%% file: main.tex
\documentclass[10pt,twocolumn,letterpaper]{article}

\usepackage[pagenumbers]{cvpr} 

\input{preamble}

\newcommand{\cmark}{\ding{51}}%
\newcommand{\xmark}{\ding{55}}%

\definecolor{cvprblue}{rgb}{0.21,0.49,0.74}
\usepackage[pagebackref,breaklinks,colorlinks,allcolors=cvprblue]{hyperref}

\def\paperID{XXXX} 
\def\confName{CVPR}
\def\confYear{2026}

\title{SearchAD: Large-Scale Rare Image Retrieval Dataset for Autonomous Driving}

\author{Felix Embacher$^{1,2}$, \quad Jonas Uhrig$^{1}$, \quad Marius Cordts$^{1}$, \quad Markus Enzweiler$^{2}$ 
\\
\\
$^{1}$Mercedes-Benz AG, \quad $^{2}$Institute for Intelligent Systems, Esslingen University of Applied Sciences
}

\begin{document}
\input{sec/0_teaser_figure}
\maketitle
\input{sec/0_abstract}    
\input{sec/1_introduction}
\input{sec/2_related_work}
\input{sec/3_searchad_dataset}
\input{sec/4_image_retrieval_benchmark}
\input{sec/5_experiments}
\input{sec/6_conclusion}

{
    \small
        \bibliographystyle{ieeenat_fullname}
    \bibliography{main}
}

\input{sec/X_suppl}

\end{document}

%% file: preamble.tex
\newcommand{\myparagraph}[1]{\vspace{.5em}\noindent\textbf{#1.}~}

\usepackage{graphicx}
\usepackage{multirow}
\usepackage{array}
\usepackage{amssymb}
\usepackage{pifont}
\usepackage{siunitx}
\usepackage{pgfplots}
\pgfplotsset{compat=1.17}
\usepackage{svg}
\usepackage{xcolor} 

\usepackage{longtable}
\usepackage{makecell}
\usepackage[export]{adjustbox} 

\makeatletter
\newcommand{\inserttitleteaser}{} 

\newcommand{\titleteaser}[2][]{%
  \renewcommand{\inserttitleteaser}{%
    \begin{center}
      \vspace{-3mm}
      \captionsetup{type=figure}
      \includesvg[width=\textwidth]{#2}%
      \vspace{-1.2mm}
      \captionof{figure}{#1}%
      \label{fig:title_teaser}
      \vspace{1.2mm}
    \end{center}%
  }%
}

\definecolor{o}{RGB}{215,155,0} 
\definecolor{Animal}{RGB}{0,100,0}           
\definecolor{Human}{RGB}{0,0,139}       
\definecolor{Marking}{RGB}{139,69,19}       
\definecolor{Object}{RGB}{255,140,0}    
\definecolor{Rideable}{RGB}{153,50,204}         
\definecolor{Scene}{RGB}{218,165,32}       
\definecolor{Sign}{RGB}{192,192,192}        %
\definecolor{Trailer}{RGB}{178,34,34}        
\definecolor{Vehicle}{RGB}{30,144,255}         

\patchcmd{\@maketitle}
  {\vspace*{12pt}
   \end{center}}
  {\vspace*{12pt}
   \end{center}
   \inserttitleteaser}
  {}{}
\makeatother

%% file: sec/0_teaser_figure.tex
\titleteaser[Qualitative examples of rare object and scene annotations from the {SearchAD} training and validation splits. References for all underlying image sources are provided in \Cref{tab:searchad_dataset_collection}. \label{fig:qualitative_examples}]{images/searchad_collage.svg}

%% file: sec/0_abstract.tex
%
%

\begin{abstract}
Retrieving rare and safety-critical driving scenarios from large-scale datasets is essential for building robust autonomous driving (AD) systems. As dataset sizes continue to grow, the key challenge shifts from collecting more data to efficiently identifying the most relevant samples. 

We introduce SearchAD, a large-scale rare image retrieval dataset for AD containing over 423k frames drawn from 11 established datasets. SearchAD provides high-quality manual annotations of more than 513k bounding boxes covering 90 rare categories. It specifically targets the “needle-in-a-haystack” problem of locating extremely rare classes, with some appearing fewer than 50 times across the entire dataset. Unlike existing benchmarks, which focused on instance-level retrieval, SearchAD emphasizes semantic image retrieval with a well-defined data split, enabling text-to-image and image-to-image retrieval, few-shot learning, and fine-tuning of multi-modal retrieval models. 

Comprehensive evaluations show that text-based methods outperform image-based ones due to stronger inherent semantic grounding. While models directly aligning spatial visual features with language achieve the best zero-shot results, and our fine-tuning baseline significantly improves performance, absolute retrieval capabilities remain unsatisfactory. With a held-out test set on a public benchmark server, SearchAD establishes the first large-scale dataset for retrieval-driven data curation and long-tail perception research in AD: 
    \href{https://iis-esslingen.github.io/searchad/}{\small\texttt{https://iis-esslingen.github.io/searchad/}}
\end{abstract}

%% file: sec/1_introduction.tex
\section{Introduction}
\label{sec:introduction}

Autonomous driving (AD) systems are critically dependent on the quality and diversity of their training data. This dependence has only deepened with the rise of end-to-end driving models~\cite{PlanningOrientedAutonomousHu2023, DualADDisentanglingDynamicDoll2024, bevad} and the integration of vision–language models (VLMs) into the AD stack~\cite{SimLingoVisionOnlyRenz2025, OrionholisticendFu2025, LMDriveClosedLoopShao2024}. While ever-larger datasets can improve coverage of real-world variability, simply scaling data volume is insufficient and often prohibitively expensive to annotate or train comprehensively. Active learning studies, however, demonstrate that comparable performance can often be achieved with a well-chosen subset of samples~\cite{ActiveadPlanningorientedLu2024}. Thus, efficient data curation, particularly for rare, safety-critical, and long-tail events, has become essential for robust and safe AD systems. Moreover, large-scale in-car data collection from a customer fleet is infeasible without effective filtering mechanisms to recognize desired objects and scenes.


Image retrieval methods provide a natural solution to this challenge by enabling targeted searches for specific driving scenarios within large-scale unlabeled datasets. Recent advances in contrastive pre-training~\cite{LearningTransferableVisualRadford2021a} have empowered VLMs to perform image retrieval in a zero-shot fashion \cite{OpenclipIlharco2021, BLIP2BootstrappingLi2023, SigmoidLossLanguageZhai2023, Metaclip2Chuang2025}, even for object categories or scene types that are typically not annotated in AD datasets.

Current state-of-the-art image retrieval models \cite{simeoni2025dinov3, Siglip2MultilingualTschannen2025, RADIOv2.5ImprovedBaselinesHeinrich2025} have primarily
been evaluated on general purpose datasets such as $\mathcal{R}$Paris6K, $\mathcal{R}$Oxford5K~\cite{RevisitingOxfordParisRadenovic2018} for instance image retrieval and MSCOCO~\cite{MicrosoftCOCOCommonLin2014} and FLICKR30K~\cite{imagedescriptionsvisualYoung2014} for text-to-image and image-to-text retrieval. 
A large-scale AD dataset for \emph{semantic} image retrieval for rare objects and scenes does not exist yet. This domain poses particular, yet underexplored, challenges. It requires identifying semantically related but visually diverse instances that occur only a few times in the dataset. For example, tractors can vary greatly in shape, size, and color, yet they belong to the same semantic class. 
In addition, retrieving very small or distant objects~\cite{FindyourNeedleGreen2025} is particularly difficult but remains critical for curating safety-relevant data such as road debris, animals on the road, or rare traffic signs.

Our main contribution is SearchAD, a large-scale dataset and benchmark for semantic image retrieval of rare objects and scenes in AD. SearchAD integrates data from 11 established AD datasets (used with permission), comprising over 423k images. All frames were manually annotated through an eight-month, multi-stage labeling process with full human quality control, resulting in more than 513k high-quality bounding boxes across 90 rare categories. See \Cref{fig:qualitative_examples} for qualitative examples. The dataset exhibits a natural long-tail distribution and preserves the original training, validation, and test splits of the source datasets. SearchAD supports both text-to-image and image-to-image retrieval, few-shot learning, and fine-tuning of multimodal models, and can easily be extended to tasks such as open-world detection and rare object retrieval. To ensure unbiased evaluation, the benchmark includes a held-out test set.
We implement and adapt zero-shot baselines for both image-to-image and text-to-image retrieval and provide a finetuning baseline to establish reference performance and foster future research on retrieval-driven data curation and long-tail perception in AD.

%% file: sec/2_related_work.tex
\begin{table}[t!]
\centering
\small
\setlength{\tabcolsep}{5pt} 
\begin{tabular}{l S[table-format=6, group-separator={,}, group-minimum-digits=4] S[table-format=3] c}

\hline
\textbf{Datasets} & \textbf{\# Images} & \textbf{\# Classes} & \textbf{Non-Curated} \\ \hline
RoadAnomaly21 \cite{SegmentMeIfYouCanBenchmarkAnomalyChan2021} & 110 & 1 & \cmark \\ 
SOS \cite{TwoVideoDataMaag2022} & 8994 & 13 & \cmark \\
Vistas-NP \cite{DenseOpensetGrcic2021} & 20000 & 4 & \cmark \\
Lost and Found \cite{LostFounddetectingPinggera2016} & 2239 & 42 & \cmark \\
RoadObstacle21 \cite{SegmentMeIfYouCanBenchmarkAnomalyChan2021} & 442 & 1 & \cmark \\
BDD-Anomaly \cite{ScalingOutDistributionHendrycks2022} & 8000 & 3 & \xmark  \\
CODA \cite{CODARealWorldLi2022} & 1500 & 34 & \xmark \\
CODA 2022 \cite{CODARealWorldLi2022} & 9768 & 43 & \xmark \\
OpenAD \cite{OpenADOpenworldXia2024} & 17256 & 206 & \xmark \\ \hline
SearchAD (ours) & 423798 & 90 & \cmark \\ \hline
\end{tabular}
\caption{Overview of publicly available datasets used for anomaly detection and open-world object detection in comparison to the proposed SearchAD dataset. The table shows the number of images and classes, and if the scenes were pre-selected using data curation methods.}
\label{tab:open_world_and_anomaly_datasets}
\vspace{-3mm}
\end{table}
\begin{table*}[t!]
\small
\centering
\begin{tabular}{l c c S[table-format=6, group-separator={,}, group-minimum-digits=4] S[table-format=3] c S[table-format=6, group-separator={,}, group-minimum-digits=4]}
\hline
\textbf{Datasets} & \textbf{Exhaustive} & \textbf{Val. Set} & \textbf{\# Frames} & \textbf{\# Original Classes} & \textbf{\# SearchAD Classes} & \textbf{\# Objects} \\ \hline
Lost and Found \cite{LostFounddetectingPinggera2016} & \cmark & \xmark & 2239 & 42 & 18 & 2098 \\
WildDash2 \cite{UnifyingPanopticSegmentationZendel2022} & \cmark & \cmark & 5068 & 26 & 80 & 5032 \\
ACDC \cite{ACDCAdverseConditionsSakaridis2021} & \cmark & \cmark & 8012 & 19* & 60 & 7471 \\
IDD Segmentation \cite{IDDDatasetExploringVarma2019} & \cmark & \cmark & 10003 & 30* & 52 & 12192 \\
KITTI \cite{VisionmeetsroboticsGeiger2013} & \xmark & \xmark & 14999 & 8 & 47 & 9840 \\
Cityscapes \cite{CityscapesDatasetSemanticCordts2016} & \cmark & \cmark & 24998 & 30* & 75 & 31037 \\
Mapillary Vistas \cite{MapillaryVistasDatasetNeuhold2017} & \cmark & \cmark & 25000 & 66* & 86 & 35093 \\
ECP \cite{EuroCityPersonsNovelBraun2019} & \cmark & \cmark & 47335 & 8 & 76 & 33081 \\
nuScenes \cite{nuScenesMultimodalDatasetCaesar2020} & \xmark & \cmark & 80314 & 32* & 56 & 166152 \\
BDD100K \cite{BDD100KDiverseDrivingYu2020} & \cmark & \cmark & 100000 & 12* & 80 & 83102 \\
Mapillary Sign \cite{MapillaryTrafficSignErtler2020} & \cmark & \cmark & 105830 & 313** & 90 & 128167 \\ \hline
\textbf{SearchAD} & Combined & \cmark & 423798 & N/A & 90 & 513265 \\ \hline
\end{tabular}
\caption{Overview of the SearchAD dataset collection. All datasets are fully included except KITTI (only keyframes from the left stereo camera) and nuScenes (only samples from front and back cameras). The SearchAD dataset combines all above datasets and introduces annotations for rare objects and scenes. The "\# SearchAD Classes" column specifies how many of the 90 SearchAD classes can be found in the training and validation splits of each dataset. The "\# Objects" column gives the number of bounding box annotations for the 344,966 images of the training and validation splits, along with their distribution across the eleven datasets.
\\ * Number of classes including background classes, \eg \emph{road}, \emph{sky}, \emph{vegetation}. ** Exclusively traffic sign classes.
}
\label{tab:searchad_dataset_collection}
\vspace{-3mm}
\end{table*}
\section{Related Work}
\label{sec:related_work}

\subsection{AD Datasets featuring Rare Classes}

In the context of rare classes, several datasets have been proposed for road-anomalies and, more recently, for open-world perception (cf. \Cref{tab:open_world_and_anomaly_datasets}). Most studies \cite{DenseOpensetGrcic2021, ScalingOutDistributionHendrycks2022} repurpose existing classes such as motorcycles and bicycles as rare classes for anomaly detection, which limits the diversity of rare categories represented in these datasets.  
A dedicated road anomaly dataset is 
CODA~\cite{CODARealWorldLi2022}, a real-world corner-case dataset designed for open-world object detection, 
carefully selected from KITTI~\cite{VisionmeetsroboticsGeiger2013}, nuScenes~\cite{nuScenesMultimodalDatasetCaesar2020} and ONCE~\cite{OneMillionScenesMao2021}.
OpenAD~\cite{OpenADOpenworldXia2024} adopts a similar pre-selection strategy as CODA~\cite{CODARealWorldLi2022} but extends it to the 3D domain, enabling 3D open-world and anomaly detection. The dataset includes 206 automatically generated classes produced through an auto-labeling pipeline. However, even in the preview release, issues such as inconsistent category definitions (\eg \textit{Traffic cone}, \textit{Cone}, \textit{Cones}, \textit{Construction Cone}) highlight the limitations of automated labeling and the importance of consistent human annotation and quality control.

However, this pre-selection makes them impractical for evaluating image retrieval methods: the class distribution becomes biased, with rare objects appearing more frequently than in natural driving data, while the large-scale property of datasets with typically more than 100,000 frames is lost. An additional drawback is the risk that pre-selection based on predefined objects and scenes may fail to identify challenging instances, resulting in their exclusion. Reintroducing the discarded scenes is infeasible without re-annotating the few remaining rare objects from the selected classes. 

In recent years, datasets with long-tail distributions have gained more attention \cite{LargeScaleLongLiu2019, LVISDatasetLargeGupta2019, DistributionBalancedLossWu2020}. Within the AD domain, the the newly released Waymo Open Dataset for End-to-End Driving~\cite{WODE2EWaymoXu2025} explicitly includes long-tail scenarios, however, focusing on scenario-level annotations and also relying on prior filtering. SearchAD complements this direction by providing a naturally long-tailed dataset focused on rare objects and scenes without scene pre-selection but instead derived directly from unfiltered real-world data with exhaustive manual annotation and quality control to ensure unbiased representation of rare categories.


\subsection{Semantic Image Retrieval}
Image retrieval addresses the challenge of finding relevant images from a large dataset based on a search query that is either visual (image-to-image retrieval) or textual (text-to-image retrieval). A key distinction lies between instance and semantic image retrieval: while instance retrieval seeks the exact same object (\eg, the \emph{Eiffel Tower}), semantic retrieval aims for images of the same category (\eg, all images containing a \emph{cat}). Despite its practical importance for environment perception, semantic image retrieval remains relatively underexplored \cite{ContentBasedImageBarz2021}. Most prior work focuses on instance image retrieval ~\cite{ObjectretrievallargePhilbin2007, LostquantizationImprovingPhilbin2008, RevisitingOxfordParisRadenovic2018, SmoothAPSmoothingBrown2020}, likely due to the subjective nature of \emph{relevance} and \emph{similarity} and a lack of suitable semantic retrieval benchmarks \cite{ContentBasedImageBarz2021}. The proposed SearchAD benchmark addresses this gap by explicitly focusing on semantic retrieval within the AD domain.
 

%% file: sec/3_searchad_dataset.tex
\section{SearchAD Dataset}
\label{sec:searchad_dataset}
\subsection{Dataset Collection} 
Instead of initiating a new and expensive global data collection, we build SearchAD by leveraging existing AD datasets selected for their diversity, prevalence, high resolution, and scale. We selected datasets with a rich set of original classes to support future research in open-world object detection, which requires identifying both common and novel (out-of-distribution) objects. \Cref{tab:searchad_dataset_collection} summarizes the eleven datasets included in SearchAD.

To preserve the original data distributions and to facilitate reuse for related research, we exhaustively annotated the full datasets with two minor exceptions: (i) To reduce redundancy and avoid over-representation of nuScenes~\cite{nuScenesMultimodalDatasetCaesar2020}, we use only the keyframes from the front and rear cameras, contributing 80,314 images in total. (ii) For KITTI~\cite{VisionmeetsroboticsGeiger2013}, we restrict the selection to keyframes from the left stereo camera for the same reason. In total, combining all eleven sources results in a large-scale dataset of 423,798 images collected from diverse geographic regions worldwide.
\begin{table*}[t]
\centering
\small
\begin{tabular}{ll}
\toprule
\textbf{Category} & \textbf{SearchAD Classes} \\
\midrule
\multirow{2}{*}{Animal} & Real: Cat, Cow, Deer, Dog, Donkey, Horse, Sheep, Wildlife \\ 
& Statue: Cow, Deer, Elephant, Horse, Lion \\
Human & Construction Worker, Firefighter, Medical, On Loading Area, Police, Refuse Collector, With Sticks or Crutches \\
Marking & Bicycle Symbol, Bus Text, Stop Text, Temporarily Invalidated, Yellow Lane Arrow \\
\multirow{2}{*}{Object} & Ball, Beacon, Euro Pallet, Hand Dolly, Hydrant, Office Chair, Pallet Truck, Platform Truck, Rollator \\
& Shopping Cart, Shopping Trolley, Suitcase Trolley, Traffic Cone, Trash Bin, Wheelbarrow \\
Rideable & Cityscooter, Police Motorcycle, Quad, Segway, Skateboard, Skates, Ski, Stroller, Three Wheeler, Toy Car, Wheelchair \\
Scene & Active Amber Lights, Active Emergency Lights, Fog, Open Door, Open Hood, Open Trunk, Snow, Tunnel \\
Sign & Animal Sign, Road Bumper Sign, Temporarily Invalidated Sign, Train Sign, Warning Triangle \\
Trailer & Agricultural Trailer, Bicycle Trailer, Boat Trailer, Car Trailer, Caravan Trailer, Carriage, Warning Trailer \\
\multirow{3}{*}{Vehicle} & \begin{tabular}[t]{@{}l@{\hspace{0.5em}}l@{}} 
                           Construction: & Concrete Mixer, Excavator, Forklift, Harvester, Loader, Steamroller, Tractor, Truck Crane \\
                           Duty: & Fire, Garbage, Medical, Military, Police, Winter\\
                           Special: & Bicycle On Back, Bicycle On Roof, Car Truck, Recreational, Train \\
                           \end{tabular} \\
\bottomrule
\end{tabular}
\caption{Overview of all 90 rare classes of SearchAD grouped in the respective object and scene categories. \emph{Animals} are further subcategorized into \emph{real animals} and \emph{animal statues} and \emph{vehicles} into \emph{construction vehicles}, \emph{duty vehicles} and \emph{special vehicles}. Qualitative examples for all 90 SearchAD classes, along with details of the labeling guidelines, are provided in the Supplementary Material.}
\label{tab:searchad_categories}
\vspace{-3mm}
\end{table*}
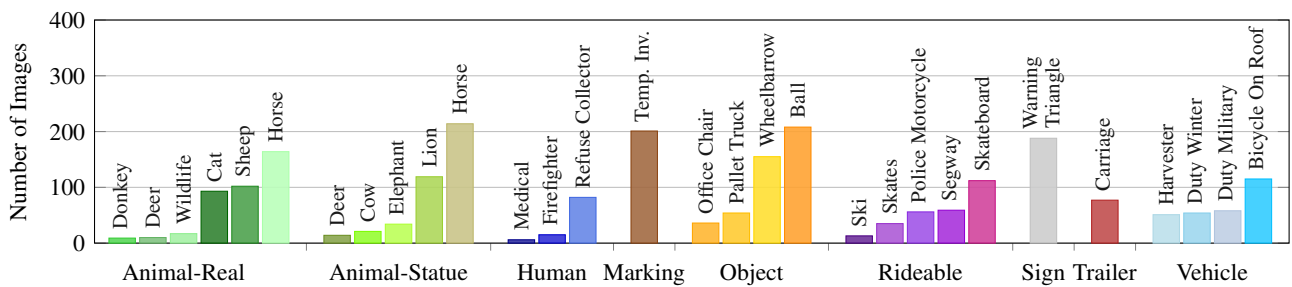
\begin{figure*}[t!]
    \centering
    \input{tikz/rare_class_distribution.tikz}
    \vspace{-2mm}
    \caption{Overview of the 30 rarest classes in the training and validation splits of SearchAD, showing the absolute number of relevant images for each class. All 30 classes appear fewer than 250 times, and some occur even less than 50 times.}
    \label{fig:twentyfive_rare_classes}
    \vspace{-3mm}
\end{figure*}
\subsection{Semantic Classes}
SearchAD focuses on rare classes. We define \textit{rarity} strictly by long-tail distribution in driving data, not global prevalence. For instance, while pets are common in daily life, they appear in very few driving frames. Yet, overlooking them poses severe risks for driving. Classes were selected based on driving impact (e.g., snowy scenes or invalid markings / signs), criticality (e.g. animals), and regulatory compliance (e.g. vulnerable road users) to challenge retrieval models. \Cref{tab:searchad_categories} summarizes the 90 SearchAD classes, which were designed to provide a comprehensive representation of nine categories relevant to AD perception. Qualitative examples are shown in \Cref{fig:qualitative_examples}. When selecting the label classes, we paid particular attention to making them as unambiguous as possible. This clear definition of relevance is crucial for constructing a high-quality semantic image retrieval benchmark where relevance plays a central role. It also helps annotators to consistently identify relevant objects, resulting in significantly improved labeling accuracy. \Cref{tab:searchad_dataset_collection} reveals that Mapillary Sign \cite{MapillaryTrafficSignErtler2020} is the only dataset to contain at least one instance of all 90 SearchAD classes. The large scale of SearchAD is therefore critical for providing sufficient representation of these rare classes.


\subsection{Dataset Splits}
To maintain the integrity of the original datasets and official benchmarks, the SearchAD test set is constructed as the union of the test splits from the eleven datasets listed in \Cref{tab:searchad_dataset_collection}. The corresponding annotations are hosted on a private benchmark server to prevent any form of test leakage. In addition, we provide training and validation splits to enable fine-tuning using the SearchAD bounding-box annotations. These splits are likewise derived from the respective train and validation partitions of the underlying datasets. Since Lost and Found \cite{LostFounddetectingPinggera2016} and KITTI \cite{VisionmeetsroboticsGeiger2013} do not include a separate validation set, they only contribute to the training and test partitions. Importantly, all 90 SearchAD classes are represented in each final dataset split. 

\subsection{Annotation Strategy}
\begin{figure}[t!]
    \centering
    \small
    \input{tikz/distribution_comparison.tikz}
    \vspace{-5mm}
    \caption{Distribution comparison showing the extreme rarity of SearchAD classes compared to BDD100K \cite{BDD100KDiverseDrivingYu2020} as a common and CODA \cite{CODARealWorldLi2022} as a corner case dataset. For each class, the class frequency (number of images containing the class) is presented per 100,000 images with CODA's counts extrapolated to this scale.
    }
    \label{fig:distribution_comparison}
    \vspace{-1mm}
\end{figure}
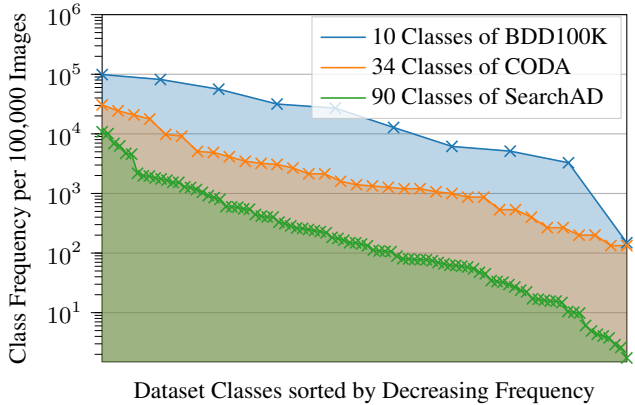

While recent corner case datasets like OpenAD \cite{OpenADOpenworldXia2024} rely on automatic corner case pre-selection and auto-labeling pipelines, our goal is to assess how effectively rare objects and scenes can be \emph{found}. To ensure a high benchmark quality and preserve the long-tail distribution (cf. \Cref{fig:distribution_comparison}), every image was manually inspected, including those that might appear irrelevant to automatic corner-case selection methods. 
Given the scale of SearchAD, this effort required an extensive eight-month annotation project conducted in collaboration with a specialized annotation company. The annotation criteria mandated that objects be visually recognizable in every frame and included if their occlusion is less than 90\%. We adopted a multi-stage annotation workflow with full (100\% of the frames) quality control by dedicated human experts. On average, annotators spent more than one minute per image, and each image was reviewed at least twice. The annotation partner was required to maintain a labeling accuracy of more than 98\% for each batch, verified by us on a randomly sampled 10\% data subset. Recognizing the challenge of identifying rare items in such a large corpus, particular emphasis was placed on maximizing recall.
\subsection{Dataset Statistics}
\begin{figure}[t!]
    \centering
    \small
    \input{tikz/box_plot_10_90.tikz}
    \vspace{-5mm}
    \caption{Object size distribution for each SearchAD category. The boxes delineate the interquartile range, whiskers indicate the 10th and 90th percentiles and red horizontal lines mark the median size. The \emph{Object} category includes the smallest instances, while scenes include the largest areas.
    }
    \label{fig:average_bbox_sizes}
    \vspace{-2mm}
\end{figure}
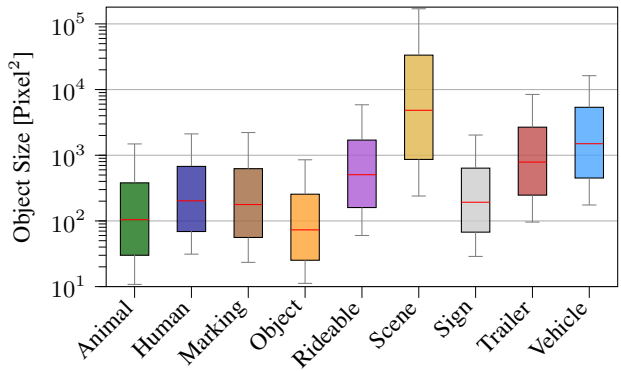

The SearchAD classes are designed to capture rare objects and scenes that would otherwise require extensive manual searching in large-scale datasets. \Cref{fig:twentyfive_rare_classes} highlights the scarcity of the 30 rarest classes in SearchAD. Notably, all these 30 classes appear in fewer than 250 of the 344,966 training and validation images, emphasizing their extreme rarity.
\Cref{fig:distribution_comparison} compares the class distribution to both a common AD dataset and a corner case dataset. Since CODA~\cite{CODARealWorldLi2022} focuses on corner case objects, the frequency of those classes is about an order of magnitude lower than the common classes of a dataset like BDD100K~\cite{BDD100KDiverseDrivingYu2020}. 
However, because scenes are pre-selected for rare events, these classes remain artificially oversampled relative to their natural prevalence. In contrast, SearchAD does not pre-select scenes and therefore preserves the natural long-tail distribution, with rare-class frequencies that are often another order of magnitude lower than in CODA.
In \Cref{fig:average_bbox_sizes}, the bounding box size distribution for each category is shown on a logarithmic scale. Across all categories, the objects only occupy a small proportion of the image, which may pose challenges for certain image retrieval methods \cite{FindyourNeedleGreen2025}. With a median bounding box size of less than $28\times28$ pixels, the category \emph{Object} comprises the smallest instances.

%% file: tikz/rare_class_distribution.tikz
\definecolor{AnimalRealCat}{RGB}{0,100,0}        
\definecolor{AnimalRealSheep}{RGB}{34,139,34}       
\definecolor{AnimalRealDonkey}{RGB}{50,205,50}         
\definecolor{AnimalRealDeer}{RGB}{100,180,100}      
\definecolor{AnimalRealExotic}{RGB}{144,238,144}        
\definecolor{AnimalRealHorse}{RGB}{170,255,170}      

\definecolor{AnimalStatueDeer}{RGB}{107,142,35}    
\definecolor{AnimalStatueCow}{RGB}{124,252,0}      
\definecolor{AnimalStatueElephant}{RGB}{173,255,47} 
\definecolor{AnimalStatueLion}{RGB}{154,205,50}    
\definecolor{AnimalStatueHorse}{RGB}{189,183,107}  

\definecolor{HumanMedical}{RGB}{0,0,139}       
\definecolor{HumanFirefighter}{RGB}{0,0,205}        
\definecolor{HumanRefuseCollector}{RGB}{65,105,225} 

\definecolor{MarkingTemporarilyInvalidated}{RGB}{139,69,19}       

\definecolor{ObjectBall}{RGB}{255,140,0}       
\definecolor{ObjectOfficeChair}{RGB}{255,165,0}    
\definecolor{ObjectPalletTruck}{RGB}{255,190,0}    
\definecolor{ObjectWheelbarrow}{RGB}{255,215,0}    

\definecolor{RideableSki}{RGB}{75,0,130}        
\definecolor{RideableSkates}{RGB}{153,50,204}         
\definecolor{RideablePoliceMotorcycle}{RGB}{138,43,226} 
\definecolor{RideableSegway}{RGB}{148,0,211}       
\definecolor{RideableSkateboard}{RGB}{199,21,133}  

\definecolor{SignWarningTriangle}{RGB}{192,192,192}        

\definecolor{TrailerCarriage}{RGB}{178,34,34}         

\definecolor{VehicleConstructionHarvester}{RGB}{173,216,230}    
\definecolor{VehicleDutyWinter}{RGB}{135,206,235}        
\definecolor{VehicleDutyMilitary}{RGB}{176,196,222}      
\definecolor{VehicleSpecialBicycleOnRoof}{RGB}{0,191,255}       

\begin{tikzpicture}
\tikzstyle{every node}=[font=\small]
\begin{axis}[
        ybar,
        width=\textwidth,
        height=4.55cm,
        xmin=-0.4,
        xmax=38.5,
        ymin=0,
        ymax=400,
        ytick={0, 100, 200, 300, 400},
        yticklabels={0, 100, 200, 300, 400},
        ylabel={Number of Images},
        xtick={2.5, 9.5, 14.5, 17.5, 21, 26.5, 30.5, 32.5, 36},
        minor xtick={6.5, 12.5, 16.5, 18.5, 23.5, 29.5, 31.5, 33.5},
        xticklabels = {
            Animal-Real,
            Animal-Statue,
            Human,
            Marking,
            Object,
            Rideable,
            Sign,
            Trailer,
            Vehicle,
        },
        major x tick style = {opacity=0},
        minor x tick num = 1, 
        xtick pos=left,
        ymajorgrids=true,
        every node near coord/.append style={
                anchor=west,
                rotate=90,
                font=\footnotesize,
        }
        ]

\addplot[bar shift=0pt, draw=AnimalRealDonkey, fill=AnimalRealDonkey!80!white, fill opacity=0.9, nodes near coords=Donkey] plot coordinates{ ( 0.5, 9 ) };
\addplot[bar shift=0pt, draw=AnimalRealDeer, fill=AnimalRealDeer!80!white, fill opacity=0.9, nodes near coords=Deer] plot coordinates{ ( 1.5, 10 ) };
\addplot[bar shift=0pt, draw=AnimalRealExotic, fill=AnimalRealExotic!80!white, fill opacity=0.9, nodes near coords=Wildlife] plot coordinates{ ( 2.5, 17 ) };
\addplot[bar shift=0pt, draw=AnimalRealCat, fill=AnimalRealCat!80!white, fill opacity=0.9, nodes near coords=Cat] plot coordinates{ ( 3.5, 93 ) };
\addplot[bar shift=0pt, draw=AnimalRealSheep, fill=AnimalRealSheep!80!white, fill opacity=0.9, nodes near coords=Sheep] plot coordinates{ ( 4.5, 102 ) };
\addplot[bar shift=0pt, draw=AnimalRealHorse, fill=AnimalRealHorse!80!white, fill opacity=0.9, nodes near coords=Horse] plot coordinates{ ( 5.5, 164 ) };

\addplot[bar shift=0pt, draw=AnimalStatueDeer, fill=AnimalStatueDeer!80!white, fill opacity=0.9, nodes near coords=Deer] plot coordinates{ ( 7.5, 14 ) };
\addplot[bar shift=0pt, draw=AnimalStatueCow, fill=AnimalStatueCow!80!white, fill opacity=0.9, nodes near coords=Cow] plot coordinates{ ( 8.5, 21 ) };
\addplot[bar shift=0pt, draw=AnimalStatueElephant, fill=AnimalStatueElephant!80!white, fill opacity=0.9, nodes near coords=Elephant] plot coordinates{ ( 9.5, 34 ) };
\addplot[bar shift=0pt, draw=AnimalStatueLion, fill=AnimalStatueLion!80!white, fill opacity=0.9, nodes near coords=Lion] plot coordinates{ ( 10.5, 119 ) };
\addplot[bar shift=0pt, draw=AnimalStatueHorse, fill=AnimalStatueHorse!80!white, fill opacity=0.9, nodes near coords=Horse] plot coordinates{ ( 11.5, 214 ) };

\addplot[bar shift=0pt, draw=HumanMedical, fill=HumanMedical!80!white, fill opacity=0.9, nodes near coords=Medical] plot coordinates{ ( 13.5, 6 ) };
\addplot[bar shift=0pt, draw=HumanFirefighter, fill=HumanFirefighter!80!white, fill opacity=0.9, nodes near coords=Firefighter] plot coordinates{ ( 14.5, 15 ) };
\addplot[bar shift=0pt, draw=HumanRefuseCollector, fill=HumanRefuseCollector!80!white, fill opacity=0.9, nodes near coords=Refuse Collector] plot coordinates{ ( 15.5, 82 ) };

\addplot[bar shift=0pt, draw=MarkingTemporarilyInvalidated, fill=MarkingTemporarilyInvalidated!80!white, fill opacity=0.9, nodes near coords={Temp. Inv.}] plot coordinates{ ( 17.5, 201 ) };

\addplot[bar shift=0pt, draw=ObjectOfficeChair, fill=ObjectOfficeChair!80!white, fill opacity=0.9, nodes near coords={Office Chair}] plot coordinates{ ( 19.5, 36 ) };
\addplot[bar shift=0pt, draw=ObjectPalletTruck, fill=ObjectPalletTruck!80!white, fill opacity=0.9, nodes near coords={Pallet Truck}] plot coordinates{ ( 20.5, 54 ) };
\addplot[bar shift=0pt, draw=ObjectWheelbarrow, fill=ObjectWheelbarrow!80!white, fill opacity=0.9, nodes near coords=Wheelbarrow] plot coordinates{ ( 21.5, 155 ) };
\addplot[bar shift=0pt, draw=ObjectBall, fill=ObjectBall!80!white, fill opacity=0.9, nodes near coords=Ball] plot coordinates{ ( 22.5, 208 ) };

\addplot[bar shift=0pt, draw=RideableSki, fill=RideableSki!80!white, fill opacity=0.9, nodes near coords=Ski] plot coordinates{ ( 24.5, 13 ) };
\addplot[bar shift=0pt, draw=RideableSkates, fill=RideableSkates!80!white, fill opacity=0.9, nodes near coords=Skates] plot coordinates{ ( 25.5, 35 ) };
\addplot[bar shift=0pt, draw=RideablePoliceMotorcycle, fill=RideablePoliceMotorcycle!80!white, fill opacity=0.9, nodes near coords={Police Motorcycle}] plot coordinates{ ( 26.5, 56 ) };
\addplot[bar shift=0pt, draw=RideableSegway, fill=RideableSegway!80!white, fill opacity=0.9, nodes near coords=Segway] plot coordinates{ ( 27.5, 59 ) };
\addplot[bar shift=0pt, draw=RideableSkateboard, fill=RideableSkateboard!80!white, fill opacity=0.9, nodes near coords=Skateboard] plot coordinates{ ( 28.5, 112 ) };

\addplot[bar shift=0pt, draw=SignWarningTriangle, fill=SignWarningTriangle!80!white, fill opacity=0.9, nodes near coords={\shortstack{Warning\\[-0.5ex]Triangle}}] plot coordinates{ ( 30.5, 188 ) };

\addplot[bar shift=0pt, draw=TrailerCarriage, fill=TrailerCarriage!80!white, fill opacity=0.9, nodes near coords=Carriage] plot coordinates{ ( 32.5, 77 ) };

\addplot[bar shift=0pt, draw=VehicleConstructionHarvester, fill=VehicleConstructionHarvester!80!white, fill opacity=0.9, nodes near coords=Harvester] plot coordinates{ ( 34.5, 51 ) };
\addplot[bar shift=0pt, draw=VehicleDutyWinter, fill=VehicleDutyWinter!80!white, fill opacity=0.9, nodes near coords=Duty Winter] plot coordinates{ ( 35.5, 54 ) };
\addplot[bar shift=0pt, draw=VehicleDutyMilitary, fill=VehicleDutyMilitary!80!white, fill opacity=0.9, nodes near coords=Duty Military] plot coordinates{ ( 36.5, 58 ) };
\addplot[bar shift=0pt, draw=VehicleSpecialBicycleOnRoof, fill=VehicleSpecialBicycleOnRoof!80!white, fill opacity=0.9, nodes near coords={Bicycle On Roof}] plot coordinates{ ( 37.5, 115 ) };

\end{axis}
\end{tikzpicture}

%% file: tikz/distribution_comparison.tikz
\begin{tikzpicture}

\definecolor{darkgray176}{RGB}{176,176,176}
\definecolor{darkorange25512714}{RGB}{255,127,14}
\definecolor{forestgreen4416044}{RGB}{44,160,44}
\definecolor{lightgray204}{RGB}{204,204,204}
\definecolor{steelblue31119180}{RGB}{31,119,180}

\begin{axis}[
width=0.49\textwidth,
height=6.2cm,
legend cell align={left},
legend style={yshift=2pt, xshift=2pt},
legend style={fill opacity=0.8, draw opacity=1, text opacity=1, draw=lightgray204},
log basis y={10},
tick align=outside,
tick pos=left,
x grid style={darkgray176},
xlabel={Dataset Classes sorted by Decreasing Frequency},
xmin=1, xmax=10,
xtick style={color=black},
xtick=\empty,
y grid style={darkgray176},
ylabel={Class Frequency per 100,000 Images},
ymajorgrids,
ymin=1.5, ymax=1000000,
ymode=log,
ytick style={color=black},
yticklabels={
  \(\displaystyle {10^{0}}\),
  \(\displaystyle {10^{1}}\),
  \(\displaystyle {10^{2}}\),
  \(\displaystyle {10^{3}}\),
  \(\displaystyle {10^{4}}\),
  \(\displaystyle {10^{5}}\),
  \(\displaystyle {10^{6}}\)
}
]
\path [draw=steelblue31119180, fill=steelblue31119180, opacity=0.3]
(axis cs:1,1)
--(axis cs:1,98688.75)
--(axis cs:2,81718.75)
--(axis cs:3,56112.5)
--(axis cs:4,31620)
--(axis cs:5,26973.75)
--(axis cs:6,12793.75)
--(axis cs:7,6151.25)
--(axis cs:8,5126.25)
--(axis cs:9,3272.5)
--(axis cs:10,148.75)
--(axis cs:10,1)
--(axis cs:10,1)
--(axis cs:9,1)
--(axis cs:8,1)
--(axis cs:7,1)
--(axis cs:6,1)
--(axis cs:5,1)
--(axis cs:4,1)
--(axis cs:3,1)
--(axis cs:2,1)
--(axis cs:1,1)
--cycle;

\path [draw=darkorange25512714, fill=darkorange25512714, opacity=0.3]
(axis cs:1,1)
--(axis cs:1,30600)
--(axis cs:1.27272727272727,24266.6666666667)
--(axis cs:1.54545454545455,20800)
--(axis cs:1.81818181818182,17733.3333333333)
--(axis cs:2.09090909090909,9800)
--(axis cs:2.36363636363636,9133.33333333333)
--(axis cs:2.63636363636364,5066.66666666667)
--(axis cs:2.90909090909091,4866.66666666667)
--(axis cs:3.18181818181818,4133.33333333333)
--(axis cs:3.45454545454545,3466.66666666667)
--(axis cs:3.72727272727273,3200)
--(axis cs:4,3066.66666666667)
--(axis cs:4.27272727272727,2666.66666666667)
--(axis cs:4.54545454545454,2133.33333333333)
--(axis cs:4.81818181818182,2133.33333333333)
--(axis cs:5.09090909090909,1600)
--(axis cs:5.36363636363636,1400)
--(axis cs:5.63636363636364,1333.33333333333)
--(axis cs:5.90909090909091,1266.66666666667)
--(axis cs:6.18181818181818,1200)
--(axis cs:6.45454545454545,1200)
--(axis cs:6.72727272727273,1066.66666666667)
--(axis cs:7,1000)
--(axis cs:7.27272727272727,866.666666666667)
--(axis cs:7.54545454545454,866.666666666667)
--(axis cs:7.81818181818182,533.333333333333)
--(axis cs:8.09090909090909,533.333333333333)
--(axis cs:8.36363636363636,400)
--(axis cs:8.63636363636364,266.666666666667)
--(axis cs:8.90909090909091,266.666666666667)
--(axis cs:9.18181818181818,200)
--(axis cs:9.45454545454545,200)
--(axis cs:9.72727272727273,133.333333333333)
--(axis cs:10,133.333333333333)
--(axis cs:10,1)
--(axis cs:10,1)
--(axis cs:9.72727272727273,1)
--(axis cs:9.45454545454545,1)
--(axis cs:9.18181818181818,1)
--(axis cs:8.90909090909091,1)
--(axis cs:8.63636363636364,1)
--(axis cs:8.36363636363636,1)
--(axis cs:8.09090909090909,1)
--(axis cs:7.81818181818182,1)
--(axis cs:7.54545454545454,1)
--(axis cs:7.27272727272727,1)
--(axis cs:7,1)
--(axis cs:6.72727272727273,1)
--(axis cs:6.45454545454545,1)
--(axis cs:6.18181818181818,1)
--(axis cs:5.90909090909091,1)
--(axis cs:5.63636363636364,1)
--(axis cs:5.36363636363636,1)
--(axis cs:5.09090909090909,1)
--(axis cs:4.81818181818182,1)
--(axis cs:4.54545454545454,1)
--(axis cs:4.27272727272727,1)
--(axis cs:4,1)
--(axis cs:3.72727272727273,1)
--(axis cs:3.45454545454545,1)
--(axis cs:3.18181818181818,1)
--(axis cs:2.90909090909091,1)
--(axis cs:2.63636363636364,1)
--(axis cs:2.36363636363636,1)
--(axis cs:2.09090909090909,1)
--(axis cs:1.81818181818182,1)
--(axis cs:1.54545454545455,1)
--(axis cs:1.27272727272727,1)
--(axis cs:1,1)
--cycle;

\path [draw=forestgreen4416044, fill=forestgreen4416044, opacity=0.3]
(axis cs:1,1)
--(axis cs:1,10828.6033985958)
--(axis cs:1.10112359550562,10084.4720929019)
--(axis cs:1.20224719101124,6885.60611770435)
--(axis cs:1.30337078651685,6280.03919226822)
--(axis cs:1.40449438202247,4698.72393221361)
--(axis cs:1.50561797752809,4564.79769020715)
--(axis cs:1.60674157303371,2179.05532719167)
--(axis cs:1.70786516853933,1978.45584782268)
--(axis cs:1.80898876404494,1960.19317845817)
--(axis cs:1.91011235955056,1816.70077630839)
--(axis cs:2.01123595505618,1753.50614263435)
--(axis cs:2.1123595505618,1667.70058498519)
--(axis cs:2.21348314606742,1536.9630630265)
--(axis cs:2.31460674157303,1535.22376118226)
--(axis cs:2.41573033707865,1288.53278294093)
--(axis cs:2.51685393258427,1256.35569882249)
--(axis cs:2.61797752808989,1161.56374831143)
--(axis cs:2.71910112359551,1043.29122290313)
--(axis cs:2.82022471910112,918.061490117867)
--(axis cs:2.92134831460674,850.518601833224)
--(axis cs:3.02247191011236,799.499081068859)
--(axis cs:3.12359550561798,600.638903544117)
--(axis cs:3.2247191011236,595.420998011398)
--(axis cs:3.32584269662921,590.203092478679)
--(axis cs:3.42696629213483,551.358684623992)
--(axis cs:3.52808988764045,544.111593606326)
--(axis cs:3.62921348314607,443.521970281129)
--(axis cs:3.73033707865169,417.432442617533)
--(axis cs:3.8314606741573,403.518027863615)
--(axis cs:3.93258426966292,402.068609660082)
--(axis cs:4.03370786516854,330.177466764841)
--(axis cs:4.13483146067416,311.045146478204)
--(axis cs:4.23595505617978,285.535386096021)
--(axis cs:4.33707865168539,263.214345761611)
--(axis cs:4.43820224719101,258.576207510305)
--(axis cs:4.53932584269663,248.140396444867)
--(axis cs:4.64044943820225,238.284352660842)
--(axis cs:4.74157303370787,229.587843439643)
--(axis cs:4.84269662921348,218.862148733498)
--(axis cs:4.9438202247191,183.206460926584)
--(axis cs:5.04494382022472,175.959369908919)
--(axis cs:5.14606741573034,164.364024280654)
--(axis cs:5.24719101123596,149.290074963909)
--(axis cs:5.34831460674157,148.42042404179)
--(axis cs:5.44943820224719,141.463216664831)
--(axis cs:5.55056179775281,132.766707443632)
--(axis cs:5.65168539325843,112.474852594169)
--(axis cs:5.75280898876404,108.416481624276)
--(axis cs:5.85393258426966,106.967063420743)
--(axis cs:5.95505617977528,105.807528857916)
--(axis cs:6.0561797752809,88.7043940562258)
--(axis cs:6.15730337078652,80.2977684757338)
--(axis cs:6.25842696629213,79.7180011943206)
--(axis cs:6.35955056179775,77.9786993500809)
--(axis cs:6.46067415730337,77.6888157093743)
--(axis cs:6.56179775280899,77.3989320686676)
--(axis cs:6.66292134831461,74.2102120208948)
--(axis cs:6.76404494382022,69.5720737695889)
--(axis cs:6.86516853932584,66.0934700811094)
--(axis cs:6.96629213483146,62.61486639263)
--(axis cs:7.06741573033708,62.0350991112168)
--(axis cs:7.1685393258427,60.295797266977)
--(axis cs:7.26966292134831,58.2666117820307)
--(axis cs:7.37078651685393,54.4981244528446)
--(axis cs:7.47191011235955,47.5409170758857)
--(axis cs:7.57303370786517,44.9319643095262)
--(axis cs:7.67415730337079,34.4961532440878)
--(axis cs:7.7752808988764,33.3366186812613)
--(axis cs:7.87640449438202,32.4669677591415)
--(axis cs:7.97752808988764,29.5681313520753)
--(axis cs:8.07865168539326,26.9591785857157)
--(axis cs:8.17977528089888,23.7704585379429)
--(axis cs:8.28089887640449,22.3210403344098)
--(axis cs:8.38202247191011,17.1031348016906)
--(axis cs:8.48314606741573,16.813251160984)
--(axis cs:8.58426966292135,16.2334838795707)
--(axis cs:8.68539325842697,15.6537165981575)
--(axis cs:8.78651685393258,15.6537165981575)
--(axis cs:8.8876404494382,14.7840656760376)
--(axis cs:8.98876404494382,10.4358110654383)
--(axis cs:9.08988764044944,10.1459274247317)
--(axis cs:9.19101123595506,9.85604378402509)
--(axis cs:9.29213483146067,6.08755645483903)
--(axis cs:9.39325842696629,4.92802189201255)
--(axis cs:9.49438202247191,4.34825461059931)
--(axis cs:9.59550561797753,4.05837096989269)
--(axis cs:9.69662921348315,3.76848732918606)
--(axis cs:9.79775280898876,2.8988364070662)
--(axis cs:9.89887640449438,2.60895276635958)
--(axis cs:10,1.73930184423972)
--(axis cs:10,1)
--(axis cs:10,1)
--(axis cs:9.89887640449438,1)
--(axis cs:9.79775280898876,1)
--(axis cs:9.69662921348315,1)
--(axis cs:9.59550561797753,1)
--(axis cs:9.49438202247191,1)
--(axis cs:9.39325842696629,1)
--(axis cs:9.29213483146067,1)
--(axis cs:9.19101123595506,1)
--(axis cs:9.08988764044944,1)
--(axis cs:8.98876404494382,1)
--(axis cs:8.8876404494382,1)
--(axis cs:8.78651685393258,1)
--(axis cs:8.68539325842697,1)
--(axis cs:8.58426966292135,1)
--(axis cs:8.48314606741573,1)
--(axis cs:8.38202247191011,1)
--(axis cs:8.28089887640449,1)
--(axis cs:8.17977528089888,1)
--(axis cs:8.07865168539326,1)
--(axis cs:7.97752808988764,1)
--(axis cs:7.87640449438202,1)
--(axis cs:7.7752808988764,1)
--(axis cs:7.67415730337079,1)
--(axis cs:7.57303370786517,1)
--(axis cs:7.47191011235955,1)
--(axis cs:7.37078651685393,1)
--(axis cs:7.26966292134831,1)
--(axis cs:7.1685393258427,1)
--(axis cs:7.06741573033708,1)
--(axis cs:6.96629213483146,1)
--(axis cs:6.86516853932584,1)
--(axis cs:6.76404494382022,1)
--(axis cs:6.66292134831461,1)
--(axis cs:6.56179775280899,1)
--(axis cs:6.46067415730337,1)
--(axis cs:6.35955056179775,1)
--(axis cs:6.25842696629213,1)
--(axis cs:6.15730337078652,1)
--(axis cs:6.0561797752809,1)
--(axis cs:5.95505617977528,1)
--(axis cs:5.85393258426966,1)
--(axis cs:5.75280898876404,1)
--(axis cs:5.65168539325843,1)
--(axis cs:5.55056179775281,1)
--(axis cs:5.44943820224719,1)
--(axis cs:5.34831460674157,1)
--(axis cs:5.24719101123596,1)
--(axis cs:5.14606741573034,1)
--(axis cs:5.04494382022472,1)
--(axis cs:4.9438202247191,1)
--(axis cs:4.84269662921348,1)
--(axis cs:4.74157303370787,1)
--(axis cs:4.64044943820225,1)
--(axis cs:4.53932584269663,1)
--(axis cs:4.43820224719101,1)
--(axis cs:4.33707865168539,1)
--(axis cs:4.23595505617978,1)
--(axis cs:4.13483146067416,1)
--(axis cs:4.03370786516854,1)
--(axis cs:3.93258426966292,1)
--(axis cs:3.8314606741573,1)
--(axis cs:3.73033707865169,1)
--(axis cs:3.62921348314607,1)
--(axis cs:3.52808988764045,1)
--(axis cs:3.42696629213483,1)
--(axis cs:3.32584269662921,1)
--(axis cs:3.2247191011236,1)
--(axis cs:3.12359550561798,1)
--(axis cs:3.02247191011236,1)
--(axis cs:2.92134831460674,1)
--(axis cs:2.82022471910112,1)
--(axis cs:2.71910112359551,1)
--(axis cs:2.61797752808989,1)
--(axis cs:2.51685393258427,1)
--(axis cs:2.41573033707865,1)
--(axis cs:2.31460674157303,1)
--(axis cs:2.21348314606742,1)
--(axis cs:2.1123595505618,1)
--(axis cs:2.01123595505618,1)
--(axis cs:1.91011235955056,1)
--(axis cs:1.80898876404494,1)
--(axis cs:1.70786516853933,1)
--(axis cs:1.60674157303371,1)
--(axis cs:1.50561797752809,1)
--(axis cs:1.40449438202247,1)
--(axis cs:1.30337078651685,1)
--(axis cs:1.20224719101124,1)
--(axis cs:1.10112359550562,1)
--(axis cs:1,1)
--cycle;

\addplot [semithick, steelblue31119180]
table {%
1 98688.75
2 81718.75
3 56112.5
4 31620
5 26973.75
6 12793.75
7 6151.25
8 5126.25
9 3272.5
10 148.75
};
\addlegendentry{10 Classes of BDD100K}
\addplot [semithick, steelblue31119180, mark=x, mark size=3, mark options={solid}, only marks, forget plot]
table {%
1 98688.75
2 81718.75
3 56112.5
4 31620
5 26973.75
6 12793.75
7 6151.25
8 5126.25
9 3272.5
10 148.75
};
\addplot [semithick, darkorange25512714]
table {%
1 30600
1.27272727272727 24266.6666666667
1.54545454545455 20800
1.81818181818182 17733.3333333333
2.09090909090909 9800
2.36363636363636 9133.33333333333
2.63636363636364 5066.66666666667
2.90909090909091 4866.66666666667
3.18181818181818 4133.33333333333
3.45454545454545 3466.66666666667
3.72727272727273 3200
4 3066.66666666667
4.27272727272727 2666.66666666667
4.54545454545454 2133.33333333333
4.81818181818182 2133.33333333333
5.09090909090909 1600
5.36363636363636 1400
5.63636363636364 1333.33333333333
5.90909090909091 1266.66666666667
6.18181818181818 1200
6.45454545454545 1200
6.72727272727273 1066.66666666667
7 1000
7.27272727272727 866.666666666667
7.54545454545454 866.666666666667
7.81818181818182 533.333333333333
8.09090909090909 533.333333333333
8.36363636363636 400
8.63636363636364 266.666666666667
8.90909090909091 266.666666666667
9.18181818181818 200
9.45454545454545 200
9.72727272727273 133.333333333333
10 133.333333333333
};
\addlegendentry{34 Classes of CODA}
\addplot [semithick, darkorange25512714, mark=x, mark size=3, mark options={solid}, only marks, forget plot]
table {%
1 30600
1.27272727272727 24266.6666666667
1.54545454545455 20800
1.81818181818182 17733.3333333333
2.09090909090909 9800
2.36363636363636 9133.33333333333
2.63636363636364 5066.66666666667
2.90909090909091 4866.66666666667
3.18181818181818 4133.33333333333
3.45454545454545 3466.66666666667
3.72727272727273 3200
4 3066.66666666667
4.27272727272727 2666.66666666667
4.54545454545454 2133.33333333333
4.81818181818182 2133.33333333333
5.09090909090909 1600
5.36363636363636 1400
5.63636363636364 1333.33333333333
5.90909090909091 1266.66666666667
6.18181818181818 1200
6.45454545454545 1200
6.72727272727273 1066.66666666667
7 1000
7.27272727272727 866.666666666667
7.54545454545454 866.666666666667
7.81818181818182 533.333333333333
8.09090909090909 533.333333333333
8.36363636363636 400
8.63636363636364 266.666666666667
8.90909090909091 266.666666666667
9.18181818181818 200
9.45454545454545 200
9.72727272727273 133.333333333333
10 133.333333333333
};
\addplot [semithick, forestgreen4416044]
table {%
1 10828.6033985958
1.10112359550562 10084.4720929019
1.20224719101124 6885.60611770435
1.30337078651685 6280.03919226822
1.40449438202247 4698.72393221361
1.50561797752809 4564.79769020715
1.60674157303371 2179.05532719167
1.70786516853933 1978.45584782268
1.80898876404494 1960.19317845817
1.91011235955056 1816.70077630839
2.01123595505618 1753.50614263435
2.1123595505618 1667.70058498519
2.21348314606742 1536.9630630265
2.31460674157303 1535.22376118226
2.41573033707865 1288.53278294093
2.51685393258427 1256.35569882249
2.61797752808989 1161.56374831143
2.71910112359551 1043.29122290313
2.82022471910112 918.061490117867
2.92134831460674 850.518601833224
3.02247191011236 799.499081068859
3.12359550561798 600.638903544117
3.2247191011236 595.420998011398
3.32584269662921 590.203092478679
3.42696629213483 551.358684623992
3.52808988764045 544.111593606326
3.62921348314607 443.521970281129
3.73033707865169 417.432442617533
3.8314606741573 403.518027863615
3.93258426966292 402.068609660082
4.03370786516854 330.177466764841
4.13483146067416 311.045146478204
4.23595505617978 285.535386096021
4.33707865168539 263.214345761611
4.43820224719101 258.576207510305
4.53932584269663 248.140396444867
4.64044943820225 238.284352660842
4.74157303370787 229.587843439643
4.84269662921348 218.862148733498
4.9438202247191 183.206460926584
5.04494382022472 175.959369908919
5.14606741573034 164.364024280654
5.24719101123596 149.290074963909
5.34831460674157 148.42042404179
5.44943820224719 141.463216664831
5.55056179775281 132.766707443632
5.65168539325843 112.474852594169
5.75280898876404 108.416481624276
5.85393258426966 106.967063420743
5.95505617977528 105.807528857916
6.0561797752809 88.7043940562258
6.15730337078652 80.2977684757338
6.25842696629213 79.7180011943206
6.35955056179775 77.9786993500809
6.46067415730337 77.6888157093743
6.56179775280899 77.3989320686676
6.66292134831461 74.2102120208948
6.76404494382022 69.5720737695889
6.86516853932584 66.0934700811094
6.96629213483146 62.61486639263
7.06741573033708 62.0350991112168
7.1685393258427 60.295797266977
7.26966292134831 58.2666117820307
7.37078651685393 54.4981244528446
7.47191011235955 47.5409170758857
7.57303370786517 44.9319643095262
7.67415730337079 34.4961532440878
7.7752808988764 33.3366186812613
7.87640449438202 32.4669677591415
7.97752808988764 29.5681313520753
8.07865168539326 26.9591785857157
8.17977528089888 23.7704585379429
8.28089887640449 22.3210403344098
8.38202247191011 17.1031348016906
8.48314606741573 16.813251160984
8.58426966292135 16.2334838795707
8.68539325842697 15.6537165981575
8.78651685393258 15.6537165981575
8.8876404494382 14.7840656760376
8.98876404494382 10.4358110654383
9.08988764044944 10.1459274247317
9.19101123595506 9.85604378402509
9.29213483146067 6.08755645483903
9.39325842696629 4.92802189201255
9.49438202247191 4.34825461059931
9.59550561797753 4.05837096989269
9.69662921348315 3.76848732918606
9.79775280898876 2.8988364070662
9.89887640449438 2.60895276635958
10 1.73930184423972
};
\addlegendentry{90 Classes of SearchAD}
\addplot [semithick, forestgreen4416044, mark=x, mark size=3, mark options={solid}, only marks, forget plot]
table {%
1 10828.6033985958
1.10112359550562 10084.4720929019
1.20224719101124 6885.60611770435
1.30337078651685 6280.03919226822
1.40449438202247 4698.72393221361
1.50561797752809 4564.79769020715
1.60674157303371 2179.05532719167
1.70786516853933 1978.45584782268
1.80898876404494 1960.19317845817
1.91011235955056 1816.70077630839
2.01123595505618 1753.50614263435
2.1123595505618 1667.70058498519
2.21348314606742 1536.9630630265
2.31460674157303 1535.22376118226
2.41573033707865 1288.53278294093
2.51685393258427 1256.35569882249
2.61797752808989 1161.56374831143
2.71910112359551 1043.29122290313
2.82022471910112 918.061490117867
2.92134831460674 850.518601833224
3.02247191011236 799.499081068859
3.12359550561798 600.638903544117
3.2247191011236 595.420998011398
3.32584269662921 590.203092478679
3.42696629213483 551.358684623992
3.52808988764045 544.111593606326
3.62921348314607 443.521970281129
3.73033707865169 417.432442617533
3.8314606741573 403.518027863615
3.93258426966292 402.068609660082
4.03370786516854 330.177466764841
4.13483146067416 311.045146478204
4.23595505617978 285.535386096021
4.33707865168539 263.214345761611
4.43820224719101 258.576207510305
4.53932584269663 248.140396444867
4.64044943820225 238.284352660842
4.74157303370787 229.587843439643
4.84269662921348 218.862148733498
4.9438202247191 183.206460926584
5.04494382022472 175.959369908919
5.14606741573034 164.364024280654
5.24719101123596 149.290074963909
5.34831460674157 148.42042404179
5.44943820224719 141.463216664831
5.55056179775281 132.766707443632
5.65168539325843 112.474852594169
5.75280898876404 108.416481624276
5.85393258426966 106.967063420743
5.95505617977528 105.807528857916
6.0561797752809 88.7043940562258
6.15730337078652 80.2977684757338
6.25842696629213 79.7180011943206
6.35955056179775 77.9786993500809
6.46067415730337 77.6888157093743
6.56179775280899 77.3989320686676
6.66292134831461 74.2102120208948
6.76404494382022 69.5720737695889
6.86516853932584 66.0934700811094
6.96629213483146 62.61486639263
7.06741573033708 62.0350991112168
7.1685393258427 60.295797266977
7.26966292134831 58.2666117820307
7.37078651685393 54.4981244528446
7.47191011235955 47.5409170758857
7.57303370786517 44.9319643095262
7.67415730337079 34.4961532440878
7.7752808988764 33.3366186812613
7.87640449438202 32.4669677591415
7.97752808988764 29.5681313520753
8.07865168539326 26.9591785857157
8.17977528089888 23.7704585379429
8.28089887640449 22.3210403344098
8.38202247191011 17.1031348016906
8.48314606741573 16.813251160984
8.58426966292135 16.2334838795707
8.68539325842697 15.6537165981575
8.78651685393258 15.6537165981575
8.8876404494382 14.7840656760376
8.98876404494382 10.4358110654383
9.08988764044944 10.1459274247317
9.19101123595506 9.85604378402509
9.29213483146067 6.08755645483903
9.39325842696629 4.92802189201255
9.49438202247191 4.34825461059931
9.59550561797753 4.05837096989269
9.69662921348315 3.76848732918606
9.79775280898876 2.8988364070662
9.89887640449438 2.60895276635958
10 1.73930184423972
};
\end{axis}

\end{tikzpicture}

%% file: tikz/box_plot_10_90.tikz
\begin{tikzpicture}

\definecolor{crimson2143940}{RGB}{214,39,40}
\definecolor{darkgray176}{RGB}{176,176,176}
\definecolor{darkorange25512714}{RGB}{255,127,14}
\definecolor{darkturquoise23190207}{RGB}{23,190,207}
\definecolor{forestgreen4416044}{RGB}{44,160,44}
\definecolor{goldenrod18818934}{RGB}{188,189,34}
\definecolor{gray}{RGB}{128,128,128}
\definecolor{gray127}{RGB}{127,127,127}
\definecolor{orchid227119194}{RGB}{227,119,194}
\definecolor{sienna1408675}{RGB}{140,86,75}
\definecolor{steelblue31119180}{RGB}{31,119,180}

\begin{axis}[
log basis y={10},
tick align=outside,
tick pos=left,
width=0.48\textwidth,
height=5.3cm,
x grid style={darkgray176},
xmin=0.5, xmax=9.5,
xtick style={color=black},
xtick={1,2,3,4,5,6,7,8,9},
xticklabel style={rotate=45.0,anchor=east},
xticklabels={Animal,Human,Marking,Object,Rideable,Scene,Sign,Trailer,Vehicle},
y grid style={darkgray176},
ylabel={Object Size [$\text{Pixel}^2$]},
ymajorgrids,
ymin=100, ymax=1800000,
ymode=log,
ytick style={color=black},
yticklabels={
  \(\displaystyle {10^{1}}\),
  \(\displaystyle {10^{2}}\),
  \(\displaystyle {10^{3}}\),
  \(\displaystyle {10^{4}}\),
  \(\displaystyle {10^{5}}\),
}
]
\path [draw=black, fill opacity=0.9,fill=Animal!80!white]
(axis cs:0.75,300)
--(axis cs:1.25,300)
--(axis cs:1.25,3801)
--(axis cs:0.75,3801)
--(axis cs:0.75,300)
--cycle;
\addplot [gray]
table {%
1 304
1 108
};
\addplot [gray]
table {%
1 3828
1 14875
};
\addplot [gray]
table {%
0.875 108
1.125 108
};
\addplot [gray]
table {%
0.875 14875
1.125 14875
};
\path [draw=black, fill opacity=0.8,fill=Human!80!white]
(axis cs:1.75,689.75)
--(axis cs:2.25,689.75)
--(axis cs:2.25,6764.25)
--(axis cs:1.75,6764.25)
--(axis cs:1.75,689.75)
--cycle;
\addplot [gray]
table {%
2 700
2 312
};
\addplot [gray]
table {%
2 6826.75
2 21150
};
\addplot [gray]
table {%
1.875 312
2.125 312
};
\addplot [gray]
table {%
1.875 21150
2.125 21150
};
\path [draw=black, fill opacity=0.8,fill=Marking!80!white]
(axis cs:2.75,560)
--(axis cs:3.25,560)
--(axis cs:3.25,6248)
--(axis cs:2.75,6248)
--(axis cs:2.75,560)
--cycle;
\addplot [gray]
table {%
3 560
3 234
};
\addplot [gray]
table {%
3 6249.5
3 22176
};
\addplot [gray]
table {%
2.875 234
3.125 234
};
\addplot [gray]
table {%
2.875 22176
3.125 22176
};
\path [draw=black, fill opacity=0.8,fill=Object!80!white]
(axis cs:3.75,252)
--(axis cs:4.25,252)
--(axis cs:4.25,2553)
--(axis cs:3.75,2553)
--(axis cs:3.75,252)
--cycle;
\addplot [gray]
table {%
4 252
4 112
};
\addplot [gray]
table {%
4 2553
4 8532
};
\addplot [gray]
table {%
3.875 112
4.125 112
};
\addplot [gray]
table {%
3.875 8532
4.125 8532
};
\path [draw=black, fill opacity=0.8,fill=Rideable!80!white]
(axis cs:4.75,1598.5)
--(axis cs:5.25,1598.5)
--(axis cs:5.25,17020.5)
--(axis cs:4.75,17020.5)
--(axis cs:4.75,1598.5)
--cycle;
\addplot [gray]
table {%
5 1598.75
5 600
};
\addplot [gray]
table {%
5 17031
5 58696
};
\addplot [gray]
table {%
4.875 600
5.125 600
};
\addplot [gray]
table {%
4.875 58696
5.125 58696
};
\path [draw=black, fill opacity=0.8,fill=Scene!80!white]
(axis cs:5.75,8640)
--(axis cs:6.25,8640)
--(axis cs:6.25,334536)
--(axis cs:5.75,334536)
--(axis cs:5.75,8640)
--cycle;
\addplot [gray]
table {%
6 8640
6 2394
};
\addplot [gray]
table {%
6 334494.5
6 1700560
};
\addplot [gray]
table {%
5.875 2394
6.125 2394
};
\addplot [gray]
table {%
5.875 1700560
6.125 1700560
};
\path [draw=black, fill opacity=0.8,fill=Sign!80!white]
(axis cs:6.75,675)
--(axis cs:7.25,675)
--(axis cs:7.25,6364)
--(axis cs:6.75,6364)
--(axis cs:6.75,675)
--cycle;
\addplot [gray]
table {%
7 675
7 288
};
\addplot [gray]
table {%
7 6364.25
7 20306
};
\addplot [gray]
table {%
6.875 288
7.125 288
};
\addplot [gray]
table {%
6.875 20306
7.125 20306
};
\path [draw=black, fill opacity=0.8,fill=Trailer!80!white]
(axis cs:7.75,2470)
--(axis cs:8.25,2470)
--(axis cs:8.25,26761.25)
--(axis cs:7.75,26761.25)
--(axis cs:7.75,2470)
--cycle;
\addplot [gray]
table {%
8 2460
8 960
};
\addplot [gray]
table {%
8 26592.5
8 84360
};
\addplot [gray]
table {%
7.875 960
8.125 960
};
\addplot [gray]
table {%
7.875 84360
8.125 84360
};
\path [draw=black, fill opacity=0.8,fill=Vehicle!80!white]
(axis cs:8.75,4500)
--(axis cs:9.25,4500)
--(axis cs:9.25,53717.5)
--(axis cs:8.75,53717.5)
--(axis cs:8.75,4500)
--cycle;
\addplot [gray]
table {%
9 4510
9 1749
};
\addplot [gray]
table {%
9 53728
9 162855
};
\addplot [gray]
table {%
8.875 1749
9.125 1749
};
\addplot [gray]
table {%
8.875 162855
9.125 162855
};
\addplot [red]
table {%
0.75 1045
1.25 1045
};
\addplot [red]
table {%
1.75 2025
2.25 2025
};
\addplot [red]
table {%
2.75 1776
3.25 1776
};
\addplot [red]
table {%
3.75 731
4.25 731
};
\addplot [red]
table {%
4.75 5056
5.25 5056
};
\addplot [red]
table {%
5.75 48385.5
6.25 48385.5
};
\addplot [red]
table {%
6.75 1927
7.25 1927
};
\addplot [red]
table {%
7.75 7872
8.25 7872
};
\addplot [red]
table {%
8.75 14980
9.25 14980
};
\end{axis}

\end{tikzpicture}

%% file: sec/4_image_retrieval_benchmark.tex
\section{SearchAD Image Retrieval Benchmark}
\label{sec:searchad_image_retrieval_benchmark}
\subsection{Benchmark Overview}
\begin{figure}[t]
    \centering
    \begingroup
    \footnotesize
    \includesvg[width=0.47\textwidth]{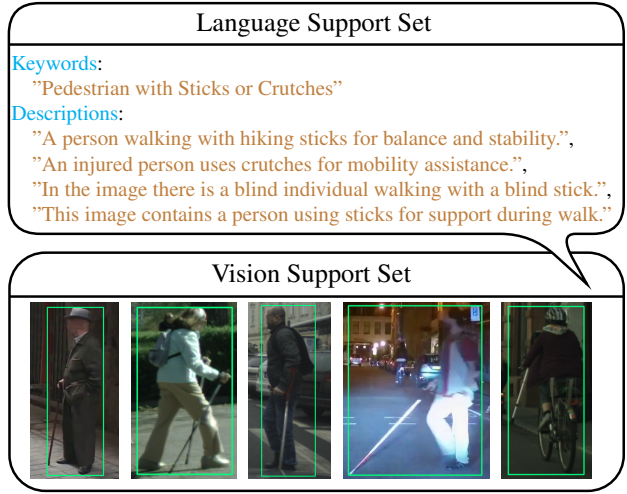}
    \endgroup
    \caption{Illustration of both the language and  support set for the class \emph{Human - With Sticks or Crutches}. Keywords offer a precise description, complemented by descriptions  provide detailed clarifications. The vision support set consists of five images selected to represent different variations of the class. Further query support sets are included in the Supplementary.
    }
    \label{fig:query_example}
    \vspace{-3mm}
\end{figure}

The SearchAD benchmark is specifically designed to evaluate methods for retrieving rare corner case objects and scenes. For each class, a query is designed that allows for both text-based and image-based search. As an example, \Cref{fig:query_example} illustrates both query modalities for the \emph{Human - With Sticks or Crutches} class. The text query consists of precise keywords defining the class-of-interest derived from the labeling guidelines of SearchAD. It also incorporates comprehensive, extended descriptions that offer a more detailed characterization of the class. For image-based retrieval, the query consists of a vision support set of 5 images, carefully chosen from the train set of SearchAD, based on size, variance and low occlusion.

\subsection{Metrics}
Unlike image retrieval benchmarks based on unique image-text pairs \cite{MicrosoftCOCOCommonLin2014, imagedescriptionsvisualYoung2014} or scene-text pairs \cite{BEVTSRTextTang2025}, common retrieval metrics such as Recall@1 are not well-suited to our setting. 
This is because Recall@1 only measures the relevance (whether the image contains the object searched for) of the top-ranked image, meaning it cannot take all relevant images into account. This constraint consequently limits the metric's interpretability and achievable upper bound~\cite{introductioninformationretrievalManning2009}. For example, if the dataset includes 10 images containing a \emph{cat}, querying for \emph{cat} yields a maximum achievable Recall@1 of $\frac{1}{10} =10$\%.
To overcome this limitation, we use R-Precision \cite{introductioninformationretrievalManning2009, OverviewSixthTextVoorhees2000} instead.

\myparagraph{R-Precision}
This metric evaluates the recall within the top $K_c$ results, where $K_c$ is the total number of relevant images for the respective class $c$ (\eg, 10 for the \emph{cat} example). Therefore, it gives the model the chance to retrieve all relevant images and to achieve an optimal R-Precision of 100\%. It is defined as:
\begin{equation}
    \text{RP}_c = \text{Recall@}K_c = \frac{r_c(K_c)}{K_c} \text{,}
\end{equation}
where $r_c(k)$ is the number of relevant images of class $c$ in the top $k$ results~\cite{introductioninformationretrievalManning2009, OverviewSixthTextVoorhees2000}. The Mean R-Precision (MRP) is then calculated as the average RP across all 90 classes~\cite{OverviewSixthTextVoorhees2000}.

\myparagraph{Mean Average Precision} 
In addition to this specific recall metric, we employ Mean Average Precision (MAP) \cite{introductioninformationretrievalManning2009} as the main metric for our image retrieval benchmark, as it assesses how well the model ranks \emph{all} images, rather than being limited to any specific rank: 
\begin{equation}
\begin{split}
\text{AP}_c &= \frac{\sum_{k=1}^{N_\text{Images}} \text{Precision}_c(k) \cdot \text{rel}_c(k)}{K_c} \\ &= \frac{\sum_{k=1}^{N_\text{Images}} r_c(k)/k \cdot \text{rel}_c(k)}{K_c} \text{,}
\end{split}
\end{equation}
where 
$\text{rel}_c(k)$ is a class-dependent indicator function that equals 1 if the image at rank $k$ is relevant and 0 otherwise~\cite{introductioninformationretrievalManning2009}. Moreover, $N_\text{Images}$ denotes the total number of images in the test set and $K_c$ is the overall number of images containing the searched class~$c$. The MAP is then again obtained by averaging AP across all classes.

\subsection{Benchmark Implementation Details}
Many image retrieval methods \cite{BLIP2BootstrappingLi2023, OpenclipIlharco2021, Metaclip2Chuang2025, Siglip2MultilingualTschannen2025} enable the offline pre-calculating of all dataset embeddings, allowing for the development of optimized search indices \cite{FAISSLIBRARYDouze2025} that significantly improve search efficiency and enable practical image retrieval applications while largely preserving accuracy (see Supplementary). However, to ensure a fair benchmark, free from any side-effects or model-specific benefits arising from an optimized search index, we opt for a naive search approach in our benchmark evaluation. Moreover, some methods can only process lower resolution images \cite{OpenclipIlharco2021, PayAttentionYourHajimiri2025, BLIP2BootstrappingLi2023}. To ensure these approaches can fairly recognize all objects, we exclude images from evaluation that contain relevant objects with an area smaller than 50 pixels. As the specific images to be excluded vary for each of the 90 classes, we ignore them selectively depending on the class being queried, rather than excluding all images from the entire benchmark.

%% file: sec/5_experiments.tex
\begin{figure}[b]
    \centering
    \begingroup
    \scriptsize
    \includesvg[width=0.475\textwidth]{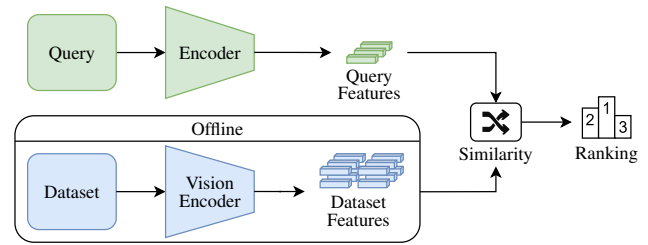}
    \endgroup
    \caption{Image retrieval framework for both VLM-based text-to-image and image-based retrieval baselines. The query is encoded and compared to all offline-calculated dataset features using cosine similarity. Text and vision queries are provided as part of the image retrieval benchmark.}
    \label{fig:method_overview}
    \vspace{-3mm}
\end{figure} 
\section{Baseline Experiments} 
\label{sec:baseline_experiments}
\subsection{Retrieval Framework} 
The baseline methods of our image retrieval benchmark are primarily VLMs pre-trained on large-scale image-text pairs. For all methods with an aligned image-text feature space, and for all image-based approaches, we use the basic framework shown in \Cref{fig:method_overview}.
 Starting from the query, we either use the vision or language support set to encode the query features. To further simplify the similarity search, and since initial investigations revealed no major disadvantages, we average the query features, for example, those from the five reference images of the vision support set. As common practice, we employ cosine similarity~\cite{CLIPBasedImagePatil2024, CORAAdaptingCLIPWu2023, EvaluationsimilaritymeasurementZhang2003} for the ranking of all images based on the query features. More precisely, the cosine distance to the query features is calculated against either a summary feature or individual patch features, depending on the underlying model and its pre-training details. Each dataset image is finally ranked based on its maximum similarity score.
\begin{figure}[b]
    \centering
    \begingroup
    \footnotesize
    \includesvg[width=0.475\textwidth]{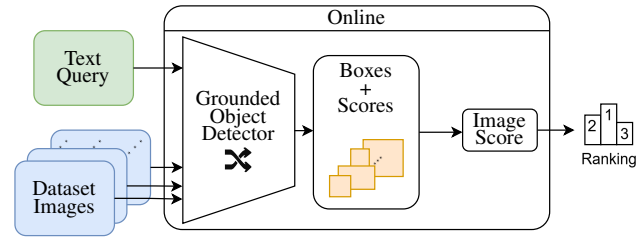}
    \endgroup
    \caption{Image retrieval framework for text-to-image retrieval using grounded object detectors. For each image, bounding boxes and scores are predicted based on the text queries that are provided as part of the image retrieval benchmark. The image score is derived from the maximum bounding box score.}
    \label{fig:gdino_overview}
    \vspace{-3mm}
\end{figure}

To evaluate the effectiveness of grounded object detectors \cite{GroundedlanguageimageLi2022, GroundingDINOMarryingLiu2025} for image retrieval, we use the framework shown in \Cref{fig:gdino_overview}. Using the main keyword from the language support set as a condition, the grounded object detector predicts bounding boxes for each dataset image. The similarity score for the image ranking is determined by the maximum score among all predicted bounding boxes with the correct class label.
Unlike VLM-based approaches that allow for offline calculation of dataset embeddings, grounded object detectors such as GLIP~\cite{GroundedlanguageimageLi2022} and GroundingDINO~\cite{GroundingDINOMarryingLiu2025} require online combination and refinement of image and text features for accurate bounding box detection. 
Consequently, only the intermediate features from the vision backbone can be pre-computed and stored offline, resulting in a significant search time disadvantage compared to text-to-image retrieval with VLMs.
\subsection{Zero-Shot Baseline Methods}
Building upon the baseline frameworks described in the previous section, we adapt state-of-the-art VLMs for both zero-shot image-to-image and text-to-image retrieval. 
\Cref{tab:model_information} summarizes input resolution, feature dimensions, and the vision backbone for all models. To enable a fair comparison across different models, we utilize the ViT-L architecture for all Vision Transformers~\cite{ImageisWorthDosovitskiy2020}, where available. Given the high-resolution images in SearchAD, we always opt for the model versions supporting the highest input resolution. For RADIO~\cite{AMRADIOAgglomerativeRanzinger2024}, we use RADIOv2.5~\cite{RADIOv2.5ImprovedBaselinesHeinrich2025} with a maximum image resolution of $1024 \times 1024$ pixels. 
Given that most VLMs exclusively support square image formats and to align with the pre-training data, we process all images accordingly. \looseness=-1
\begin{table}[b]
    \setlength{\tabcolsep}{3pt} 
    \small
    \centering
    \begin{tabular}{lrrr}
        \toprule
        Model & Input Res. & Feature Dim. & Vis. Enc. \\
        \midrule
        GroundingDINO \cite{GroundingDINOMarryingLiu2025} & [1333, 600] & [882, 1024] & Swin-B \\
        NACLIP \cite{PayAttentionYourHajimiri2025} & [336, 336] & [576, 768] & ViT-L \\
        OpenCLIP \cite{OpenclipIlharco2021} & [336, 336] & [1, 1024] & ViT-L \\
        RADIO \cite{RADIOv2.5ImprovedBaselinesHeinrich2025} & [1024, 1024] & [1, 1024] & ViT-L \\
        BLIP2 \cite{BLIP2BootstrappingLi2023} & [224, 224] & [32, 768] & ViT-L \\
        SigLIP2 \cite{Siglip2MultilingualTschannen2025} & [512, 512] & [1, 1024] & ViT-L \\
        MetaCLIP2 \cite{Metaclip2Chuang2025} & [378, 378] & [1, 1024] & ViT-H \\
        NARADIO \cite{RayFrontsOpenSetAlama2025, PayAttentionYourHajimiri2025} & [1024, 1024] & [4096, 1152] & ViT-L \\
        \bottomrule
    \end{tabular}
    \caption{Overview of baseline models for zero-shot image-to-image and text-to-image retrieval. Main differences are the feature dimensionality and the maximum input resolution.}
    \label{tab:model_information}
    \vspace{-3mm}
\end{table}

\begin{table*}[ht] 
    \sisetup{
      table-align-uncertainty=true,
      separate-uncertainty=true,
    }
    \renewrobustcmd{\bfseries}{\fontseries{b}\selectfont}

    \setlength{\tabcolsep}{4pt} 
    \small
    \centering
    \begin{tabular}{l l S[table-format=2.2, detect-weight=true, mode=text] S[table-format=2.2, detect-weight=true, mode=text] *{9}{S[table-format=2.2, detect-weight=true, mode=text]}}
        \toprule
        & {\multirow{3}{*}{Model}} & {\multirow{3}{*}{\centering MAP [\%]}} & {\multirow{3}{*}{\centering MRP [\%]}} & \multicolumn{9}{c}{Category-wise MAP [\%]} \\
        \cmidrule(lr){5-13} 
        \multicolumn{2}{l}{} & & & {\centering Animal} & {\centering Human} & {\centering Marking} & {\centering Object} & {\centering Rideable} & {\centering Scene} & {\centering Sign} & {\centering Trailer} & {\centering Vehicle} \\
        \midrule
        \multirow{8}{*}{\rotatebox[origin=c]{90}{\textbf{Text-to-Image}}} & GroundingDINO \cite{GroundingDINOMarryingLiu2025} & 5.25 & 6.49 & 11.87 & 0.59 & 1.45 & 9.49 & 5.99 & 1.76 & 0.67 & 3.14 & 3.11 \\
        & OpenCLIP \cite{OpenclipIlharco2021} & 7.45 & 10.17 & 4.96 & 7.38 & 3.70 & 5.12 & 8.45 & 23.24 & 1.28 & 3.84 & 7.70 \\
        & SigLIP2 \cite{Siglip2MultilingualTschannen2025} & 8.57 & 11.24 & 5.10 & 7.16 & 3.74 & 6.55 & 8.95 & 20.64 & 3.50 & 5.32 & 11.56 \\
        & BLIP2 \cite{BLIP2BootstrappingLi2023} & 9.14 & 11.90 & 8.71 & 9.00 & 6.12 & 5.71 & 8.23 & 24.73 & 1.49 & 4.13 & 10.83 \\
        & MetaCLIP2 \cite{Metaclip2Chuang2025} & 9.41 & 12.66 & 8.72 & 8.96 & 3.91 & 7.51 & 11.15 & 19.23 & 2.71 & 4.49 & 11.43 \\
        & RADIO \cite{RADIOv2.5ImprovedBaselinesHeinrich2025} & 9.49 & 11.86 & 9.03 & \bfseries 11.77 & 3.25 & 7.18 & 11.44 & 22.34 & 1.61 & 3.33 & 10.23 \\
        & NACLIP \cite{PayAttentionYourHajimiri2025} & 9.59 & 11.92 & 10.10 & 5.21 & 7.02 & 7.34 & 7.32 & \bfseries 25.95 & 1.39 & 4.23 & 11.84 \\
        & NARADIO \cite{RayFrontsOpenSetAlama2025, PayAttentionYourHajimiri2025} & \bfseries 14.27 & \bfseries 17.91 & \bfseries 14.60 & 9.74 & \bfseries 13.23 & \bfseries 14.11 & \bfseries 15.14 & 15.37 & \bfseries 4.12 & \bfseries 16.02 & \bfseries 17.20 \\
        \midrule 
        \multirow{8}{*}{\rotatebox[origin=c]{90}{\textbf{Image-to-Image}}} & OpenCLIP \cite{OpenclipIlharco2021} & 3.98 & 5.55 & 5.60 & 1.26 & 2.12 & 3.22 & 1.82 & 12.66 & 0.73 & 2.96 & 3.80 \\
        & RADIO \cite{RADIOv2.5ImprovedBaselinesHeinrich2025} & 4.34 & 6.00 & 5.60 & 0.95 & 1.48 & 3.30 & 3.24 & 10.46 & 1.15 & 2.50 & 5.88 \\
        & MetaCLIP2 \cite{Metaclip2Chuang2025} & 5.10 & 6.64 & 5.72 & 1.61 & 1.93 & 3.22 & 7.06 & 9.70 & 0.58 & 4.93 & 6.47 \\
        & SigLIP2 \cite{Siglip2MultilingualTschannen2025} & 6.04 & 7.97 & 5.76 & 2.18 & 1.97 & 4.95 & 5.94 & 10.67 & 1.93 & 5.78 & 8.85 \\
        & NACLIP \cite{PayAttentionYourHajimiri2025} & 6.56 & 9.30 & \bfseries 12.78 & 5.95 & 3.54 & 6.75 & 5.57 & 2.48 & 1.32 & 5.63 & 7.21 \\
        & GroundingDINO \cite{GroundingDINOMarryingLiu2025} & 7.62 & 10.45 & 7.73 & 2.78 & \bfseries 5.97 & \bfseries 10.15 & 4.92 & 12.32 & \bfseries 2.35 & 7.14 & 8.91 \\
        & BLIP2 \cite{BLIP2BootstrappingLi2023} & 7.95 & \bfseries 10.82 & 7.44 & \bfseries 8.92 & 5.60 & 6.63 & \bfseries 6.85 & \bfseries 18.29 & 1.20 & 5.83 & 8.44 \\
        & NARADIO \cite{RayFrontsOpenSetAlama2025, PayAttentionYourHajimiri2025} & \bfseries 8.31 & 10.61 & 6.10 & 3.80 & 2.68 & 7.56 & 5.20 & 10.41 & 1.80 & \bfseries 13.61 & \bfseries 14.25 \\
        \bottomrule
    \end{tabular}
    \caption{Evaluation of retrieval methods on the SearchAD test set. The table presents Mean R-Precision (MRP) and Mean Average Precision (MAP) over all classes, and category-wise MAP, for both text-to-image and image-to-image retrieval methods. Bold values highlight the top performance for each metric and modality. Class-wise evaluation and MAP for all 90 classes are in Supplementary.}
    \label{tab:average_map_values_categories_text_and_image}
    \vspace{-4mm}
\end{table*}
\subsection{Text-to-Image Retrieval}
The upper part of \Cref{tab:average_map_values_categories_text_and_image} shows the results for text-based retrieval methods, including a breakdown of MAP per category (cf. \Cref{tab:searchad_categories}). Except for the \textit{scene} category, NARADIO~\cite{RayFrontsOpenSetAlama2025, PayAttentionYourHajimiri2025}, the neighbor-aware (NA) version of RADIO, outperforms all other baselines with an average MAP of 14.27\% (best in 49 out of 90 classes and 7 out of 9 categories). 
Originally designed to enable open-vocabulary semantic segmentation with VLMs, 
NARADIO comes with two advantages:
First, the lightweight language adapters can be adjusted to transform the spatial features to the text-aligned space. 
Second, the gaussian kernel introduced in \cite{PayAttentionYourHajimiri2025} and adapted for RADIO in \cite{RayFrontsOpenSetAlama2025} enforces that the exact spatial localization of the vision encoder patches is preserved, which is crucial for dense and object-level prediction tasks. While NACLIP~\cite{PayAttentionYourHajimiri2025} extended CLIP~\cite{LearningTransferableVisualRadford2021a} with this neighbor-awareness, NARADIO comes with the additional advantage that it can handle high-resolution input images, 
which is extremely beneficial for small instances, such as objects, animals and signs. 
In contrast to NACLIP and NARADIO, the vision heads of MetaCLIP2~\cite{Metaclip2Chuang2025} and SigLIP2~\cite{Siglip2MultilingualTschannen2025} are optimized to process pooled features of the vision backbone, rather than patch features. Consequently, incorporating their spatial features proved ineffective. \\
Interestingly, GroundingDINO~\cite{GroundingDINOMarryingLiu2025} performs worse than the VLM-based methods, except for the \emph{Object} and \emph{Animal} categories. This demonstrates that the high input resolution can be beneficial, \eg leading to the best results for the classes \emph{Hydrant} and \emph{Stroller}. However, long descriptions lead to false positive detections with high similarity scores for parts of the descriptions. Therefore, using only the first and most precise keyword led to the best results. Details can be found in the Supplementary Material.
\begin{figure}[b]
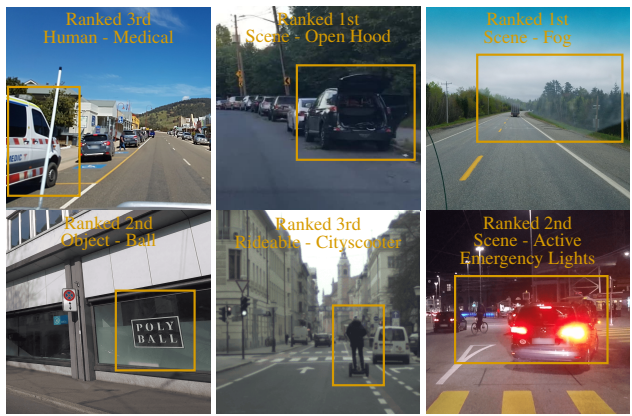

    \begingroup
    \small
    \centering
    
    \begin{subfigure}[b]{0.155\textwidth}
        \centering
        \scriptsize
        \includesvg[width=\textwidth]{images/human_medical_des.svg}
        \label{fig:bild1}
    \end{subfigure}
    \begin{subfigure}[b]{0.155\textwidth}
        \centering
        \scriptsize
        \includesvg[width=\textwidth]{images/scene_open_hood_des.svg}
        \label{fig:bild2}
    \end{subfigure}
    \begin{subfigure}[b]{0.155\textwidth}
        \centering
        \scriptsize
        \includesvg[width=\textwidth]{images/scene_fog_des.svg}
        \label{fig:bild3}
    \end{subfigure}
    \\
    \vspace{-0.8\baselineskip}    
    \begin{subfigure}[b]{0.155\textwidth}
        \centering
        \scriptsize
        \includesvg[width=\textwidth]{images/object_ball_des.svg}
        \caption{Wrong category.}
        \label{fig:bild4}
    \end{subfigure}
    \begin{subfigure}[b]{0.155\textwidth}
        \centering
        \scriptsize
        \includesvg[width=\textwidth]{images/rideable_cityscooter_des.svg}
        \caption{Wrong class.}
        \label{fig:bild5}
    \end{subfigure}
    \begin{subfigure}[b]{0.155\textwidth}
        \centering
        \scriptsize
        \includesvg[width=\textwidth]{images/scene_active_emergency_lights_des.svg}
        \caption{Global context gap.}
        \label{fig:bild6}
    \end{subfigure}
    \caption{Qualitative examples of three common failure cases found in the top 3 ranked results of NARADIO \cite{RayFrontsOpenSetAlama2025, PayAttentionYourHajimiri2025}.}
    \label{fig:failure_cases_naradio}
    \endgroup
    \vspace{-3mm}
\end{figure}

\subsection{Image-to-Image Retrieval}
In the lower part of \Cref{tab:average_map_values_categories_text_and_image}, the results for the vision-based approaches are shown. For each baseline, we used the feature matching strategy that yielded best results. For SigLIP2~\cite{SigmoidLossLanguageZhai2023} and MetaCLIP2~\cite{Metaclip2Chuang2025}, we calculate the similarity between the query and the non-spatial dataset features. Unlike BLIP2 and NARADIO, where patch features can be effectively compared to the existing summary feature, GroundingDINO and NACLIP achieve the best results by creating a summary feature from the spatial features. For each query image, a gaussian kernel is used to reduce the spatial features to a single feature vector, prioritizing the more representative central patches as described in \cite{NeuralCodesImageBabenko2014}. \\ 
While GroundingDINO with its Swin Transformer~\cite{SwinTransformerHierarchicalLiu2021} achieves the highest R-Precision, NARADIO attains the best MAP at 8.31\%. In contrast to text-based methods, no single method achieves the best results across all categories.

\subsection{Fine-tuning Baseline}
Leveraging SearchAD training annotations and textual descriptions from the language support sets, we fine-tuned BLIP2~\cite{BLIP2BootstrappingLi2023} using standard image-text matching and contrastive losses. Without hyperparameter optimization or box-/region-level supervision, image-level fine-tuning already demonstrated significant potential. As shown in \Cref{tab:finetuning_baseline}, this approach achieved an MAP of 12.06\% for text-to-image retrieval, establishing a first fine-tuning baseline for future research. The model's Top-25 results for the most biased class \textit{Caravan Trailer} (comprising 76\% of training images from Mapillary-Sign~\cite{MapillaryTrafficSignErtler2020}) included samples from 6 of 11 datasets, underscoring its robust generalization across diverse domain styles. Remarkably, fine-tuning without explicitly incorporating the vision support set also enhances image-to-image retrieval to an MAP of 11.13\%.
\begin{table}[t]
    \sisetup{
      table-align-uncertainty=true,
      separate-uncertainty=true,
    }
    \renewrobustcmd{\bfseries}{\fontseries{b}\selectfont}
    \setlength{\tabcolsep}{4pt} 
    \small
    \centering
    \begin{tabular}{l S[detect-weight=true, mode=text] S[detect-weight=true, mode=text] S[detect-weight=true, mode=text] S[detect-weight=true, mode=text]}
        \toprule
        & \multicolumn{2}{c}{Text-to-Image} & \multicolumn{2}{c}{Image-to-Image} \\
        \cmidrule(lr){2-3} \cmidrule(lr){4-5}
        BLIP2~\cite{BLIP2BootstrappingLi2023} & {MAP [\%]} & {MRP [\%]} & {MAP [\%]} & {MRP [\%]} \\
        \midrule
        Baseline & 9.14 & 11.90 & 7.95 & 10.82 \\
        Fine-tuned & \bfseries 12.06 & \bfseries 15.60 & \bfseries 11.13 & \bfseries 13.83 \\
        \bottomrule
    \end{tabular}
    \vspace{-2mm}
    \caption{Fine-tuning baseline based on BLIP2~\cite{BLIP2BootstrappingLi2023}, trained on the SearchAD train set and evaluated on the test set.}
    \label{tab:finetuning_baseline}
    \vspace{-2mm}
\end{table}

\subsection{Failure Mode Analysis}
Among the zero-shot methods, NARADIO achieved best performance. However, with an MAP of 14.27\%, significant room for improvement remains. As illustrated in \Cref{fig:failure_cases_naradio}, the top 3 search results are negatively impacted by the inclusion of irrelevant images, demonstrating a clear lack of semantic precision. Failure cases include: (a) retrieval of correct semantic class, but within the wrong object category (\eg, \emph{medical vehicles} retrieved when querying for \emph{medical staff}), (b) difficulty in precisely identifying subclasses (\eg, when querying for \emph{open hoods}, vehicles with both \emph{open trunk} and \emph{door} exhibiting a higher similarity than those with an \emph{open hood}), and (c) a trade-off between local and global information, where local image features may be misinterpreted (\eg, \emph{dirty windshields} as \emph{fog}).\looseness=-1

%% file: sec/6_conclusion.tex
\section{Conclusion}
In this paper, we introduced SearchAD, a novel large-scale image retrieval dataset for rare objects and scenes. The dataset comprises over 423k frames and 90 manually annotated rare classes, with 30 appearing fewer than 250 times in the training and validation splits. We proposed a novel semantic image retrieval benchmark for autonomous driving that evaluates a wide range of currently available baselines for zero-shot text-to-image and image-to-image retrieval, thereby revealing the extreme difficulties inherent in this task. Even the most advanced method, NARADIO~\cite{RayFrontsOpenSetAlama2025, PayAttentionYourHajimiri2025} (the neighbor-aware variant of RADIO~\cite{RADIOv2.5ImprovedBaselinesHeinrich2025}), achieves a mean average precision of only 14.27\%. Its persistent failure cases and incomplete semantic understanding clearly indicate substantial room for future improvement.

Our detailed analysis revealed that text-based methods, particularly those leveraging spatial features aligned with text, exhibited the best results among the evaluated baselines. 
Moreover, the ability to process high-resolution images proved effective in identifying small object categories. 

Future research could explore fine-tuning VLMs using the bounding box annotations provided by SearchAD. Additional promising directions include developing few-shot or multi-modal approaches and extending the benchmark to rare object retrieval by incorporating the localization task. Moreover, combining the rare annotations of SearchAD with existing common classes could establish new long-tail and open-world perception benchmarks.

\myparagraph{Acknowledgments} This work is a result of the joint research project STADT:up (19A22006O). The project is supported by the German Federal Ministry for Economic Affairs and Energy (BMWE), based on a decision of the German Bundestag. The author is solely responsible for the content of this publication.



%% file: sec/X_suppl.tex
\clearpage
\setcounter{page}{1}
\maketitlesupplementary

\renewcommand{\thesubsection}{\Alph{subsection}} 


\subsection{SearchAD Dataset}
\begin{figure*}[t]
    \centering
    \includegraphics[width=\textwidth]{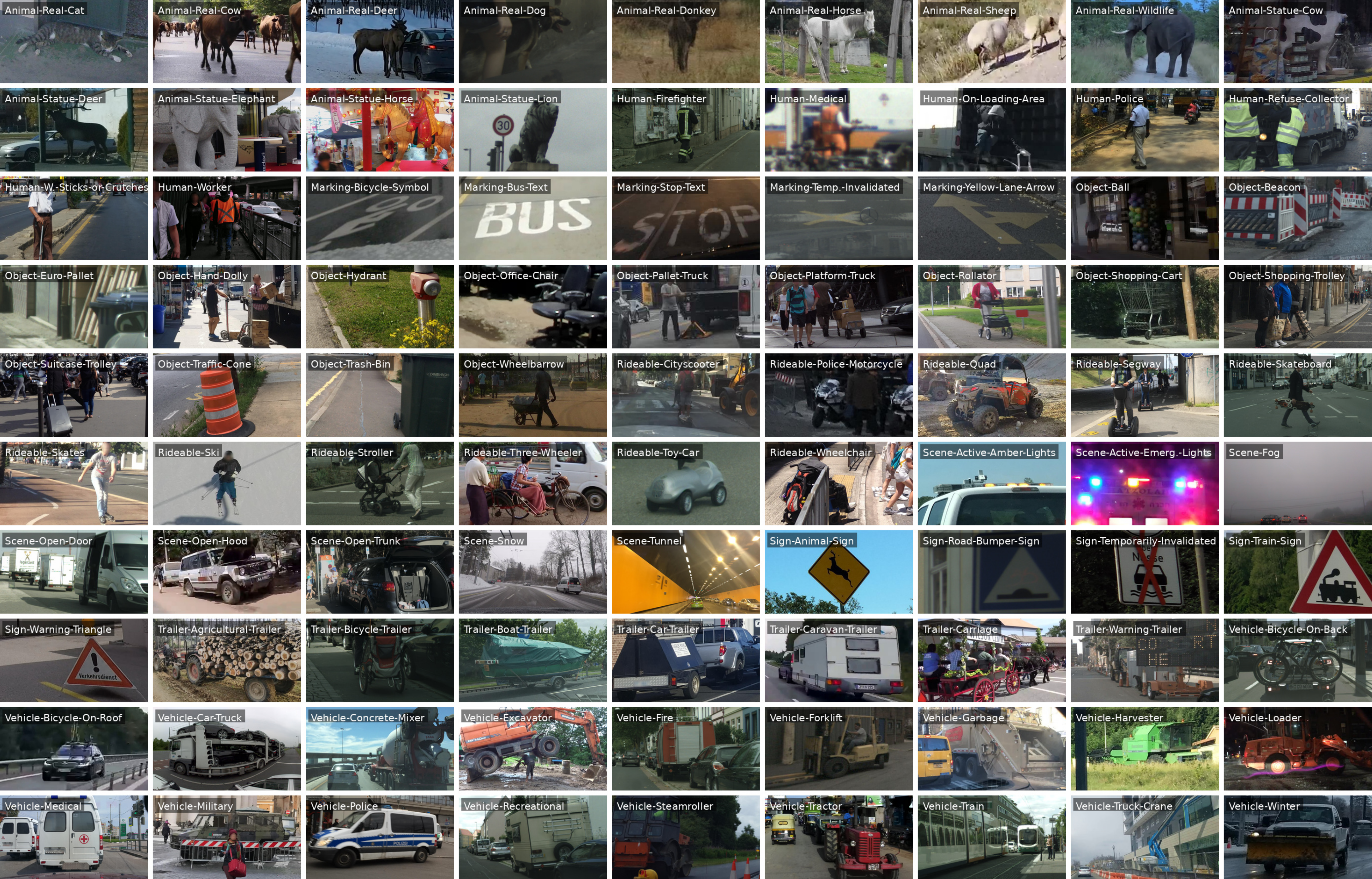}
    \caption{Overview of the 90 classes of SearchAD. Cropped instances of objects and scenes are provided for each class, extracted from the SearchAD training split. References for all underlying image sources are provided in \Cref{tab:searchad_dataset_collection}.}
    \label{fig:examples_for_all_classes}
\end{figure*}

In this section, we provide more details on the SearchAD dataset (cf. \Cref{sec:searchad_dataset}) and its labeling project. In \Cref{fig:examples_for_all_classes}, we showcase a cropped instance for each of the 90 semantic classes of \Cref{tab:searchad_categories}, with all objects and scenes originating from the training split. Similar crops are also used for the vision support set of the search queries.

In Table \ref{tab:labeling_guidelines}, we illustrate the labeling guidelines for each of the nine categories using one representative class as an example. The complete labeling guidelines will be provided on the project page to maintain the scope of this supplementary material. Each entry provides a detailed label description, along with positive and negative examples. To clearly delineate relevant from irrelevant instances, the label description provides specific examples. A ball on a poster, for instance, is considered a false positive for the \emph{Ball} class. Likewise, for \emph{Fog}, strong rainy or snowy scenes are included as false positives, as their poor visibility results from conditions other than fog.

Furthermore, we maintained quality through weekly checks and by addressing label supplier questions, ensuring continuous adherence to our requirements. For challenging samples, we provided support to the label supplier upon request to determine whether the object or scene should be labeled. Moreover, this feedback also served as training material for the human experts.

Beyond the 90 official classes, SearchAD also provides catch-all classes for several categories, namely: \emph{Animal-Real-Other}, \emph{Animal-Statue-Other}, \emph{Human-Duty-Other}, \emph{Object-Movable-Other} and \emph{Vehicle-Construction-Other}.

\subsection{SearchAD Image Retrieval Benchmark}
This section complements \Cref{sec:searchad_image_retrieval_benchmark} by providing further details and more examples of the search queries, alongside a comparison of the test and validation splits.

Five more search queries, including their vision and language support sets, are illustrated in \Cref{tab:more_search_query_examples}. This structure is standard for all remaining search queries within the SearchAD image retrieval benchmark, forming its default vision and language support sets. Specifically, language support sets contain up to three keywords and a total of four descriptions. The vision support set invariably consists of five images. Although some models might handle certain formulations better, we opted not to optimize the language support sets based on a specific subset of models. Instead, we defined them as precisely as possible, in alignment with the labeling guidelines. All 90 search queries, including their language and vision support sets are part of the benchmark and will be provided via the SearchAD devkit.

In \Cref{tab:test_and_validation_split_comparison}, we compare the results for all baseline methods on the validation split with those on the test split. Overall, the results are reasonably balanced between the two:  7 out of 16 methods show slightly better MAP performance on the test data, and 9 out of 16 methods exhibit slightly better MRP performance on this split. While text-based methods show a slight advantage on the test data, image-based methods exhibit a minor bias towards better performance on the validation split.
\begin{table}[t] 
    \sisetup{
      table-align-uncertainty=true,
      separate-uncertainty=true,
    }
    \renewrobustcmd{\bfseries}{\fontseries{b}\selectfont}

    \setlength{\tabcolsep}{1.5pt} 
    \small
    \centering
    \begin{tabular}{l l S[table-format=2.2, detect-weight=true, mode=text] S[table-format=2.2, detect-weight=true, mode=text] S[table-format=2.2, detect-weight=true, mode=text] S[table-format=2.2, detect-weight=true, mode=text]}
        \toprule
        & {\multirow{2}{*}{Model}} & \multicolumn{2}{c}{\centering Test Split} & \multicolumn{2}{c}{\centering Val Split} \\
        \cmidrule(lr){3-4} \cmidrule(lr){5-6} 
        & & {\centering MAP [\%]} & {\centering MRP [\%]} & {\centering MAP [\%]} & {\centering MRP [\%]} \\
        \midrule
        \multirow{8}{*}{\rotatebox[origin=c]{90}{\textbf{Text-to-Image}}} & GDINO \cite{GroundingDINOMarryingLiu2025} & \bfseries 5.25 & \bfseries 6.49 & 5.14 & 6.28 \\
        & OpenCLIP \cite{OpenclipIlharco2021} & 7.45 & \bfseries 10.17 & \bfseries 7.90 & 9.82 \\
        & SigLIP2 \cite{Siglip2MultilingualTschannen2025} & \bfseries 8.57 & \bfseries 11.24 & 8.22 & 10.03 \\
        & BLIP2 \cite{BLIP2BootstrappingLi2023} & 9.14 & \bfseries 11.90 & \bfseries 10.14 & 11.77 \\
        & MetaCLIP2 \cite{Metaclip2Chuang2025} & \bfseries 9.41 & \bfseries 12.66 & 8.74 & 11.23 \\
        & RADIO \cite{RADIOv2.5ImprovedBaselinesHeinrich2025} & 9.49 & 11.86 & \bfseries 10.44 & \bfseries 12.18 \\
        & NACLIP \cite{PayAttentionYourHajimiri2025} & \bfseries 9.59 & 11.92 & 9.42 & \bfseries 12.28 \\
        & NARADIO \cite{RayFrontsOpenSetAlama2025, PayAttentionYourHajimiri2025} & \bfseries 14.27 & \bfseries 17.91 & 13.22 & 15.60 \\
        \midrule 
        \multirow{8}{*}{\rotatebox[origin=c]{90}{\textbf{Image-to-Image}}} & OpenCLIP \cite{OpenclipIlharco2021} & 3.98 & 5.55 & \bfseries 4.16 & \bfseries 5.75 \\
        & RADIO \cite{RADIOv2.5ImprovedBaselinesHeinrich2025} & 4.34 & 6.00 & \bfseries 4.64 & \bfseries 6.09 \\
        & MetaCLIP2 \cite{Metaclip2Chuang2025} & \bfseries 5.10 & \bfseries 6.64 & 4.49 & 5.64 \\
        & SigLIP2 \cite{Siglip2MultilingualTschannen2025} & \bfseries 6.04 & \bfseries 7.97 & 5.42 & 7.39 \\
        & NACLIP \cite{PayAttentionYourHajimiri2025} & 6.56 & 9.30 & \bfseries 8.07 & \bfseries 10.19 \\
        & GDINO \cite{GroundingDINOMarryingLiu2025} & 7.62 & \bfseries 10.45 & \bfseries 8.41 & 10.30 \\
        & BLIP2 \cite{BLIP2BootstrappingLi2023} & 7.95 & 10.82 & \bfseries 9.20 & \bfseries 11.19 \\
        & NARADIO \cite{RayFrontsOpenSetAlama2025, PayAttentionYourHajimiri2025} & 8.31 & 10.61 & \bfseries 10.04 & \bfseries 12.10 \\
        \bottomrule
    \end{tabular}
    \caption{Evaluation of retrieval methods on the SearchAD test and validation split. The table presents Mean R-Precision (MRP) and Mean Average Precision (MAP) for both splits, and for both text-to-image and image-to-image retrieval methods. Bold values highlight the top performance for each metric (MAP and MRP) within each row (model) across both splits. The results show that the test and validation splits are relatively well balanced.}
    \label{tab:test_and_validation_split_comparison}
    \vspace{-4mm}
\end{table}

\myparagraph{Text-to-Image Retrieval}
Since the MAP for RADIO~\cite{RADIOv2.5ImprovedBaselinesHeinrich2025}, BLIP2~\cite{BLIP2BootstrappingLi2023}, MetaCLIP2~\cite{Metaclip2Chuang2025} and NACLIP~\cite{PayAttentionYourHajimiri2025} all span from 9.14\% up to 9.59\%, their ranking varies in the validation split. While RADIO, MetaCLIP2, and BLIP2 successfully retrieved the image, including a group of firefighters, at the very top, achieving an AP of 100\% for \emph{Firefighter}, other models encountered difficulties with this challenging object. Notably, NACLIP~\cite{PayAttentionYourHajimiri2025} only achieved an AP of 3.33\% for \emph{Firefighter}, which allowed BLIP2 to even surpass NACLIP on the validation set. To better validate the somewhat underrepresented classes, the training split can be leveraged for zero-shot methods.

\myparagraph{Image-to-Image Retrieval}
For the image-based methods, the ranking on the validation split is almost the same as the ranking on the test split. Nevertheless, MetaCLIP2 shows reduced performance on the validation set, primarily driven by a significant gap in AP for the two rideable vehicles \emph{Police Motorcycle} and \emph{Toy Car} between the validation and test set. 

\subsection{Detailed Baseline Results}
The following section presents detailed class-wise baseline results, qualitative examples, and additional insights into the challenges GroundingDINO~\cite{GroundingDINOMarryingLiu2025} encounters in text-to-image retrieval.

\begin{figure*}[ht!]
    \begingroup
    \small
    \centering
    
    \begin{subfigure}[b]{0.3\textwidth}
        \centering
        \scriptsize
        \includegraphics[width=\textwidth]{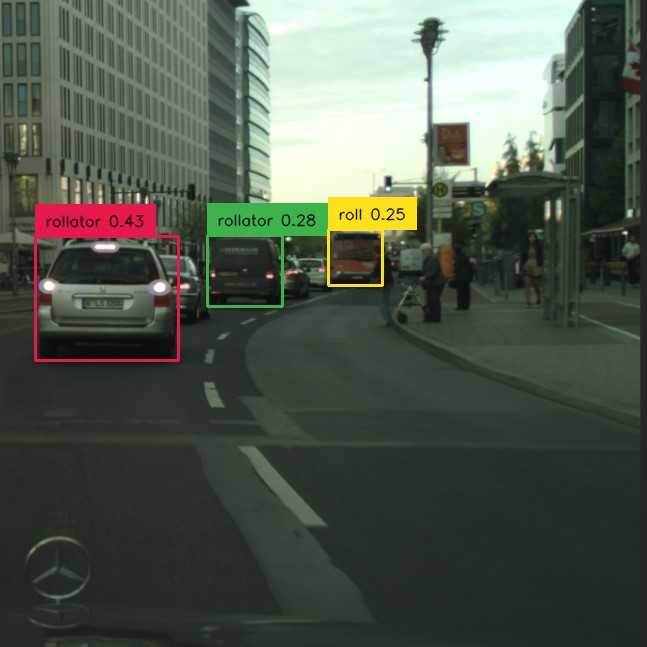}
        \caption{First Keyword: "Rollator"}
        \label{fig:gdino_first_keyword}
    \end{subfigure}
    \begin{subfigure}[b]{0.3\textwidth}
        \centering
        \scriptsize
        \includegraphics[width=\textwidth]{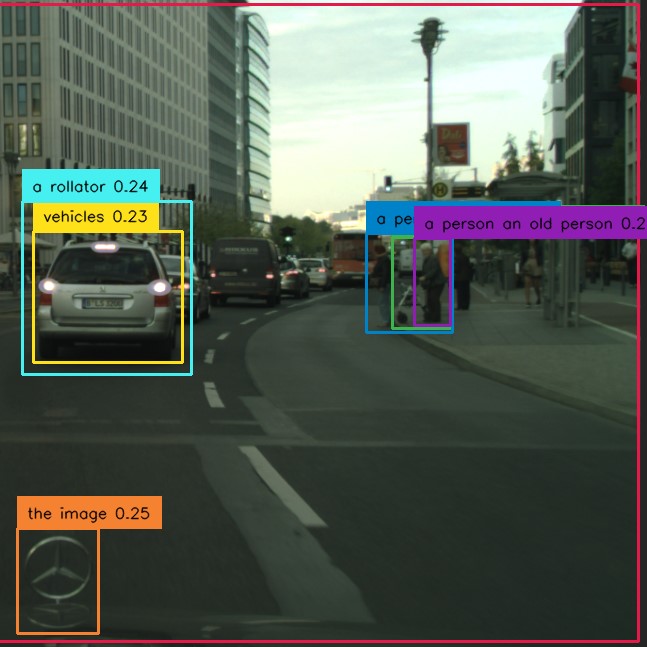}
        \caption{Full Descriptions, cf. \Cref{tab:more_search_query_examples}}
        \label{fig:gdino_descriptions}
    \end{subfigure}
    \begin{subfigure}[b]{0.3\textwidth}
        \centering
        \scriptsize
        \includegraphics[width=\textwidth]{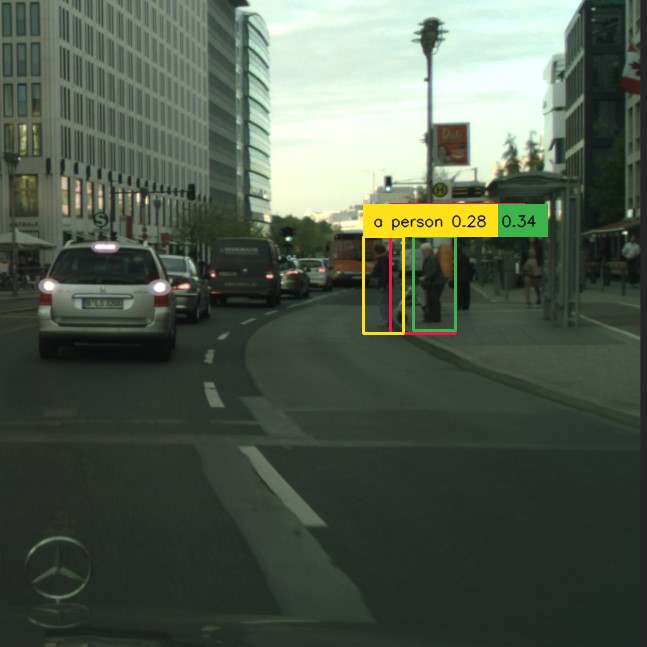}
        \caption{"A person walking with their rollator."}
        \label{fig:gdino_first_description}
    \end{subfigure}
    \caption{Visualization of the conflicting choice of a suitable text prompt for GroundingDINO \cite{GroundingDINOMarryingLiu2025}. The bounding box predictions for the class \emph{Object-Rollator} are shown for different text inputs. Both box threshold and text threshold are set to 0.23 for better visualization. The search query of \emph{Object-Rollator} along with its language support set can be found in \Cref{tab:more_search_query_examples}.}
    \label{fig:groundingdino_visu}
    \endgroup
\end{figure*}
\subsubsection{Class-wise Image Retrieval Results}
\myparagraph{Text-to-Image Retrieval}
\Cref{tab:map_class_wise_text_to_image_retrieval} shows the class-wise Average Precisions (APs) for text-based image retrieval across all baseline methods. The category-wise Mean Average Precisions (MAPs) can be derived by taking the mean of the respective class-wise AP scores, \eg, for NARADIO~\cite{RayFrontsOpenSetAlama2025, PayAttentionYourHajimiri2025}:

\begin{equation*}
\begin{split}
        \text{MAP}_{\text{\emph{Marking}}} 
        &= \frac{31.45\!+\!16.12\!+\!9.72\!+\!0.99\!+\!7.88}{5} \\ 
        &= 13.23\%
\end{split}
\end{equation*}
The overall MAP is shown in the last row of the table and is calculated by directly averaging the AP scores across all classes.
\Cref{tab:map_class_wise_text_to_image_retrieval} demonstrates that NARADIO~\cite{RayFrontsOpenSetAlama2025, PayAttentionYourHajimiri2025} outperforms the other methods for 49 out of the 90 classes. However, the results also demonstrate that each model achieves the best performance for at least one class, underscoring the unique strengths and weaknesses of individual models and their pre-training. This highlights the importance of employing a diverse set of semantic classes for a comprehensive evaluation of zero-shot image retrieval capabilities. 

Furthermore, the table reveals that more frequent classes, such as \emph{Traffic Cone}, \emph{Hydrant} or \emph{Tunnel}, are generally easier to retrieve than rare classes, highlighting the key challenge of our proposed benchmark. However, not only rare, but also certain specific search queries, such as the scene classes \emph{Open Door} and \emph{Open Trunk}, also prove challenging. Even with well-described language support (\eg, \emph{Scene-Open-Trunk} in \Cref{tab:more_search_query_examples}), models struggle to accurately interpret and recognize their precise semantic meaning. Finally, models lacking text-aligned spatial features face significant difficulties with small objects. This is evident in the APs for the \emph{Animal}, \emph{Marking} and \emph{Object} categories, where spatial and high-resolution (cf. \Cref{tab:model_information}) methods NARADIO and GroundingDINO~\cite{GroundingDINOMarryingLiu2025} consistently outperform other methods.

\myparagraph{Image-to-Image Retrieval}
In \Cref{tab:map_class_wise_image_to_image_retrieval}, the class-wise results for the image-based methods are shown. In general, text-to-image retrieval methods demonstrate superior performance compared to image-to-image retrieval. Image-to-image retrieval outperforms or matches the best text-to-image method for only 29 out of 90 classes. This superior performance does not appear to follow a clear categorical pattern, instead occurring somewhat randomly across various classes. Notably, for the two classes \emph{Train} and \emph{Police Motorcycle}, the best performing image-to-image method achieved an AP more than 10\% higher than the best performing text-to-image retrieval method.

For image-to-image retrieval, NARADIO again emerges as the overall best performing model and leads in many individual classes. However, unlike in text-to-image retrieval, other methods such as BLIP2~\cite{BLIP2BootstrappingLi2023} and GroundingDINO show comparable strong performance. For example, GroundingDINO achieves the best results for 8 out of the 15 \emph{Object} classes. Conversely, the image-based OpenCLIP~\cite{OpenclipIlharco2021} method do not achieve the best performance for any of the 90 classes.

\subsubsection{Qualitative Image Retrieval Results}
Based on the retrieval results on the validation split of SearchAD, we showcase the top 5 ranked images of all eight text-to-image retrieval methods.  The corresponding search queries, along with their language support sets, can be found in \Cref{tab:more_search_query_examples}. Notably, some retrieved images might appear quite similar, as certain datasets include temporally correlated recordings (e.g., video frames rather than individual keyframes).

\Cref{fig:vehicle_tractor_results} illustrates the top retrieval results when searching for \emph{Tractors}. The qualitative results confirm the MAP scores shown in \Cref{tab:map_class_wise_text_to_image_retrieval}, demonstrating that models with text-aligned spatial features generally achieve superior results. Notably, GroundingDINO~\cite{GroundingDINOMarryingLiu2025} struggles to retrieve the correct vehicles in this context. OpenCLIP~\cite{OpenclipIlharco2021} incorrectly ranks a semantically similar \emph{harvester} as second, an image that should ideally be retrieved when searching for the very rare \emph{Vehicle-Harvester} class. While all models exhibit at least one false positive image within their top 5 results, NARADIO~\cite{RayFrontsOpenSetAlama2025, PayAttentionYourHajimiri2025} is the only method to achieve a perfect precision of 100\% within its top 5 results (Precision@5 \cite{introductioninformationretrievalManning2009}).

However, \Cref{fig:animal_cow_results} and \Cref{fig:marking_bus_text_results} highlight a persistent challenge: all models struggle to retrieve precisely described semantic classes. The results highlight a common failure mode where models correctly identify the semantic class but also retrieve images from incorrect categories (cf. \Cref{fig:failure_cases_naradio}). Specifically, all eight methods in \Cref{fig:animal_cow_results} incorrectly retrieve at least one image featuring a traffic sign symbolizing a cow. For \emph{Marking-Bus-Text} in \Cref{fig:marking_bus_text_results}, all models except for NARDIO and NACLIP~\cite{PayAttentionYourHajimiri2025} tend to retrieve generic road markings right next to the ego vehicle instead of the specific 'BUS' text. Addressing this issue by better conditioning specific categories remains an area for future research.

The results in \Cref{fig:scene_open_trunk_results} highlight another limitation: Models struggle to completely comprehend the semantic descriptions for specific search queries. For instance, when searching for scenes like \emph{Open Trunk}, the presence of a large trunk seems to be prioritized over the critical detail of it being open. While most models struggle to retrieve any correct images within their top 5 results, NARADIO once again achieves a Precision@5 of 100\%.


\begin{figure}[t]
    \hspace{0.1cm}
    \small
    \begin{subfigure}[b]{0.1\textwidth} 
        \centering
        \input{tikz/confusion_matrix.tikz}
        \label{fig:matrix_a_single_colorbar}
    \end{subfigure}
    \hspace{1.8cm}
    \begin{subfigure}[b]{0.1\textwidth} 
        \centering
        \input{tikz/confusion_matrix2.tikz}
        \label{fig:matrix_b_single_colorbar}
    \end{subfigure}
    \vspace{-2em}
    \caption{Retrieval confusion matrices based on the Top-50 results of NARADIO. Queries: (i) Animal-Real-Cow, Animal-Real-Horse, Sign-Animal-Sign and (ii) Scene-Open-Door, Scene-Open-Hood, Scene-Open-Trunk.}
    \label{fig:confusion_matrices}
    \vspace{-3mm}
\end{figure}
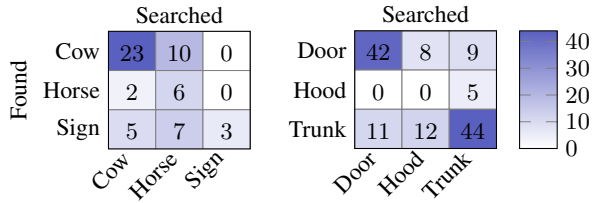

\subsubsection{Text Prompt Sensitivity in GroundingDINO}
We chose GroundingDINO~\cite{GroundingDINOMarryingLiu2025} as one of our baseline methods due to its excellent ability to recognize open-world objects, which represent some of the most challenging retrieval instances in our benchmark. As detailed in \Cref{tab:map_class_wise_text_to_image_retrieval}, GroundingDINO also demonstrates the best performance across all models mainly for object and animal classes. GroundingDINO excels particularly with classic and simple object and animal classes, such as \emph{Dog}, \emph{Horse}, \emph{Hydrant}, or \emph{Stroller}. 

However, its performance significantly degrades for other classes. This decline is primarily attributed to its sensitivity to text prompt choices, as exemplified by the \emph{Object-Rollator} class in \Cref{fig:groundingdino_visu}. While GroundingDINO generally achieves its best results with precise and short descriptions, issues arise with less optimal prompts. For instance, when using only the first keyword, as shown in \Cref{fig:gdino_first_keyword}, "Rollator" is often unknown to the model, resulting in a similarity score below the threshold of 0.23 for the object-of-interest. Meanwhile, the car in front is detected with a high similarity score of 0.43, leading to numerous false positives in the top retrieval results.

Conversely, employing a comprehensive description can also introduce false positives. For example, the Mercedes-Benz star might be erroneously detected with the label "the image" (similarity score of 0.25), and the car in front of the ego vehicle is again misidentified as a rollator. For the \emph{Rollator} class, the most effective prompt was found to be "A person walking with their rollator." which directs the model's attention more appropriately towards pedestrians.

In contrast, many other VLM-based text-to-image retrieval methods show significantly greater robustness to prompt engineering as they achieve a similar overall MAP regardless of whether using only keywords or both keywords and descriptions. This robustness, combined with the ability to pre-calculate and store image features in optimized search indices like FAISS \cite{FAISSLIBRARYDouze2025}, makes VLM-based methods considerably more suitable for real-time search and practical applications demanding rapid results.

\begin{table}[t] 
    \sisetup{
      table-align-uncertainty=true,
      separate-uncertainty=true, 
    }
    \renewrobustcmd{\bfseries}{\fontseries{b}\selectfont}
    \setlength{\tabcolsep}{1.5pt} 
    \small
    \centering
    \begin{tabular}{c l @{} S[detect-weight=true, mode=text] @{} S[detect-weight=true, mode=text]} 
        \toprule
        & Model - BLIP2~\cite{BLIP2BootstrappingLi2023} & MAP [\%] & MRP [\%] \\
        \midrule
        \multirow{4}{*}{\rotatebox[origin=c]{90}{Language}}
        & BLIP2 - Baseline & 9.14 & 11.90 \\
        & BLIP2 - Max. Query & 6.20 & 8.06 \\
        & BLIP2 - Eval Small & 1.45 & 1.62 \\
        & BLIP2 - Eval Large & 13.51 & 15.75 \\
        \hline
        \multirow{4}{*}{\rotatebox[origin=c]{90}{Vision}} & BLIP2 - Baseline & 7.95 & 10.82 \\
        & BLIP2 - Random Sets & 5.36(0.23) & 7.20(0.22) \\
        & BLIP2 - Max. Query & 5.73 & 7.88 \\
        & BLIP2 - RPN & 12.46 & 16.15 \\
        \bottomrule
    \end{tabular}
    \vspace{-2mm}
    \caption{Ablation studies based on SearchAD test set.}
    \label{tab:ablation_studies}
    \vspace{-2mm}
\end{table}

\subsubsection{Quantitative Failure Mode Analysis}
We extend our NARADIO~\cite{RayFrontsOpenSetAlama2025, PayAttentionYourHajimiri2025} results and quantify failure modes via Top-50 confusion matrices (\Cref{fig:confusion_matrices}). The analysis reveals semantic ambiguity (e.g., \textit{Cow} queries retrieve \textit{Animal Signs} more frequently than explicit sign queries do) and visual similarity (e.g., \textit{Open Trunk} retrieves \textit{Open Hoods}). Derived from the confusion matrices, the highest $\text{Precision}@50$ achieved is $88\%$ for \textit{Open Trunk}.

\subsection{Ablation Studies based on BLIP2}
\subsubsection{Ablations of Vision and Language Support Sets}
Queries were optimized for superior validation accuracy across all baselines, without model-specific support set optimization. Leveraging insights from labeling QCs, vision support images were manually selected for high variance, representativeness, and low occlusion. We validated our curation against 20 random selections from the training set (cf. \Cref{tab:ablation_studies} - Baseline vs. Random Sets). Mean queries outperform maximal similarity across single queries (cf. \Cref{tab:ablation_studies} - Baseline vs. Max. Query), highlighting their robustness against confusing correlations.

\subsubsection{Class-Agnostic Region Proposal Preprocessing}
While our main paper intentionally established plain baselines, we validated incorporating Faster R-CNN's RPN~\cite{FasterRCNNRen2017} as a pre-selection stage for plausible region proposals. As a second step, each region will be processed separately by BLIP2~\cite{BLIP2BootstrappingLi2023}. Unlike refined detectors, which suppress unknown classes (e.g., road markings) as background, we opt for raw RPN proposals to maintain high recall for rare objects. Augmenting these proposals with the entire image (crucial for scenes like \emph{Fog}) yields a substantial MAP boost from 7.95\% to 12.46\% (cf. \Cref{tab:ablation_studies}). This confirms that region-based processing is effective, though more an engineering solution.
\subsubsection{Size-Dependent Evaluation}
We isolated object size as a critical performance bottleneck: As show in \Cref{tab:ablation_studies} MAP collapses from $13.51\%$ (largest $33\%$ of objects) to $1.45\%$ (smallest $33\%$), quantitatively confirming the challenge of small object retrieval.
\subsubsection{Optimized Search Index}
\Cref{tab:faiss_index_searchtime} compares both retrieval accuracy and runtime for VLM-based methods based on BLIP2~\cite{BLIP2BootstrappingLi2023} with and without index and grounded object detectors. The results are based on the entire SearchAD test set. While the MAP only reduces by 0.4\%, both search time and setup time are significantly reduced by the index. When scaling the dataset to millions of images, the index will be inevitable to enable AD developers fast search across AD databases. Moreover, the search time of more than 20 hours for GroundingDINO~\cite{GroundingDINOMarryingLiu2025} demonstrates that grounded object detectors are generally not suitable for real-world applications.
\begin{table}[t] 
    \sisetup{
      table-align-uncertainty=true,
      separate-uncertainty=true,
    }
    \renewrobustcmd{\bfseries}{\fontseries{b}\selectfont}

    \setlength{\tabcolsep}{3pt} 
    \small
    \centering
    \begin{tabular}{l c S[table-format=3.2, detect-weight=true, mode=text] S[table-format=2.2, detect-weight=true, mode=text] S[table-format=2.2, detect-weight=true, mode=text]}
        \toprule
        Search Time Optimization & $t_\text{Search}$ & $t_\text{Setup}$ &  {\centering MAP [\%]} \\
        \midrule
        BLIP2 - No Search Index~\cite{BLIP2BootstrappingLi2023} & 5.55s & 179.29s & \bfseries 9.14 \\
        BLIP2 - Faiss Index~\cite{FAISSLIBRARYDouze2025} & \bfseries 0.28s & \bfseries 28.23s & 8.74 \\
        GroundingDINO~\cite{GroundingDINOMarryingLiu2025} & 21\,-\,23h & \textbf{{-}} & 5.25 \\
        \bottomrule
    \end{tabular}
    \vspace{-2mm}
    \caption{MAP and search time for VLM-based methods (with and without index) and grounded object detectors on SearchAD Test.}
    \label{tab:faiss_index_searchtime}
\vspace{-4mm}
\end{table}

{
\onecolumn
\small
\centering

\begin{longtable}{>{\centering}p{0.6cm} p{6.0cm} p{8.0cm}}
    \toprule
    \textbf{Class} & \textbf{Language Support Set} & \textbf{Vision Support Set} \\
    \midrule
    \endfirsthead

    \multicolumn{3}{c}{\tablename\ \thetable\ -- Continued from previous page} \\
    \toprule
    \textbf{Class} & \textbf{Language Support Set} & \textbf{Vision Support Set} \\
    \midrule
    \endhead

    \midrule
    \multicolumn{3}{r}{\tablename\ \thetable\ -- \textit{Continued on next page}} \\
    \endfoot

    \midrule
    \multicolumn{3}{r}{Continued on next page} \\
    \endfoot

    \endlastfoot

    \vspace{0cm}
    \rotatebox{90}{\textbf{Animal-Real-Cow}} &
    \vspace{0cm}
    \textcolor{cyan}{Keywords}:
    \begin{itemize}
        \item[] \textcolor{brown}{"Cow"},
        \item[] \textcolor{brown}{"Cattle"},
        \item[] \textcolor{brown}{"Bovine"},
    \end{itemize}
    \textcolor{cyan}{Descriptions}:
    \begin{itemize}
        \item[] \textcolor{brown}{"A cow next to the street or standing on the street."},
        \item[] \textcolor{brown}{"The cow is standing in front of the vehicle or below the vehicle."},
        \item[] \textcolor{brown}{"There is a cow in the driving scene on the road or on the sidewalk."},
        \item[] \textcolor{brown}{"There are cows next to the road on a pasture."}
    \end{itemize}
    &
    {
    \vspace{0cm}
    \begin{tabular}{ccc}
        \includegraphics[height=4cm]{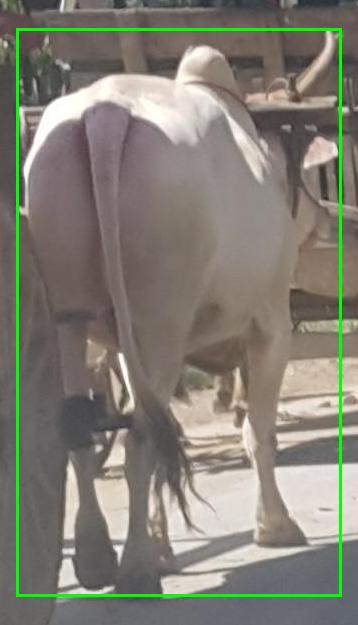} &
        \includegraphics[height=4cm]{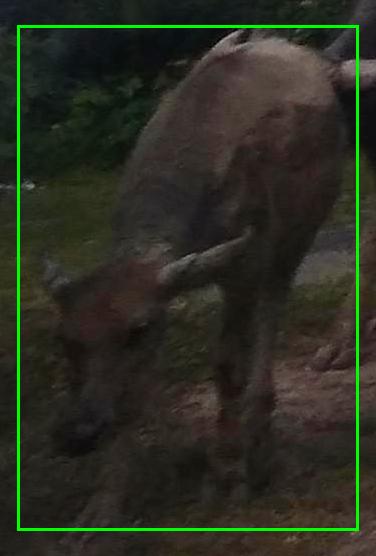} &
        \includegraphics[height=4cm]{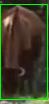} \\
        \includegraphics[height=2.4cm]{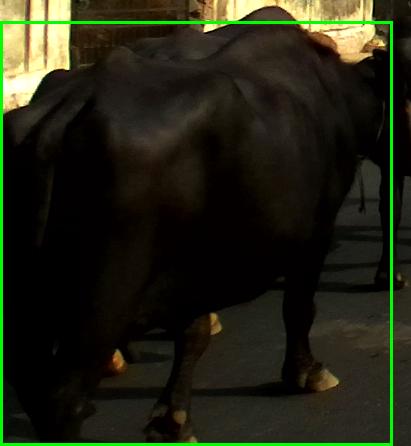} &
        \includegraphics[height=2.4cm]{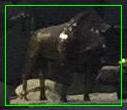} \\
    \end{tabular}
    } \\
    \hline
    
    \vspace{0cm}
    \rotatebox{90}{\textbf{Marking-Bus-Text}} &
    \vspace{0cm}
    \textcolor{cyan}{Keywords}:
    \begin{itemize}
        \item[] \textcolor{brown}{"Bus Road Marking"},
        \item[] \textcolor{brown}{"Bus Text Pattern"},
    \end{itemize}
    \textcolor{cyan}{Descriptions}:
    \begin{itemize}
        \item[] \textcolor{brown}{"The word bus is painted on the street in white color."},
        \item[] \textcolor{brown}{"The word bus is painted on the street in yellow color."},
        \item[] \textcolor{brown}{"The bus text pattern on the road indicates the bus lane."},
        \item[] \textcolor{brown}{"The image contains a bus road marking that may indicate a bus stop."}
    \end{itemize}
    &
    {
    \vspace{0cm}
    \begin{tabular}{cc}
        \includegraphics[width=0.2\textwidth]{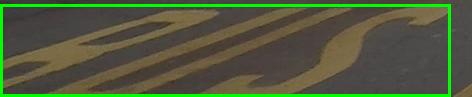} &
        \includegraphics[width=0.2\textwidth]{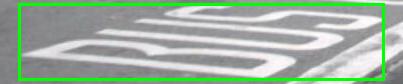} \\
        \includegraphics[width=0.2\textwidth]{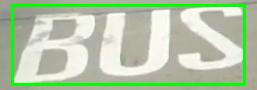} &
        \includegraphics[width=0.2\textwidth]{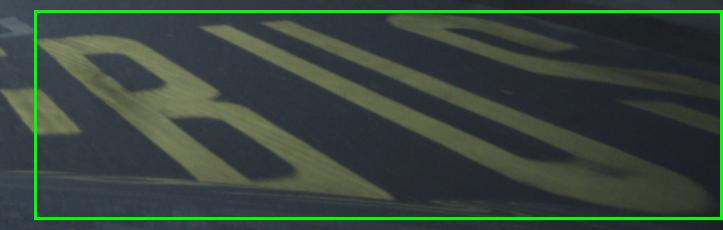} \\
        \includegraphics[width=0.2\textwidth]{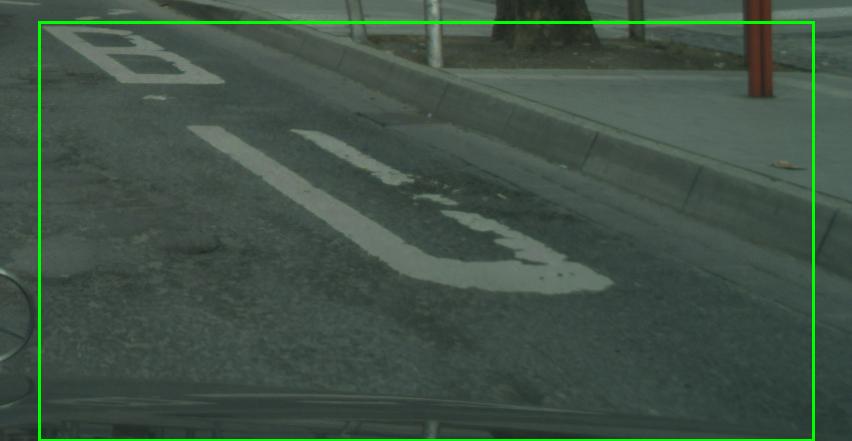} & \\
    \end{tabular}
    } \\
    \hline

    \vspace{0cm}
    \rotatebox{90}{\textbf{Object-Rollator}} &
    \vspace{0cm}
    \textcolor{cyan}{Keywords}:
    \begin{itemize}
        \item[] \textcolor{brown}{"Rollator"},
        \item[] \textcolor{brown}{"Walking Aid"},
        \item[] \textcolor{brown}{"Mobility Walker"}
    \end{itemize}
    \textcolor{cyan}{Descriptions}:
    \begin{itemize}
        \item[] \textcolor{brown}{"A person walking with their rollator."},
        \item[] \textcolor{brown}{"A rollator is visible in the image, requiring vehicles to be aware of potential pedestrians with mobility issues."},
        \item[] \textcolor{brown}{"The image shows a walking aid, indicating a need for increased awareness of vulnerable road users."},
        \item[] \textcolor{brown}{"There is an old person using a rollator for mobility assistance."}
    \end{itemize}
    &
    {
    \vspace{0cm}
    \begin{tabular}{ccc}
        \includegraphics[width=0.11\textwidth]{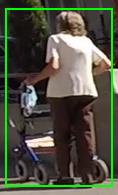} &
        \includegraphics[width=0.112\textwidth]{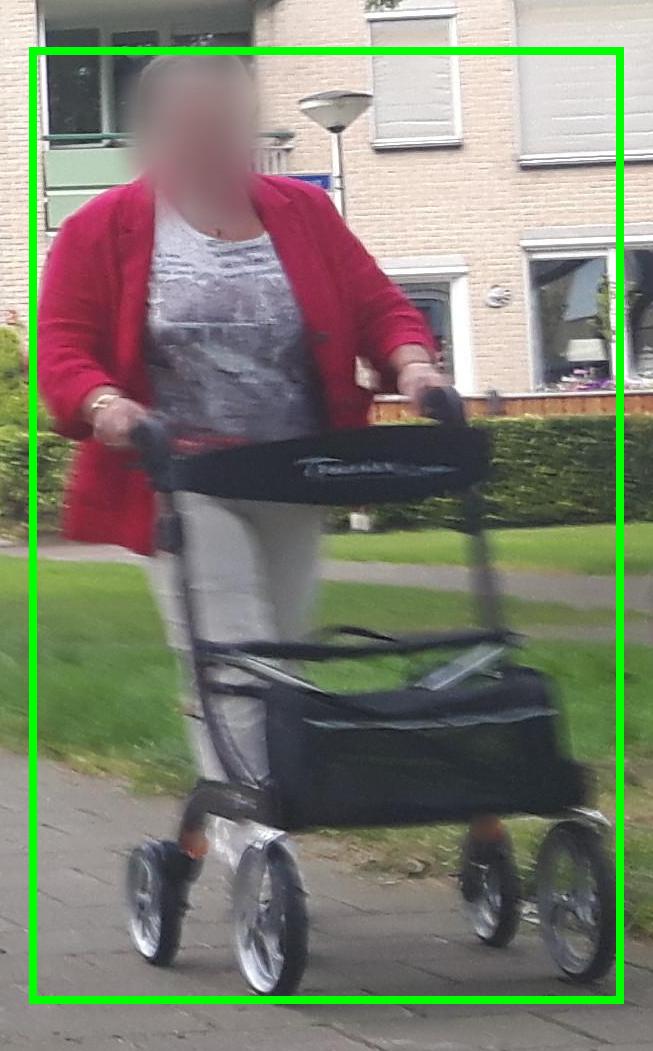} &
        \includegraphics[width=0.095\textwidth]{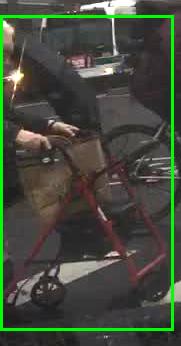} \\
        \includegraphics[width=0.11\textwidth]{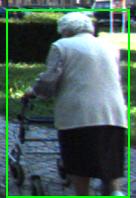} &
        \includegraphics[width=0.13\textwidth]{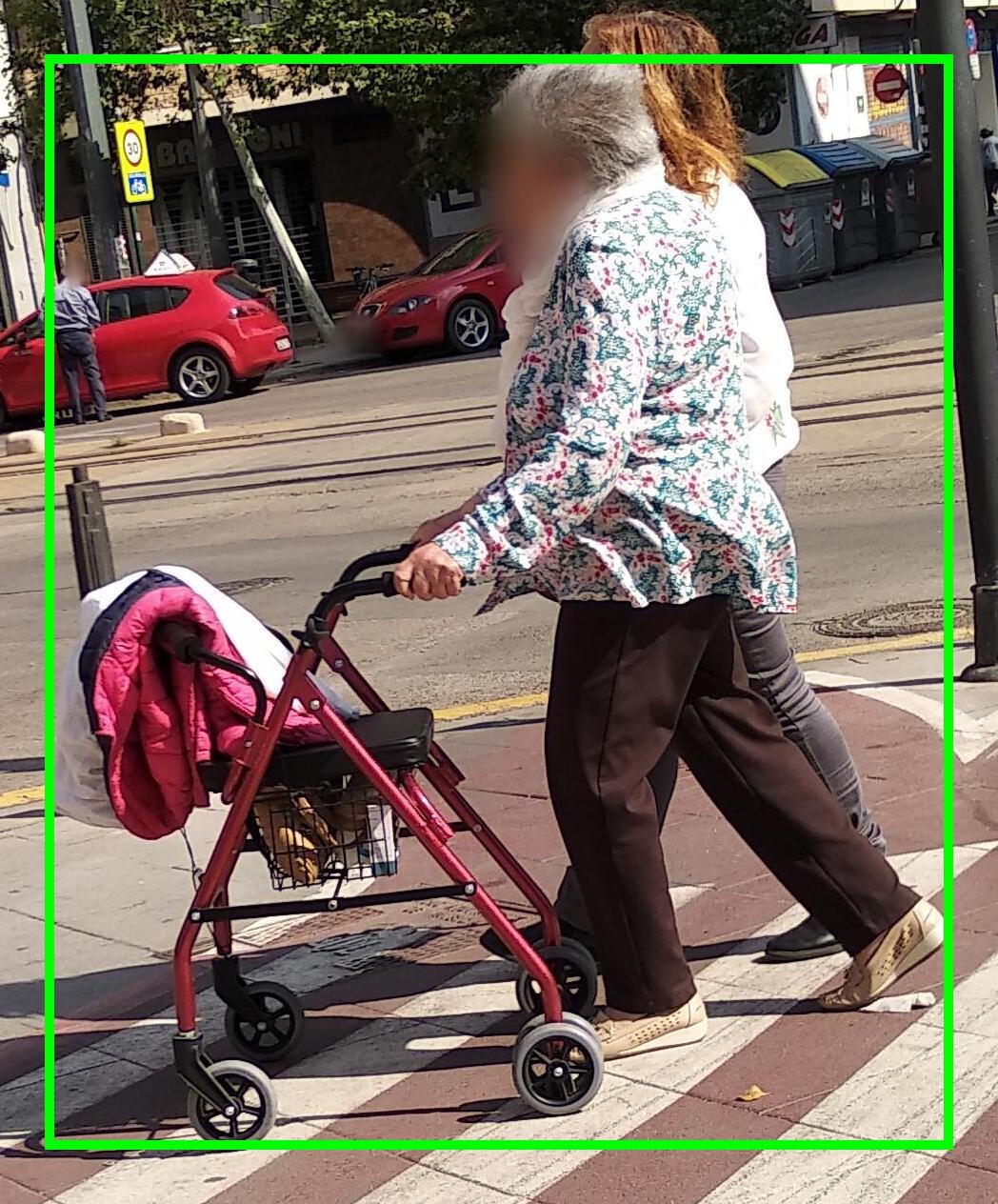} \\
    \end{tabular}
    } \\
    
    \vspace{0cm}
    \rotatebox{90}{\textbf{Scene-Open-Trunk}} &
    \vspace{0cm}
    \textcolor{cyan}{Keywords}:
    \begin{itemize}
        \item[] \textcolor{brown}{"Open Car Trunk"},
        \item[] \textcolor{brown}{"Vehicle with open trunk."},
        \item[] \textcolor{brown}{"Van or truck with open trunk doors."},
    \end{itemize}
    \textcolor{cyan}{Descriptions}:
    \begin{itemize}
        \item[] \textcolor{brown}{"The image shows a passenger car, van or truck with an open trunk."},
        \item[] \textcolor{brown}{"The van has its trunk doors to the back of the vehicle open."},
        \item[] \textcolor{brown}{"The image contains a vehicle with an open trunk."},
        \item[] \textcolor{brown}{"The open trunk of the vehicle allows to look into the trunk or the loading area of the truck."}
    \end{itemize}
    &
    {
    \vspace{0cm}
    \begin{tabular}{cc}
        \includegraphics[width=0.2\textwidth]{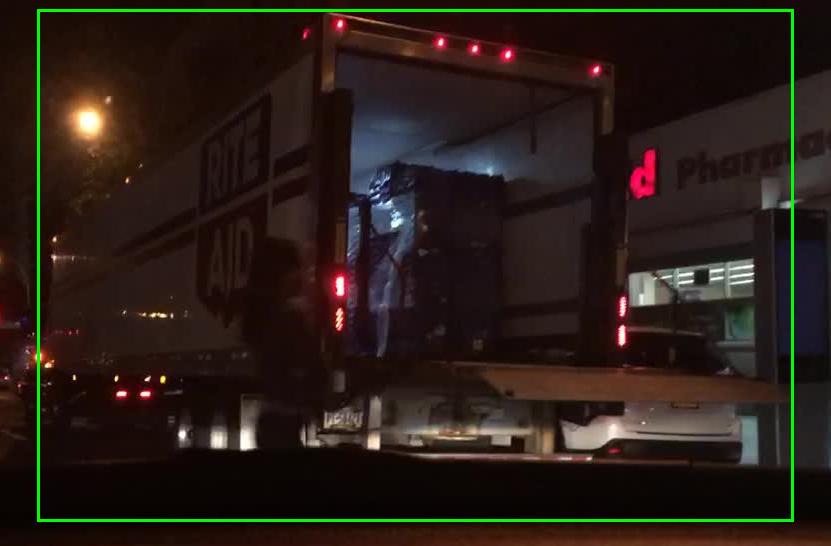} &
        \includegraphics[width=0.2\textwidth]{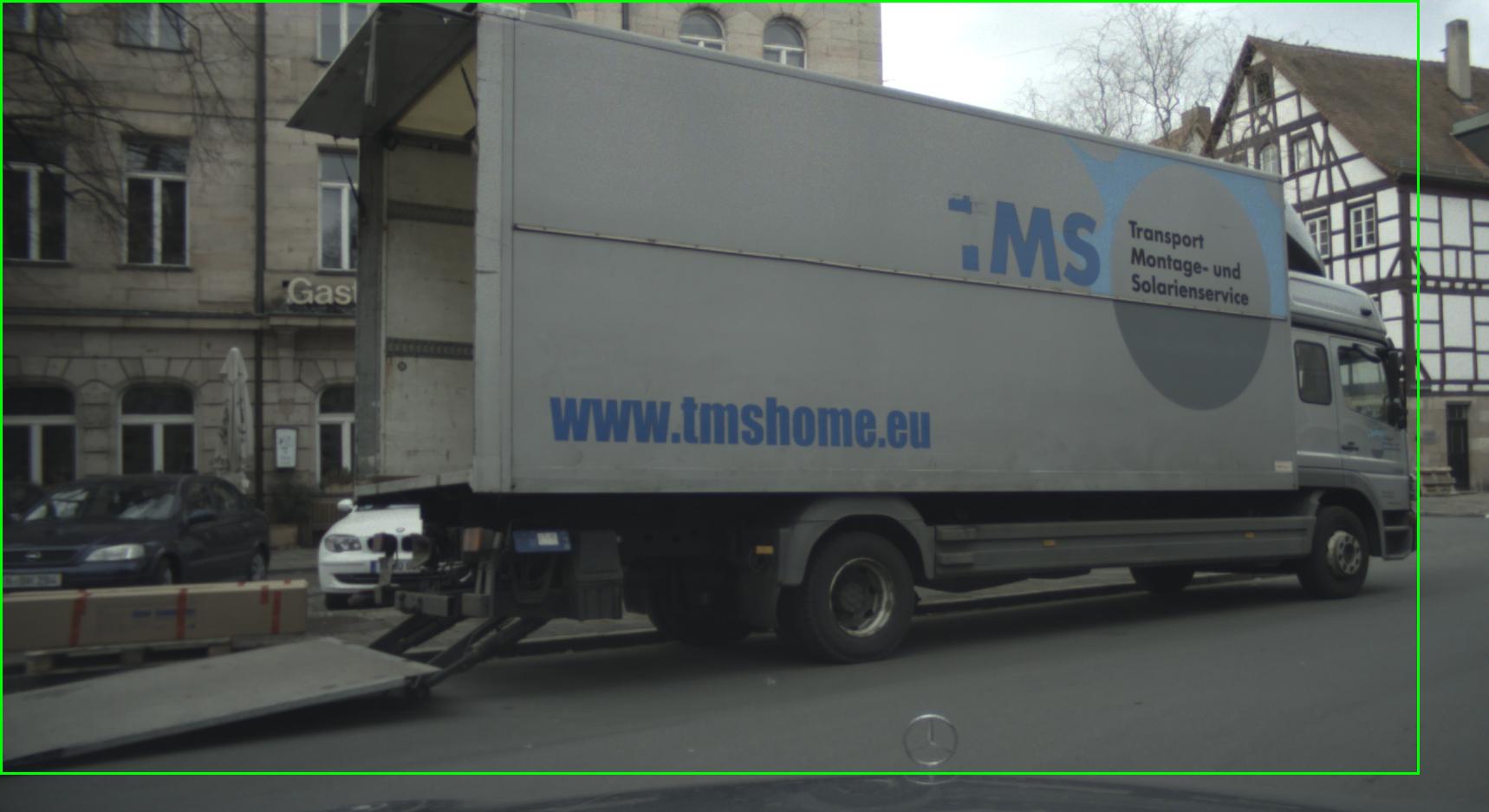} \\
        \includegraphics[width=0.2\textwidth]{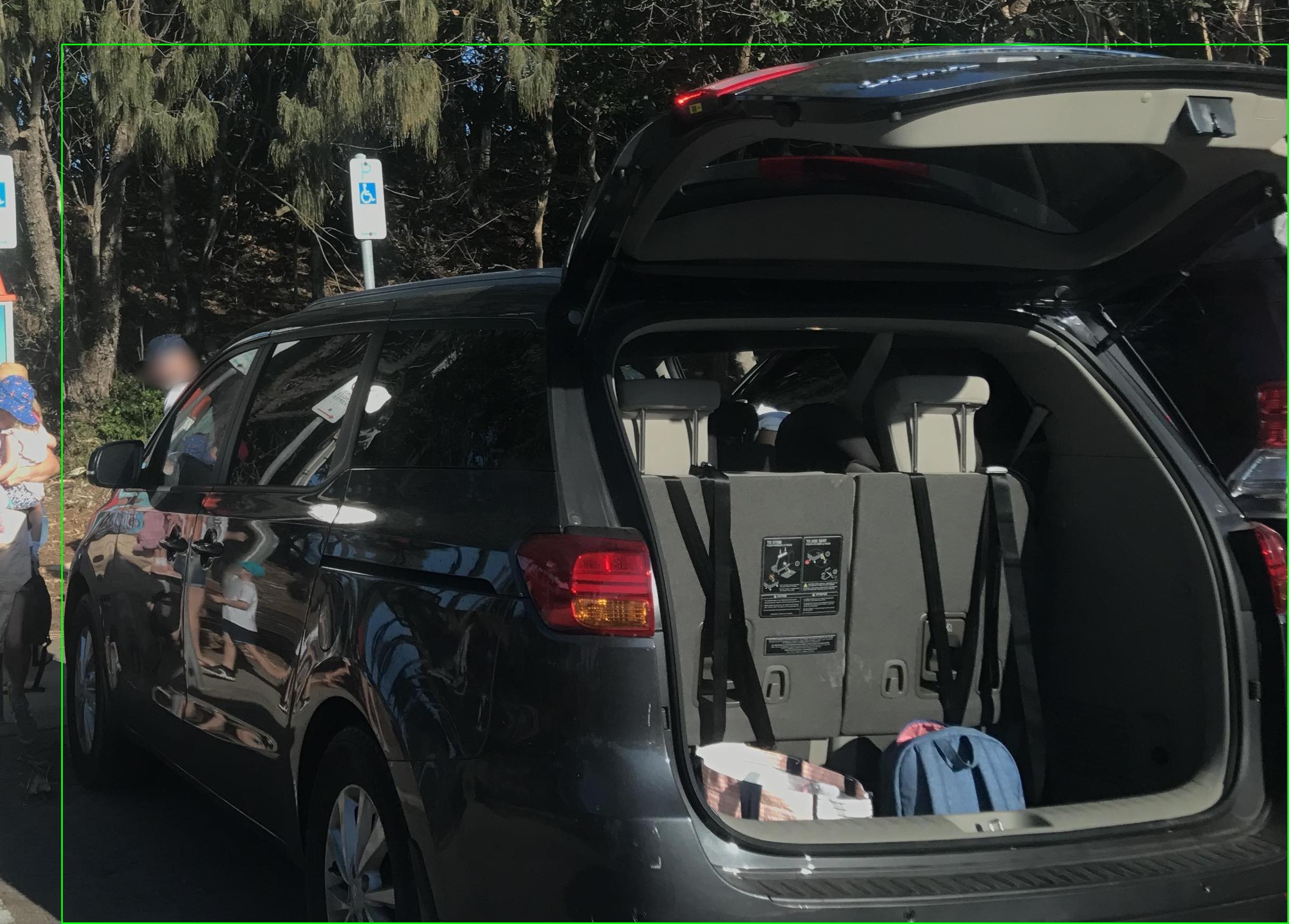} &
        \includegraphics[width=0.2\textwidth]{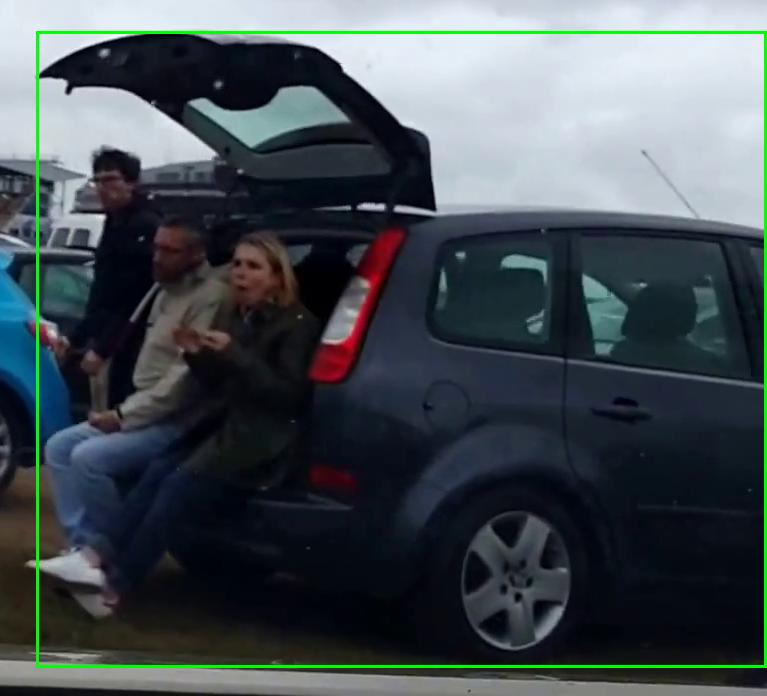} \\
        \includegraphics[width=0.2\textwidth]{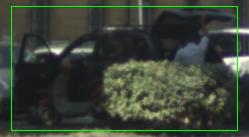} & \\
    \end{tabular}
    } \\
    \hline
    
    \vspace{0cm}
    \rotatebox{90}{\textbf{Vehicle-Tractor}} &
    \vspace{0cm}
    \textcolor{cyan}{Keywords}:
    \begin{itemize}
        \item[] \textcolor{brown}{"Tractor"},
        \item[] \textcolor{brown}{"Farm Tractor"},
        \item[] \textcolor{brown}{"Agricultural Tractor"},
    \end{itemize}
    \textcolor{cyan}{Descriptions}:
    \begin{itemize}
        \item[] \textcolor{brown}{"A tractor is a versatile vehicle designed to deliver a high tractive effort at slow speeds, used for plowing, planting, and other farm tasks."},
        \item[] \textcolor{brown}{"A tractor is visible in the image, requiring vehicles to be aware of its presence on rural roads and fields."},
        \item[] \textcolor{brown}{"The image contains a tractor on the country road or on the field in the background next to the road."},
        \item[] \textcolor{brown}{"The image shows a large green or red agricultural tractor with a front shovel."}
    \end{itemize}
    &
    {
    \vspace{0cm}
    \begin{tabular}{cc}
        \includegraphics[width=0.18\textwidth]{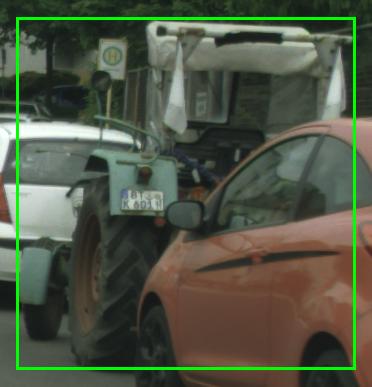} &
        \includegraphics[width=0.18\textwidth]{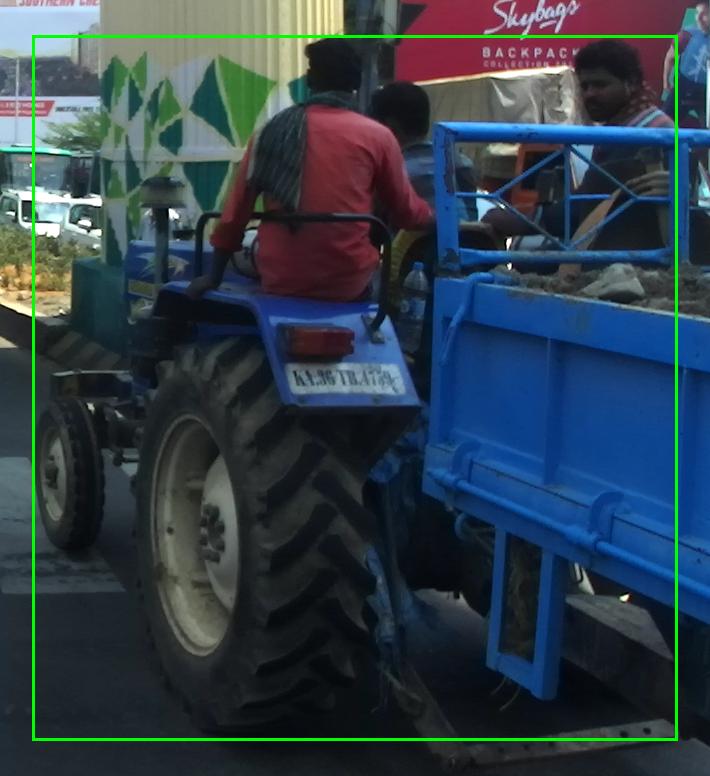} \\
        \includegraphics[width=0.18\textwidth]{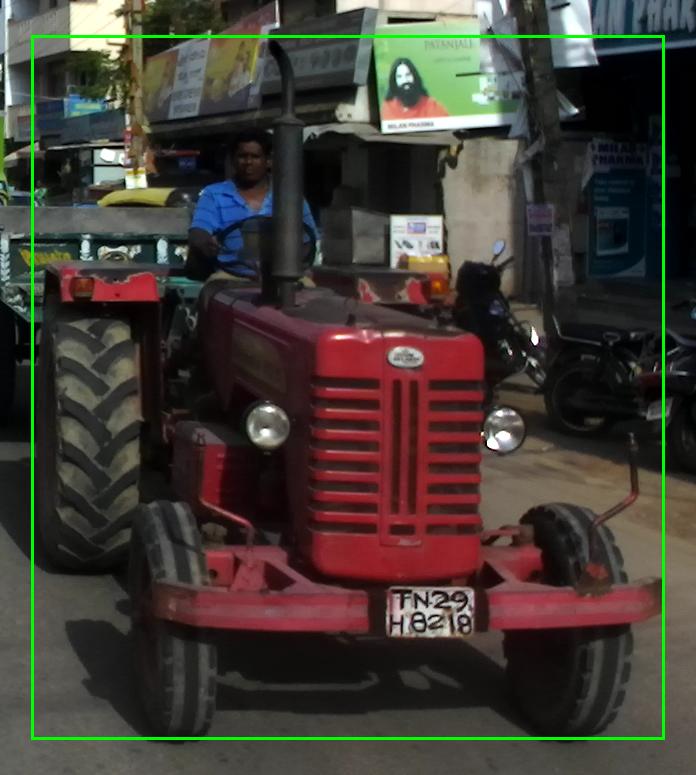} &
        \includegraphics[width=0.18\textwidth]{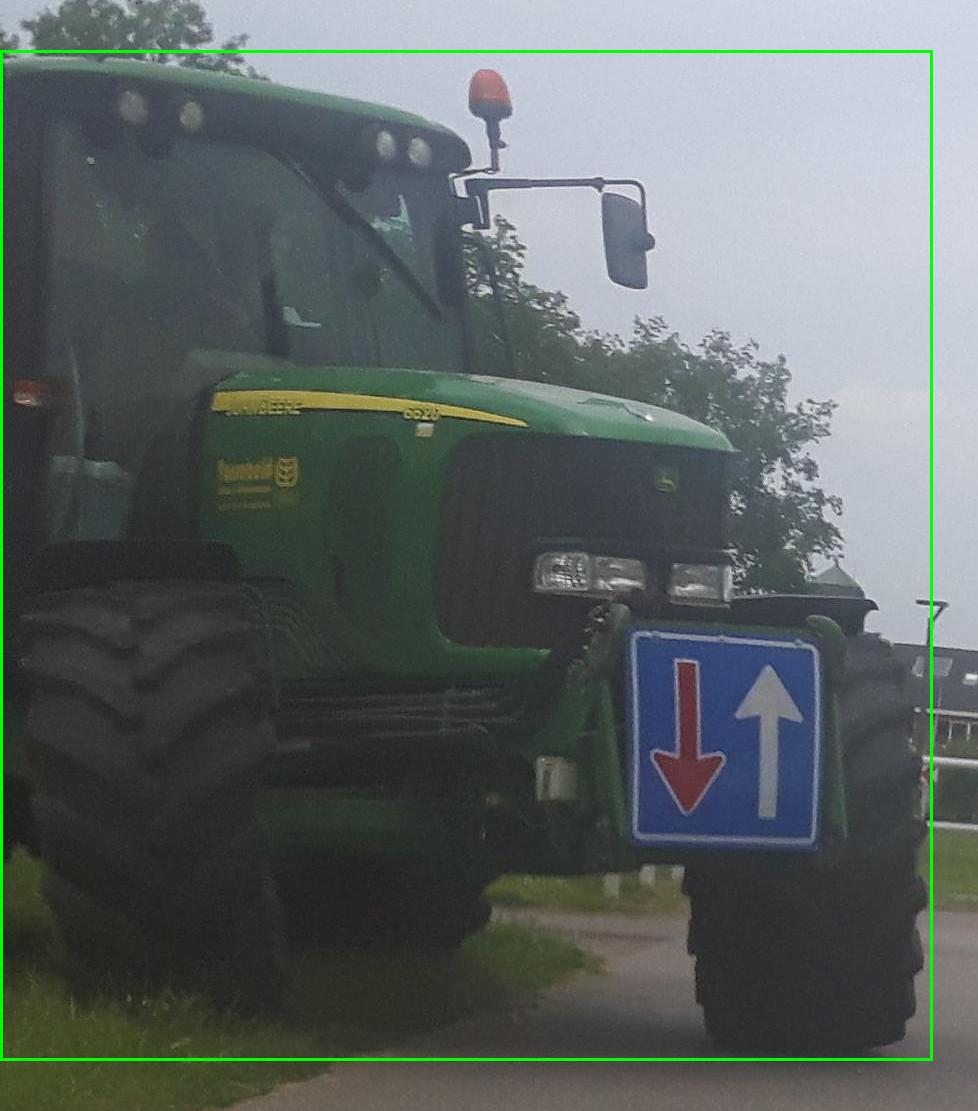} \\
        \includegraphics[width=0.18\textwidth]{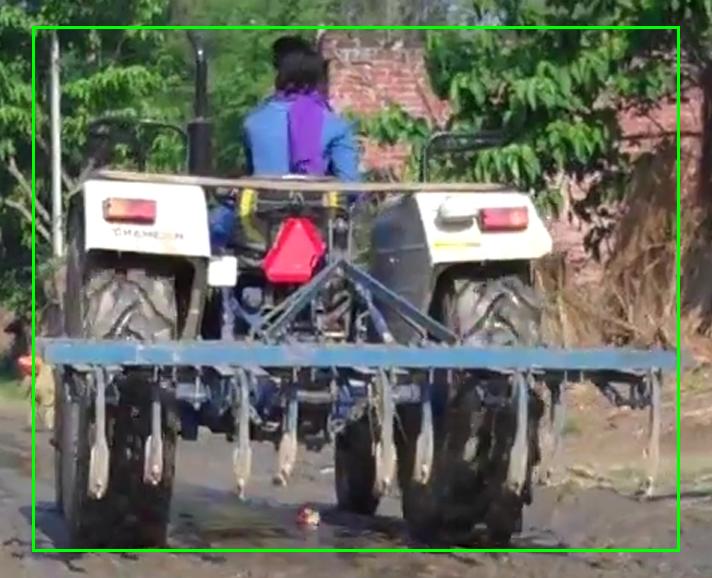} & \\
    \end{tabular}
    } \\
    \hline
    
    \caption{Illustration of five search queries from different categories including their vision and language support sets.}
    \label{tab:more_search_query_examples}
\end{longtable}
}

{
    \onecolumn
    \small
    \setlength{\tabcolsep}{3pt} 
    \centering
    
    \begin{longtable}{l *{8}{S[table-format=2.2, detect-weight=true, mode=text]}}
        \toprule
    
        \sisetup{
          table-align-uncertainty=true,
          separate-uncertainty=true,
        }
        \renewrobustcmd{\bfseries}{\fontseries{b}\selectfont}
        \centering 
    
        {\multirow{3}{*}{Class}} & \multicolumn{8}{c}{Class-wise Average Precision (AP) [\%]} \\ 
        \cmidrule(lr){2-9}
        & {\centering GDINO} & {\centering OpenCLIP} & {\centering SigLIP2} & {\centering BLIP2} & {\centering MetaCLIP2} & {\centering RADIO} & {\centering NACLIP} & {\centering NARADIO} \\ 
        & {\centering \cite{GroundingDINOMarryingLiu2025}} & {\centering \cite{OpenclipIlharco2021}} & {\centering \cite{Siglip2MultilingualTschannen2025}} & {\centering \cite{BLIP2BootstrappingLi2023}} & {\centering \cite{Metaclip2Chuang2025}} & {\centering \cite{RADIOv2.5ImprovedBaselinesHeinrich2025}} & {\centering \cite{PayAttentionYourHajimiri2025}} & {\centering \cite{RayFrontsOpenSetAlama2025, PayAttentionYourHajimiri2025}} \\ 
        \midrule
        \endfirsthead
    
        \multicolumn{9}{c}{\tablename\ \thetable\ -- Continued from previous page} \\
        \toprule
        {\multirow{3}{*}{Class}} & \multicolumn{8}{c}{Class-wise Average Precision (AP) [\%]} \\ 
        \cmidrule(lr){2-9}
        & {\centering GDINO} & {\centering OpenCLIP} & {\centering SigLIP2} & {\centering BLIP2} & {\centering MetaCLIP2} & {\centering RADIO} & {\centering NACLIP} & {\centering NARADIO} \\ 
        & {\centering \cite{GroundingDINOMarryingLiu2025}} & {\centering \cite{OpenclipIlharco2021}} & {\centering \cite{Siglip2MultilingualTschannen2025}} & {\centering \cite{BLIP2BootstrappingLi2023}} & {\centering \cite{Metaclip2Chuang2025}} & {\centering \cite{RADIOv2.5ImprovedBaselinesHeinrich2025}} & {\centering \cite{PayAttentionYourHajimiri2025}} & {\centering \cite{RayFrontsOpenSetAlama2025, PayAttentionYourHajimiri2025}} \\ 
        \midrule
        \endhead
    
        \bottomrule
        \multicolumn{9}{c}{\tablename\ \thetable\ -- Continued on next page} \\
        \endfoot
    
        \endlastfoot
    
        Animal-Real-Cat & 0.11 & 0.18 & 0.09 & 0.03 & 0.03 & 0.07 & 0.11 & \bfseries 0.14 \\
        Animal-Real-Cow & 16.86 & 7.97 & 16.55 & 9.22 & 14.21 & 13.40 & 27.96 & \bfseries 28.25 \\
        Animal-Real-Deer & \bfseries 48.04 & 6.67 & 6.69 & 39.40 & 33.53 & 38.49 & 10.43 & 40.90 \\
        Animal-Real-Dog & \bfseries 29.74 & 8.95 & 13.71 & 12.78 & 16.06 & 10.67 & 20.39 & 28.23 \\
        Animal-Real-Donkey & 0.00 & 0.01 & 0.07 & 0.02 & 0.03 & 0.15 & \bfseries 0.22 & 0.10 \\
        Animal-Real-Horse & \bfseries 14.92 & 1.31 & 1.36 & 2.40 & 0.97 & 3.76 & 3.44 & 4.83 \\
        Animal-Real-Sheep & 10.71 & 3.41 & 17.66 & 10.42 & 10.47 & 12.82 & 11.87 & \bfseries 28.97 \\
        Animal-Real-Wildlife & 0.02 & 0.01 & 0.19 & 0.01 & 0.08 & 1.57 & \bfseries 16.70 & 4.22 \\
        Animal-Statue-Cow & 0.00 & 0.00 & 0.00 & \bfseries 0.02 & 0.00 & 0.00 & 0.00 & 0.01 \\
        Animal-Statue-Deer & 33.34 & 33.34 & 6.69 & 33.34 & 33.34 & \bfseries 33.36 & 33.34 & 33.34 \\
        Animal-Statue-Elephant & 0.01 & 0.02 & 0.06 & 0.07 & 0.01 & 0.03 & 0.08 & \bfseries 0.27 \\
        Animal-Statue-Horse & 0.17 & 0.33 & 1.41 & 1.19 & 0.49 & 1.04 & 0.87 & \bfseries 6.59 \\
        Animal-Statue-Lion & 0.36 & 2.30 & 1.83 & 4.29 & 4.13 & 1.99 & 5.90 & \bfseries 13.93 \\
        \hline
        Human-Construction-Worker & 1.18 & 12.91 & 16.27 & 15.64 & 16.57 & 14.45 & 13.14 & \bfseries 18.12 \\
        Human-Firefighter & 0.03 & 26.59 & 7.66 & 27.39 & 6.94 & \bfseries 41.53 & 4.98 & 7.34 \\
        Human-Medical & 0.01 & 0.14 & 0.04 & 0.00 & 0.04 & 0.00 & 3.85 & \bfseries 7.15 \\
        Human-On-Loading-Area & 0.41 & 1.69 & 2.10 & 1.19 & 1.47 & 1.67 & 0.73 & \bfseries 2.51 \\
        Human-Police & 1.39 & 1.90 & 8.35 & 6.07 & \bfseries 18.68 & 12.52 & 7.56 & 15.56 \\
        Human-Refuse-Collector & 0.04 & 7.63 & 10.18 & 11.04 & \bfseries 18.16 & 11.44 & 3.92 & 0.95 \\
        Human-With-Sticks-or-Crutches & 1.07 & 0.82 & 5.56 & 1.64 & 0.84 & 0.80 & 2.33 & \bfseries 16.56 \\
        \hline
        Marking-Bicycle-Symbol & 4.29 & 12.06 & 10.21 & 16.91 & 10.92 & 7.78 & 19.14 & \bfseries 31.45 \\
        Marking-Bus-Text & 1.67 & 3.31 & 4.74 & 6.28 & 5.76 & 4.39 & 11.52 & \bfseries 16.12 \\
        Marking-Stop-Text & 0.53 & 2.36 & 2.50 & 5.33 & 2.05 & 2.87 & 3.10 & \bfseries 9.72 \\
        Marking-Temporarily-Invalidated & 0.03 & 0.06 & 0.42 & 0.10 & 0.05 & 0.14 & 0.07 & \bfseries 0.99 \\
        Marking-Yellow-Lane-Arrow & 0.74 & 0.72 & 0.82 & 1.94 & 0.78 & 1.06 & 1.27 & \bfseries 7.88 \\
        \hline
        Object-Ball & 14.69 & 3.09 & 11.41 & 3.98 & 5.36 & 6.89 & 3.98 & \bfseries 21.69 \\
        Object-Beacon & 5.21 & 5.76 & 8.04 & 7.89 & \bfseries 12.00 & 9.09 & 6.28 & 8.38 \\
        Object-Euro-Pallet & 0.60 & 4.07 & 12.66 & 3.95 & 19.10 & 13.72 & 9.29 & \bfseries 24.98 \\
        Object-Hand-Dolly & 0.19 & 2.49 & 5.10 & 1.15 & \bfseries 10.58 & 9.62 & 1.53 & 5.87 \\
        Object-Hydrant & \bfseries 43.25 & 14.82 & 10.08 & 13.15 & 14.73 & 16.23 & 13.21 & 20.97 \\
        Object-Office-Chair & 0.21 & 0.02 & 0.01 & 0.01 & 0.01 & 0.03 & 0.01 & \bfseries 0.04 \\
        Object-Pallet-Truck & 0.19 & 0.40 & 7.17 & 0.46 & 2.31 & \bfseries 7.19 & 0.34 & 6.75 \\
        Object-Platform-Truck & 0.33 & 0.46 & 0.55 & 0.48 & 0.94 & \bfseries 1.69 & 0.33 & 0.51 \\
        Object-Rollator & 0.10 & 0.26 & 0.52 & 0.40 & 2.68 & 0.29 & 0.89 & \bfseries 10.85 \\
        Object-Shopping-Cart & 0.55 & 6.42 & 1.21 & 3.69 & 2.21 & 3.02 & 6.31 & \bfseries 6.65 \\
        Object-Shopping-Trolley & 0.57 & 0.53 & 0.60 & 0.91 & 0.54 & 0.44 & 1.16 & \bfseries 2.29 \\
        Object-Suitcase-Trolley & 3.90 & 1.30 & 1.72 & 2.26 & 2.11 & 2.16 & 2.28 & \bfseries 6.79 \\
        Object-Traffic-Cone & \bfseries 60.69 & 22.34 & 22.42 & 28.62 & 26.94 & 28.26 & 45.75 & 59.11 \\
        Object-Trash-Bin & 10.77 & 10.47 & 16.00 & 8.32 & 12.27 & 8.44 & 12.18 & \bfseries 29.48 \\
        Object-Wheelbarrow & 1.10 & 4.43 & 0.75 & \bfseries 10.33 & 0.87 & 0.66 & 6.61 & 7.31 \\
        \hline
        Rideable-Cityscooter & 0.29 & 0.72 & 1.06 & 0.48 & 2.49 & 1.00 & 0.94 & \bfseries 4.44 \\
        Rideable-Police-Motorcycle & 0.58 & \bfseries 44.78 & 7.87 & 34.78 & 10.69 & 10.85 & 8.41 & 5.86 \\
        Rideable-Quad & 0.20 & 10.89 & 11.05 & 11.05 & 10.99 & 11.34 & 10.88 & \bfseries 13.00 \\
        Rideable-Segway & 10.47 & 0.61 & 0.03 & 0.03 & 13.92 & 0.05 & 0.38 & \bfseries 14.95 \\
        Rideable-Skateboard & 2.28 & 2.45 & 3.92 & 2.38 & 3.07 & 2.27 & \bfseries 4.90 & 4.05 \\
        Rideable-Skates & 0.34 & 2.01 & 1.26 & 2.25 & 4.01 & 20.02 & 1.44 & \bfseries 33.41 \\
        Rideable-Ski & 0.31 & 1.21 & 8.34 & 0.22 & 0.51 & 17.69 & \bfseries 22.50 & 5.50 \\
        Rideable-Stroller & \bfseries 30.63 & 5.37 & 5.78 & 6.18 & 4.16 & 5.65 & 8.48 & 18.99 \\
        Rideable-Three-Wheeler & 17.80 & 9.99 & 14.55 & 24.85 & \bfseries 36.59 & 26.89 & 12.07 & 27.04 \\
        Rideable-Toy-Car & 0.65 & 13.63 & \bfseries 41.64 & 6.98 & 34.53 & 26.77 & 9.06 & 33.61 \\
        Rideable-Wheelchair & 2.31 & 1.28 & 2.96 & 1.40 & 1.69 & 3.28 & 1.48 & \bfseries 5.65 \\
        \hline
        Scene-Active-Amber-Lights & 0.12 & 0.51 & 0.38 & 0.81 & 0.83 & \bfseries 1.21 & 0.43 & 0.80 \\
        Scene-Active-Emergency-Lights & 0.10 & 0.23 & 0.49 & \bfseries 2.11 & 0.22 & 1.35 & 0.77 & 1.93 \\
        Scene-Fog & 0.87 & 86.36 & 71.19 & \bfseries 87.33 & 68.92 & 61.45 & 86.86 & 1.58 \\
        Scene-Open-Door & 1.71 & 2.26 & 3.22 & 2.47 & 2.70 & 2.62 & 3.00 & \bfseries 7.58 \\
        Scene-Open-Hood & 0.07 & 0.09 & \bfseries 6.86 & 3.00 & 0.59 & 0.35 & 0.06 & 0.25 \\
        Scene-Open-Trunk & 1.85 & 2.93 & 4.29 & 3.20 & 3.57 & 3.30 & 2.70 & \bfseries 7.58 \\
        Scene-Snow & 6.63 & 60.34 & 54.60 & 66.47 & 56.87 & 70.17 & \bfseries 74.79 & 61.49 \\
        Scene-Tunnel & 2.74 & 33.25 & 24.11 & 32.46 & 20.13 & 38.29 & 39.01 & \bfseries 41.71 \\
        \hline
        Sign-Animal-Sign & 1.41 & 2.35 & 4.72 & 3.37 & \bfseries 6.15 & 3.91 & 2.96 & 3.66 \\
        Sign-Road-Bumper-Sign & 1.10 & 1.94 & 2.44 & 1.51 & 2.04 & 1.96 & 1.35 & \bfseries 2.55 \\
        Sign-Temporarily-Invalidated-Sign & 0.05 & 0.10 & 0.12 & \bfseries 0.15 & 0.09 & 0.10 & 0.11 & 0.13 \\
        Sign-Train-Sign & 0.62 & 1.87 & \bfseries 3.82 & 2.11 & 1.57 & 1.87 & 1.51 & 1.34 \\
        Sign-Warning-Triangle & 0.17 & 0.14 & 6.40 & 0.31 & 3.70 & 0.21 & 1.02 & \bfseries 12.93 \\
        \hline
        Trailer-Agricultural-Trailer & 0.17 & 0.33 & 0.50 & 0.71 & \bfseries 1.40 & 1.34 & 0.56 & \bfseries 1.40 \\
        Trailer-Bicycle-Trailer & 0.11 & 0.15 & 0.88 & 1.14 & 0.83 & 0.41 & 0.50 & \bfseries 4.20 \\
        Trailer-Boat-Trailer & 0.10 & 12.41 & 7.86 & 0.53 & 3.25 & 0.47 & 6.77 & \bfseries 28.22 \\
        Trailer-Car-Trailer & 4.59 & 2.44 & 4.45 & 5.25 & 4.87 & 5.15 & 6.46 & \bfseries 18.21 \\
        Trailer-Caravan-Trailer & 0.32 & 2.57 & 1.59 & 2.11 & 2.88 & 1.57 & 4.42 & \bfseries 11.52 \\
        Trailer-Carriage & 16.19 & 8.09 & 20.22 & 18.01 & 15.16 & 12.52 & 9.92 & \bfseries 46.44 \\
        Trailer-Warning-Trailer & 0.47 & 0.92 & 1.78 & 1.13 & \bfseries 3.01 & 1.85 & 1.00 & 2.11 \\
        \hline
        Vehicle-Concrete-Mixer & 3.23 & 18.80 & 26.23 & 20.50 & 25.29 & 18.49 & 24.30 & \bfseries 32.84 \\
        Vehicle-Excavator & 7.05 & 7.65 & 14.91 & 12.84 & 16.97 & 14.20 & 23.02 & \bfseries 31.37 \\
        Vehicle-Forklift & 3.02 & 6.11 & 8.86 & 4.52 & 2.34 & 5.55 & \bfseries 15.15 & 15.03 \\
        Vehicle-Harvester & 0.01 & 0.91 & 0.04 & 0.12 & 0.01 & 0.04 & \bfseries 0.75 & 0.34 \\
        Vehicle-Loader & 0.38 & 3.38 & 3.08 & 5.75 & 4.90 & 5.33 & 9.36 & \bfseries 19.26 \\
        Vehicle-Steamroller & 0.18 & 1.44 & 4.18 & 2.26 & 5.31 & 6.28 & 4.38 & \bfseries 15.85 \\
        Vehicle-Tractor & 0.59 & 6.97 & 8.61 & 5.68 & 8.34 & 5.54 & 11.16 & \bfseries 25.01 \\
        Vehicle-Truck-Crane & 3.65 & 9.62 & 11.89 & 10.21 & 11.65 & 11.05 & 15.45 & \bfseries 19.14 \\
        Vehicle-Fire & 4.97 & 14.81 & 12.77 & \bfseries 16.94 & 12.75 & 12.32 & 6.00 & 10.98 \\
        Vehicle-Garbage & 1.59 & 16.79 & 17.84 & 13.04 & \bfseries 21.21 & 9.17 & 19.62 & 5.38 \\
        Vehicle-Medical & 6.58 & 14.25 & 17.46 & 16.28 & 15.84 & 8.14 & 14.48 & \bfseries 18.45 \\
        Vehicle-Military & 0.05 & 4.25 & 27.98 & 24.90 & \bfseries 29.89 & 30.12 & 2.70 & 16.83 \\
        Vehicle-Police & 0.57 & 2.67 & 4.70 & 6.89 & 4.98 & 6.70 & 4.76 & \bfseries 27.49 \\
        Vehicle-Winter & 0.01 & 6.78 & 0.37 & 1.10 & 0.60 & 1.46 & \bfseries 13.10 & 0.26 \\
        Vehicle-Bicycle-On-Back & 0.16 & 1.26 & 7.43 & 2.18 & 2.40 & 3.08 & 0.33 & \bfseries 11.41 \\
        Vehicle-Bicycle-On-Roof & 0.02 & 0.01 & \bfseries 0.19 & 0.02 & 0.01 & 0.02 & 0.01 & 0.05 \\
        Vehicle-Car-Truck & 0.07 & 0.82 & 6.70 & 11.56 & \bfseries 11.94 & 9.60 & 0.91 & 10.57 \\
        Vehicle-Recreational & 1.16 & 5.83 & 9.75 & 14.45 & 24.48 & 15.86 & 22.51 & \bfseries 28.12 \\
        Vehicle-Train & 25.79 & 24.03 & 36.72 & 36.48 & 18.20 & 31.35 & 36.90 & \bfseries 38.35 \\
        \midrule
        \bfseries Mean over all classes (MAP) & 5.25 & 7.45 & 8.57 & 9.14 & 9.41 & 9.49 & 9.59 & \bfseries 14.27 \\
        \bottomrule
        \caption{Evaluation of text-to-image retrieval methods on the SearchAD test set: Class-wise Average Precision (AP), and the Mean Average Precision (MAP) over all 90 classes. Bold values indicate the highest scores across the models. The models (columns) are ordered ascending by their overall MAP.}
        \label{tab:map_class_wise_text_to_image_retrieval}
    \end{longtable}

    \newpage
    \begin{longtable}{l *{8}{S[table-format=2.2, detect-weight=true, mode=text]}}
        \toprule
    
        \sisetup{
          table-align-uncertainty=true,
          separate-uncertainty=true,
        }
        \renewrobustcmd{\bfseries}{\fontseries{b}\selectfont}
        \centering 
    
        {\multirow{3}{*}{Class}} & \multicolumn{8}{c}{Class-wise Average Precision (AP) [\%]} \\ 
        \cmidrule(lr){2-9}
        & {\centering OpenCLIP} & {\centering RADIO} & {\centering MetaCLIP2} & {\centering SigLIP2} & {\centering NACLIP} & {\centering GDINO} & {\centering BLIP2} & {\centering NARADIO} \\ 
        & {\centering \cite{OpenclipIlharco2021}} & {\centering \cite{RADIOv2.5ImprovedBaselinesHeinrich2025}} & {\centering \cite{Metaclip2Chuang2025}} & {\centering \cite{Siglip2MultilingualTschannen2025}} & {\centering \cite{PayAttentionYourHajimiri2025}} & {\centering \cite{GroundingDINOMarryingLiu2025}} & {\centering \cite{BLIP2BootstrappingLi2023}} & {\centering \cite{RayFrontsOpenSetAlama2025, PayAttentionYourHajimiri2025}} \\ 
        \midrule
        \endfirsthead
    
        \multicolumn{9}{c}{\tablename\ \thetable\ -- Continued from previous page} \\
        \toprule
        {\multirow{3}{*}{Class}} & \multicolumn{8}{c}{Class-wise Average Precision (AP) [\%]} \\ 
        \cmidrule(lr){2-9}
        & {\centering OpenCLIP} & {\centering RADIO} & {\centering MetaCLIP2} & {\centering SigLIP2} & {\centering NACLIP} & {\centering GDINO} & {\centering BLIP2} & {\centering NARADIO} \\ 
        & {\centering \cite{OpenclipIlharco2021}} & {\centering \cite{RADIOv2.5ImprovedBaselinesHeinrich2025}} & {\centering \cite{Metaclip2Chuang2025}} & {\centering \cite{Siglip2MultilingualTschannen2025}} & {\centering \cite{PayAttentionYourHajimiri2025}} & {\centering \cite{GroundingDINOMarryingLiu2025}} & {\centering \cite{BLIP2BootstrappingLi2023}} & {\centering \cite{RayFrontsOpenSetAlama2025, PayAttentionYourHajimiri2025}} \\ 
        \midrule
        \endhead
    
        \bottomrule
        \multicolumn{9}{c}{\tablename\ \thetable\ -- Continued on next page} \\
        \endfoot
    
        \endlastfoot
    
        Animal-Real-Cat & 0.02 & 0.02 & 0.01 & 0.02 & 0.03 & \bfseries 0.07 & 0.06 & 0.02 \\
        Animal-Real-Cow & 0.39 & 0.19 & 2.19 & 0.25 & \bfseries 17.63 & 10.51 & 13.06 & 0.26 \\
        Animal-Real-Deer & 33.39 & 33.48 & 33.62 & 33.34 & \bfseries 69.70 & 21.11 & 33.57 & 33.35 \\
        Animal-Real-Dog & 5.02 & 1.47 & 2.11 & 4.70 & 18.85 & \bfseries 20.73 & 9.16 & 6.91 \\
        Animal-Real-Donkey & 0.01 & 0.03 & 0.02 & 0.01 & 0.01 & \bfseries 0.05 & 0.00 & 0.01 \\
        Animal-Real-Horse & 0.03 & 0.03 & 0.02 & 0.02 & \bfseries 4.27 & 2.28 & 0.57 & 0.10 \\
        Animal-Real-Sheep & 0.04 & 0.13 & 0.03 & 0.05 & \bfseries 11.20 & 0.96 & 4.47 & 3.80 \\
        Animal-Real-Wildlife & 0.04 & 0.02 & 0.00 & 0.01 & 0.11 & \bfseries 5.57 & 0.00 & 0.00 \\
        Animal-Statue-Cow & \bfseries 0.00 & \bfseries 0.00 & \bfseries 0.00 & \bfseries 0.00 & \bfseries 0.00 & \bfseries 0.00 & \bfseries 0.00 & \bfseries 0.00 \\
        Animal-Statue-Deer & 33.34 & 33.35 & 33.34 & 33.34 & \bfseries 33.36 & 33.34 & 33.35 & 33.34 \\
        Animal-Statue-Elephant & 0.02 & 1.24 & 0.01 & 0.08 & 0.01 & 0.26 & 0.01 & \bfseries 0.08 \\
        Animal-Statue-Horse & 0.40 & 2.77 & 2.93 & \bfseries 3.00 & 2.52 & 1.93 & 0.56 & 1.39 \\
        Animal-Statue-Lion & 0.05 & 0.05 & 0.05 & 0.05 & \bfseries 8.39 & 3.73 & 1.93 & 0.07 \\
        \hline
        Human-Construction-Worker & 1.38 & 2.79 & 1.92 & 2.38 & \bfseries 11.68 & 9.77 & 8.94 & 9.70 \\
        Human-Firefighter & 4.57 & 0.02 & 0.03 & 0.03 & 7.48 & 0.11 & \bfseries 27.14 & 0.05 \\
        Human-Medical & 0.01 & 0.00 & 0.01 & 0.00 & \bfseries 0.02 & 0.01 & 0.00 & 0.00 \\
        Human-On-Loading-Area & 0.62 & 0.45 & 0.55 & 0.50 & \bfseries 2.11 & 0.95 & 1.61 & 0.46 \\
        Human-Police & 0.78 & 2.44 & 7.57 & 0.84 & \bfseries 17.00 & 6.76 & 14.77 & 15.55 \\
        Human-Refuse-Collector & 0.73 & 0.54 & 0.58 & \bfseries 10.75 & 2.59 & 1.19 & 9.03 & 0.27 \\
        Human-With-Sticks-or-Crutches & 0.74 & 0.45 & 0.64 & 0.80 & 0.73 & 0.68 & \bfseries 0.96 & 0.55 \\
        \hline
        Marking-Bicycle-Symbol & 5.98 & 4.24 & 5.07 & 6.12 & 8.16 & \bfseries 17.42 & 14.63 & 5.57 \\
        Marking-Bus-Text & 2.85 & 1.71 & 3.41 & 2.53 & 4.46 & 6.57 & \bfseries 7.94 & 5.40 \\
        Marking-Stop-Text & 1.15 & 0.88 & 0.56 & 0.70 & 0.44 & 0.92 & \bfseries 3.46 & 1.44 \\
        Marking-Temporarily-Invalidated & 0.10 & 0.04 & 0.11 & 0.06 & 0.18 & \bfseries 0.32 & 0.16 & 0.06 \\
        Marking-Yellow-Lane-Arrow & 0.52 & 0.56 & 0.47 & 0.47 & 4.45 & \bfseries 4.59 & 1.80 & 0.92 \\
        \hline
        Object-Ball & 0.15 & 0.09 & 0.14 & 0.41 & 0.09 & \bfseries 22.52 & 1.86 & 0.93 \\
        Object-Beacon & 6.45 & 7.94 & 7.46 & 12.02 & 17.57 & \bfseries 25.95 & 25.93 & 20.13 \\
        Object-Euro-Pallet & 5.00 & 3.04 & 4.20 & \bfseries 19.90 & 4.36 & 2.62 & 12.63 & 9.73 \\
        Object-Hand-Dolly & 1.44 & 0.33 & 0.34 & 1.11 & 1.86 & \bfseries 2.09 & 0.38 & 0.39 \\
        Object-Hydrant & 7.19 & 10.54 & 8.67 & 10.54 & 8.10 & \bfseries 13.27 & 7.87 & 8.10 \\
        Object-Office-Chair & 0.01 & 0.04 & 0.01 & 0.01 & 0.00 & \bfseries 0.40 & 0.01 & 0.01 \\
        Object-Pallet-Truck & 0.36 & 0.52 & 0.39 & \bfseries 0.60 & 0.20 & 0.08 & 0.54 & 0.36 \\
        Object-Platform-Truck & 0.38 & 0.58 & 0.41 & 0.38 & \bfseries 1.21 & 0.57 & 0.44 & 0.36 \\
        Object-Rollator & 0.22 & 0.11 & 0.18 & 0.22 & 1.07 & \bfseries 1.24 & 0.44 & 0.61 \\
        Object-Shopping-Cart & 0.61 & 3.44 & 1.16 & 0.20 & 6.79 & 5.73 & 1.71 & \bfseries 9.02 \\
        Object-Shopping-Trolley & 0.60 & \bfseries 1.06 & 0.41 & 0.77 & 0.83 & 0.62 & 0.53 & 0.93 \\
        Object-Suitcase-Trolley & 2.12 & 2.13 & 2.60 & 2.82 & 2.12 & 2.75 & 2.90 & \bfseries 5.46 \\
        Object-Traffic-Cone & 10.68 & 13.77 & 10.89 & 13.28 & 36.88 & \bfseries 51.95 & 27.12 & 41.18 \\
        Object-Trash-Bin & 12.44 & 5.89 & 11.12 & 11.75 & 6.75 & \bfseries 19.78 & 14.86 & 16.13 \\
        Object-Wheelbarrow & 0.59 & 0.08 & 0.31 & 0.31 & \bfseries 13.39 & 2.70 & 2.20 & 0.09 \\
        \hline
        Rideable-Cityscooter & 0.38 & 0.31 & 0.63 & 0.29 & 1.29 & 0.46 & \bfseries 0.78 & 0.37 \\
        Rideable-Police-Motorcycle & 2.73 & 0.07 & \bfseries 59.10 & 0.64 & 13.11 & 4.98 & 34.95 & 0.33 \\
        Rideable-Quad & 2.70 & 0.05 & 9.72 & \bfseries 11.02 & 10.94 & 8.32 & 7.21 & 0.08 \\
        Rideable-Segway & 0.58 & 0.01 & 0.48 & \bfseries 8.75 & 0.02 & 3.32 & 0.04 & 0.05 \\
        Rideable-Skateboard & 0.58 & 0.20 & 1.94 & \bfseries 2.24 & 0.12 & 0.16 & 0.30 & 0.81 \\
        Rideable-Skates & 0.03 & 0.92 & 0.01 & 0.43 & 0.03 & 0.51 & \bfseries 0.97 & 0.06 \\
        Rideable-Ski & 6.65 & \bfseries 25.10 & 0.17 & 12.51 & 0.10 & 0.05 & 8.34 & 21.67 \\
        Rideable-Stroller & 2.00 & 0.93 & 2.31 & 1.75 & 11.06 & \bfseries 12.09 & 4.08 & 1.31 \\
        Rideable-Three-Wheeler & 1.44 & 5.91 & 1.38 & 1.87 & 19.47 & 20.89 & 7.92 & \bfseries 21.33 \\
        Rideable-Toy-Car & 2.35 & 1.06 & 1.75 & \bfseries 24.20 & 2.91 & 1.45 & 9.64 & 2.48 \\
        Rideable-Wheelchair & 0.51 & 1.14 & 0.17 & 1.66 & 2.26 & 1.91 & 1.12 & \bfseries 8.67 \\
        \hline
        Scene-Active-Amber-Lights & 0.24 & 1.13 & 0.37 & 0.60 & 0.78 & 0.60 & 1.14 & \bfseries 3.05 \\
        Scene-Active-Emergency-Lights & 0.11 & 0.13 & 0.17 & 0.13 & 0.11 & 0.16 & \bfseries 0.30 & 0.15 \\
        Scene-Fog & 71.57 & 23.94 & 46.04 & 53.21 & 2.79 & \bfseries 77.07 & 70.22 & 58.32 \\
        Scene-Open-Door & 2.50 & 2.04 & 2.12 & 2.73 & 1.83 & \bfseries 2.96 & 2.47 & 2.53 \\
        Scene-Open-Hood & 0.15 & 0.12 & 0.08 & \bfseries 0.20 & 0.08 & 0.08 & 0.11 & 0.10 \\
        Scene-Open-Trunk & 2.98 & 2.15 & 2.52 & 2.81 & 2.50 & 3.16 & \bfseries 3.20 & 2.20 \\
        Scene-Snow & 10.58 & 34.12 & 7.32 & 9.98 & 7.43 & 9.95 & \bfseries 41.99 & 11.07 \\
        Scene-Tunnel & 13.18 & 20.07 & 19.02 & 15.69 & 4.32 & 4.55 & \bfseries 26.92 & 5.87 \\
        \hline
        Sign-Animal-Sign & 1.13 & \bfseries 3.48 & 0.55 & 1.34 & 2.96 & 1.24 & 1.41 & 2.79 \\
        Sign-Road-Bumper-Sign & 1.08 & 1.07 & 0.94 & 1.08 & 1.03 & \bfseries 5.66 & 2.06 & 2.73 \\
        Sign-Temporarily-Invalidated-Sign & 0.17 & 0.11 & 0.11 & 0.15 & 0.11 & \bfseries 0.84 & 0.20 & 0.31 \\
        Sign-Train-Sign & 1.01 & 0.59 & 0.90 & 1.20 & 1.57 & \bfseries 2.02 & 1.60 & 0.64 \\
        Sign-Warning-Triangle & 0.25 & 0.50 & 0.38 & \bfseries 5.87 & 0.92 & 2.00 & 0.74 & 2.55 \\
        \hline
        Trailer-Agricultural-Trailer & 4.31 & 3.22 & 7.29 & \bfseries 9.39 & 2.05 & 0.52 & 3.31 & 2.12 \\
        Trailer-Bicycle-Trailer & 0.24 & 0.11 & \bfseries 2.86 & 0.36 & 0.28 & 0.24 & 0.70 & 3.01 \\
        Trailer-Boat-Trailer & 0.66 & 0.09 & 3.55 & 8.63 & 18.68 & 5.98 & 7.12 & \bfseries 31.67 \\
        Trailer-Car-Trailer & 3.89 & 2.42 & \bfseries 6.60 & 4.06 & 1.88 & 3.96 & 4.33 & 6.53 \\
        Trailer-Caravan-Trailer & 0.59 & 0.91 & 0.69 & 0.85 & 1.77 & 3.66 & 2.23 & \bfseries 6.90 \\
        Trailer-Carriage & 7.64 & 6.44 & 8.64 & 13.45 & 13.78 & 34.38 & 20.63 & \bfseries 42.04 \\
        Trailer-Warning-Trailer & 3.41 & 4.31 & \bfseries 4.85 & 3.76 & 0.96 & 1.24 & 2.51 & 2.99 \\
        \hline
        Vehicle-Concrete-Mixer & 2.80 & 6.30 & 13.51 & 8.74 & 10.20 & \bfseries 24.30 & 16.05 & 23.86 \\
        Vehicle-Excavator & 2.09 & 6.84 & 2.40 & 8.27 & 9.32 & 16.14 & 10.79 & \bfseries 31.34 \\
        Vehicle-Forklift & 0.44 & 6.34 & 0.53 & 6.42 & 9.33 & 5.26 & 12.46 & \bfseries 13.66 \\
        Vehicle-Harvester & 0.03 & 0.03 & 0.02 & 0.03 & 0.27 & \bfseries 0.38 & 0.13 & 0.29 \\
        Vehicle-Loader & 2.27 & 4.20 & 1.07 & 2.74 & 12.15 & \bfseries 17.32 & 5.16 & 17.54 \\
        Vehicle-Steamroller & 0.29 & 3.61 & 0.25 & 0.75 & 5.33 & \bfseries 10.86 & 3.49 & 15.66 \\
        Vehicle-Tractor & 3.05 & 2.67 & 4.17 & 8.82 & 7.16 & 6.61 & 7.84 & \bfseries 25.15 \\
        Vehicle-Truck-Crane & 7.20 & 10.06 & 3.28 & 9.65 & 6.03 & 4.64 & 11.06 & \bfseries 14.83 \\
        Vehicle-Fire & 2.20 & 4.05 & 4.78 & 2.98 & 8.26 & 9.35 & \bfseries 9.67 & 2.44 \\
        Vehicle-Garbage & 7.31 & 2.95 & \bfseries 14.32 & 14.04 & 2.48 & 3.49 & 6.09 & 8.83 \\
        Vehicle-Medical & 5.26 & 5.35 & 5.14 & 7.47 & 3.08 & 3.99 & 5.24 & \bfseries 22.92 \\
        Vehicle-Military & 6.07 & 1.11 & 30.85 & \bfseries 35.16 & 0.92 & 2.36 & 18.96 & 0.80 \\
        Vehicle-Police & 4.40 & 3.74 & 3.96 & 6.14 & 1.76 & 9.03 & 6.84 & \bfseries 27.24 \\
        Vehicle-Winter & 0.41 & 0.18 & 0.11 & \bfseries 2.15 & 0.20 & 0.15 & 0.48 & 0.42 \\
        Vehicle-Bicycle-On-Back & 0.31 & 1.30 & 0.22 & \bfseries 2.77 & 0.37 & 0.23 & 0.22 & 0.91 \\
        Vehicle-Bicycle-On-Roof & 0.01 & 0.02 & 0.02 & 0.03 & 0.07 & 0.02 & 0.03 & \bfseries 0.08 \\
        Vehicle-Car-Truck & 0.91 & 8.06 & 9.98 & \bfseries 10.94 & 2.17 & 5.62 & 6.65 & 1.74 \\
        Vehicle-Recreational & 9.73 & 7.69 & 9.94 & \bfseries 11.07 & 9.13 & 6.67 & 6.33 & 10.34 \\
        Vehicle-Train & 17.34 & 37.19 & 18.34 & 29.91 & 48.81 & 42.90 & 32.84 & \bfseries 52.66 \\
        \midrule
        \bfseries Mean over all classes (MAP) & 3.98 & 4.34 & 5.10 & 6.04 & 6.56 & 7.62 & 7.95 & \bfseries 8.31 \\
        \bottomrule
        \caption{Evaluation of image-to-image retrieval methods on the SearchAD test set: Class-wise Average Precision (AP), and the Mean Average Precision (MAP) over all 90 classes. Bold values indicate the highest scores across the models. The models (columns) are ordered ascending by their overall MAP.}
        \label{tab:map_class_wise_image_to_image_retrieval}
    \end{longtable}
}

{
\def\rankOne{\bfseries Ranked 1st}
\def\rankTwo{\bfseries Ranked 2nd}
\def\rankThree{\bfseries Ranked 3rd}
\def\rankFour{\bfseries Ranked 4th}
\def\rankFive{\bfseries Ranked 5th}

\captionsetup[subfigure]{labelformat=empty}

\newlength{\subfigwidth}
\setlength{\subfigwidth}{0.18\textwidth} 

\begin{figure*}[ht]
    \begingroup
    \centering 
    \small
    \rotatebox[origin=left]{90}{\hspace{0.02cm} \textbf{GDINO}} 
    \begin{subfigure}[b]{\subfigwidth}
        \caption{\rankOne}
        \includegraphics[width=\textwidth]{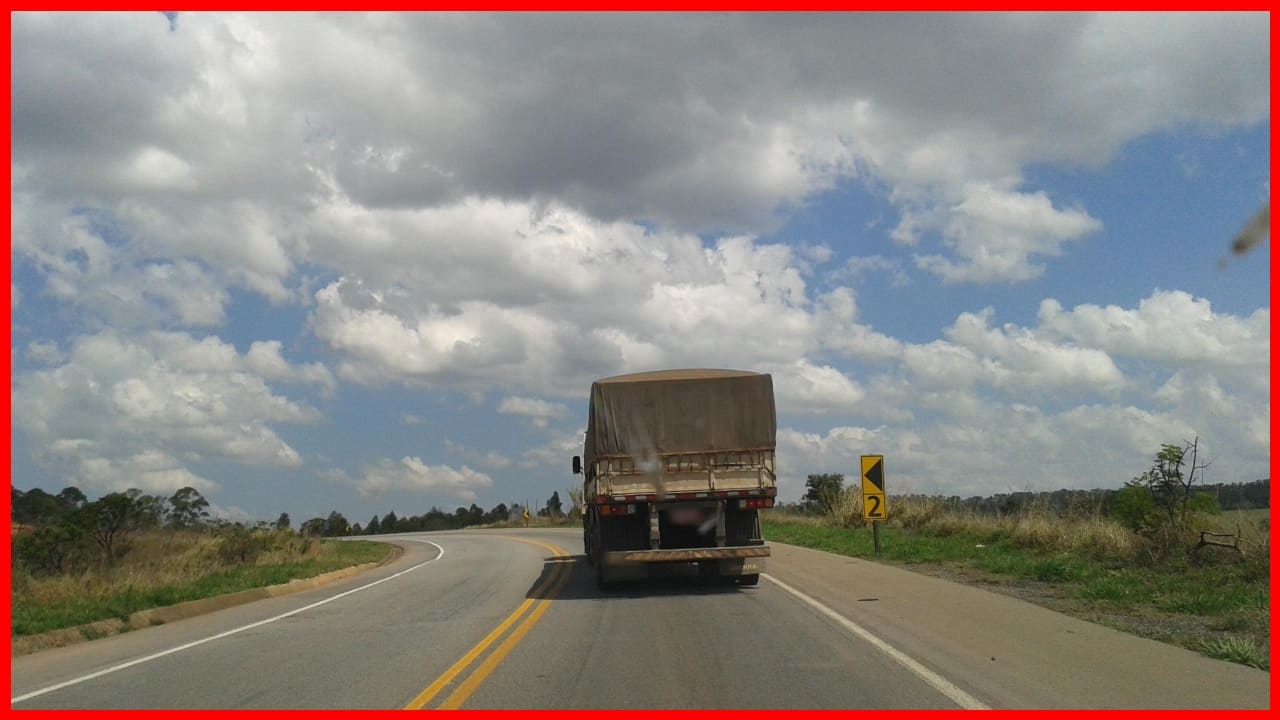}
    \end{subfigure}
    \begin{subfigure}[b]{\subfigwidth}
        \caption{\rankTwo}
        \includegraphics[width=\textwidth]{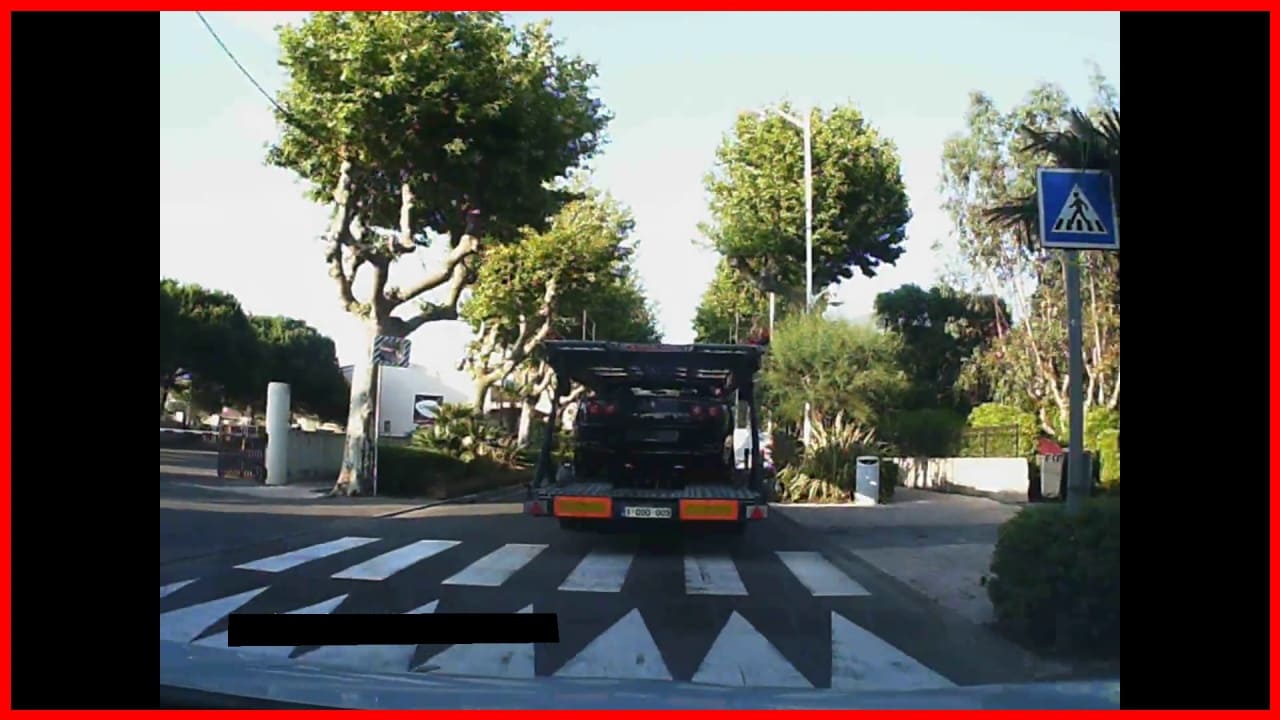}
    \end{subfigure}
    \begin{subfigure}[b]{\subfigwidth}
        \caption{\rankThree}
        \includegraphics[width=\textwidth]{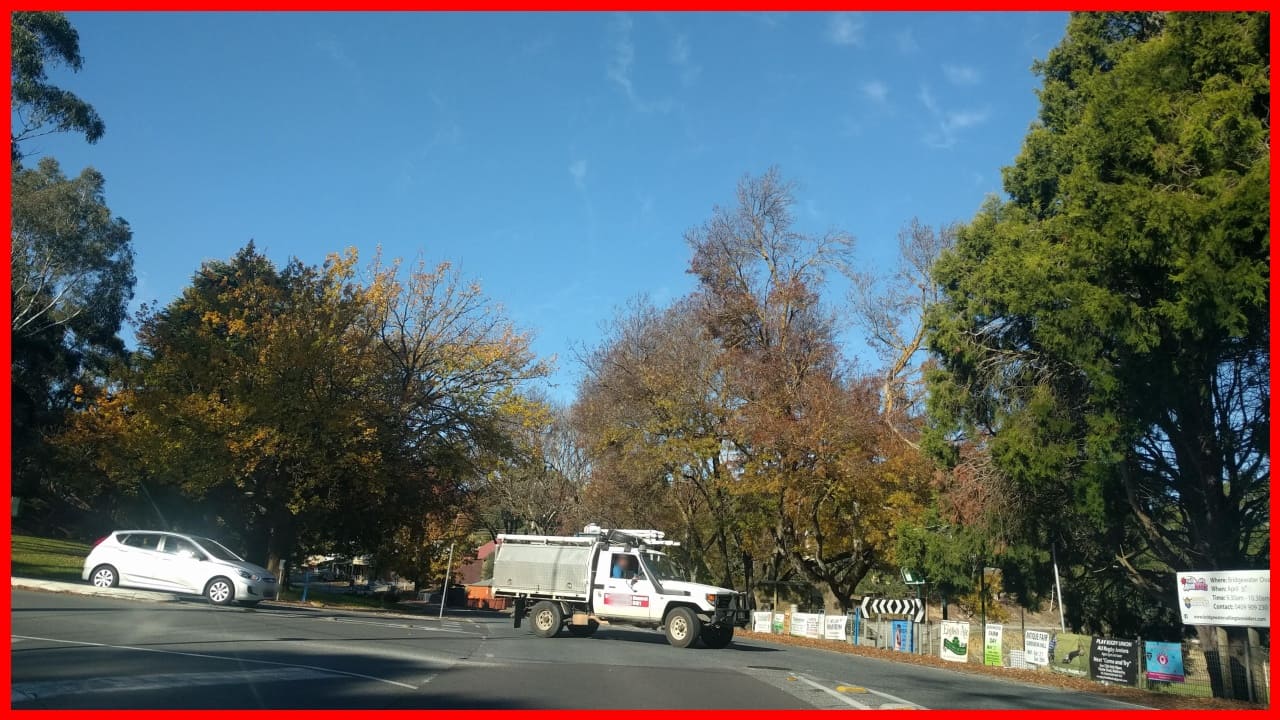}
    \end{subfigure}
    \begin{subfigure}[b]{\subfigwidth}
        \caption{\rankFour}
        \includegraphics[width=\textwidth]{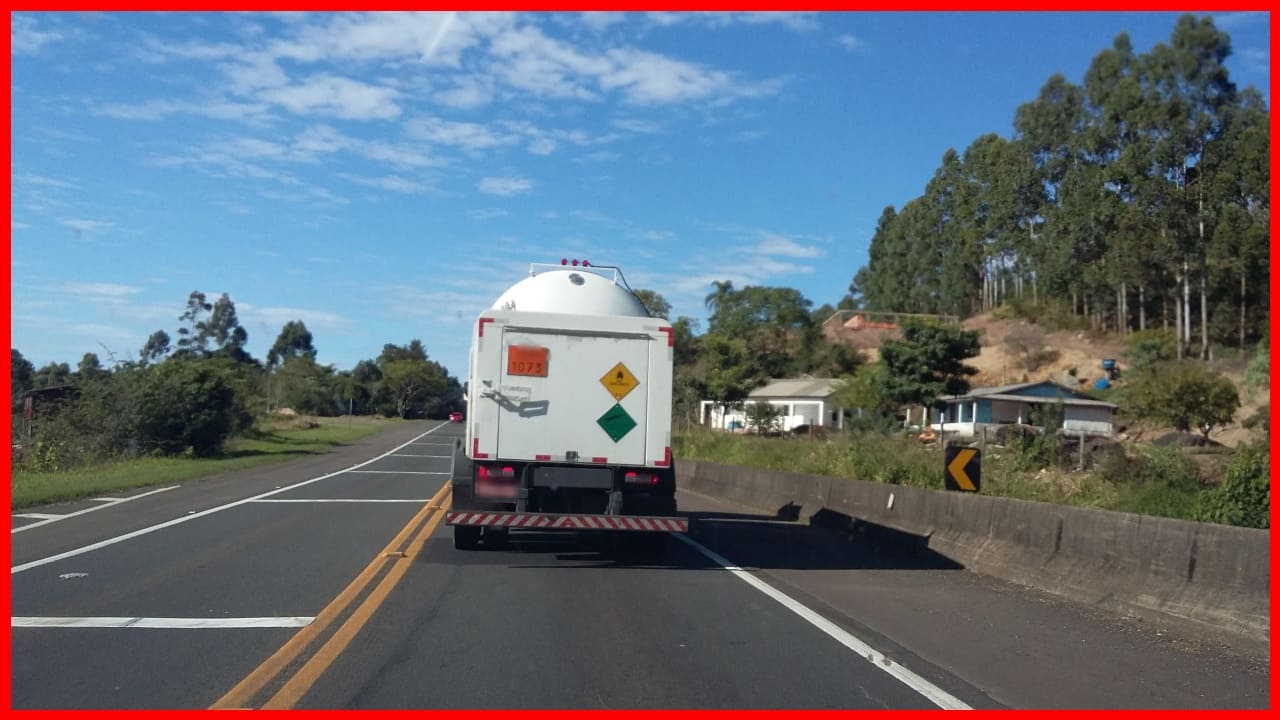}
    \end{subfigure}
    \begin{subfigure}[b]{\subfigwidth}
        \caption{\rankFive}
        \includegraphics[width=\textwidth]{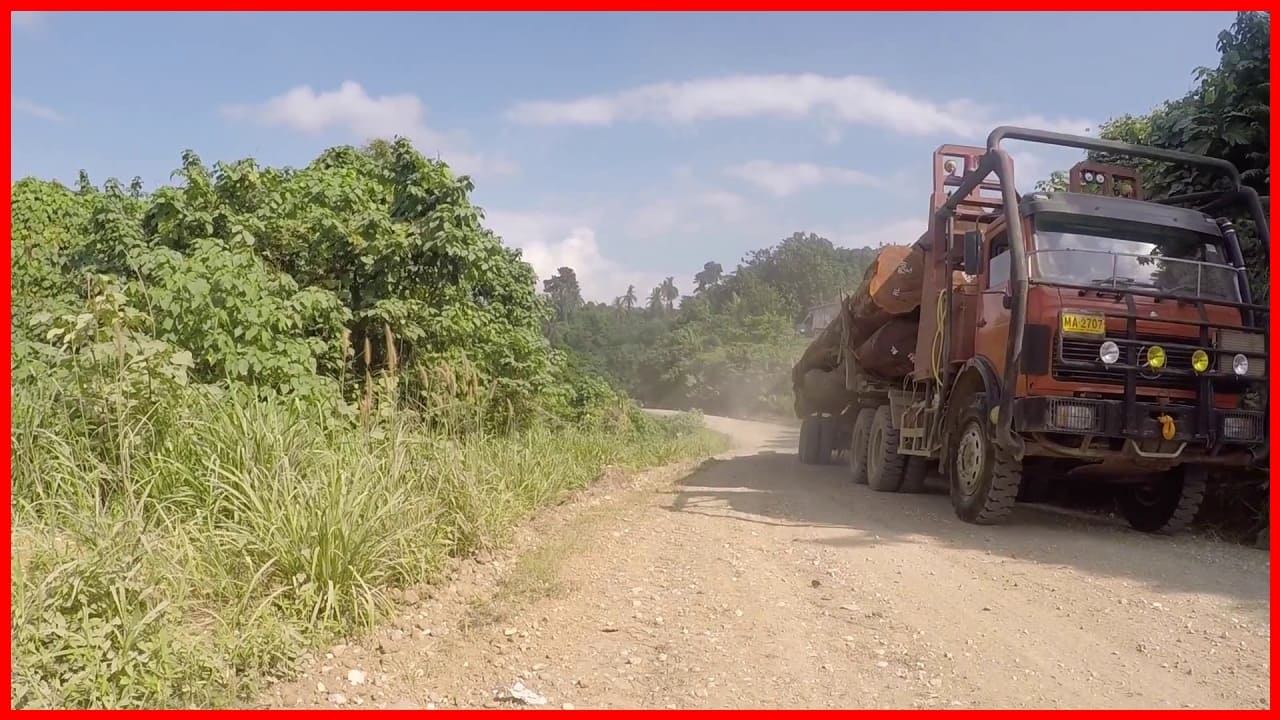}
    \end{subfigure} \\
    
    \rotatebox[origin=left]{90}{\hspace{0.17cm} \textbf{CLIP}} 
    \begin{subfigure}[b]{\subfigwidth}
        \includegraphics[width=\textwidth]{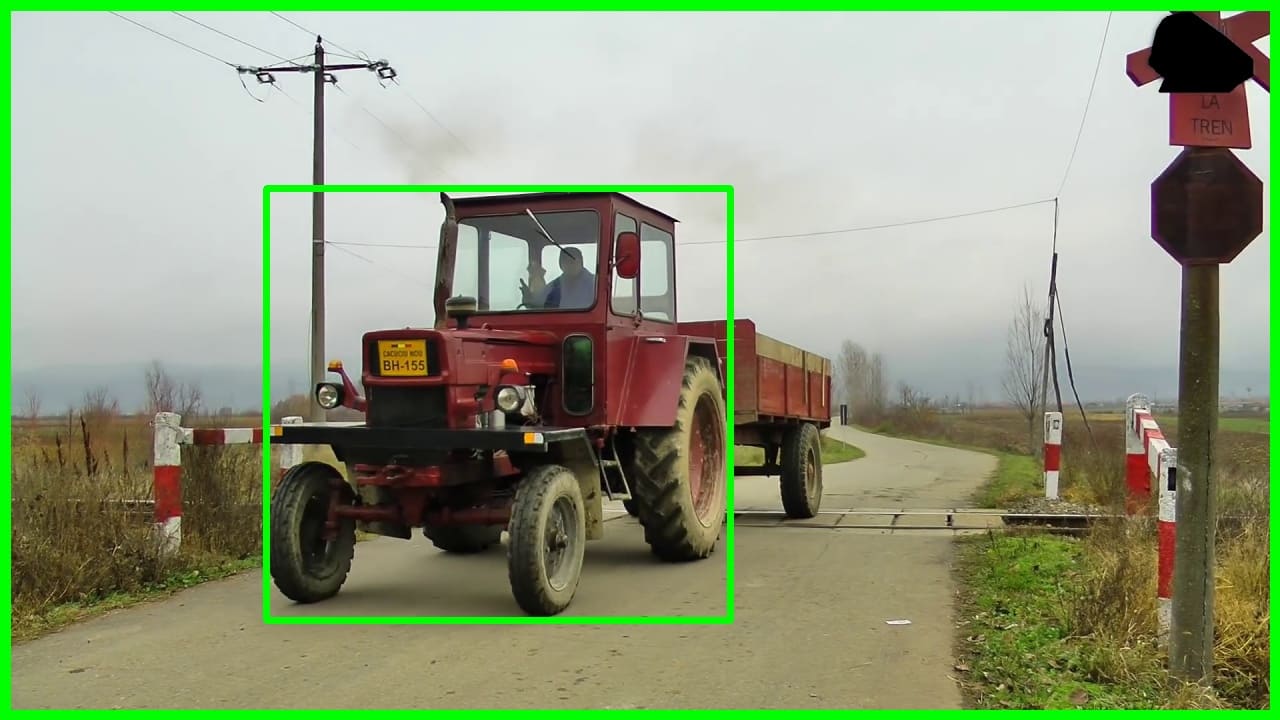}
    \end{subfigure}
    \begin{subfigure}[b]{\subfigwidth}
        \includegraphics[width=\textwidth]{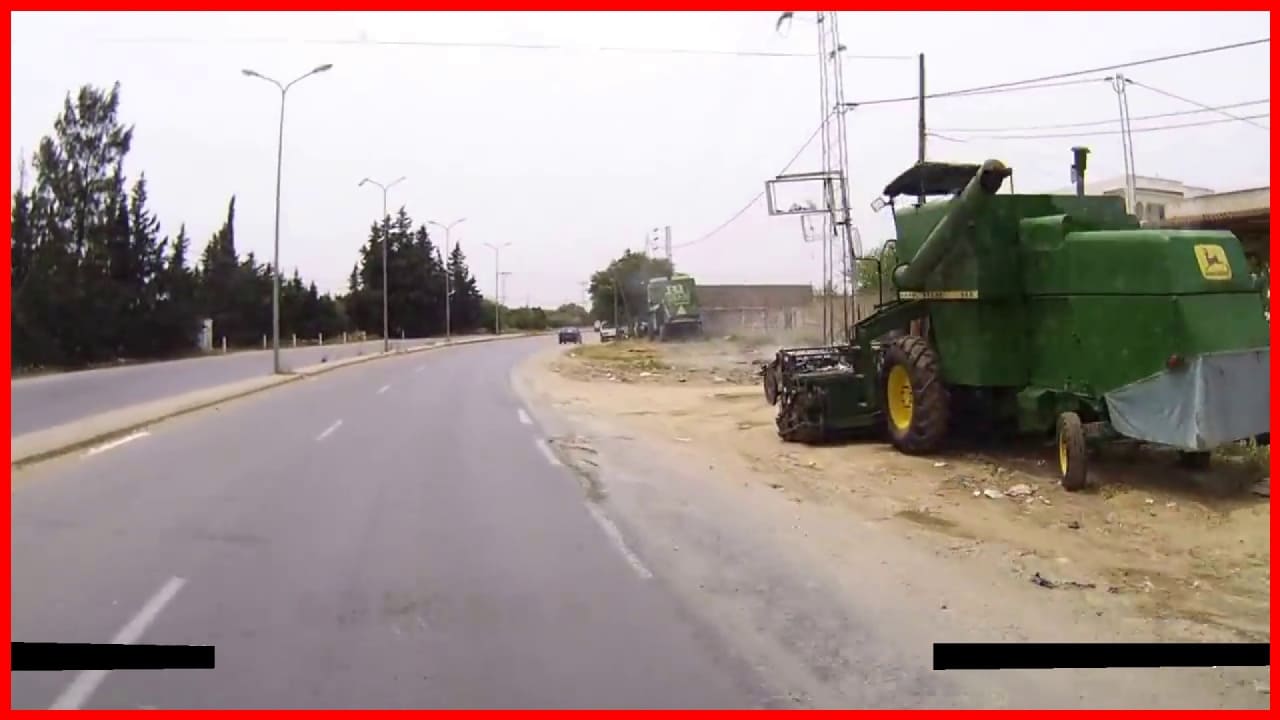}
    \end{subfigure}
    \begin{subfigure}[b]{\subfigwidth}
        \includegraphics[width=\textwidth]{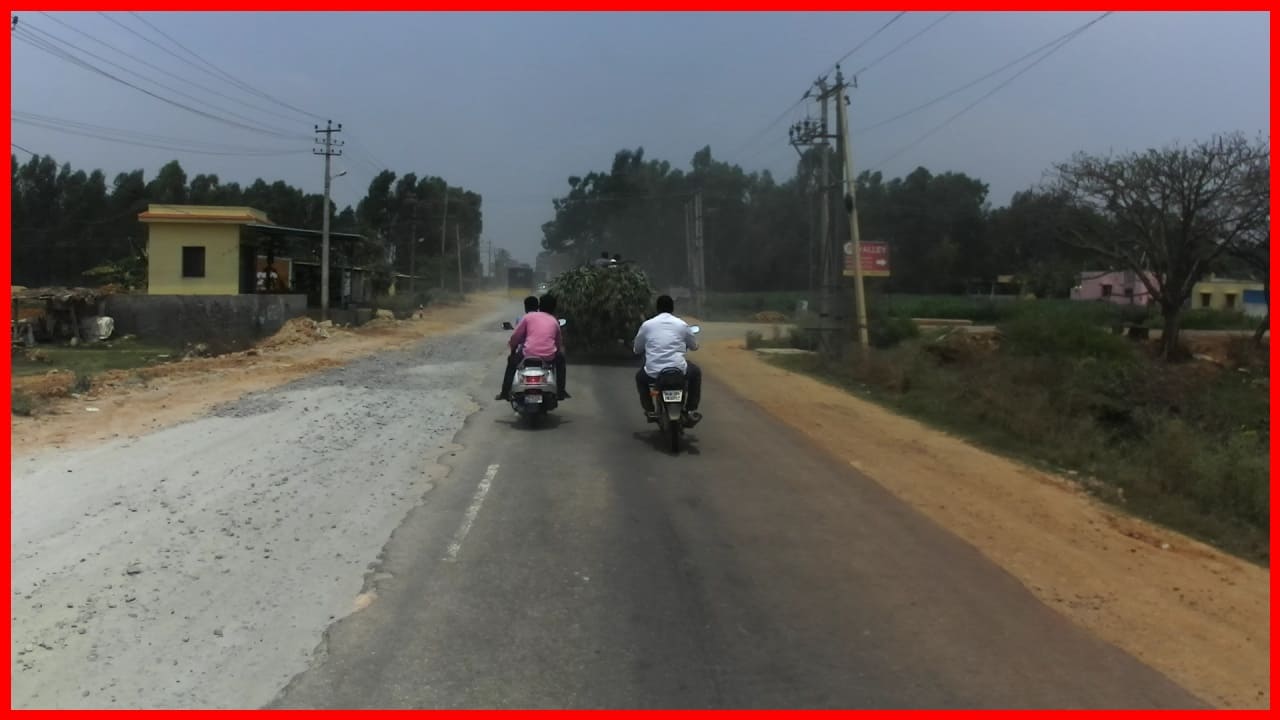}
    \end{subfigure}
    \begin{subfigure}[b]{\subfigwidth}
        \includegraphics[width=\textwidth]{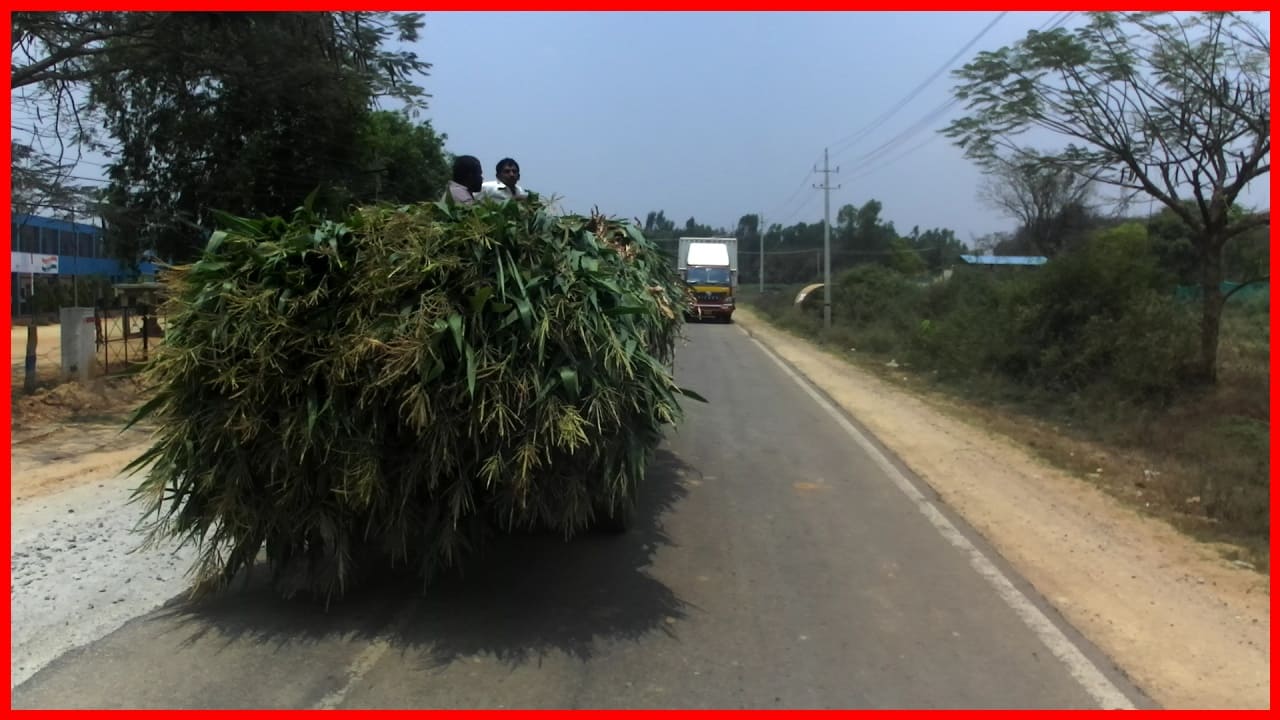}
    \end{subfigure}
    \begin{subfigure}[b]{\subfigwidth}
        \includegraphics[width=\textwidth]{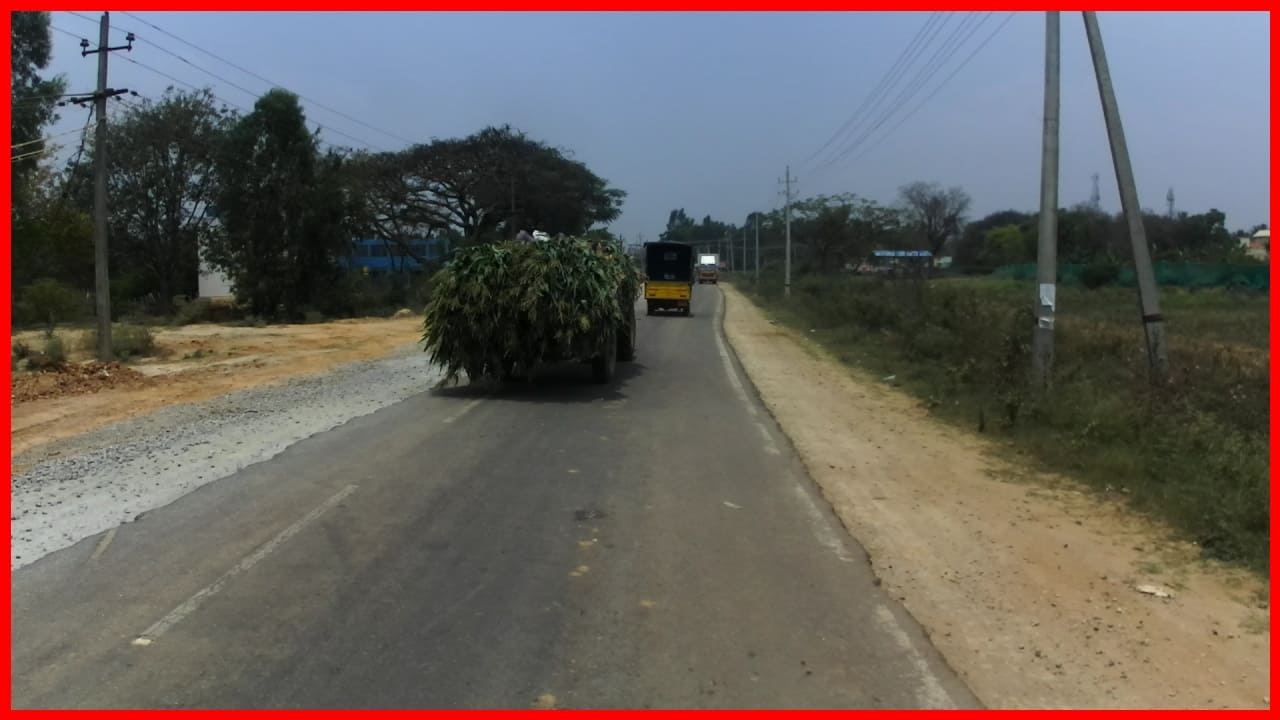}
    \end{subfigure} \\

    \rotatebox[origin=left]{90}{\hspace{0.07cm} \textbf{RADIO}} 
    \begin{subfigure}[b]{\subfigwidth}
        \includegraphics[width=\textwidth]{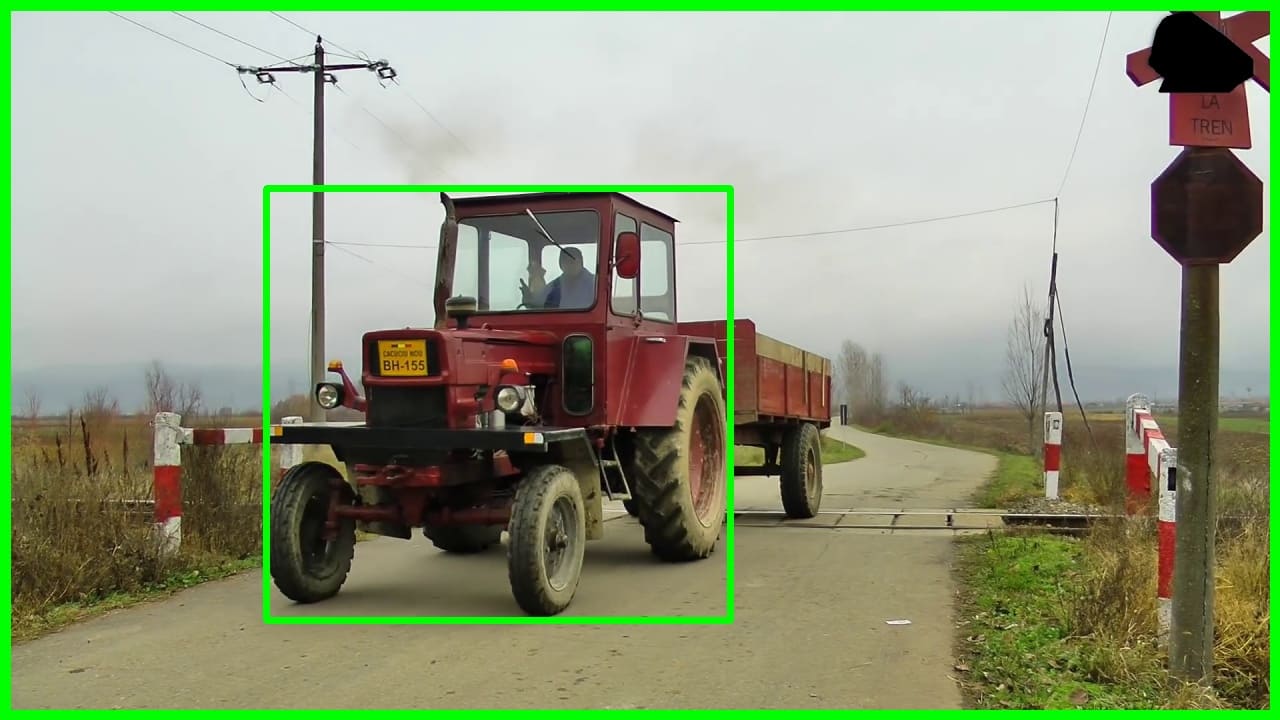}
    \end{subfigure}
    \begin{subfigure}[b]{\subfigwidth}
        \includegraphics[width=\textwidth]{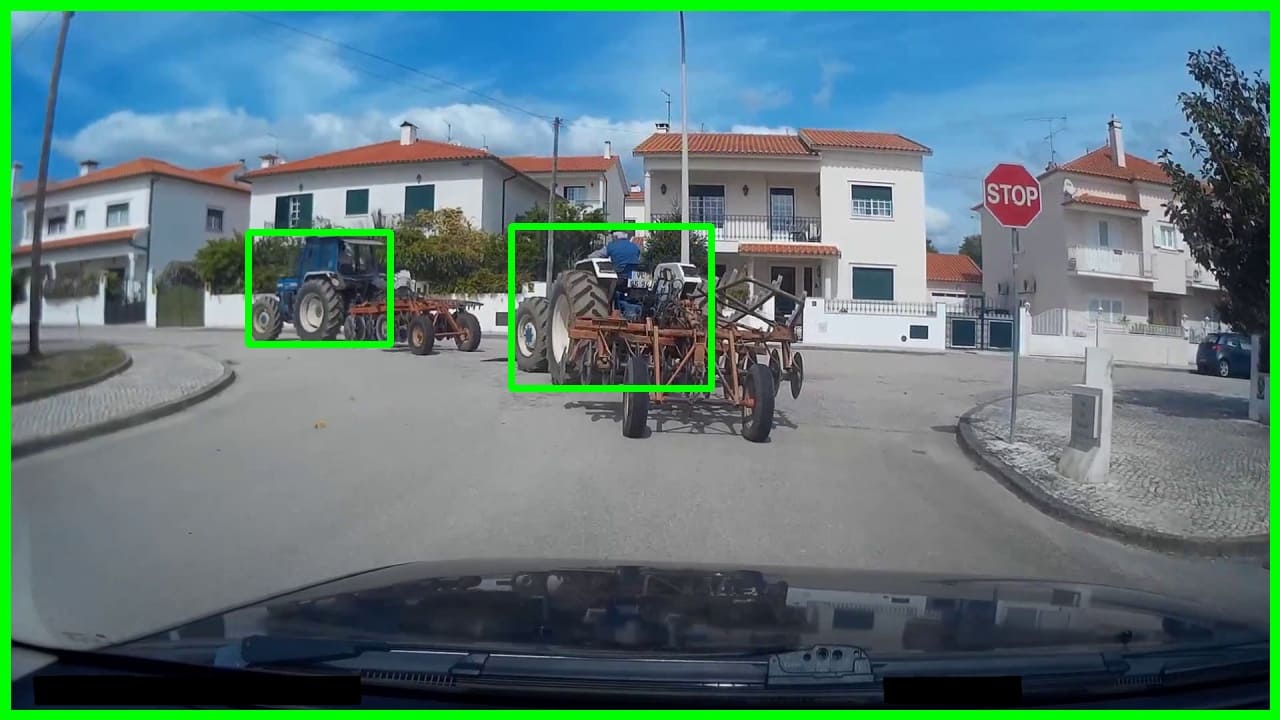}
    \end{subfigure}
    \begin{subfigure}[b]{\subfigwidth}
        \includegraphics[width=\textwidth]{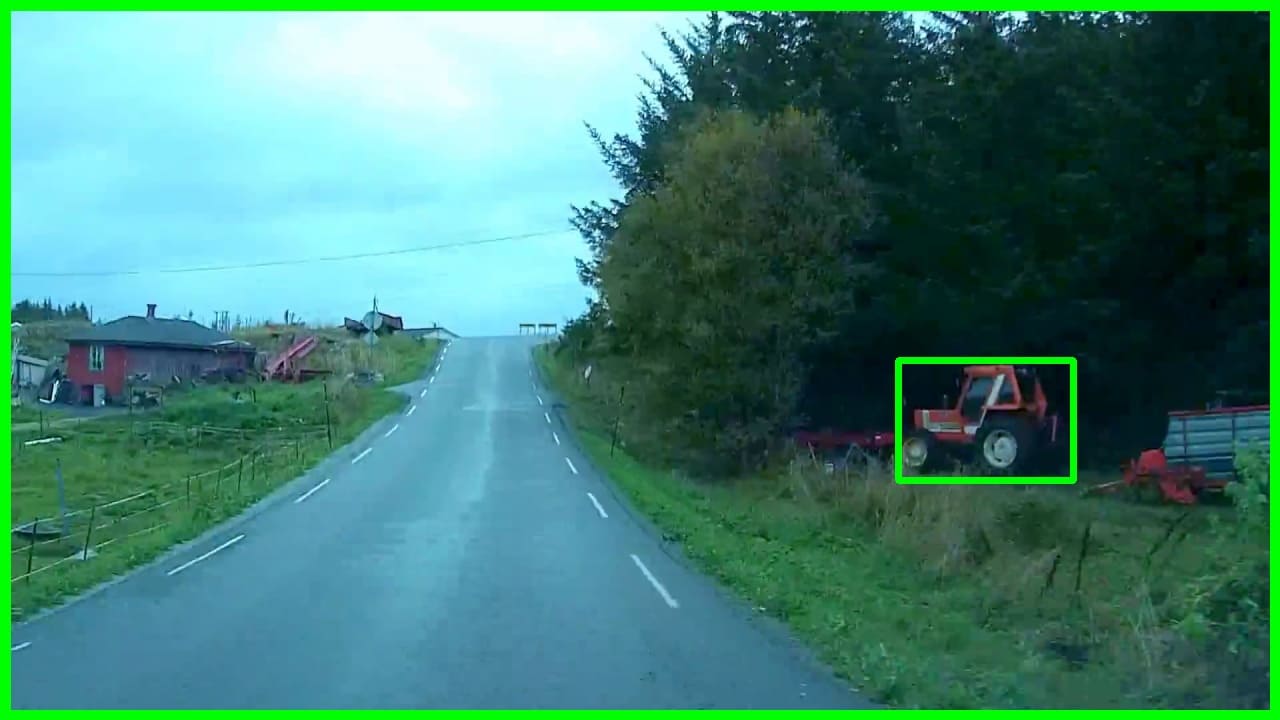}
    \end{subfigure}
    \begin{subfigure}[b]{\subfigwidth}
        \includegraphics[width=\textwidth]{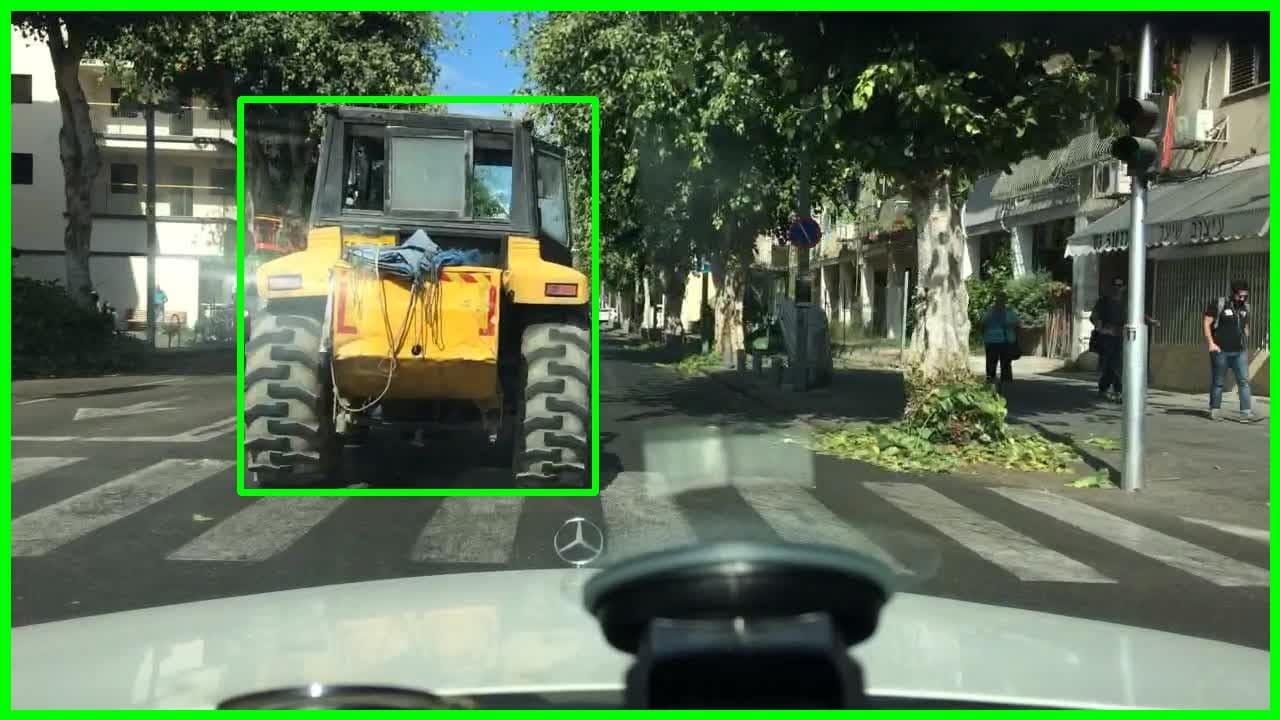}
    \end{subfigure}
    \begin{subfigure}[b]{\subfigwidth}
        \includegraphics[width=\textwidth]{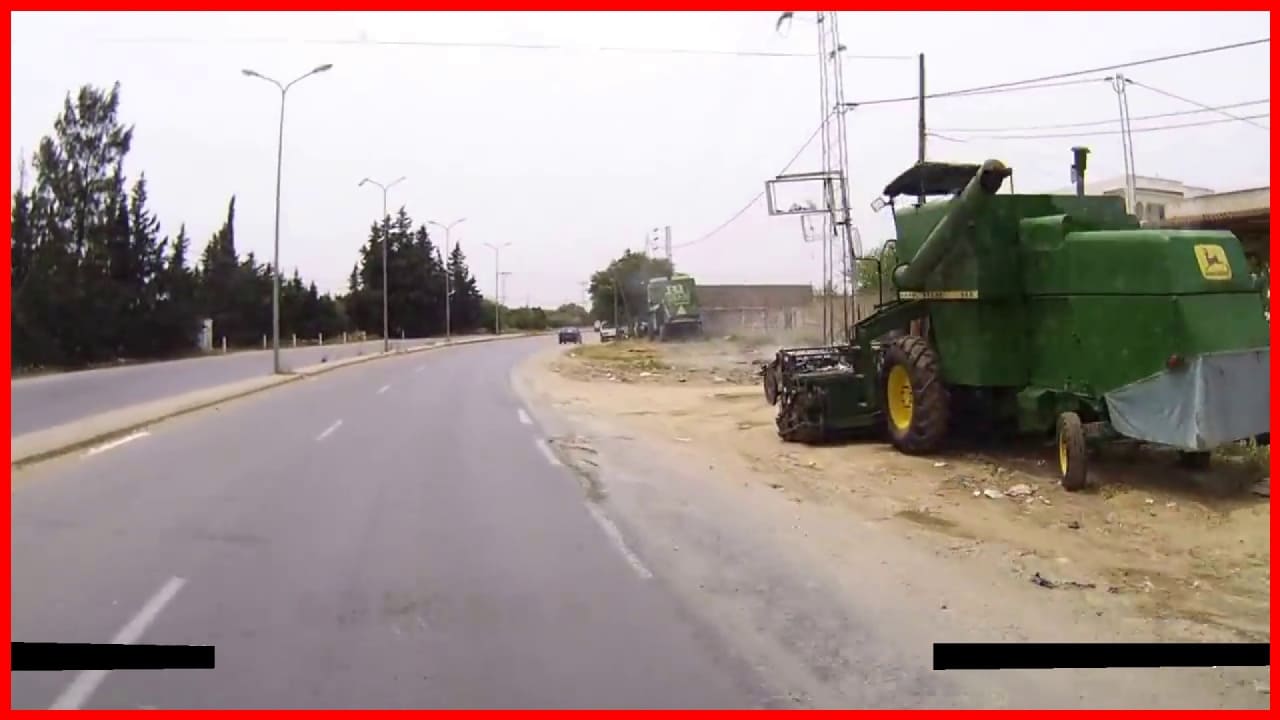}
    \end{subfigure} \\

    \rotatebox[origin=left]{90}{\hspace{0.15cm} \textbf{BLIP2}} 
    \begin{subfigure}[b]{\subfigwidth}
        \includegraphics[width=\textwidth]{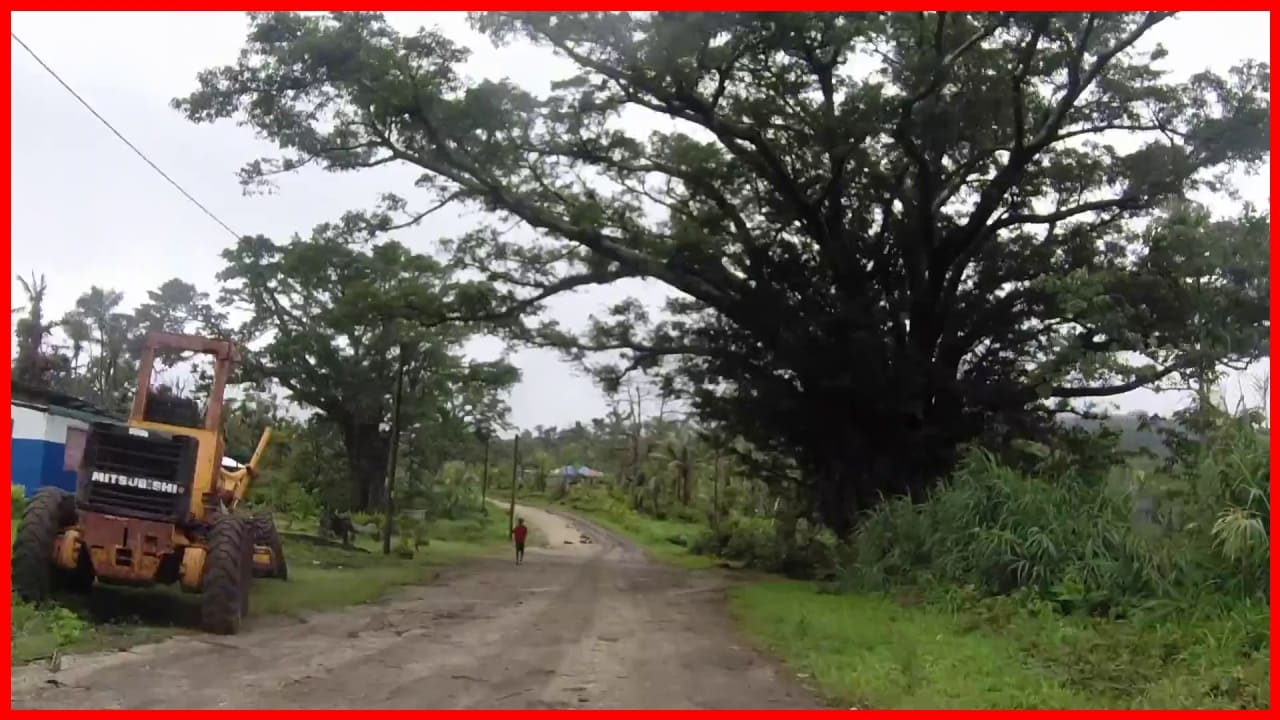}
    \end{subfigure}
    \begin{subfigure}[b]{\subfigwidth}
        \includegraphics[width=\textwidth]{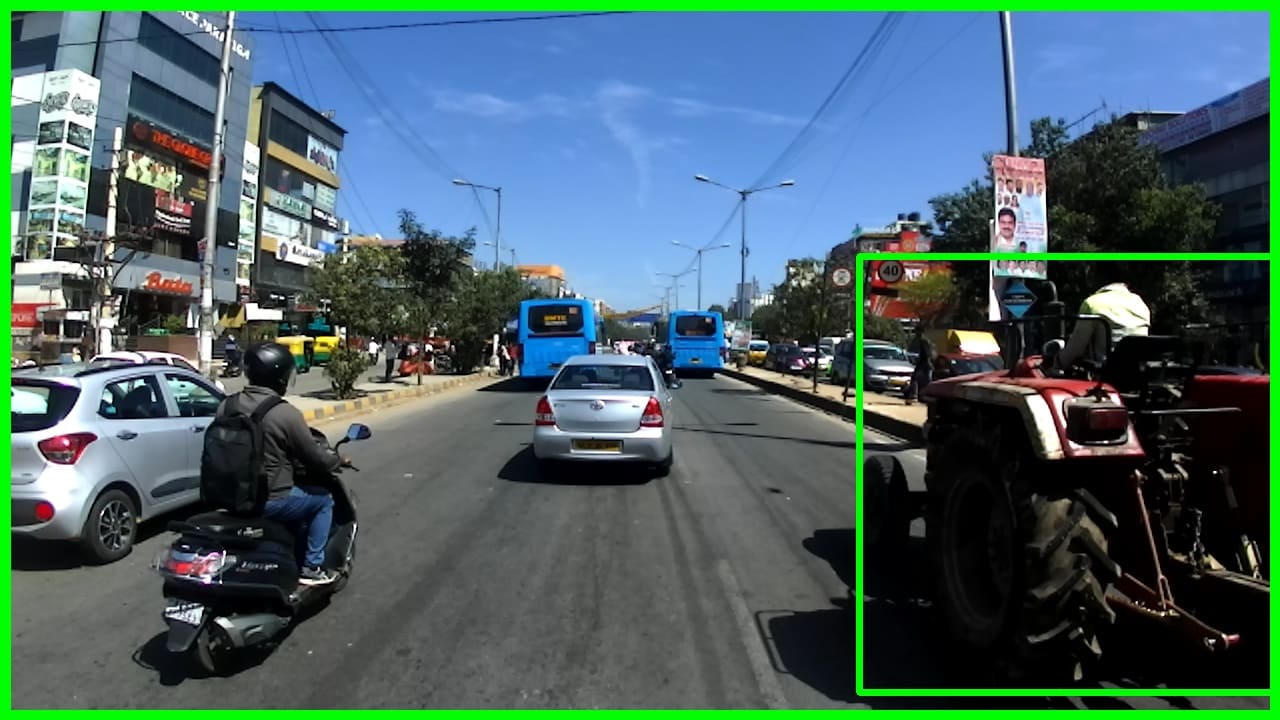}
    \end{subfigure}
    \begin{subfigure}[b]{\subfigwidth}
        \includegraphics[width=\textwidth]{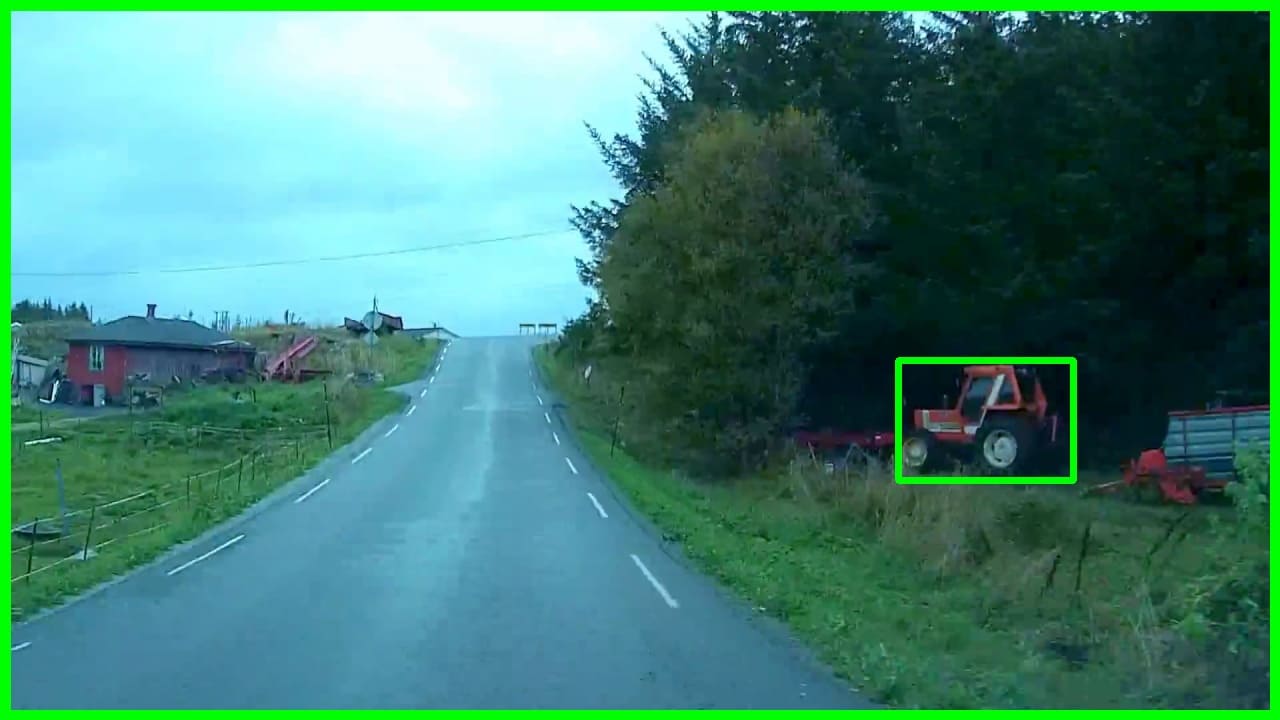}
    \end{subfigure}
    \begin{subfigure}[b]{\subfigwidth}
        \includegraphics[width=\textwidth]{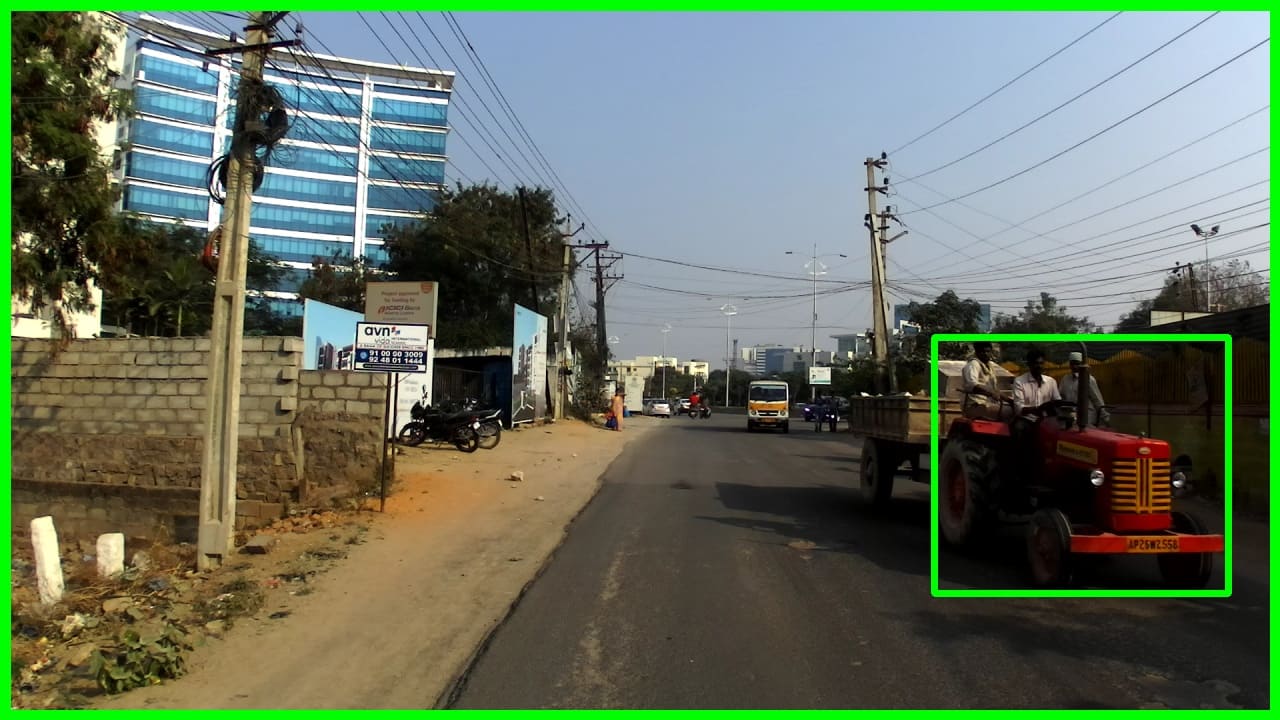}
    \end{subfigure}
    \begin{subfigure}[b]{\subfigwidth}
        \includegraphics[width=\textwidth]{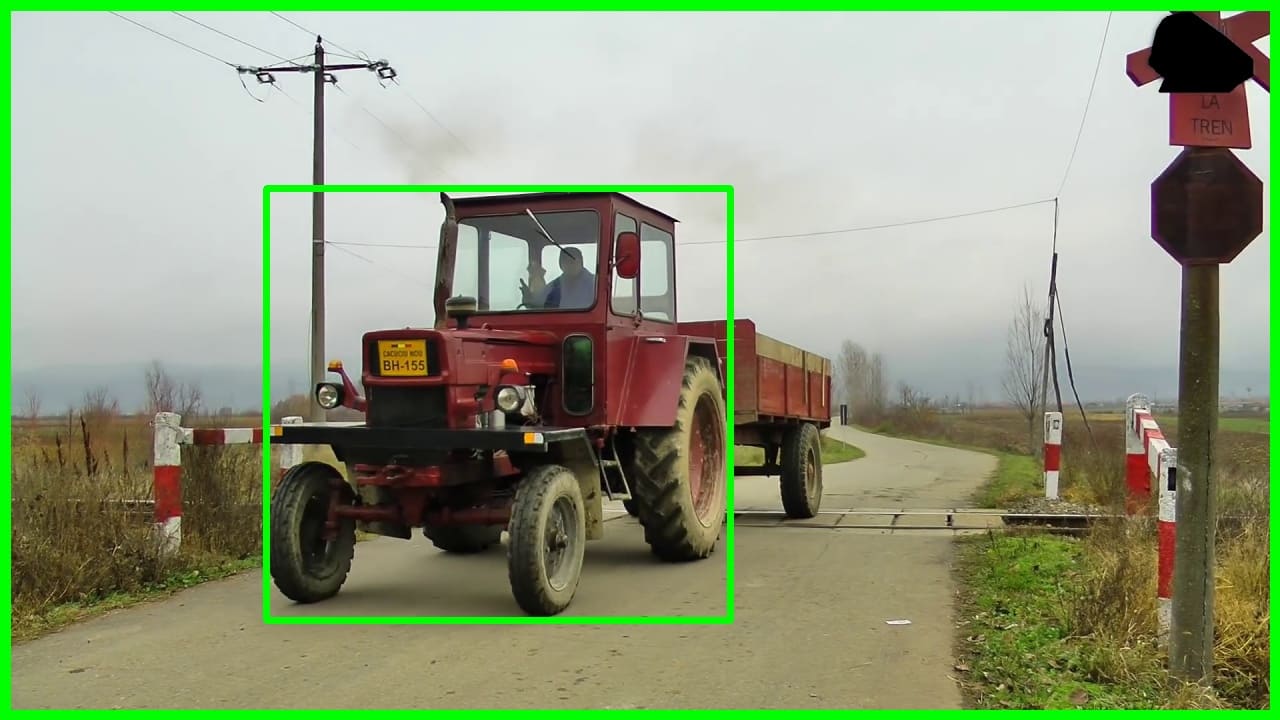}
    \end{subfigure} \\

    \rotatebox[origin=left]{90}{\hspace{-0.29cm} \textbf{METACLIP2}} 
    \begin{subfigure}[b]{\subfigwidth}
        \includegraphics[width=\textwidth]{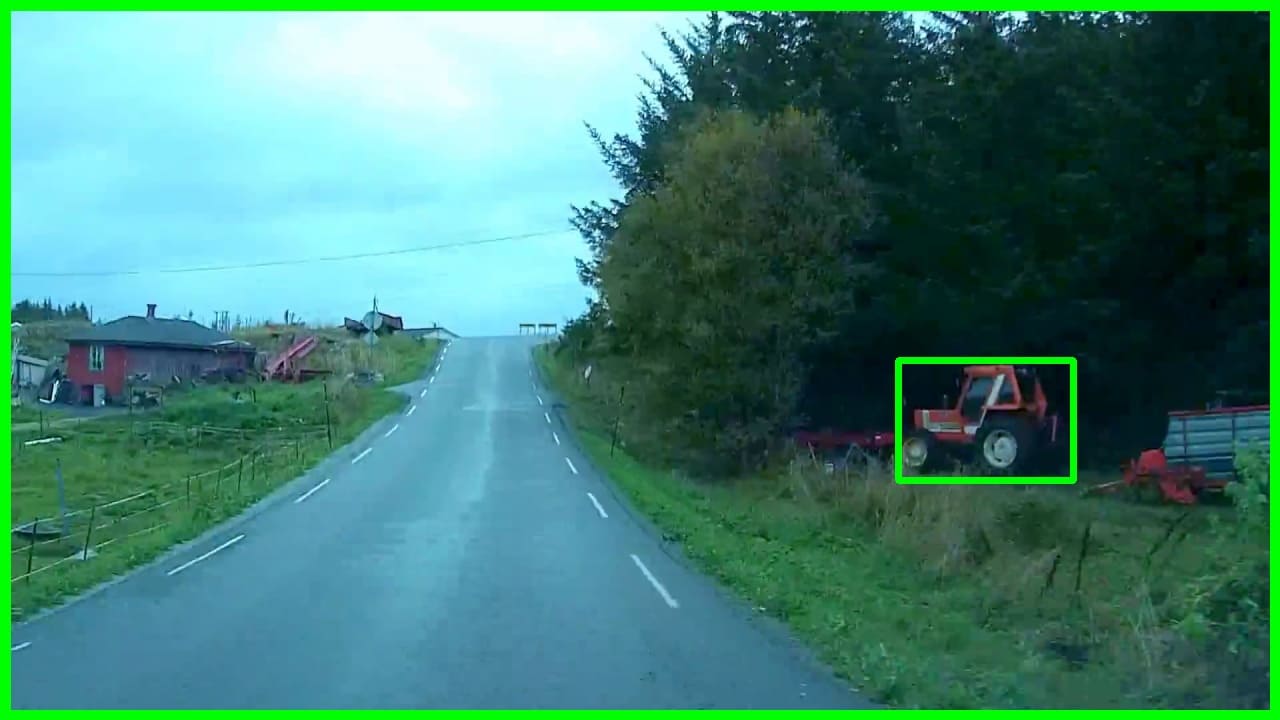}
    \end{subfigure}
    \begin{subfigure}[b]{\subfigwidth}
        \includegraphics[width=\textwidth]{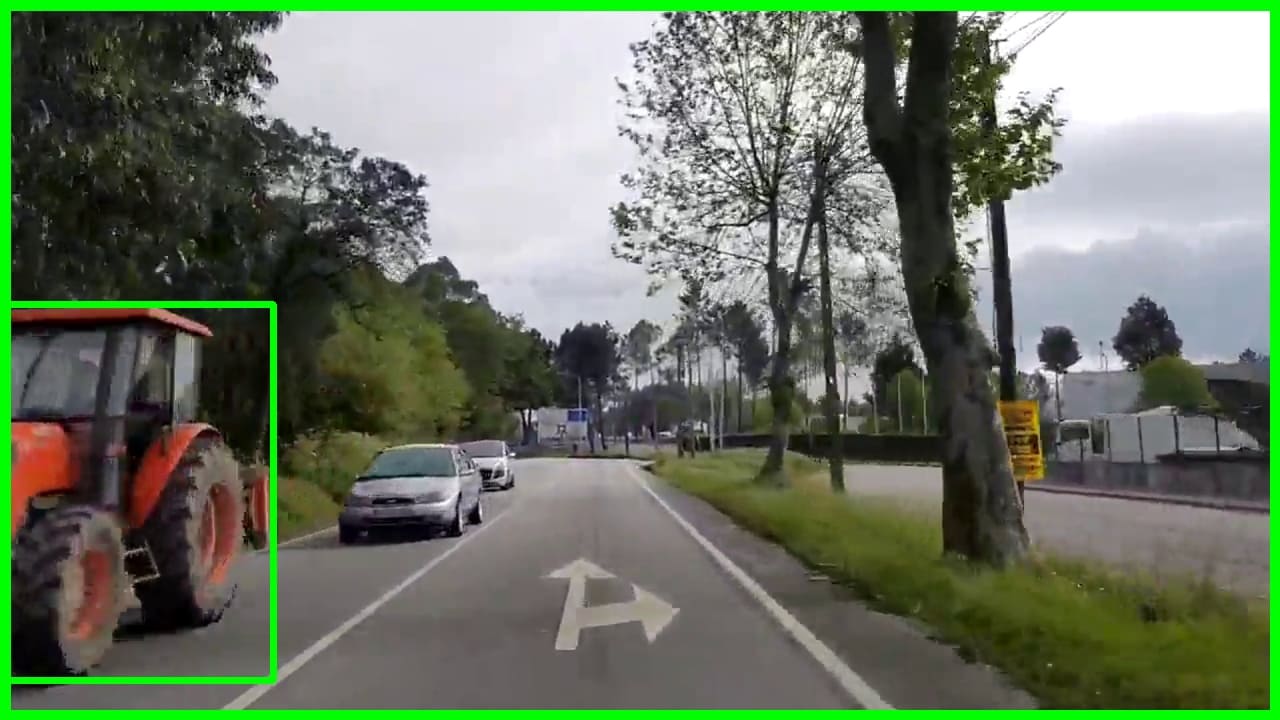}
    \end{subfigure}
    \begin{subfigure}[b]{\subfigwidth}
        \includegraphics[width=\textwidth]{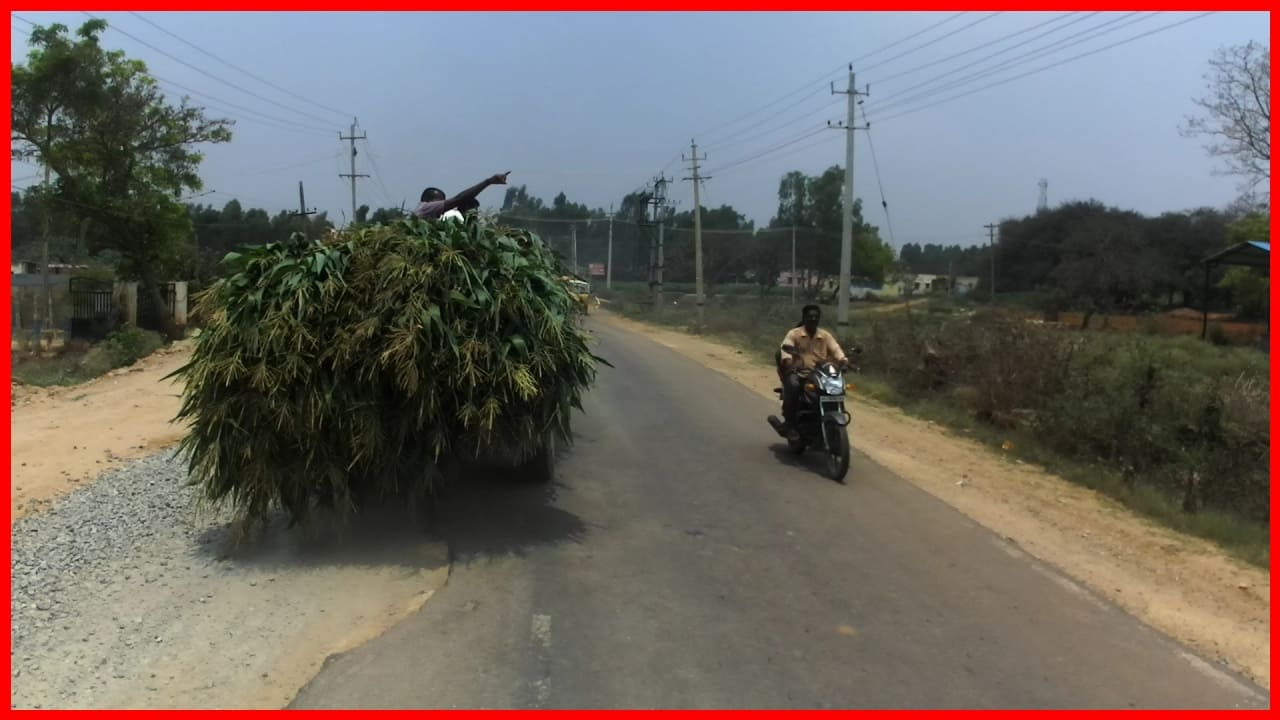}
    \end{subfigure}
    \begin{subfigure}[b]{\subfigwidth}
        \includegraphics[width=\textwidth]{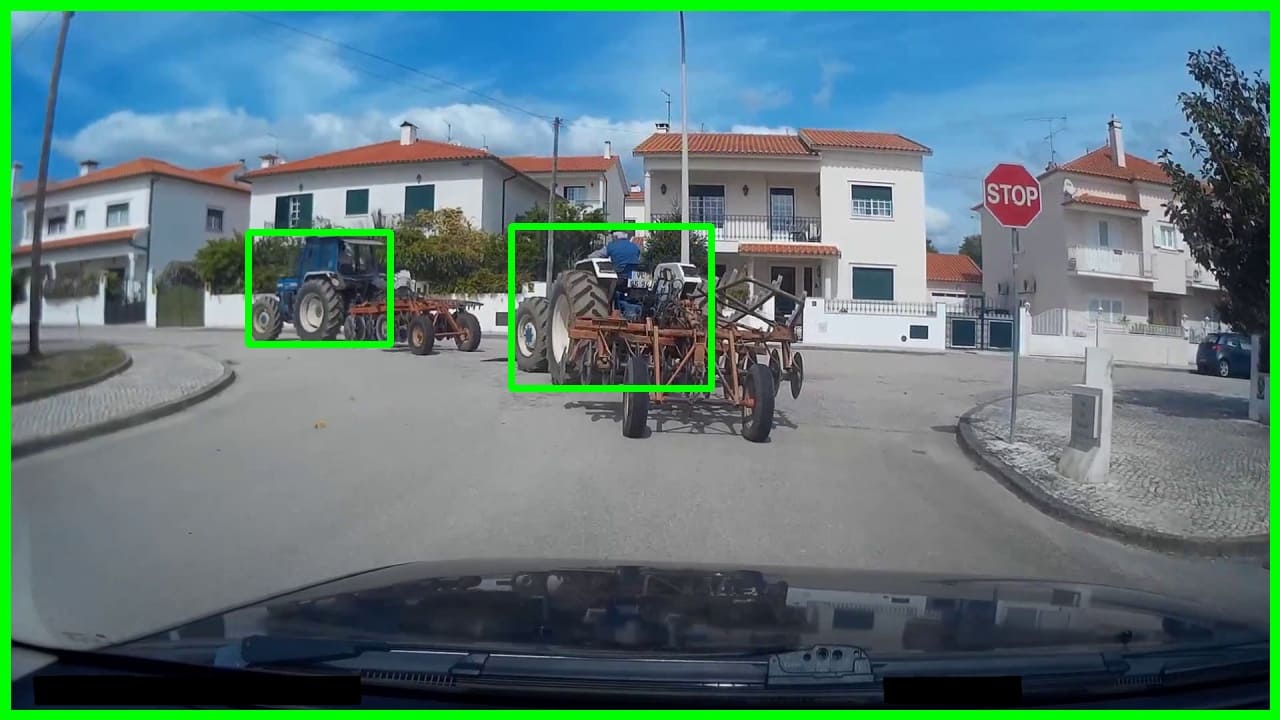}
    \end{subfigure}
    \begin{subfigure}[b]{\subfigwidth}
        \includegraphics[width=\textwidth]{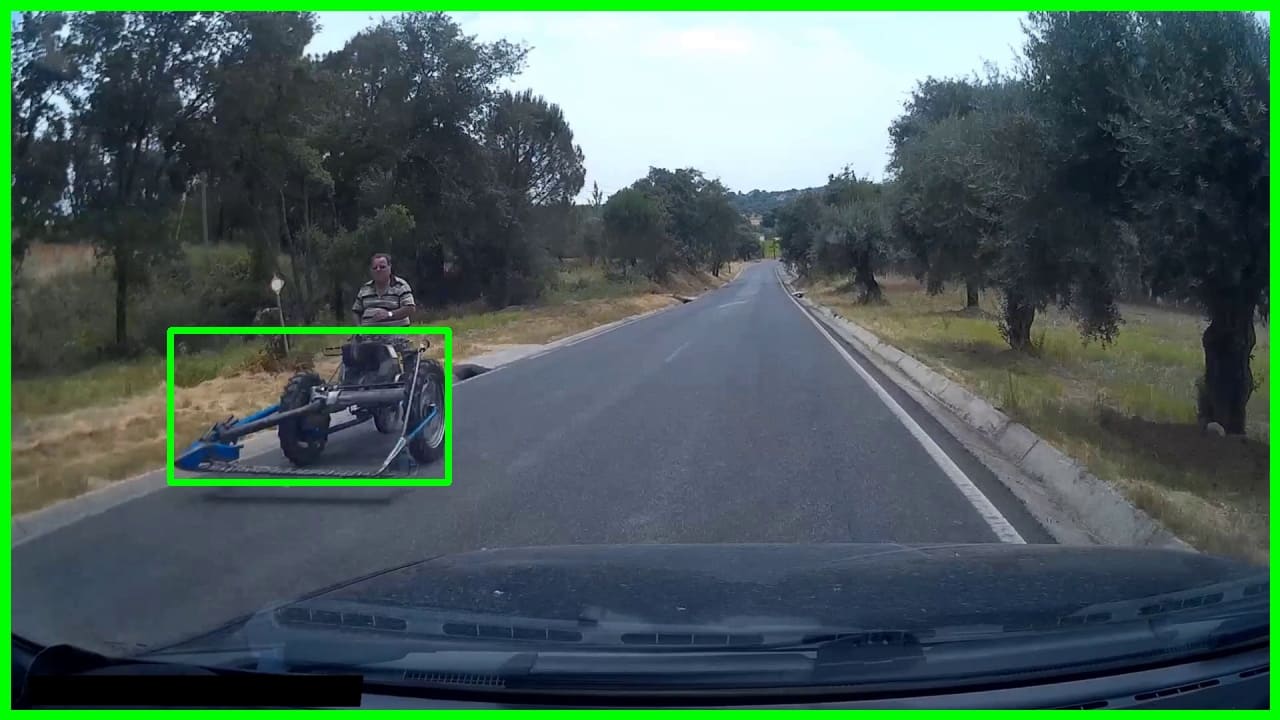}
    \end{subfigure} \\

    \rotatebox[origin=left]{90}{\hspace{0.02cm} \textbf{SIGLIP2}} 
    \begin{subfigure}[b]{\subfigwidth}
        \includegraphics[width=\textwidth]{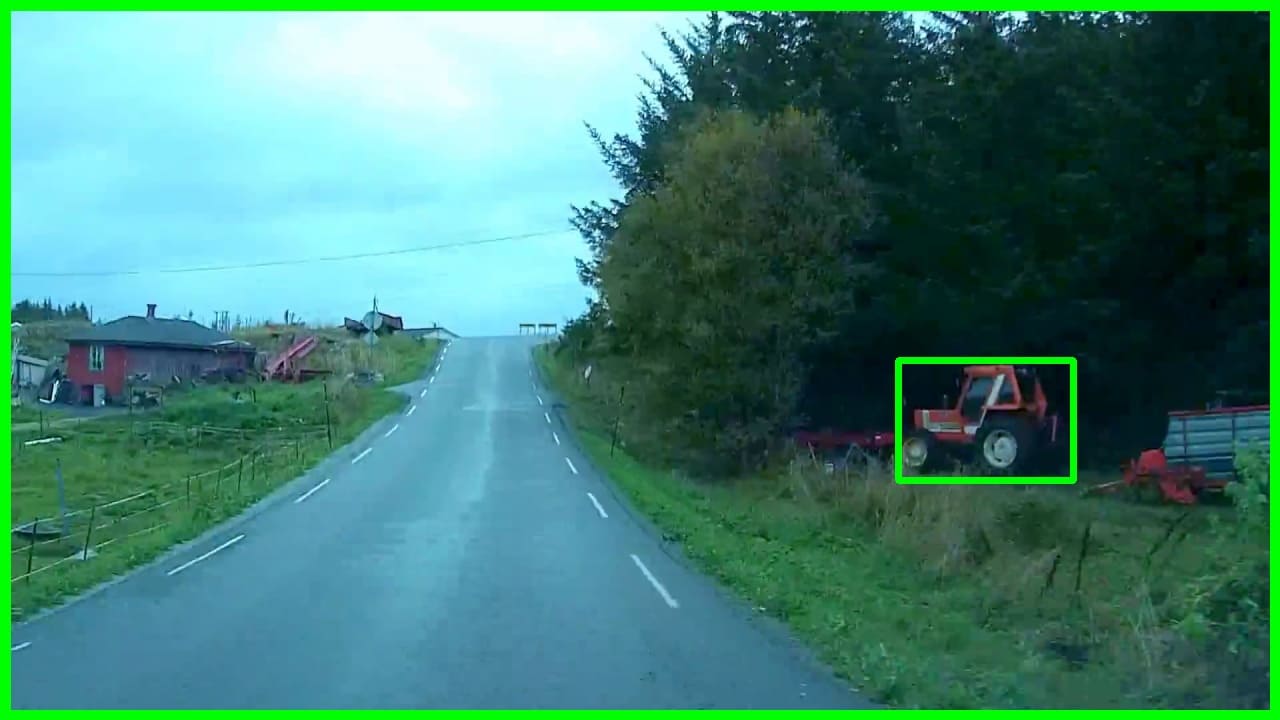}
    \end{subfigure}
    \begin{subfigure}[b]{\subfigwidth}
        \includegraphics[width=\textwidth]{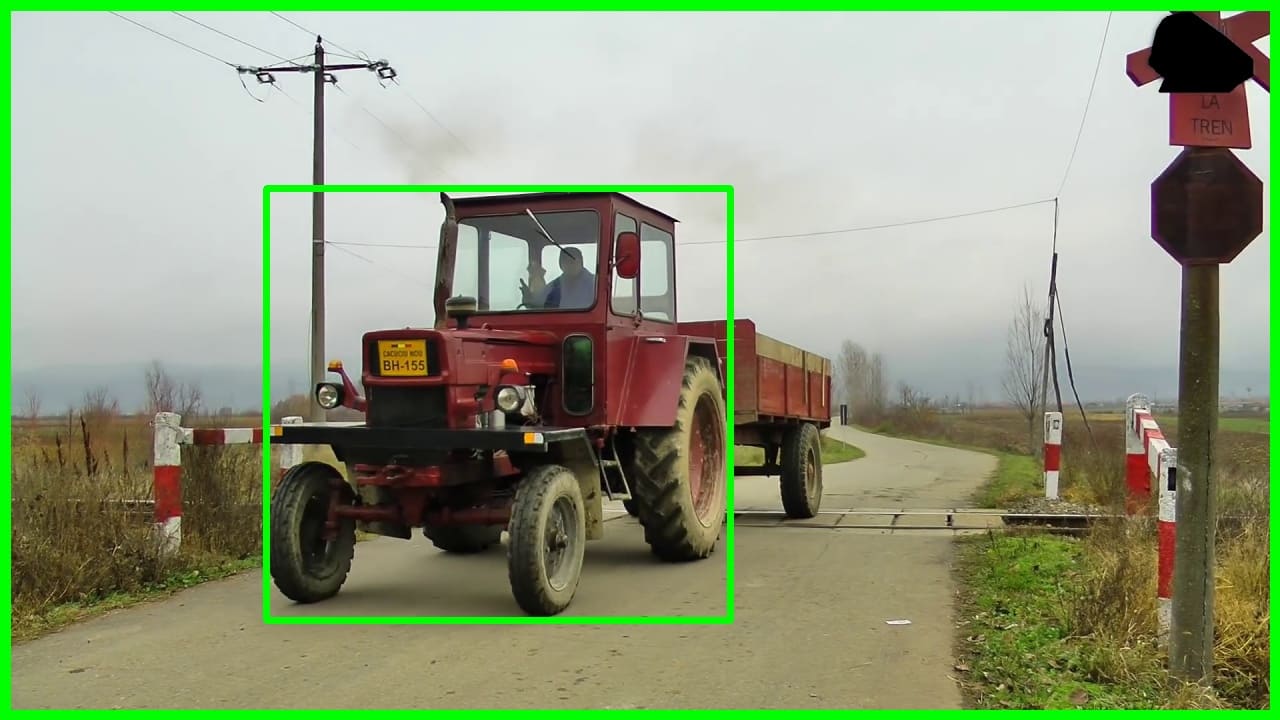}
    \end{subfigure}
    \begin{subfigure}[b]{\subfigwidth}
        \includegraphics[width=\textwidth]{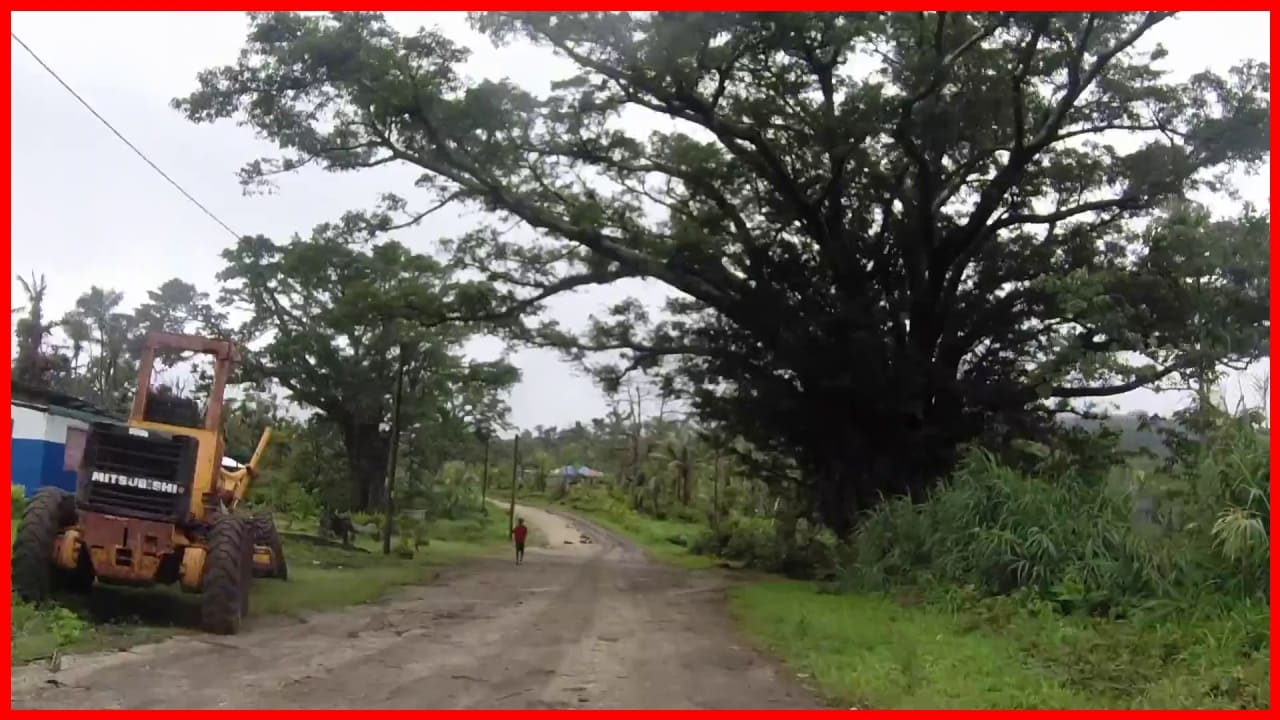}
    \end{subfigure}
    \begin{subfigure}[b]{\subfigwidth}
        \includegraphics[width=\textwidth]{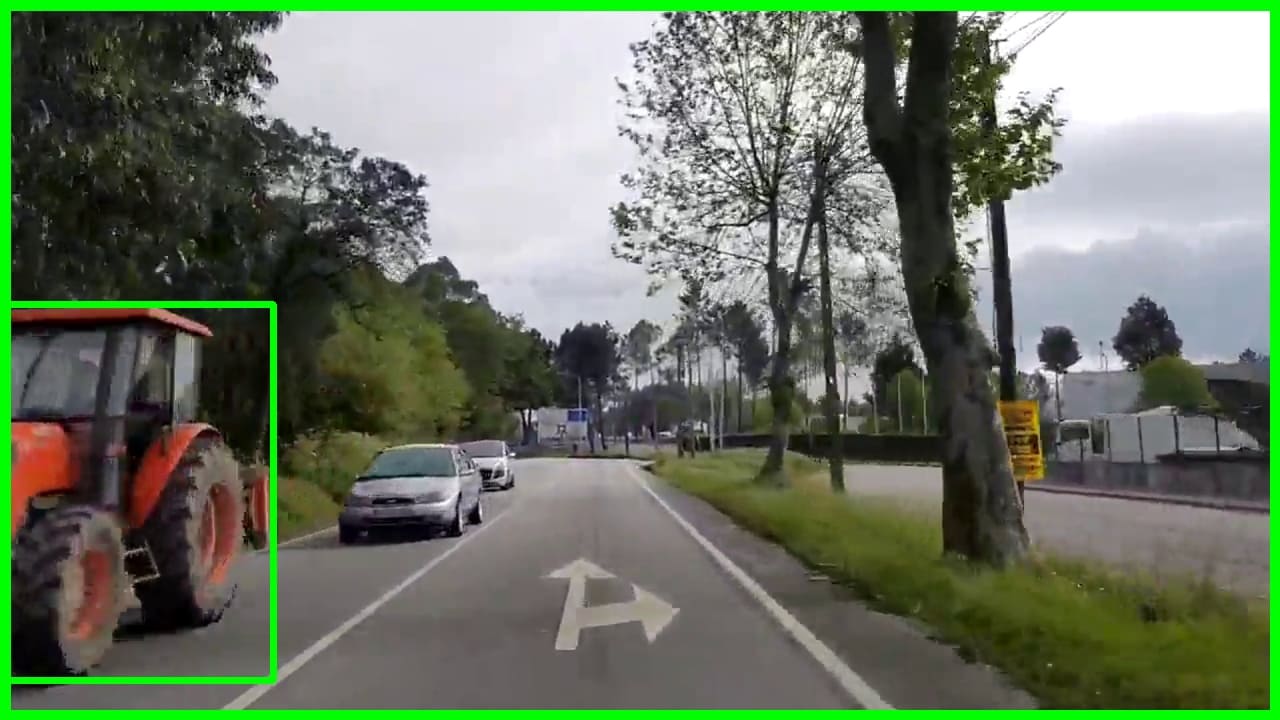}
    \end{subfigure}
    \begin{subfigure}[b]{\subfigwidth}
        \includegraphics[width=\textwidth]{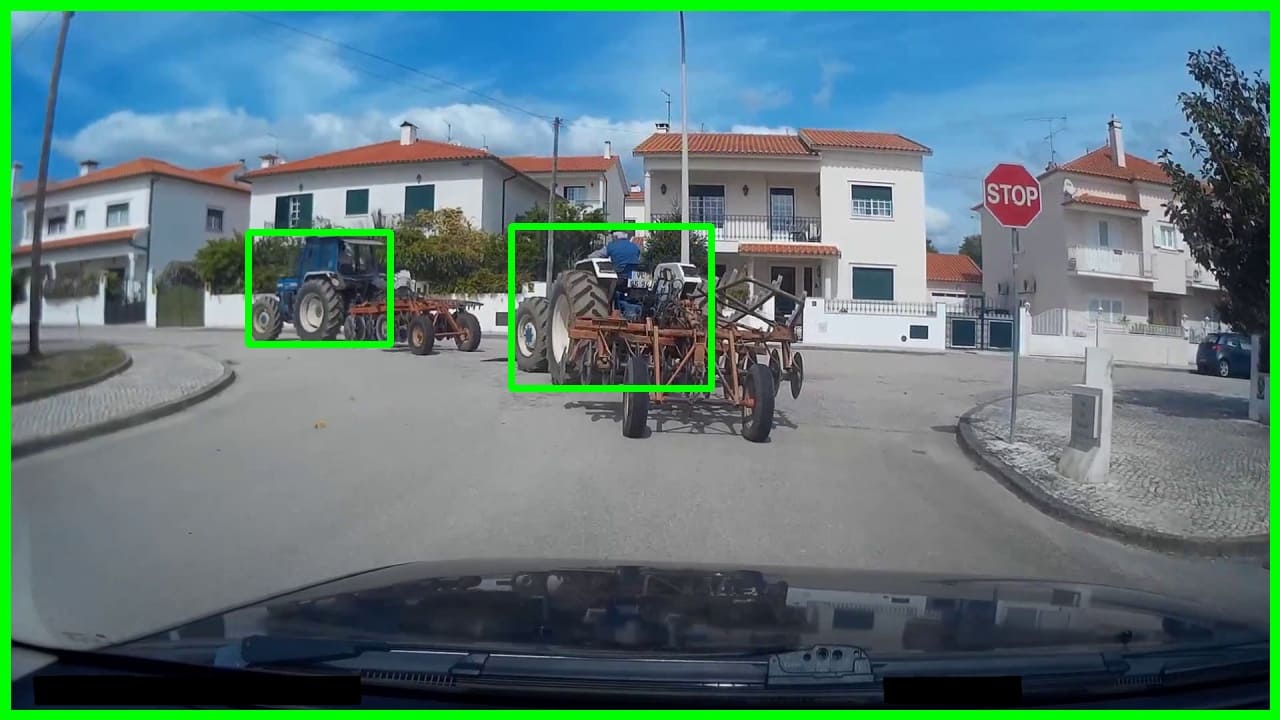}
    \end{subfigure} \\

    \rotatebox[origin=left]{90}{\hspace{0.04cm} \textbf{NACLIP}} 
    \begin{subfigure}[b]{\subfigwidth}
        \includegraphics[width=\textwidth]{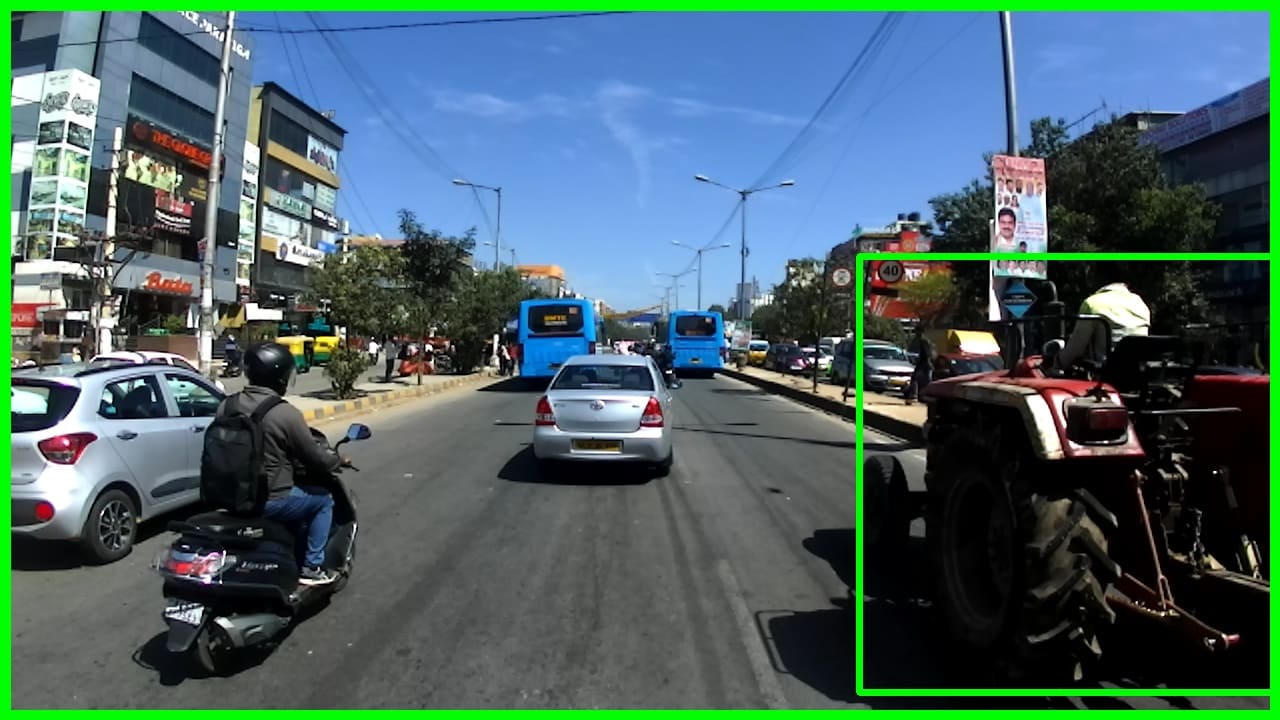}
    \end{subfigure}
    \begin{subfigure}[b]{\subfigwidth}
        \includegraphics[width=\textwidth]{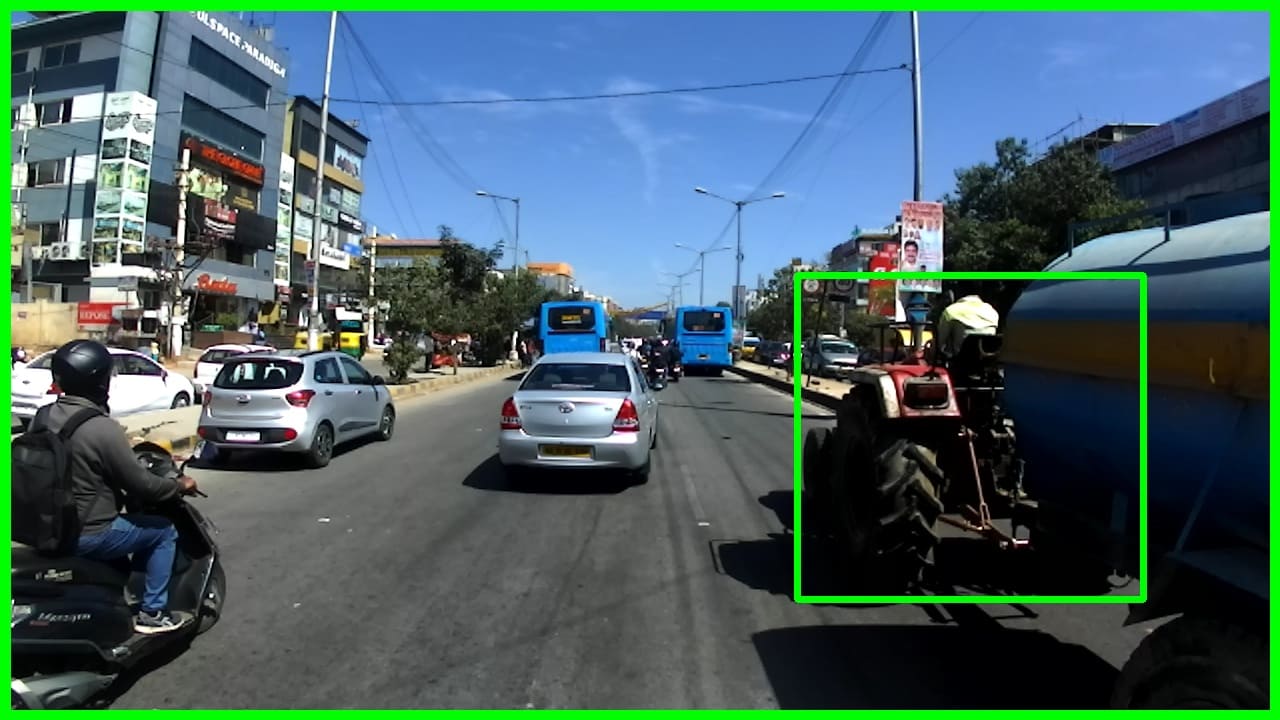}
    \end{subfigure}
    \begin{subfigure}[b]{\subfigwidth}
        \includegraphics[width=\textwidth]{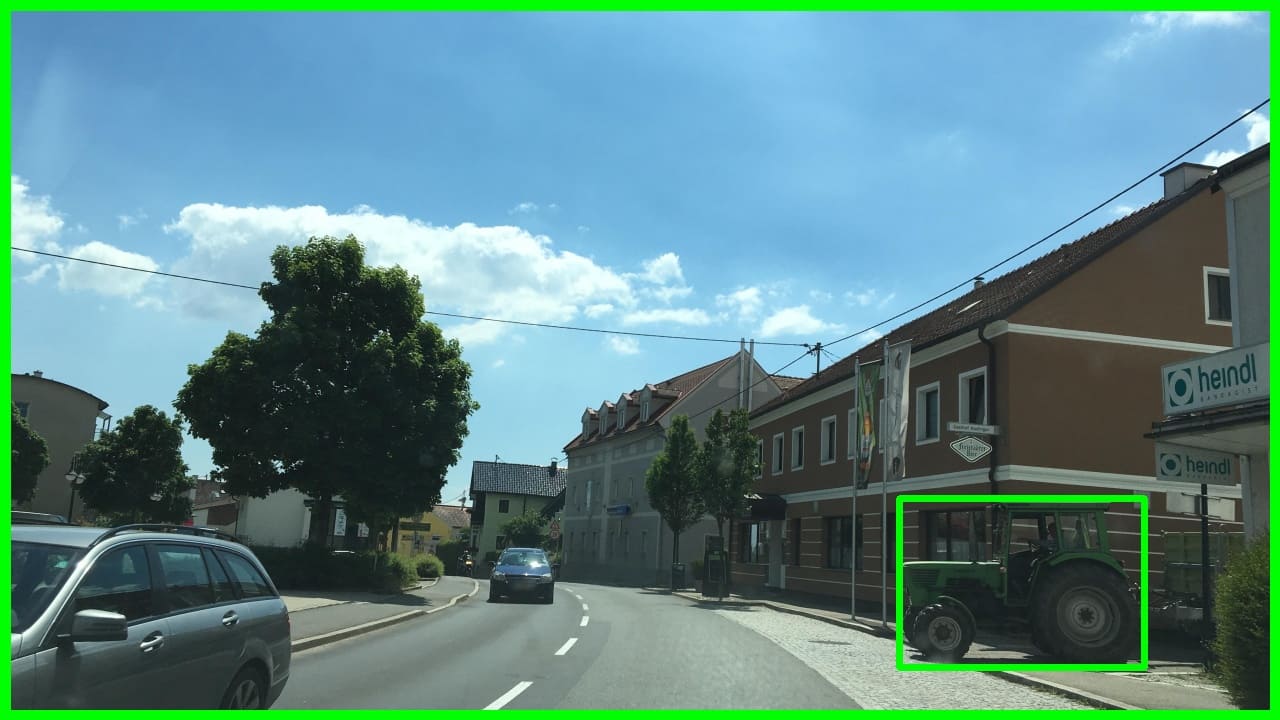}
    \end{subfigure}
    \begin{subfigure}[b]{\subfigwidth}
        \includegraphics[width=\textwidth]{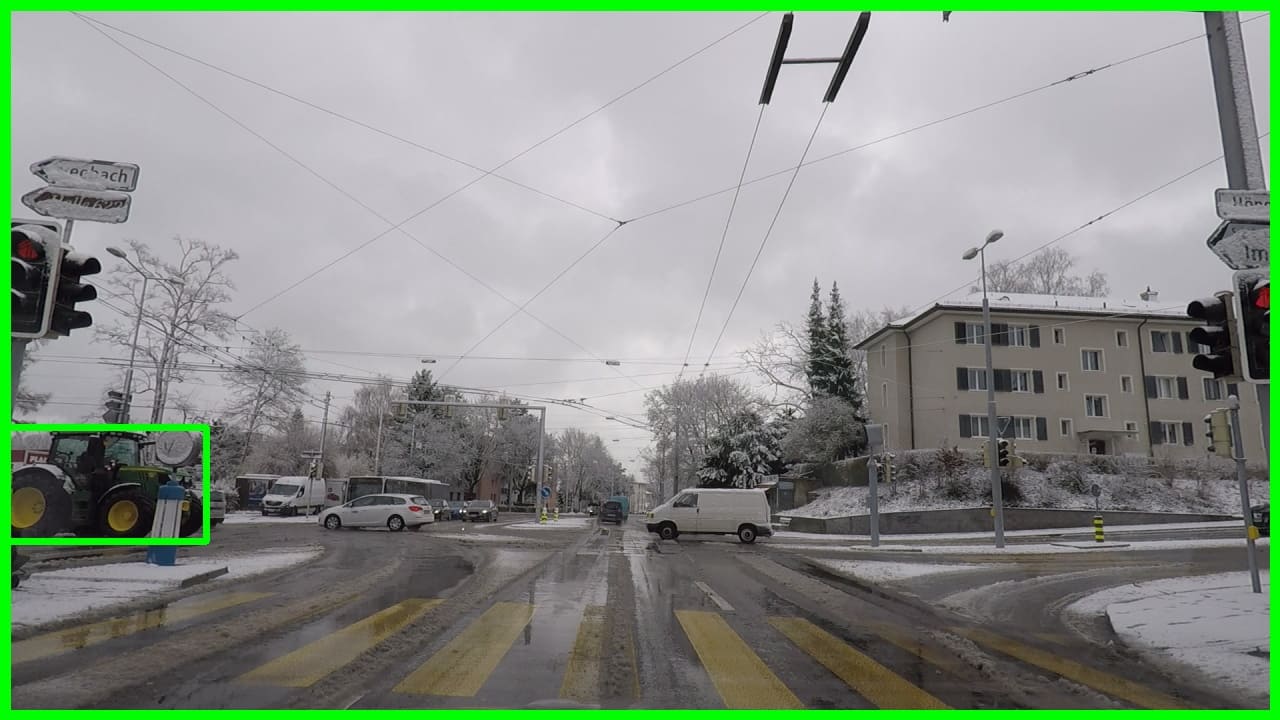}
    \end{subfigure}
    \begin{subfigure}[b]{\subfigwidth}
        \includegraphics[width=\textwidth]{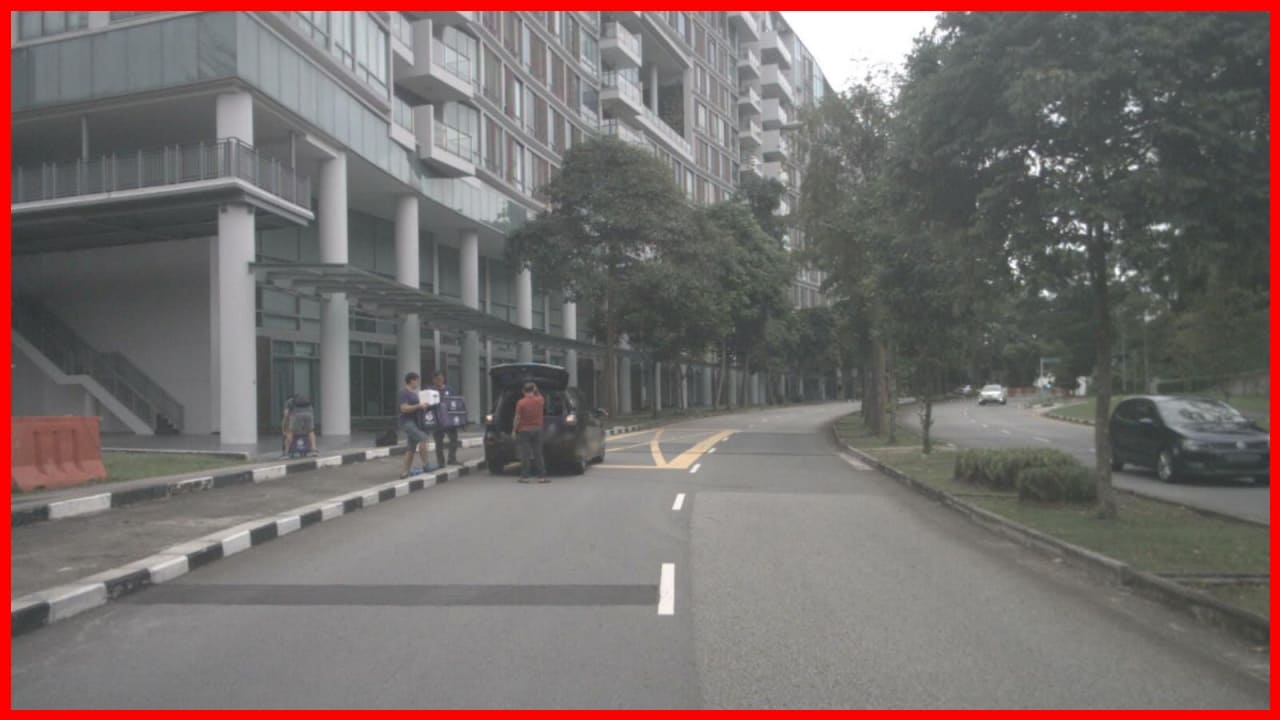}
    \end{subfigure} \\

    \rotatebox[origin=left]{90}{\hspace{-0.15cm} \textbf{NARADIO}} 
    \begin{subfigure}[b]{\subfigwidth}
        \includegraphics[width=\textwidth]{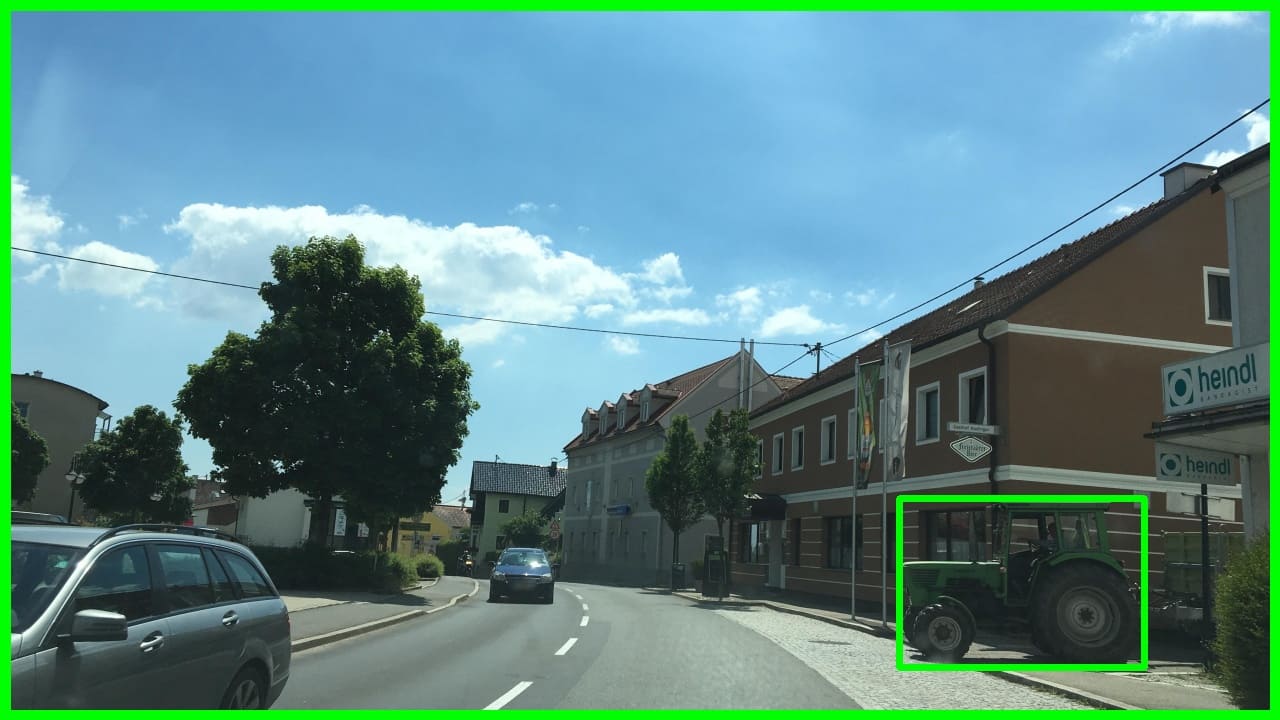}
    \end{subfigure}
    \begin{subfigure}[b]{\subfigwidth}
        \includegraphics[width=\textwidth]{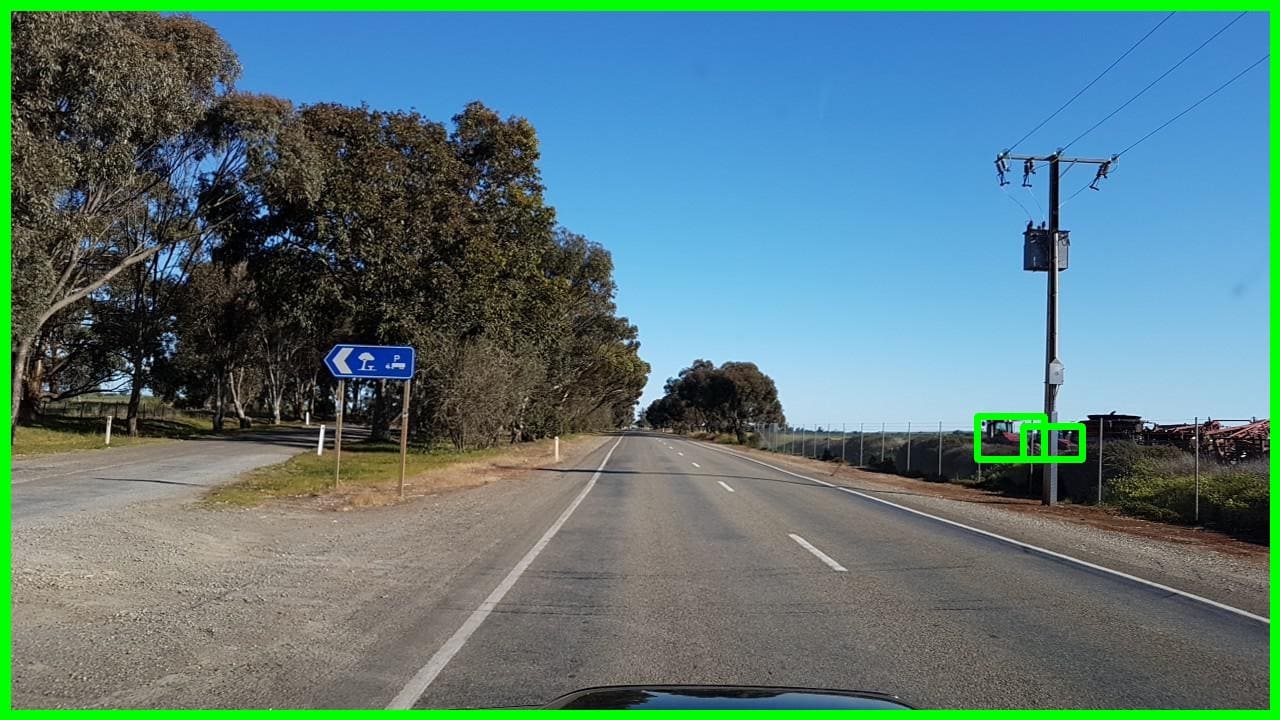}
    \end{subfigure}
    \begin{subfigure}[b]{\subfigwidth}
        \includegraphics[width=\textwidth]{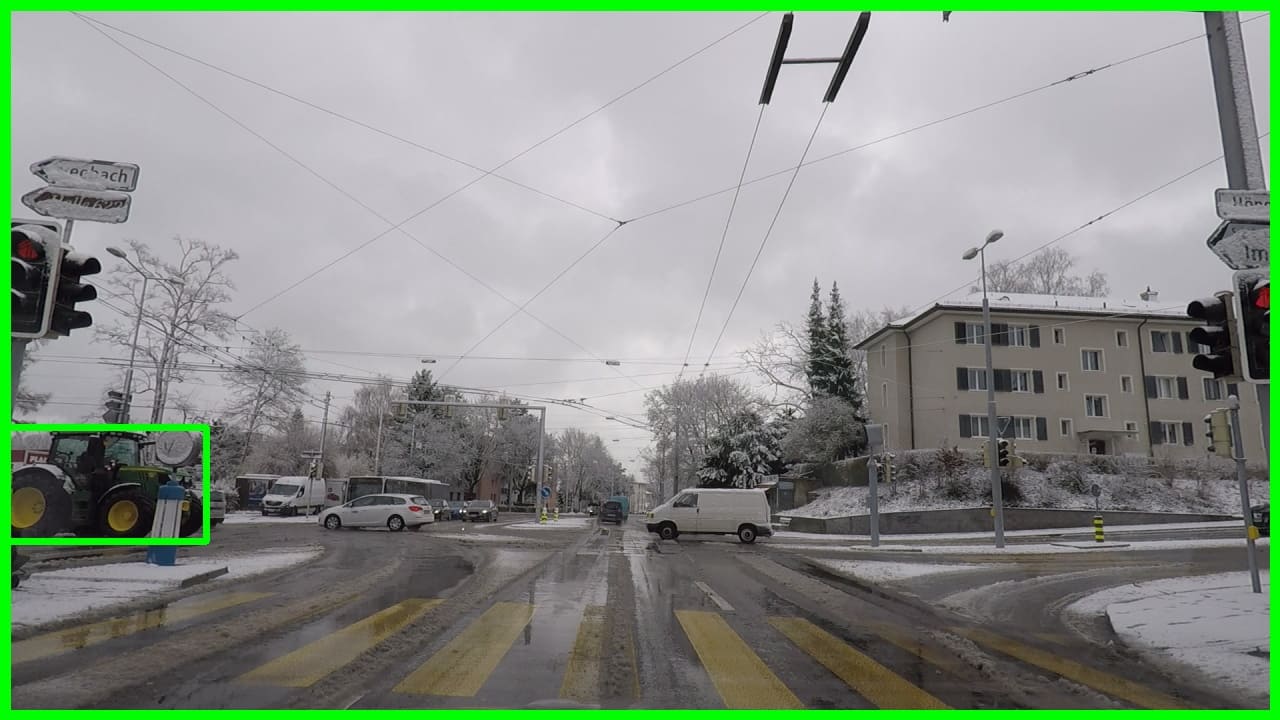}
    \end{subfigure}
    \begin{subfigure}[b]{\subfigwidth}
        \includegraphics[width=\textwidth]{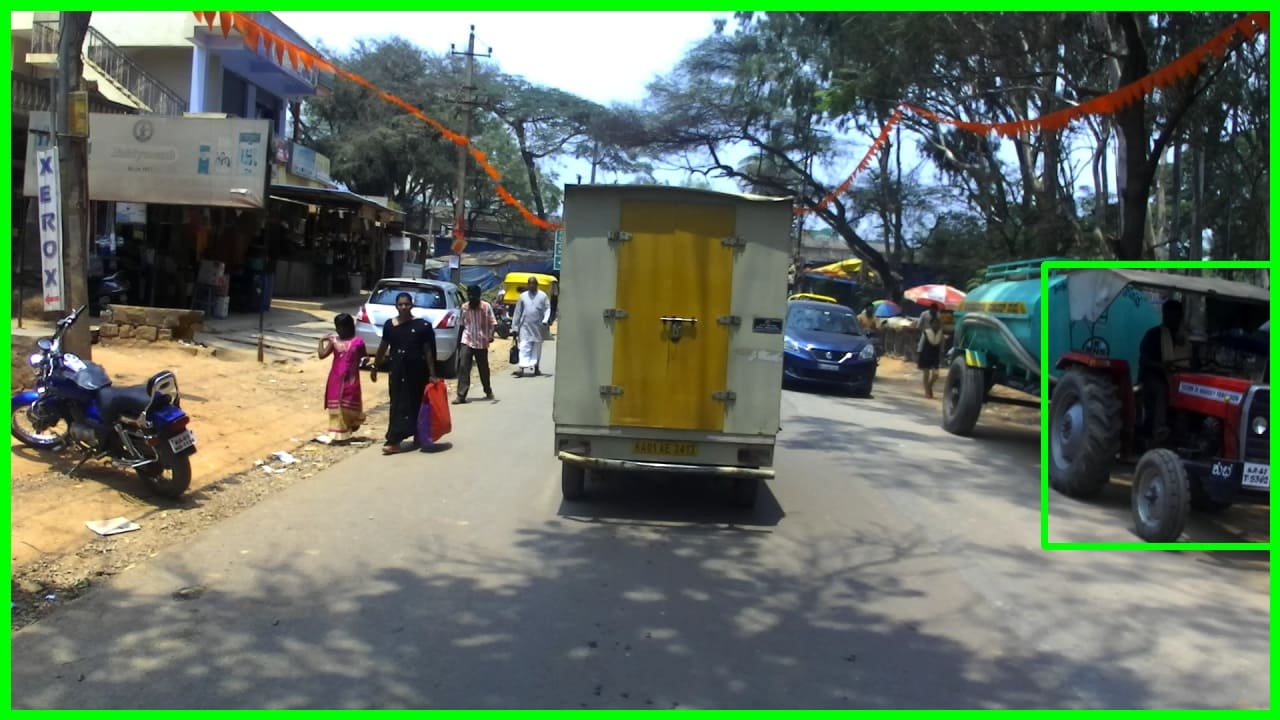}
    \end{subfigure}
    \begin{subfigure}[b]{\subfigwidth}
        \includegraphics[width=\textwidth]{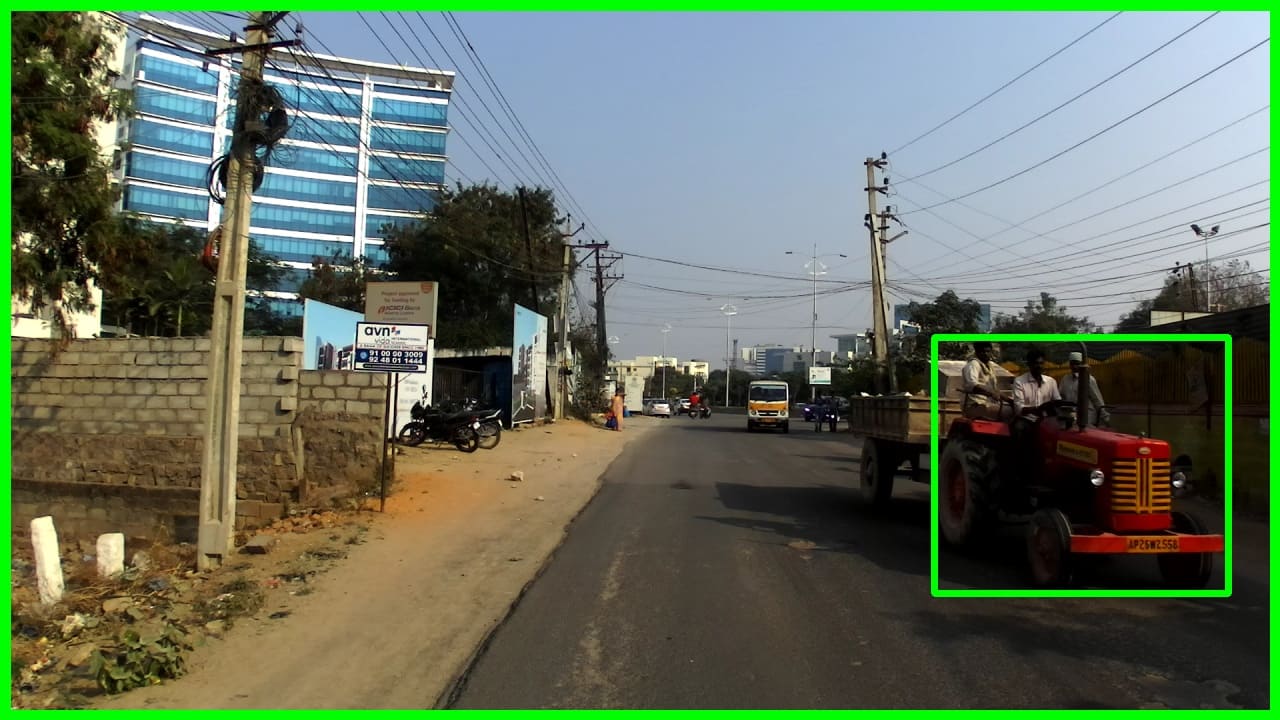}
    \end{subfigure} \\
    \caption{Text-based retrieval results on the validation split for the \emph{Vehicle-Tractor} class, showing top 5 ranked images for each model. Model references can be found in \Cref{tab:map_class_wise_text_to_image_retrieval}.}
    \label{fig:vehicle_tractor_results}
    \endgroup
\end{figure*}
}

{
\def\rankOne{\bfseries Ranked 1st}
\def\rankTwo{\bfseries Ranked 2nd}
\def\rankThree{\bfseries Ranked 3rd}
\def\rankFour{\bfseries Ranked 4th}
\def\rankFive{\bfseries Ranked 5th}

\captionsetup[subfigure]{labelformat=empty}

\setlength{\subfigwidth}{0.18\textwidth} 

\begin{figure*}[ht]
    \begingroup
    \centering 
    \small
    \rotatebox[origin=left]{90}{\hspace{0.02cm} \textbf{GDINO}} 
    \begin{subfigure}[b]{\subfigwidth}
        \caption{\rankOne}
        \includegraphics[width=\textwidth]{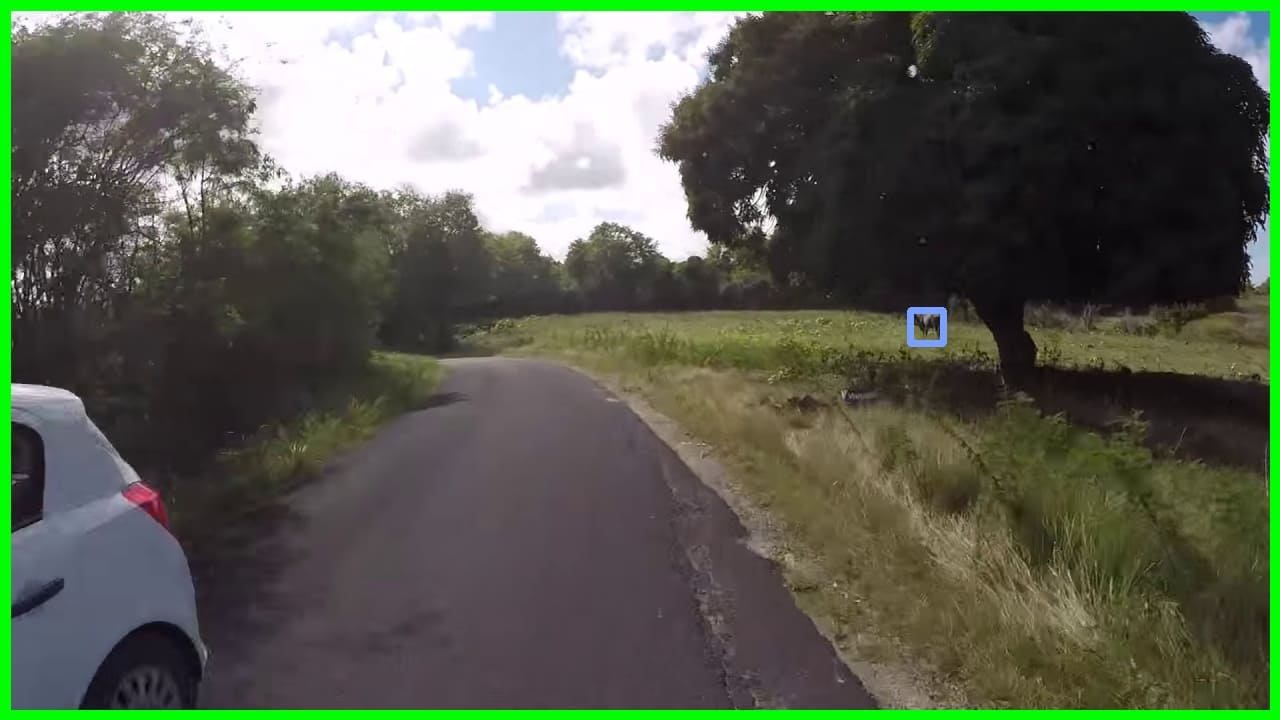}
    \end{subfigure}
    \begin{subfigure}[b]{\subfigwidth}
        \caption{\rankTwo}
        \includegraphics[width=\textwidth]{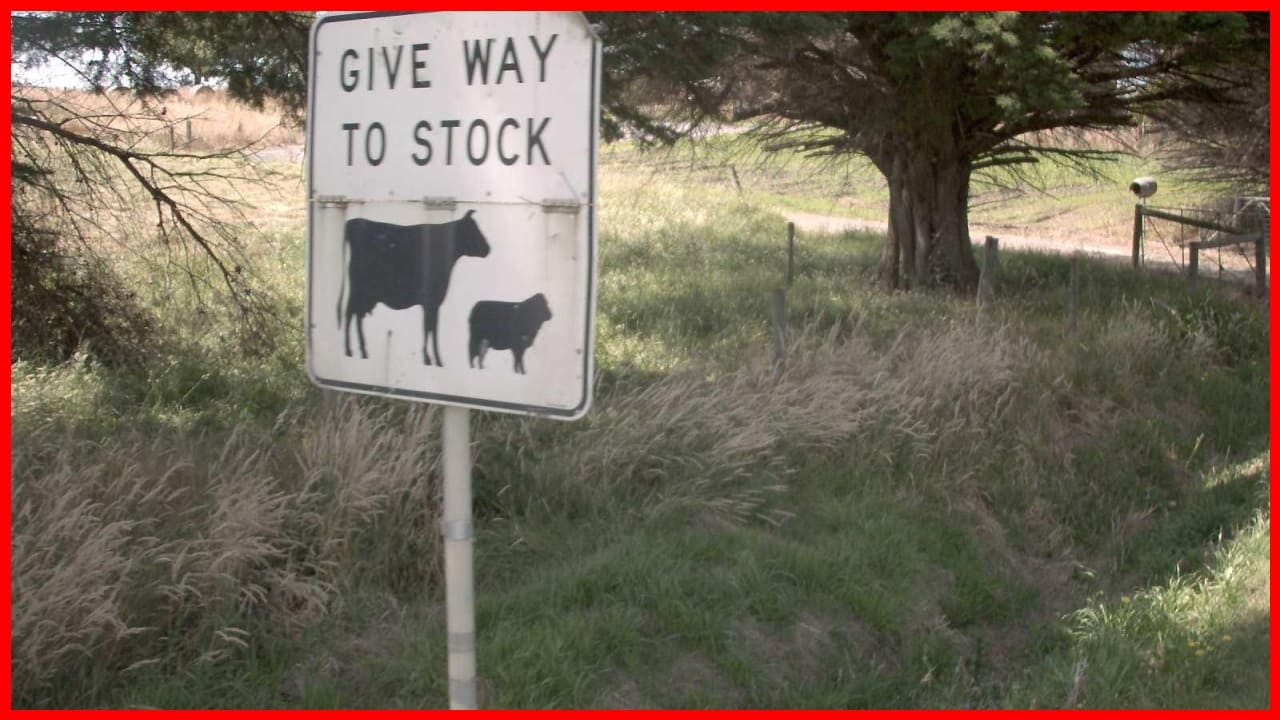}
    \end{subfigure}
    \begin{subfigure}[b]{\subfigwidth}
        \caption{\rankThree}
        \includegraphics[width=\textwidth]{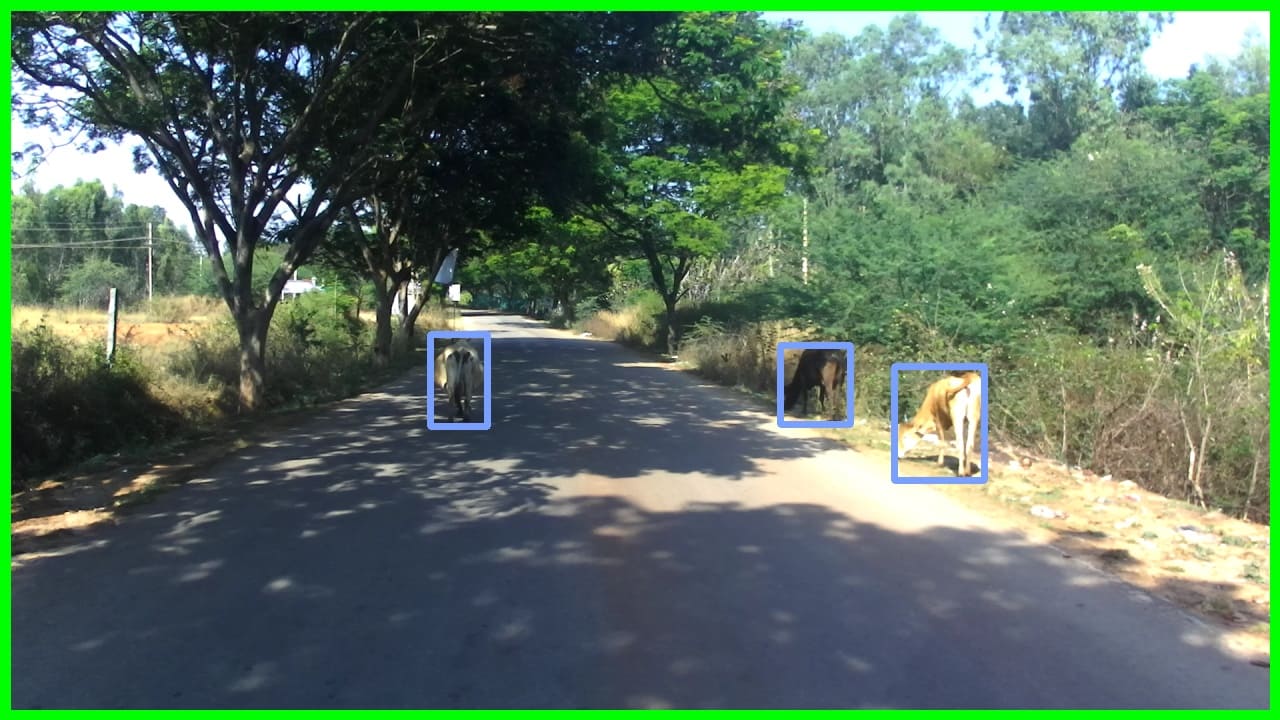}
    \end{subfigure}
    \begin{subfigure}[b]{\subfigwidth}
        \caption{\rankFour}
        \includegraphics[width=\textwidth]{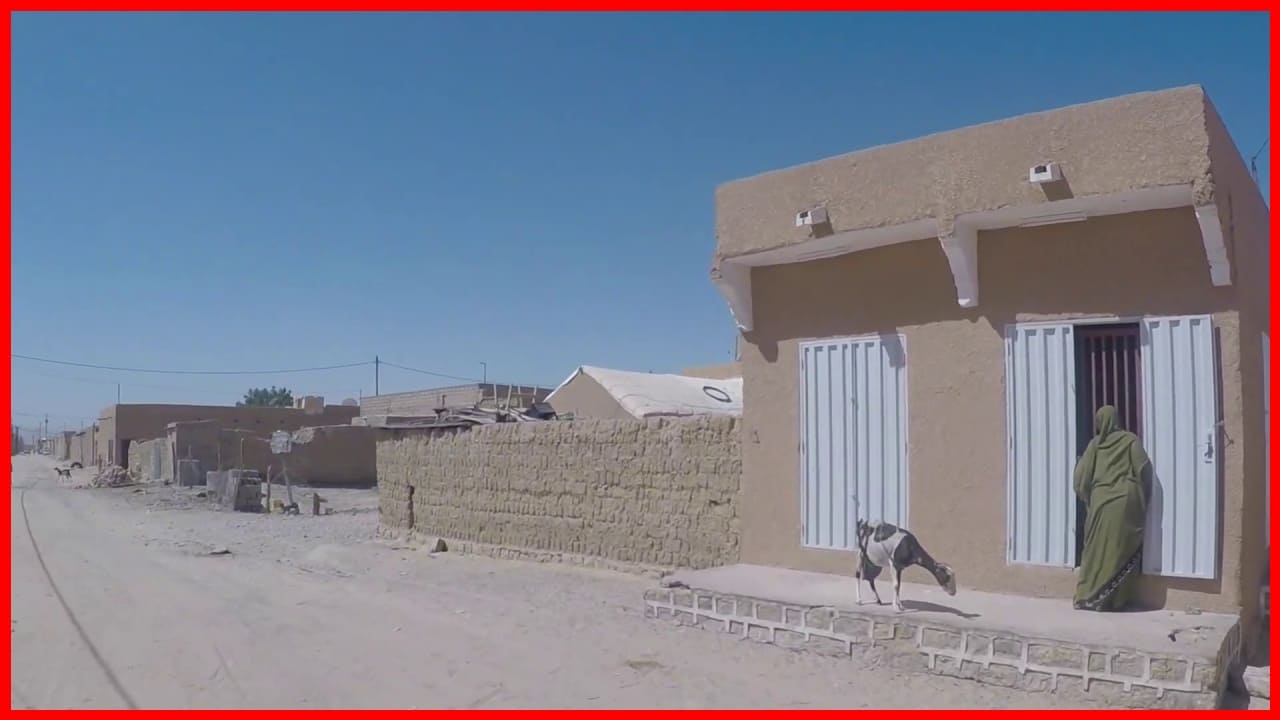}
    \end{subfigure}
    \begin{subfigure}[b]{\subfigwidth}
        \caption{\rankFive}
        \includegraphics[width=\textwidth]{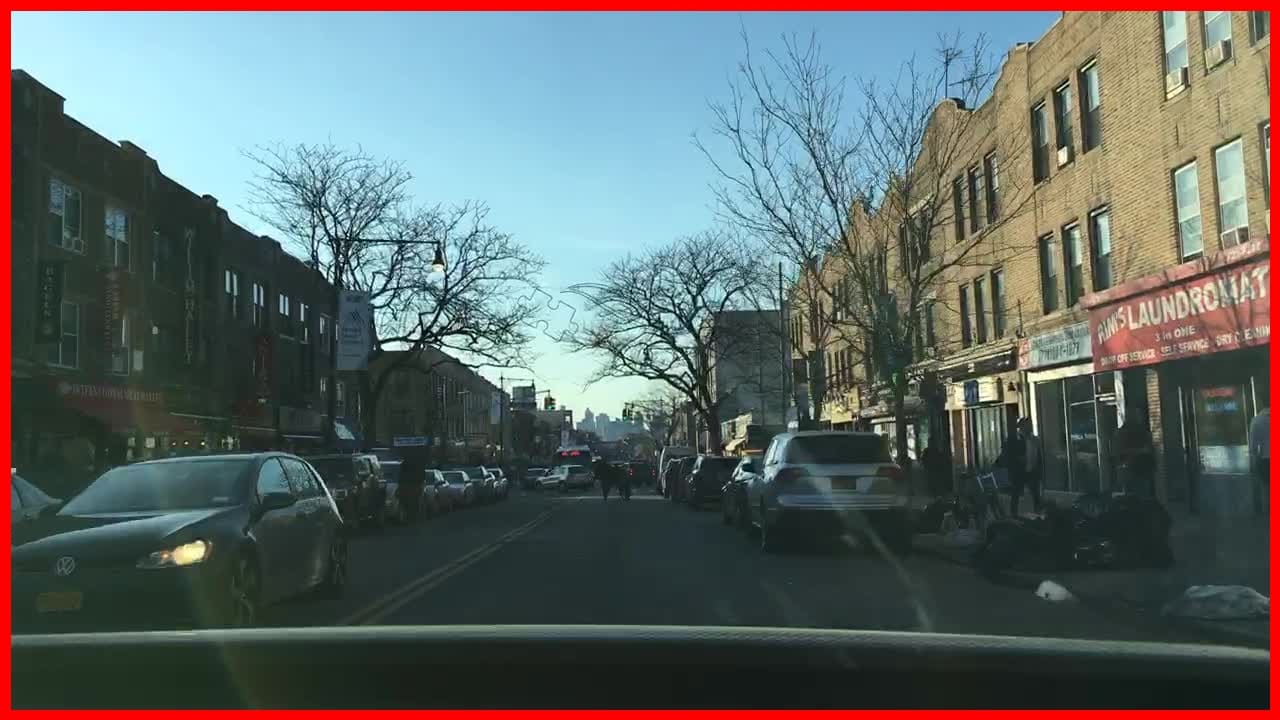}
    \end{subfigure} \\
    
    \rotatebox[origin=left]{90}{\hspace{0.17cm} \textbf{CLIP}} 
    \begin{subfigure}[b]{\subfigwidth}
        \includegraphics[width=\textwidth]{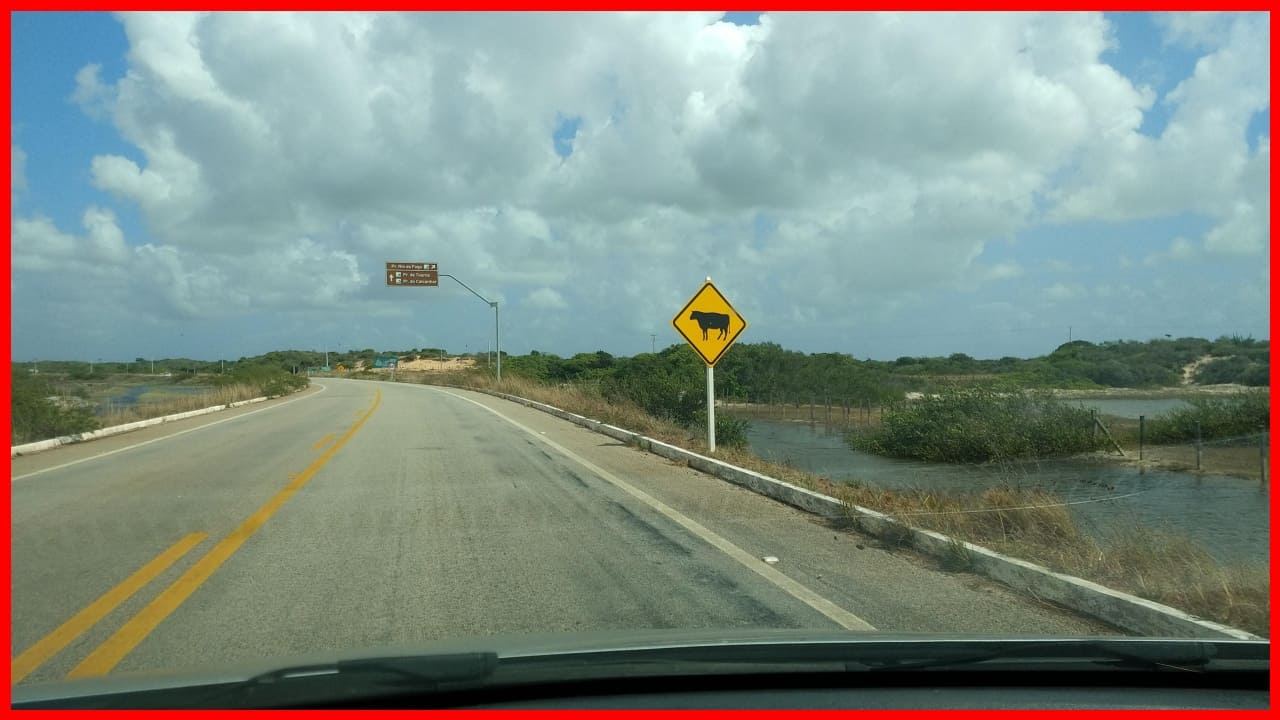}
    \end{subfigure}
    \begin{subfigure}[b]{\subfigwidth}
        \includegraphics[width=\textwidth]{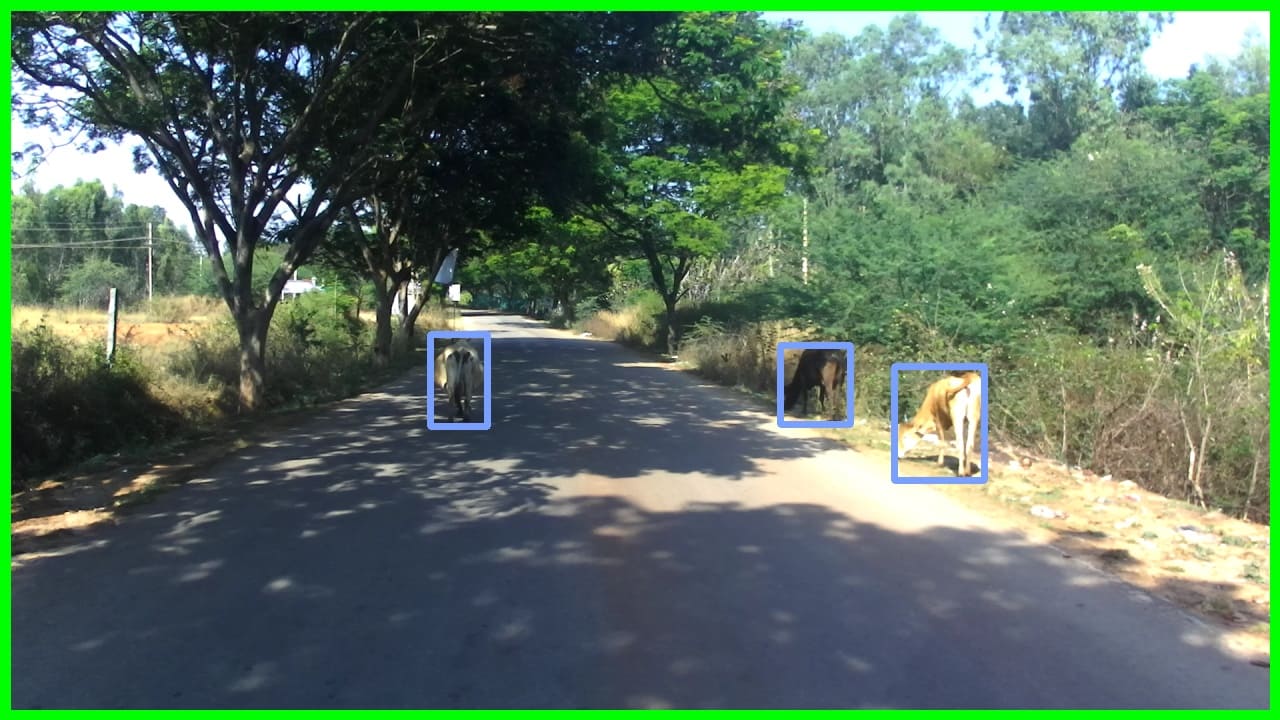}
    \end{subfigure}
    \begin{subfigure}[b]{\subfigwidth}
        \includegraphics[width=\textwidth]{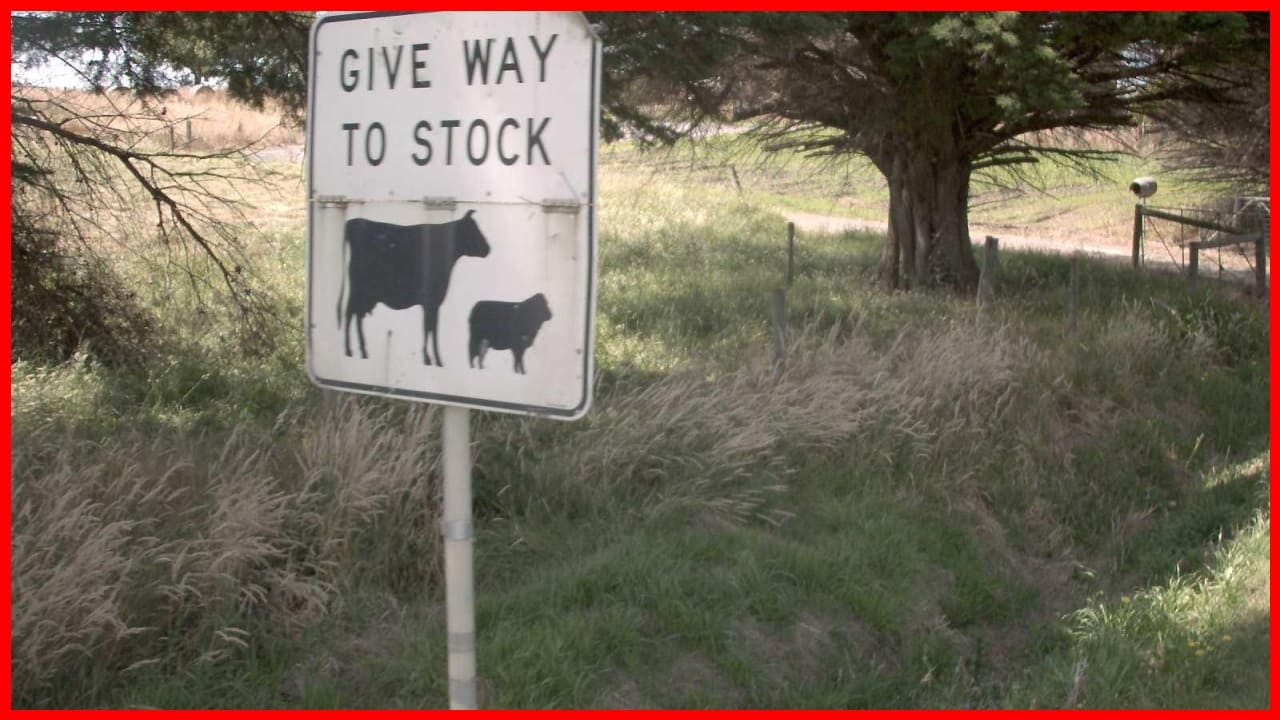}
    \end{subfigure}
    \begin{subfigure}[b]{\subfigwidth}
        \includegraphics[width=\textwidth]{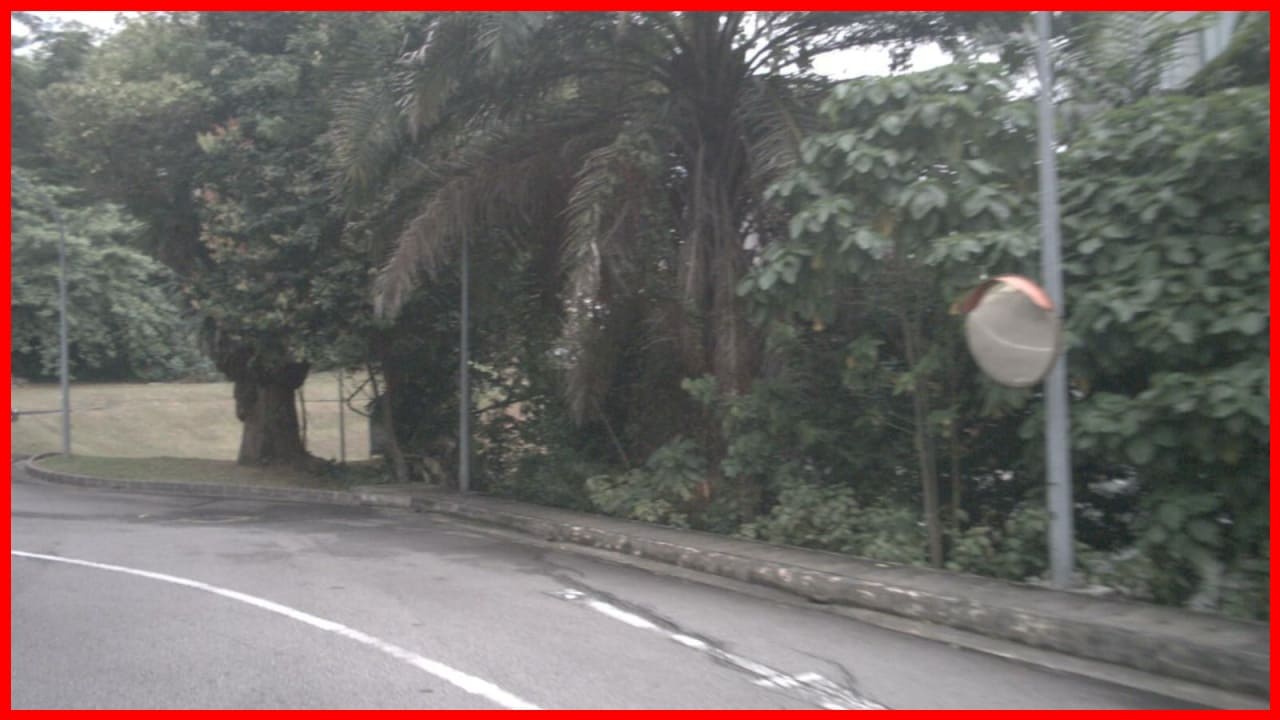}
    \end{subfigure}
    \begin{subfigure}[b]{\subfigwidth}
        \includegraphics[width=\textwidth]{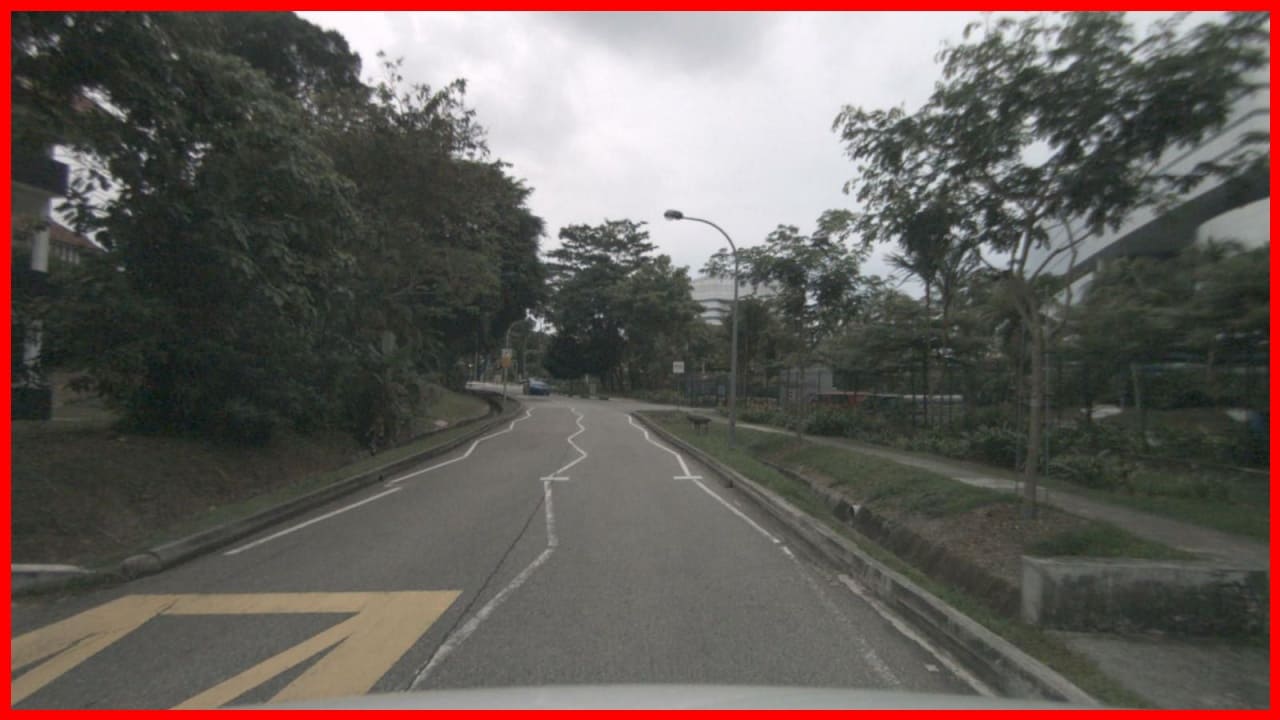}
    \end{subfigure} \\

    \rotatebox[origin=left]{90}{\hspace{0.07cm} \textbf{RADIO}} 
    \begin{subfigure}[b]{\subfigwidth}
        \includegraphics[width=\textwidth]{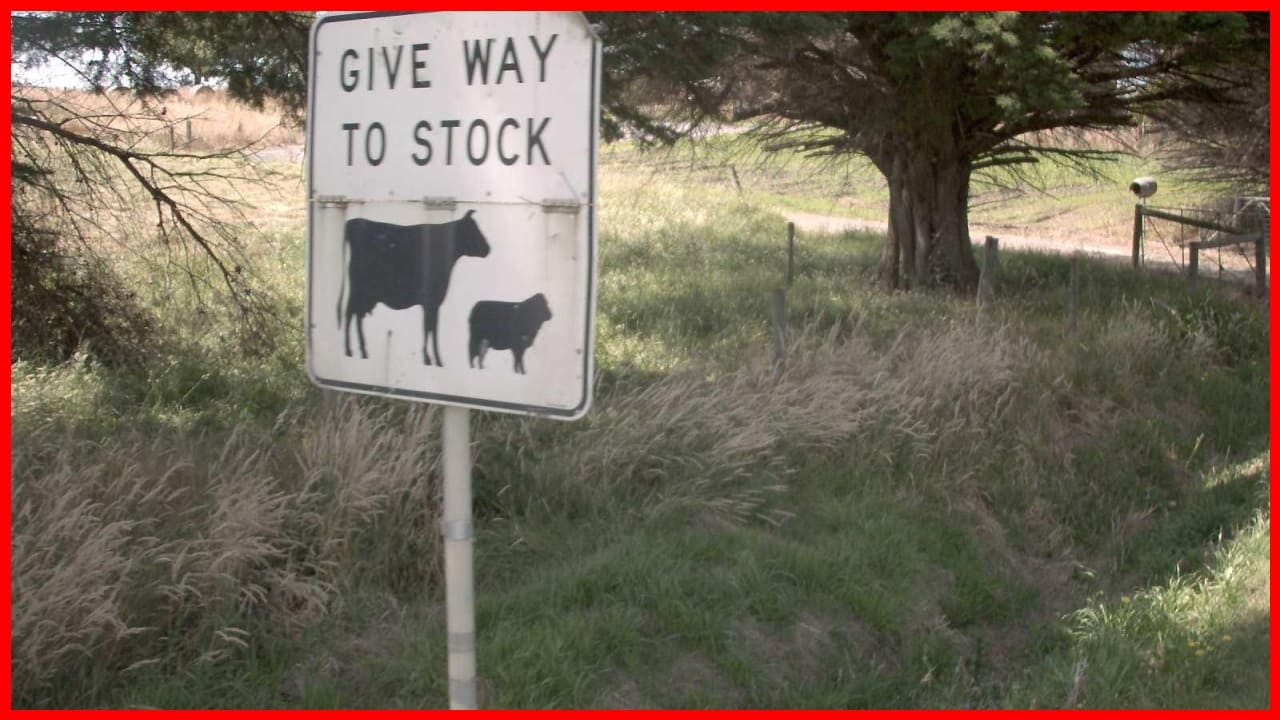}
    \end{subfigure}
    \begin{subfigure}[b]{\subfigwidth}
        \includegraphics[width=\textwidth]{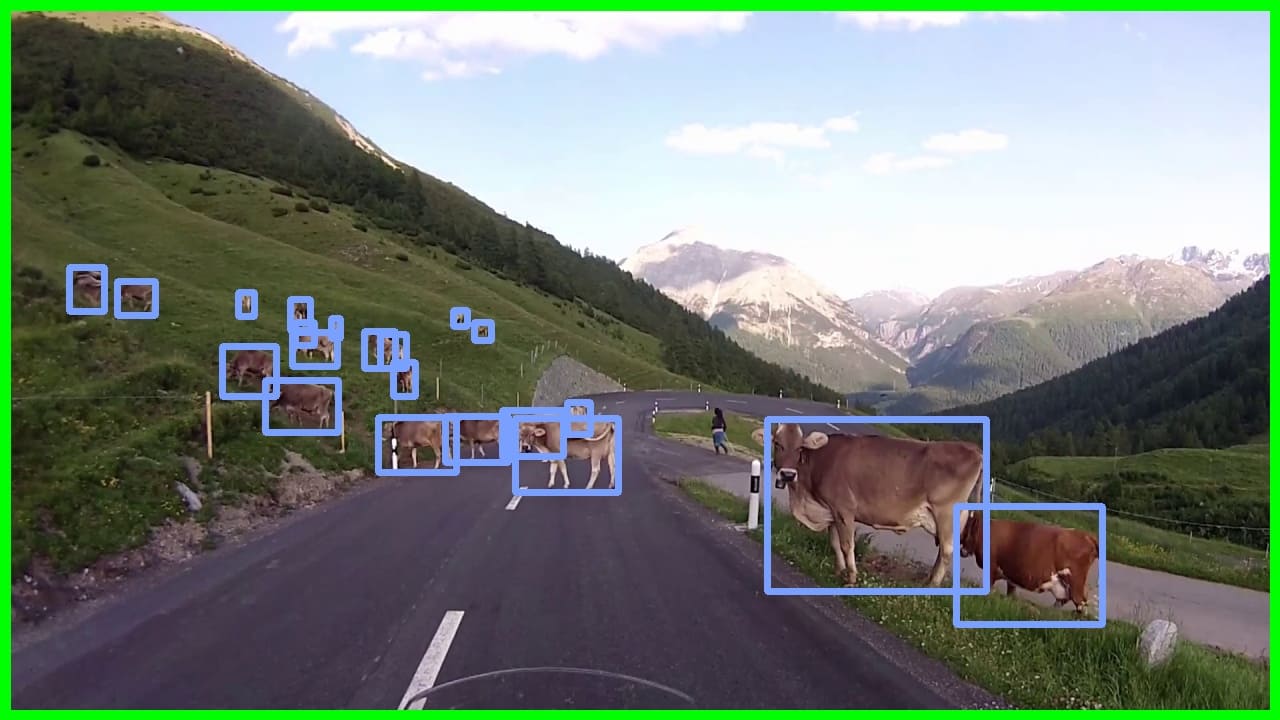}
    \end{subfigure}
    \begin{subfigure}[b]{\subfigwidth}
        \includegraphics[width=\textwidth]{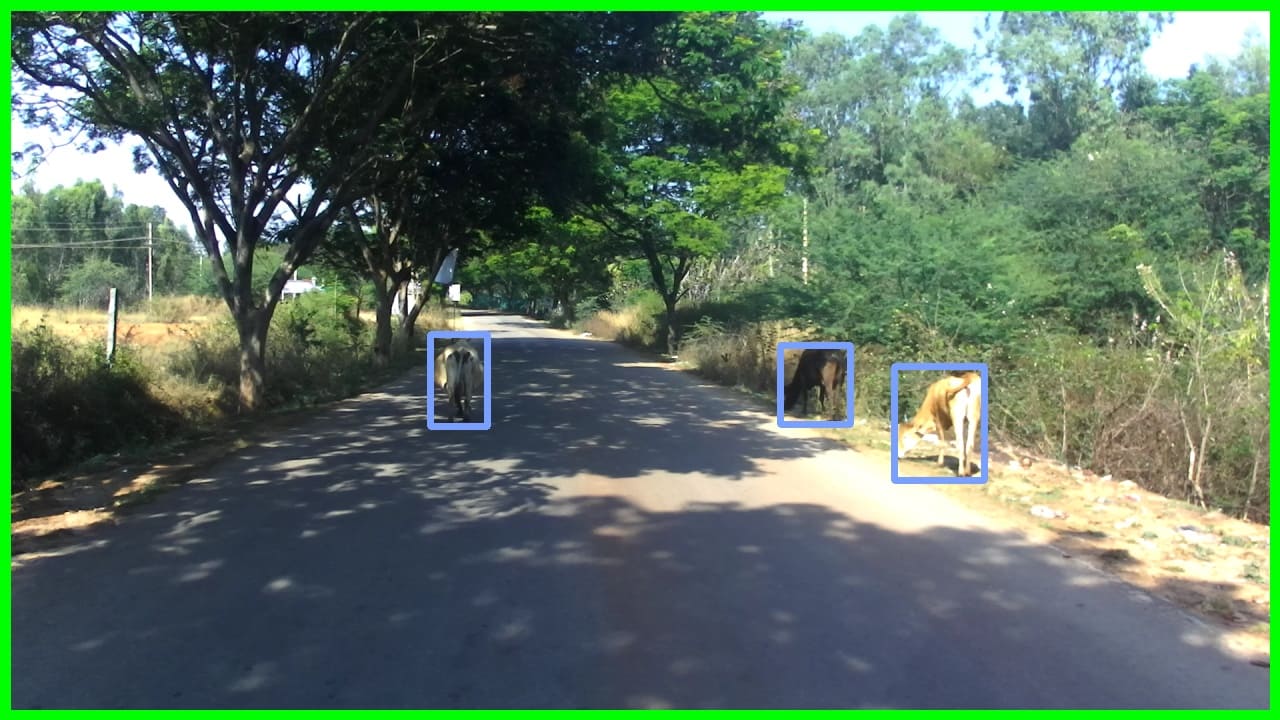}
    \end{subfigure}
    \begin{subfigure}[b]{\subfigwidth}
        \includegraphics[width=\textwidth]{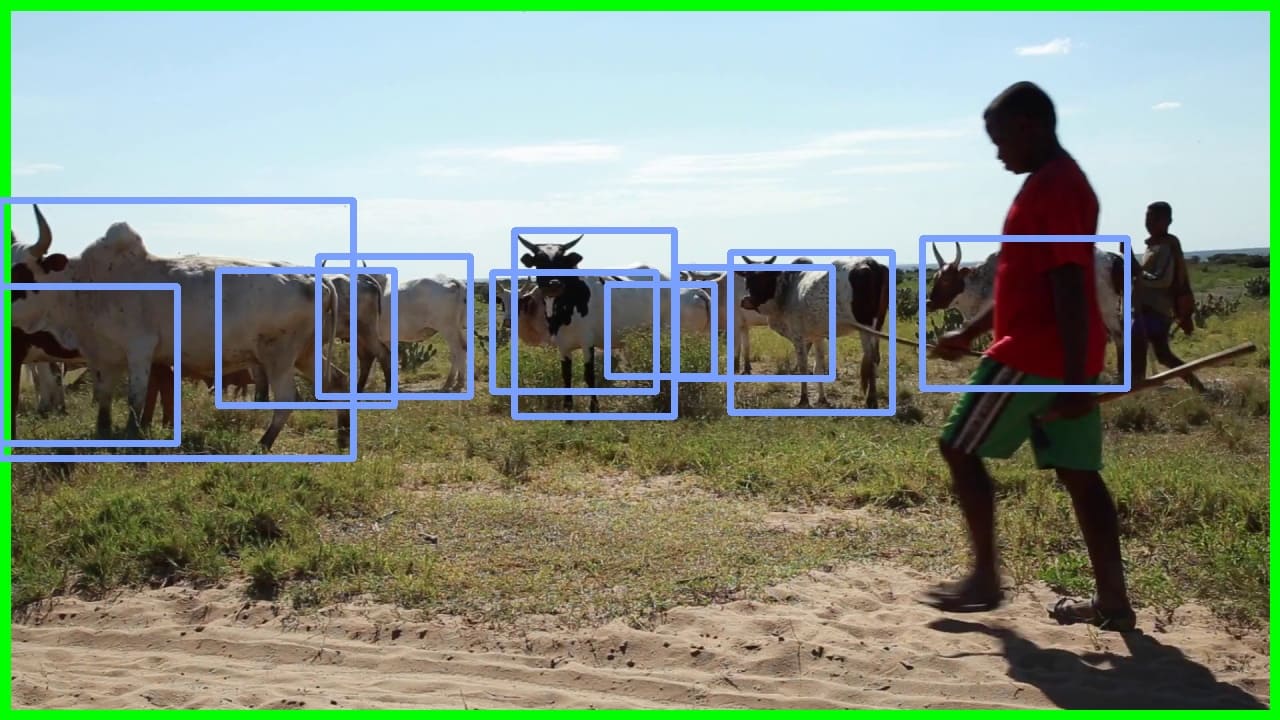}
    \end{subfigure}
    \begin{subfigure}[b]{\subfigwidth}
        \includegraphics[width=\textwidth]{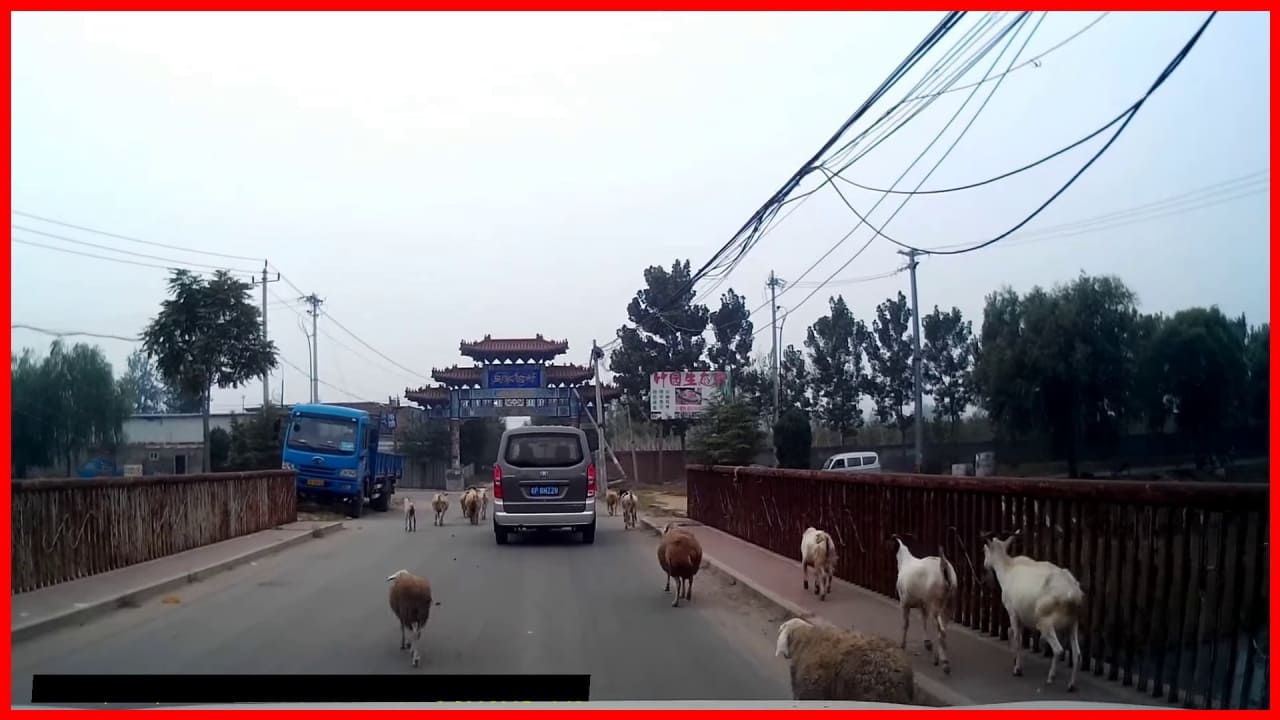}
    \end{subfigure} \\

    \rotatebox[origin=left]{90}{\hspace{0.15cm} \textbf{BLIP2}} 
    \begin{subfigure}[b]{\subfigwidth}
        \includegraphics[width=\textwidth]{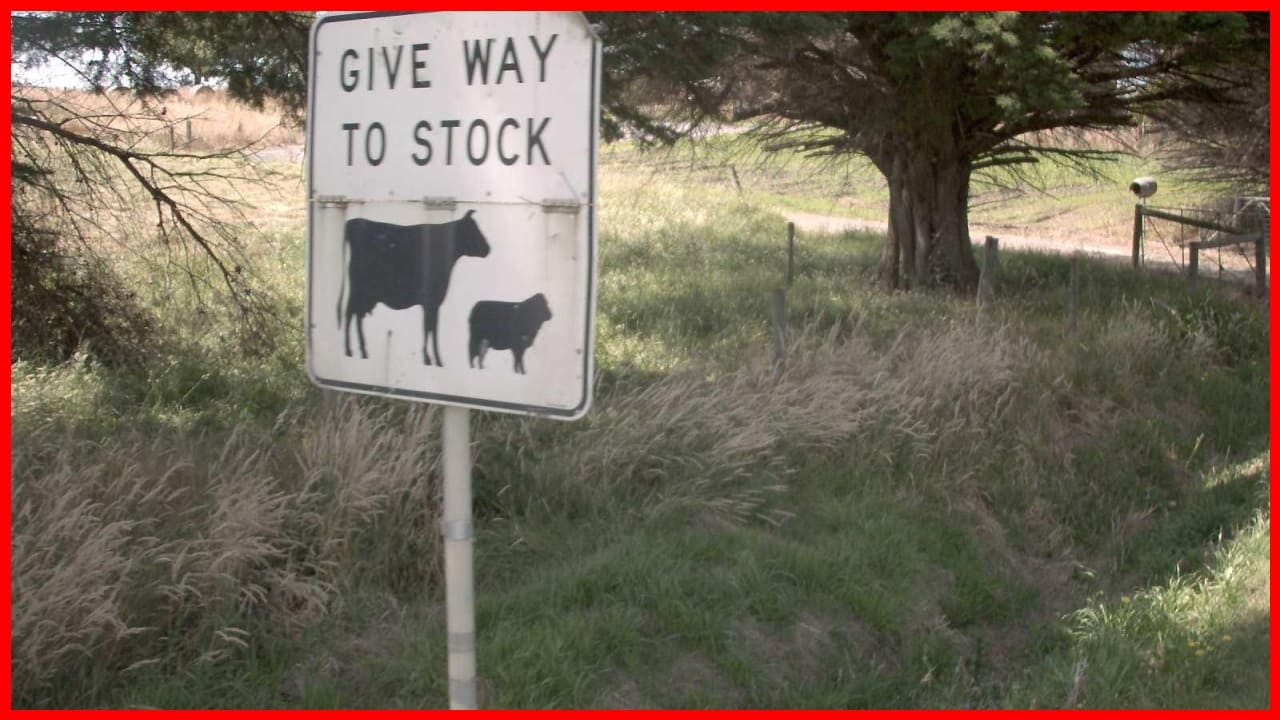}
    \end{subfigure}
    \begin{subfigure}[b]{\subfigwidth}
        \includegraphics[width=\textwidth]{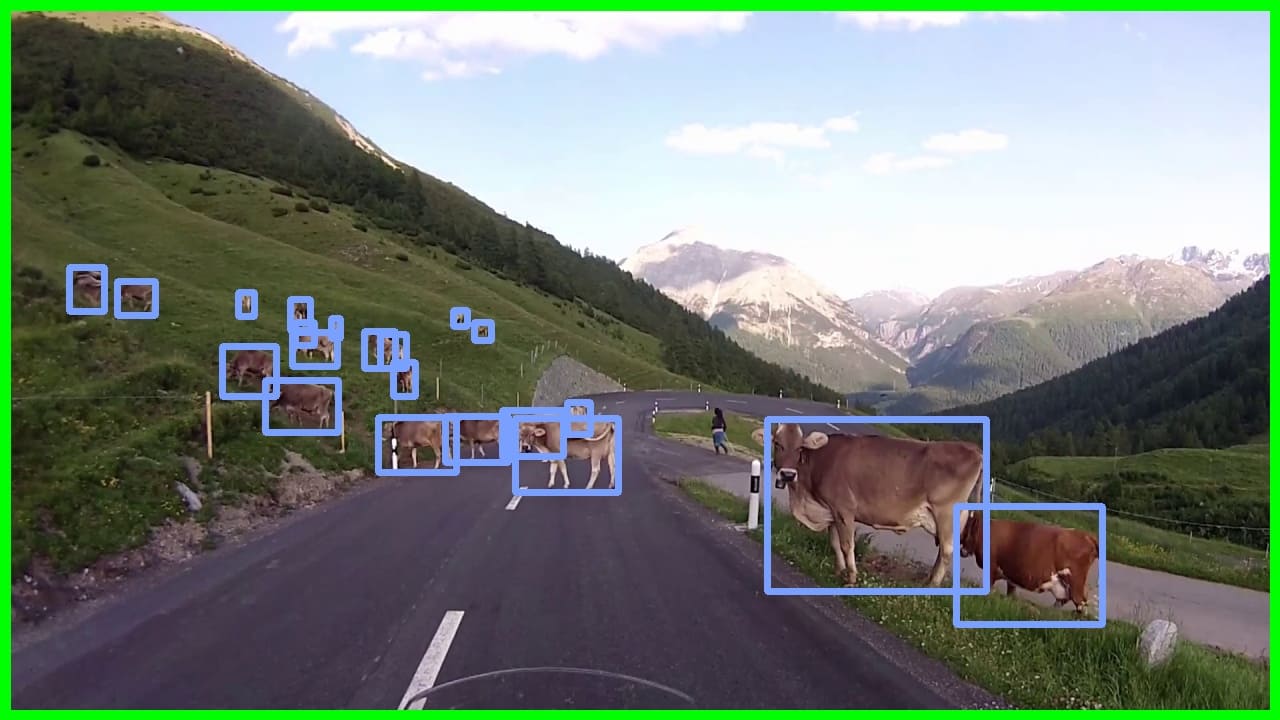}
    \end{subfigure}
    \begin{subfigure}[b]{\subfigwidth}
        \includegraphics[width=\textwidth]{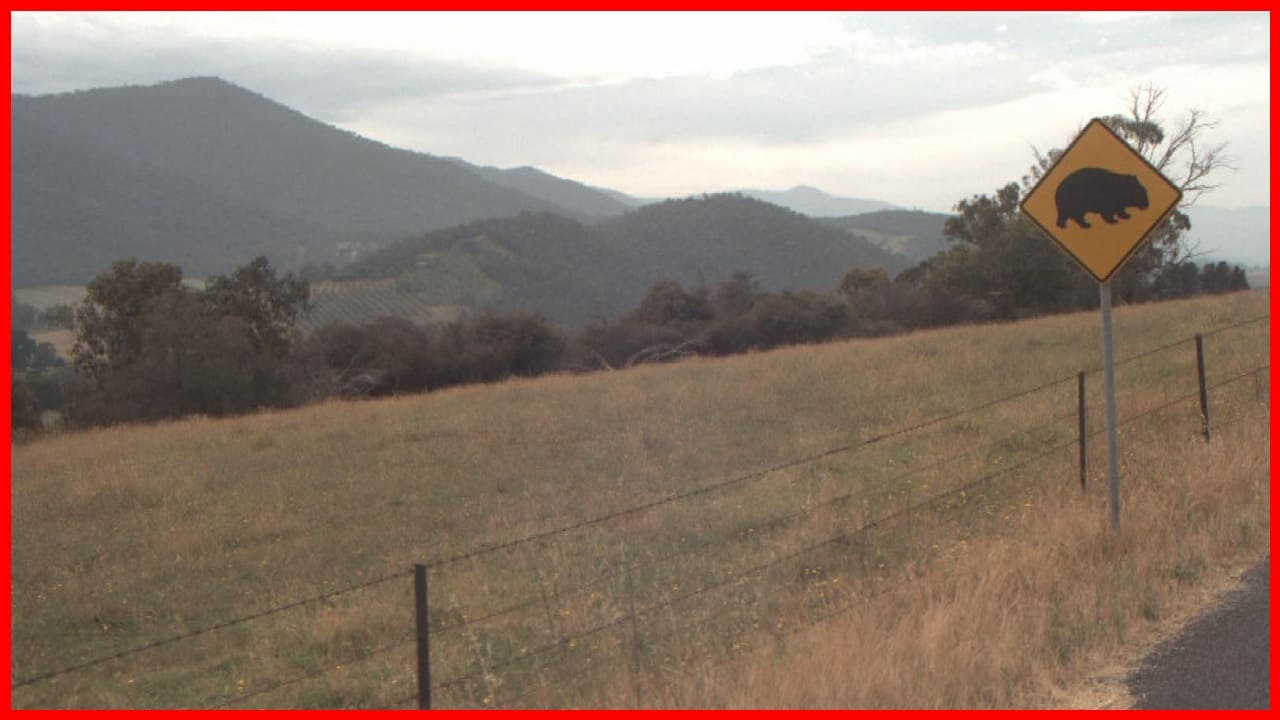}
    \end{subfigure}
    \begin{subfigure}[b]{\subfigwidth}
        \includegraphics[width=\textwidth]{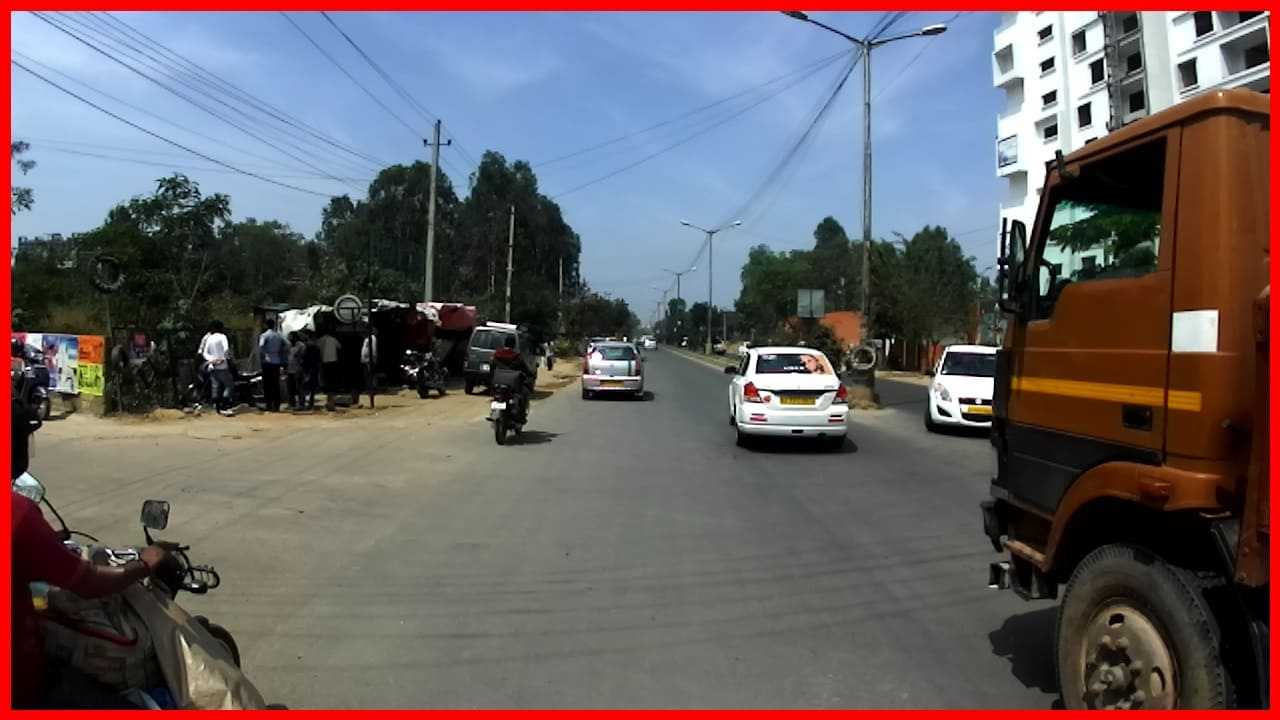}
    \end{subfigure}
    \begin{subfigure}[b]{\subfigwidth}
        \includegraphics[width=\textwidth]{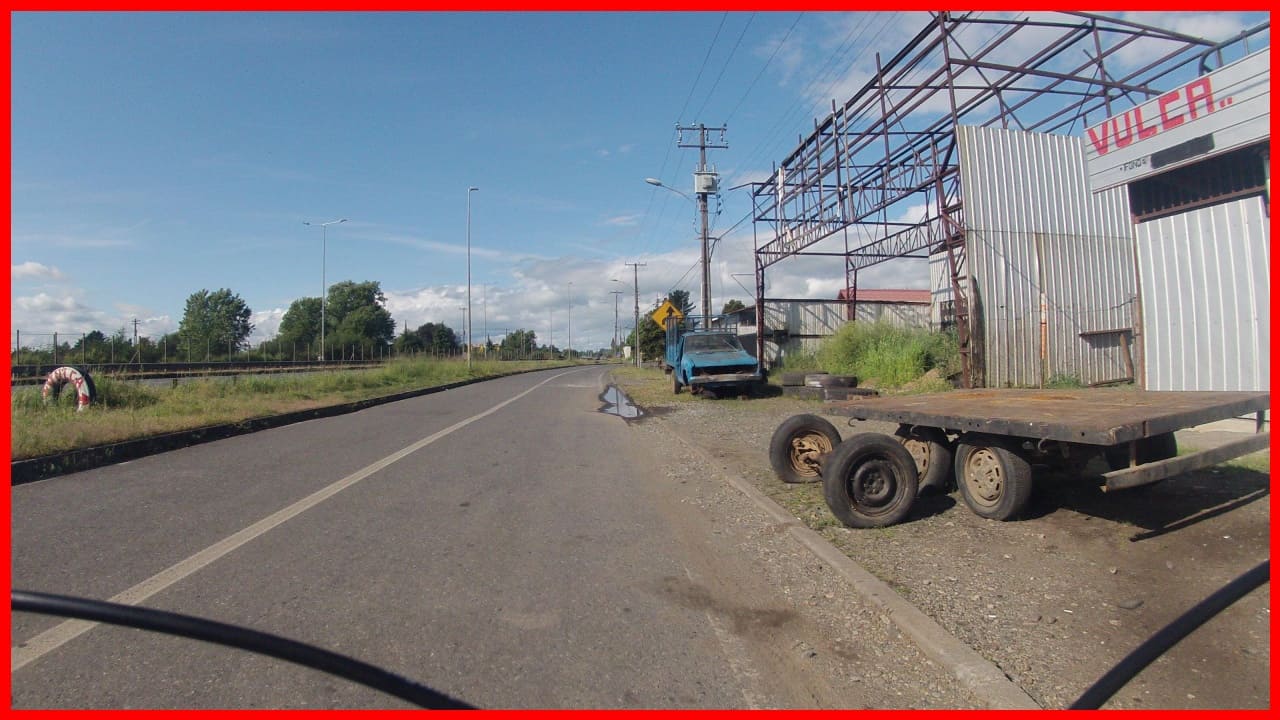}
    \end{subfigure} \\

    \rotatebox[origin=left]{90}{\hspace{-0.29cm} \textbf{METACLIP2}} 
    \begin{subfigure}[b]{\subfigwidth}
        \includegraphics[width=\textwidth]{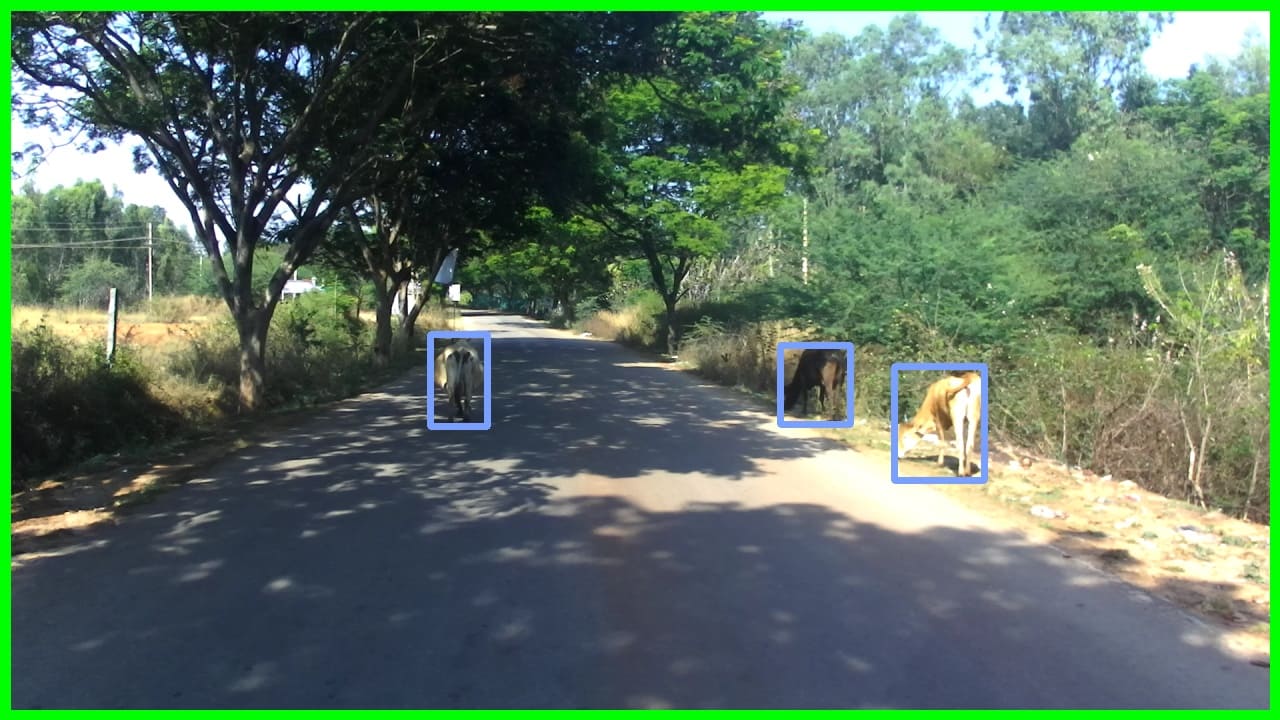}
    \end{subfigure}
    \begin{subfigure}[b]{\subfigwidth}
        \includegraphics[width=\textwidth]{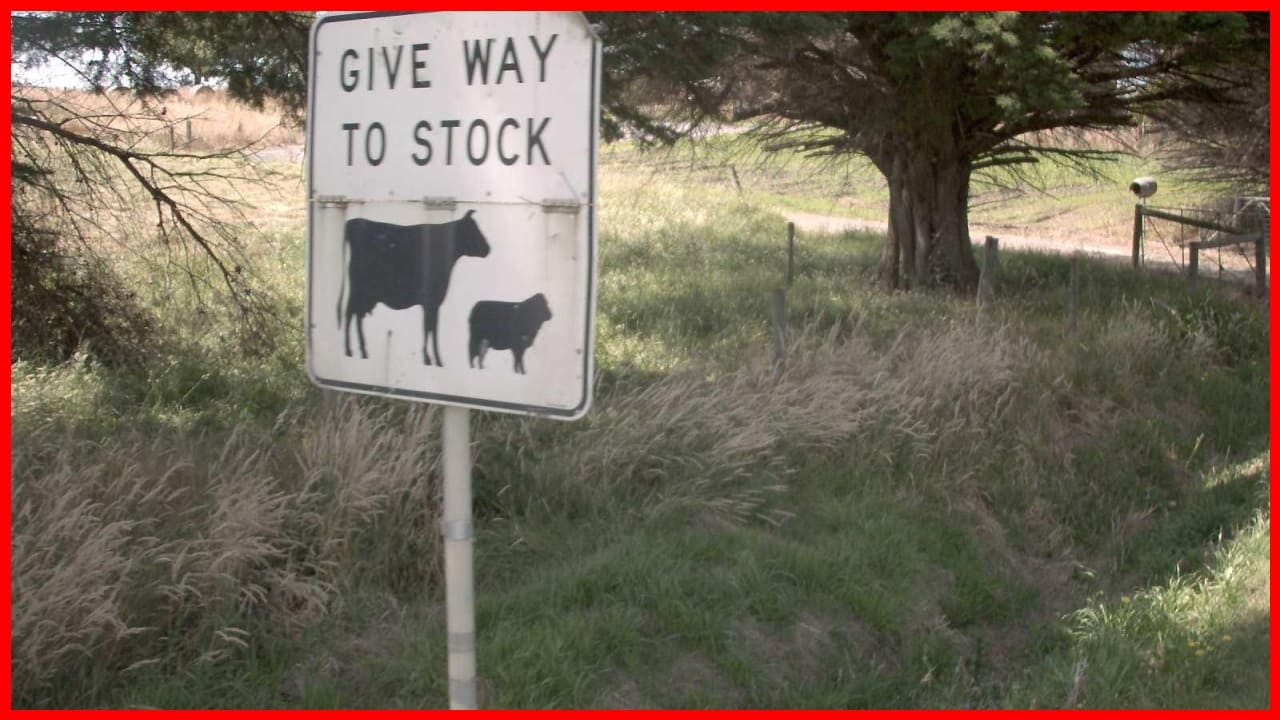}
    \end{subfigure}
    \begin{subfigure}[b]{\subfigwidth}
        \includegraphics[width=\textwidth]{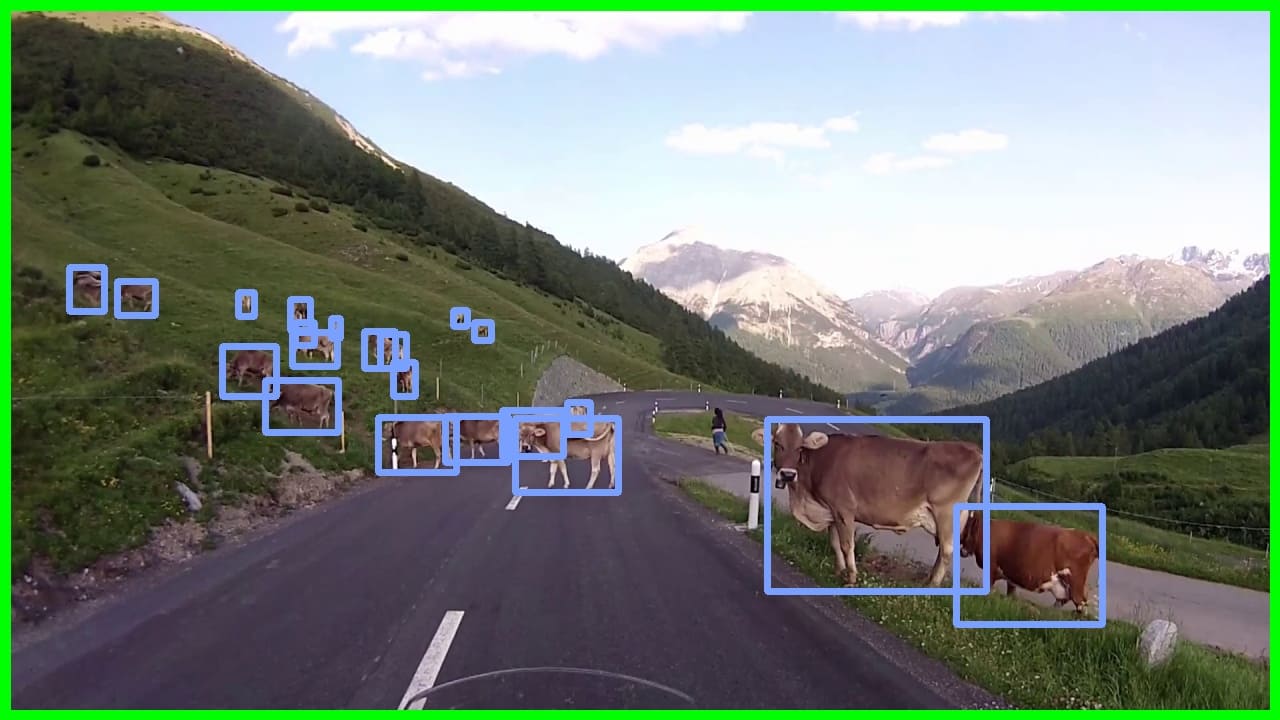}
    \end{subfigure}
    \begin{subfigure}[b]{\subfigwidth}
        \includegraphics[width=\textwidth]{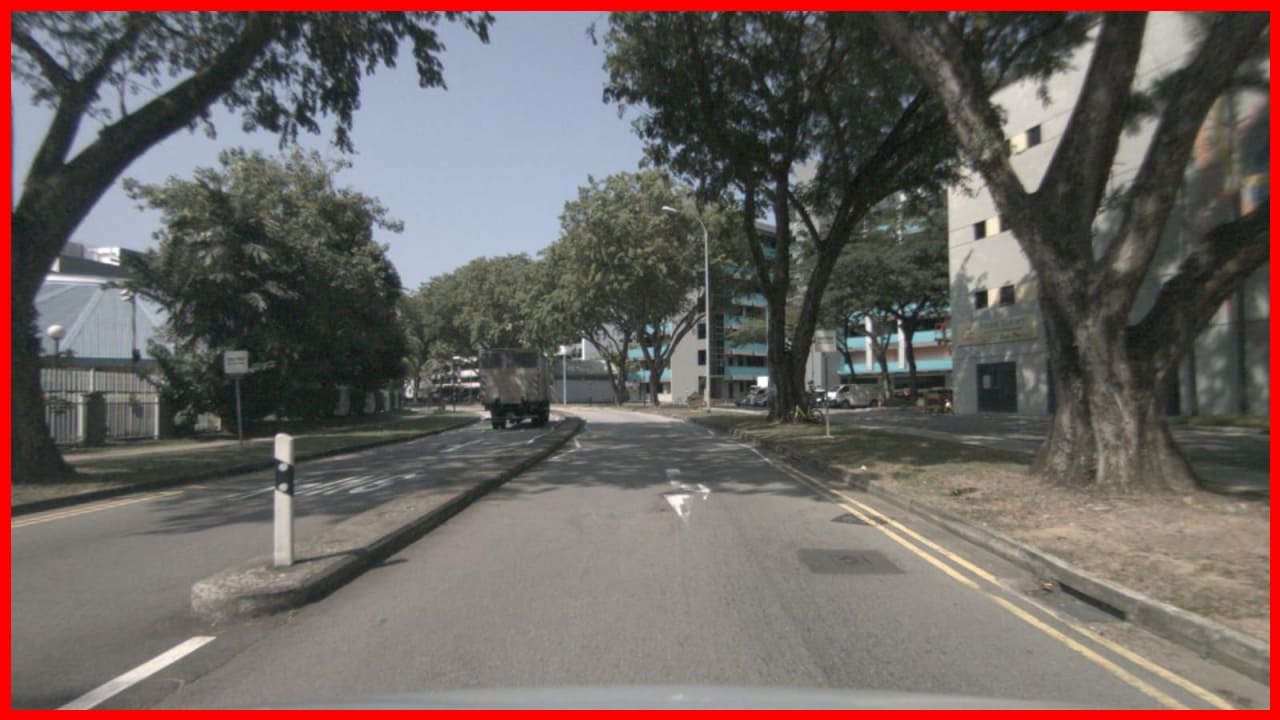}
    \end{subfigure}
    \begin{subfigure}[b]{\subfigwidth}
        \includegraphics[width=\textwidth]{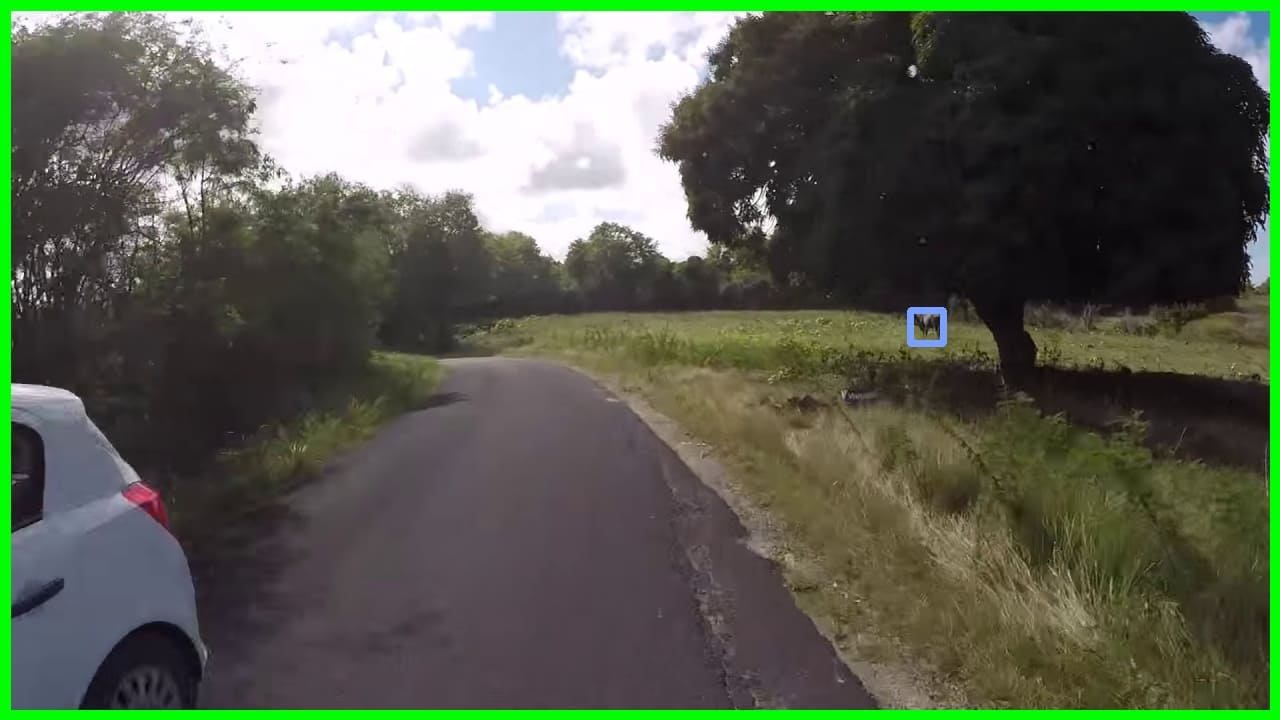}
    \end{subfigure} \\

    \rotatebox[origin=left]{90}{\hspace{0.02cm} \textbf{SIGLIP2}} 
    \begin{subfigure}[b]{\subfigwidth}
        \includegraphics[width=\textwidth]{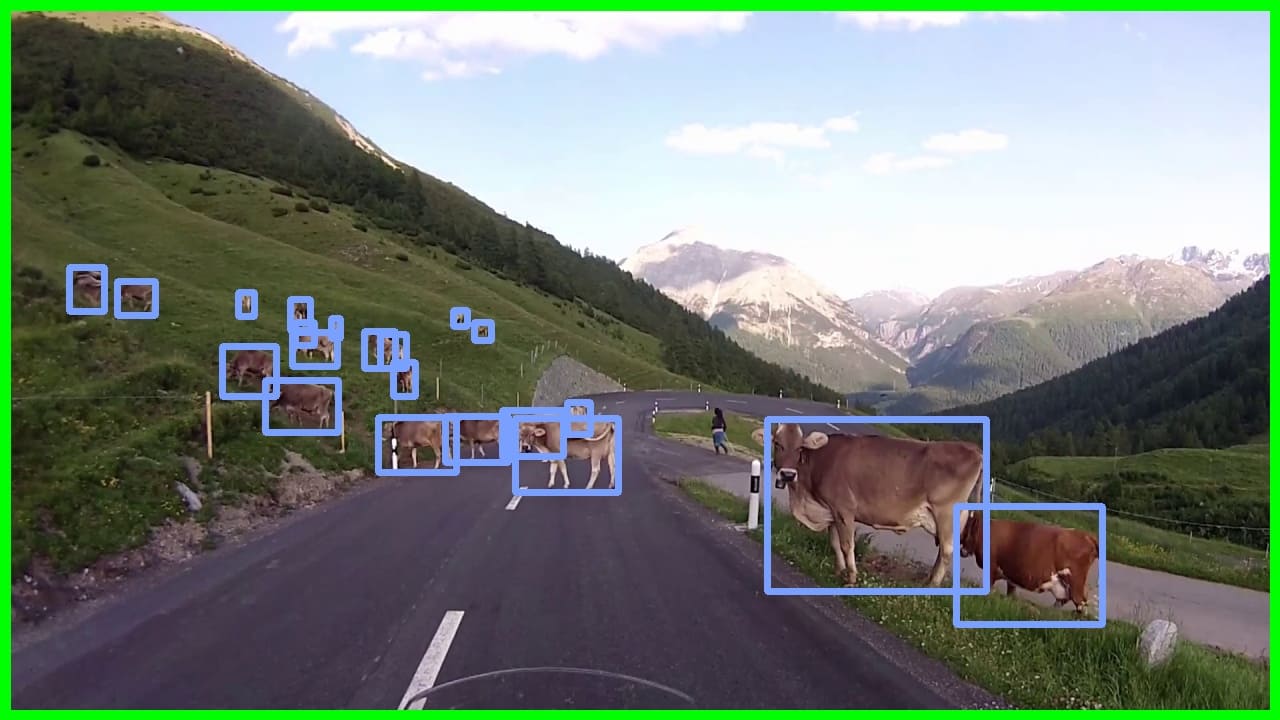}
    \end{subfigure}
    \begin{subfigure}[b]{\subfigwidth}
        \includegraphics[width=\textwidth]{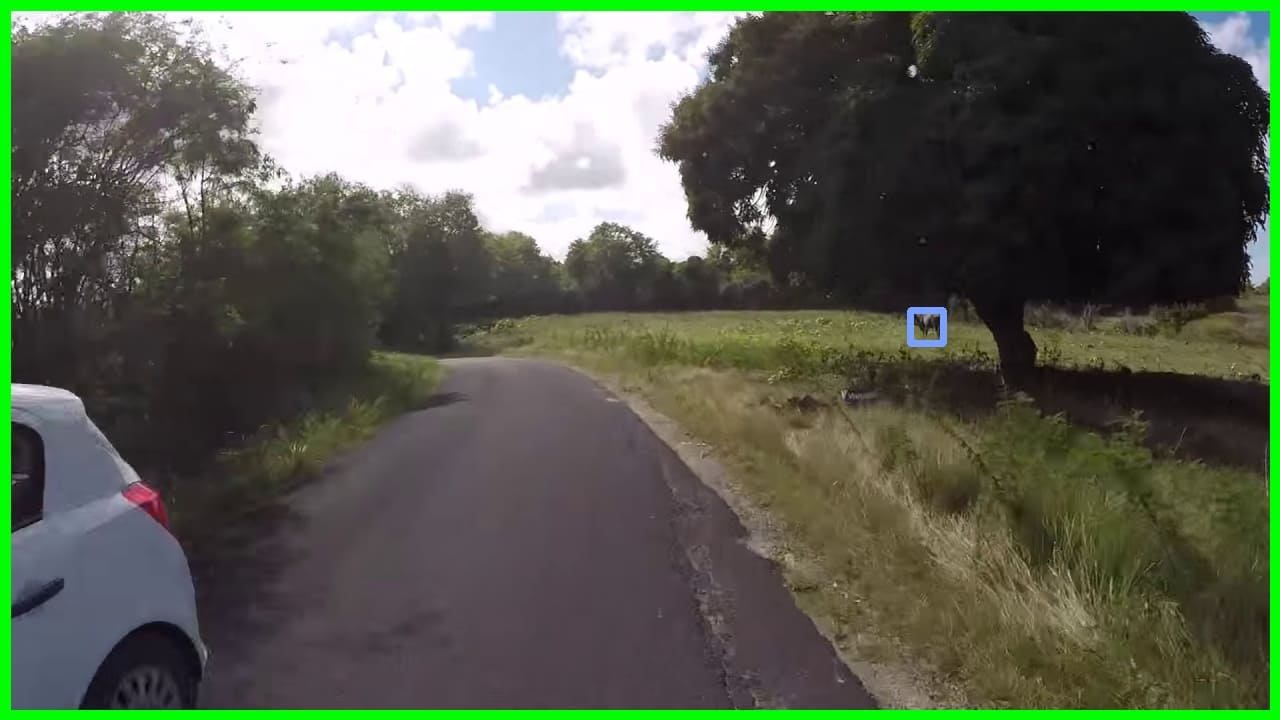}
    \end{subfigure}
    \begin{subfigure}[b]{\subfigwidth}
        \includegraphics[width=\textwidth]{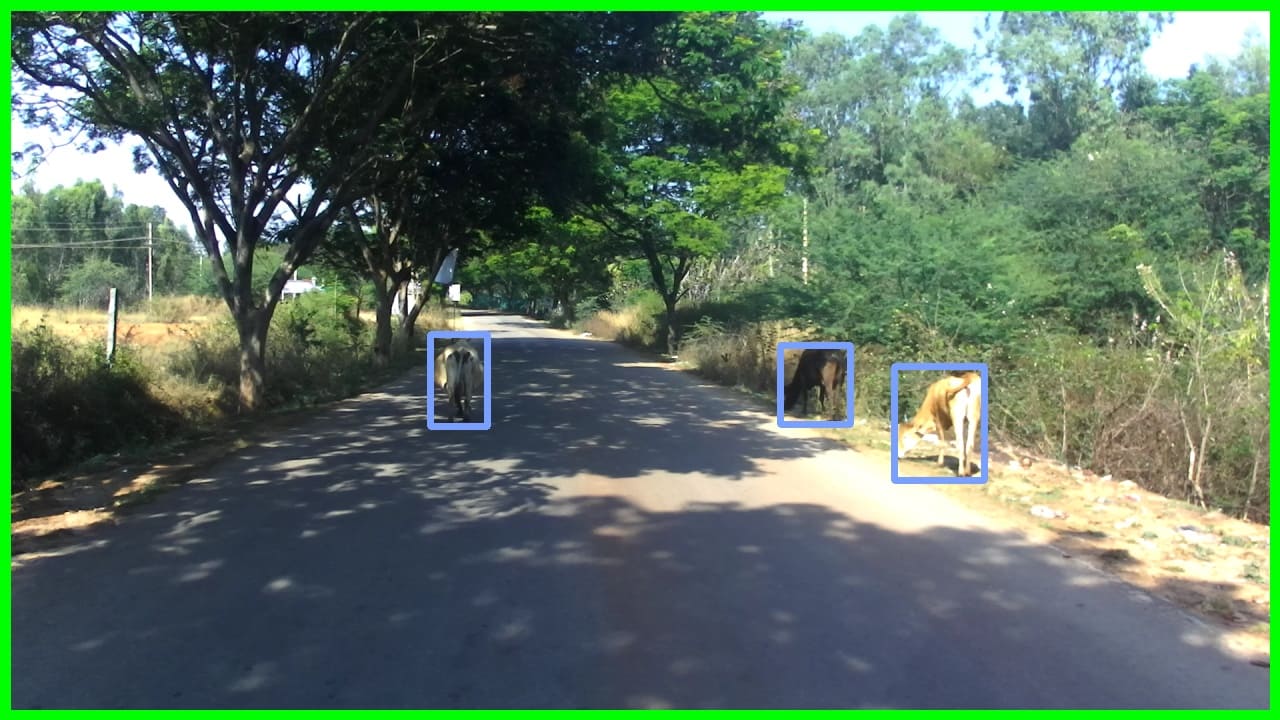}
    \end{subfigure}
    \begin{subfigure}[b]{\subfigwidth}
        \includegraphics[width=\textwidth]{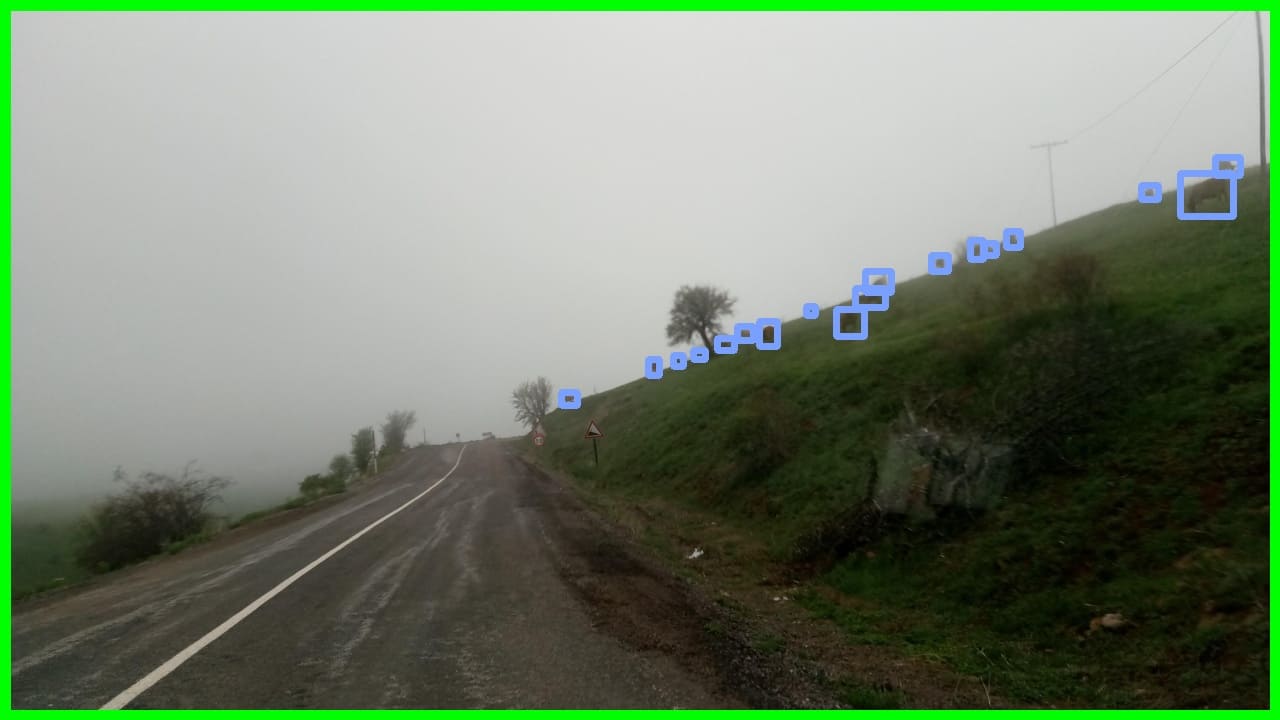}
    \end{subfigure}
    \begin{subfigure}[b]{\subfigwidth}
        \includegraphics[width=\textwidth]{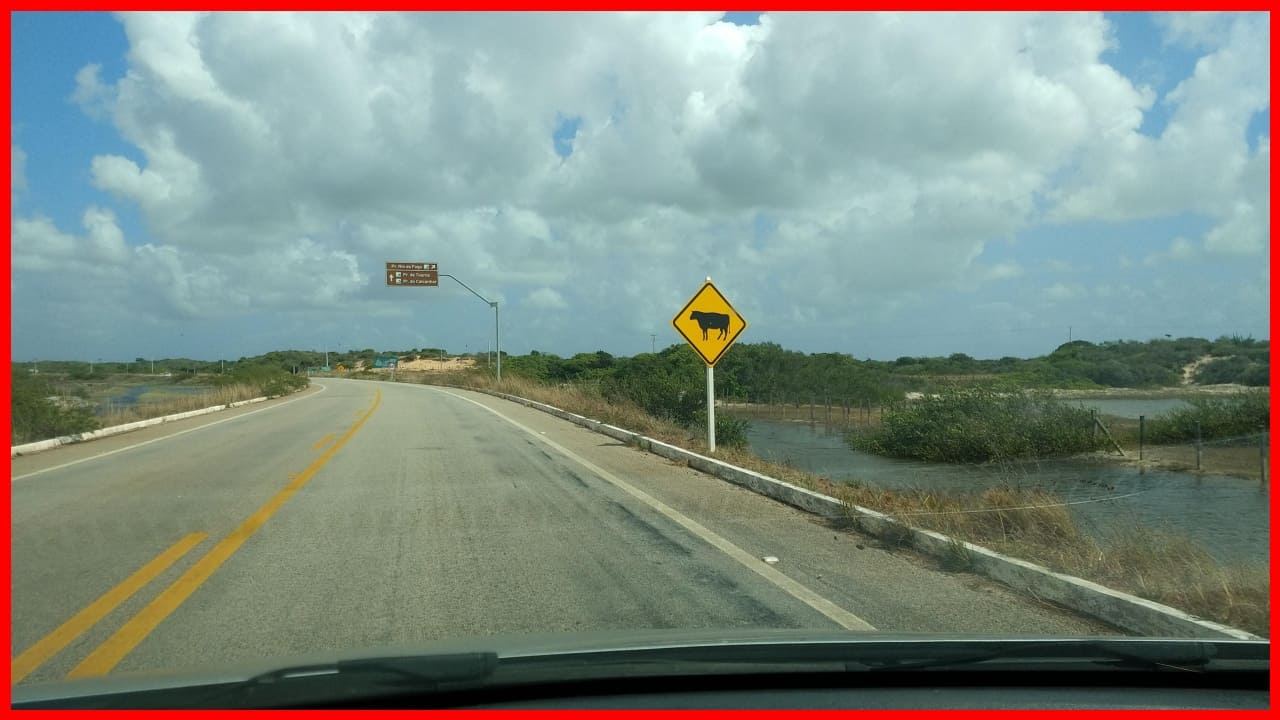}
    \end{subfigure} \\

    \rotatebox[origin=left]{90}{\hspace{0.04cm} \textbf{NACLIP}} 
    \begin{subfigure}[b]{\subfigwidth}
        \includegraphics[width=\textwidth]{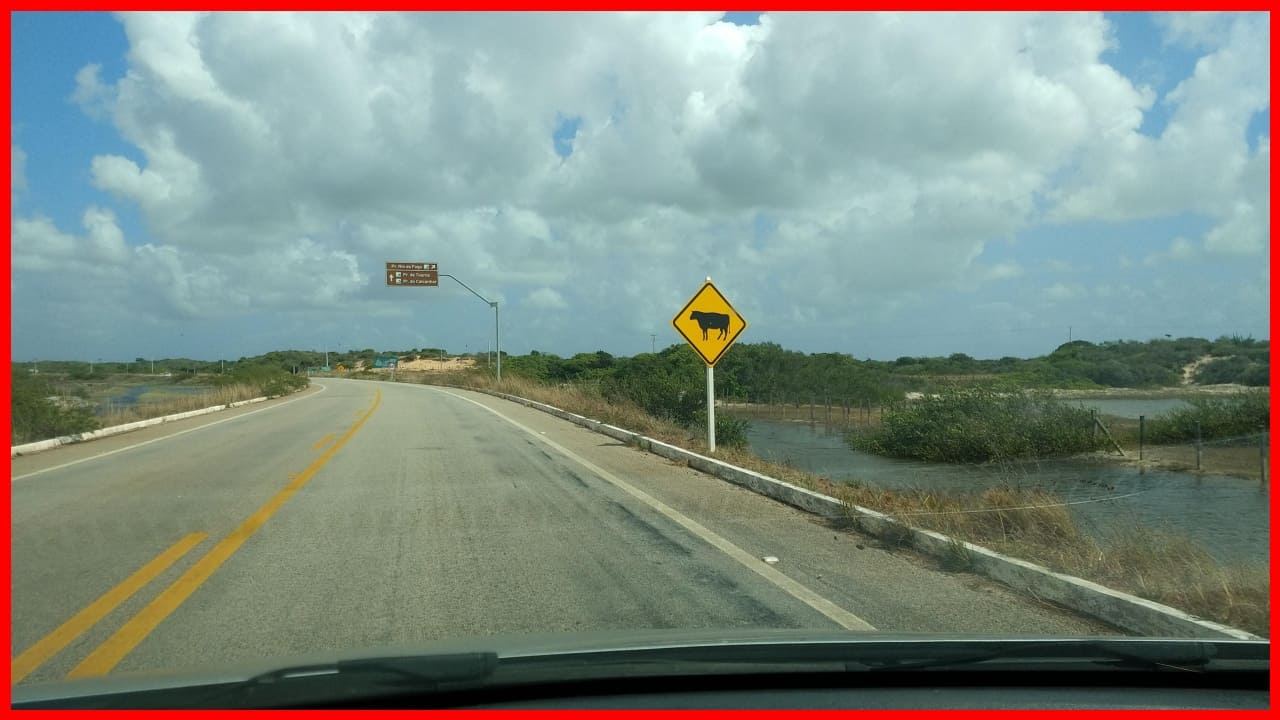}
    \end{subfigure}
    \begin{subfigure}[b]{\subfigwidth}
        \includegraphics[width=\textwidth]{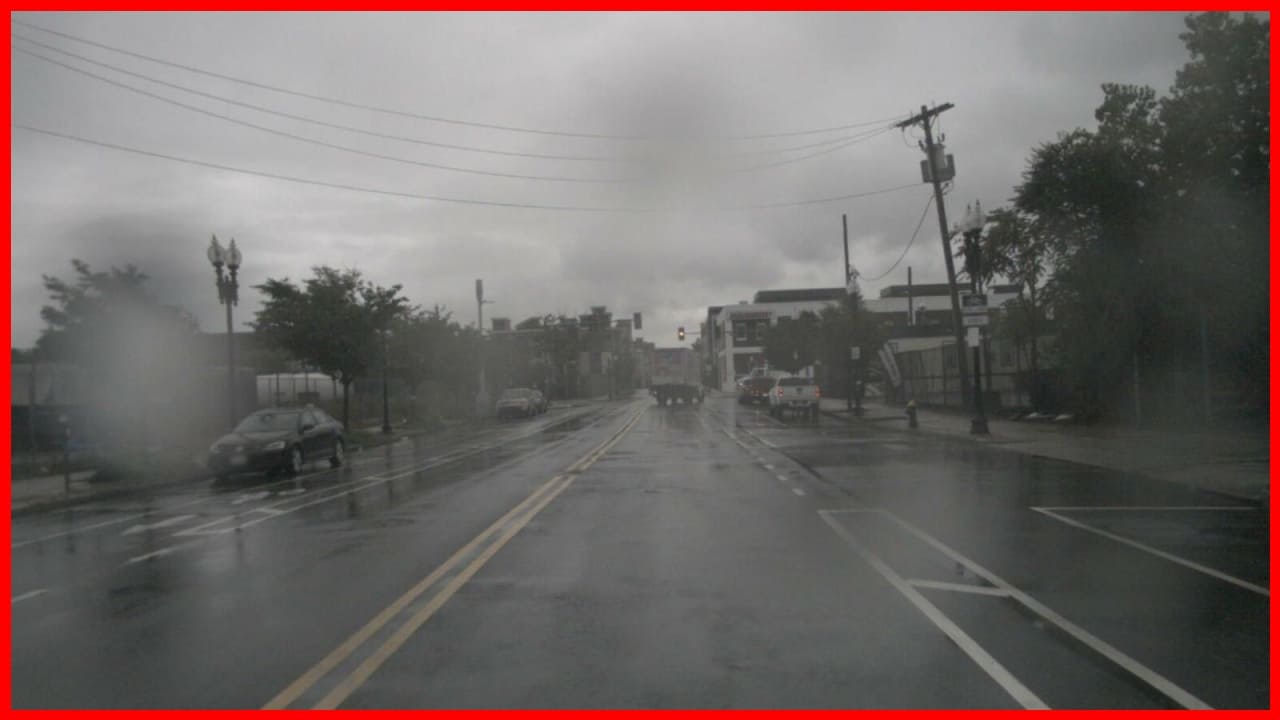}
    \end{subfigure}
    \begin{subfigure}[b]{\subfigwidth}
        \includegraphics[width=\textwidth]{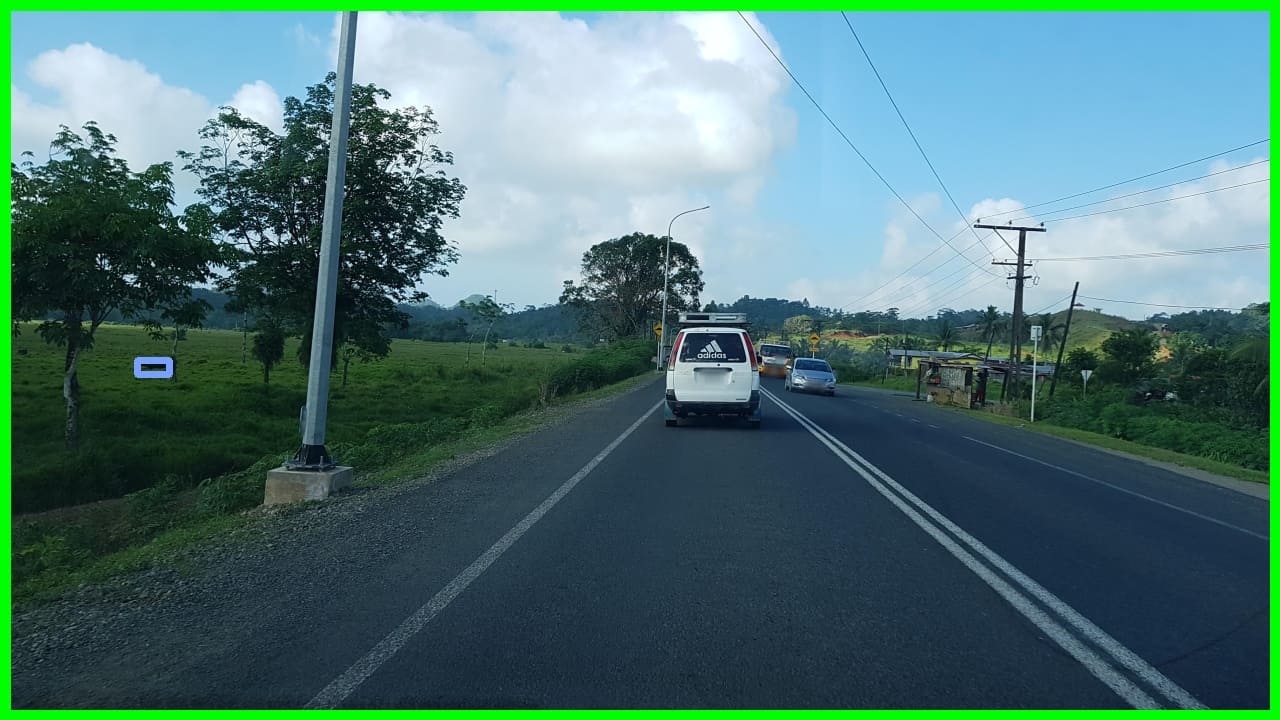}
    \end{subfigure}
    \begin{subfigure}[b]{\subfigwidth}
        \includegraphics[width=\textwidth]{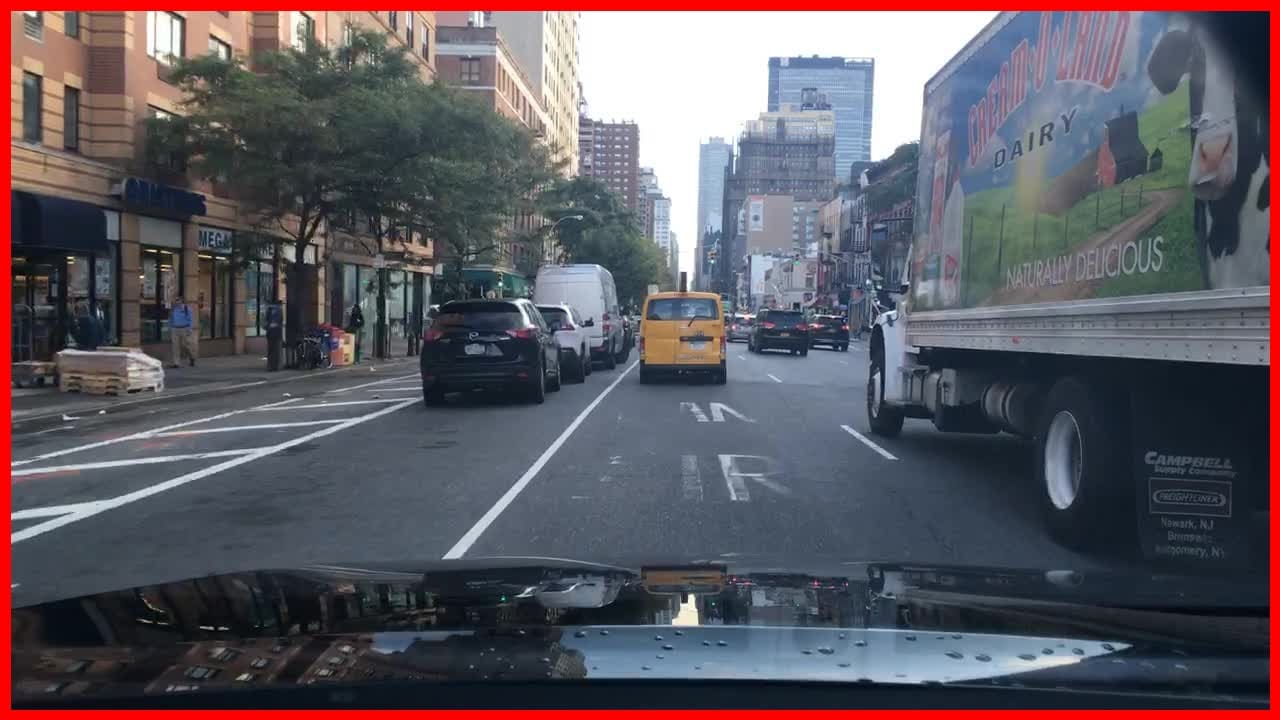}
    \end{subfigure}
    \begin{subfigure}[b]{\subfigwidth}
        \includegraphics[width=\textwidth]{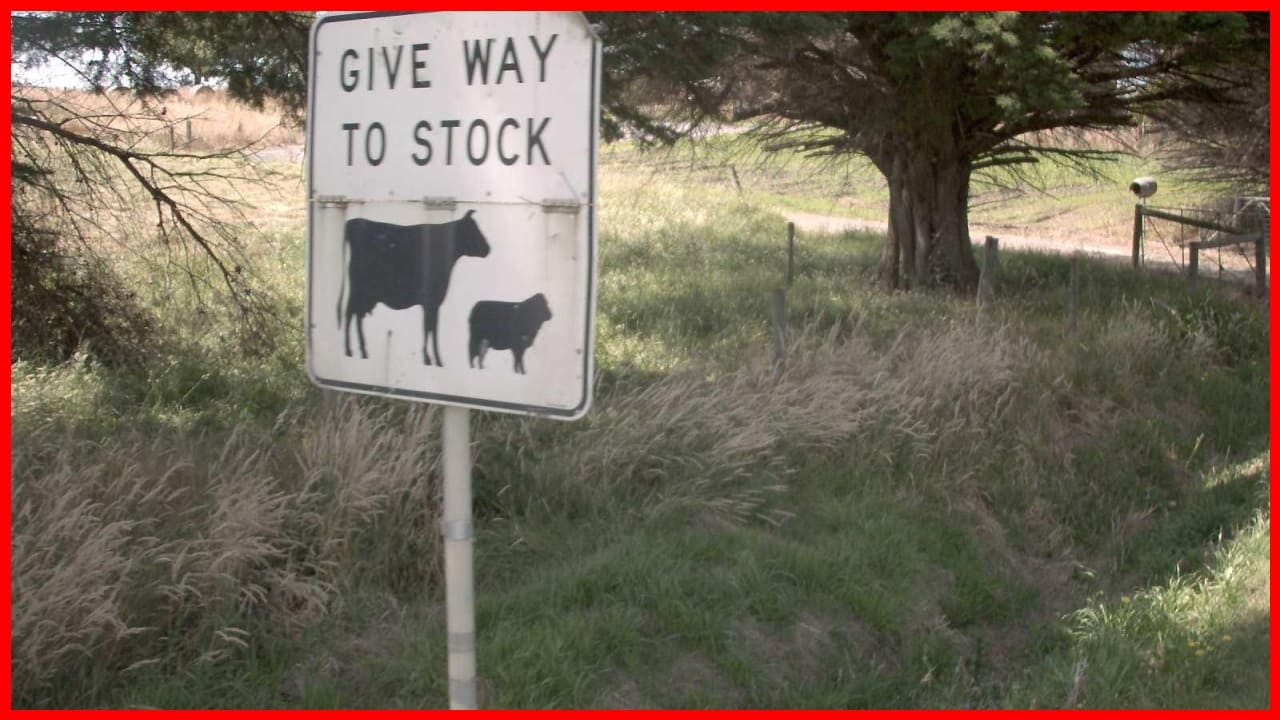}
    \end{subfigure} \\

    \rotatebox[origin=left]{90}{\hspace{-0.15cm} \textbf{NARADIO}} 
    \begin{subfigure}[b]{\subfigwidth}
        \includegraphics[width=\textwidth]{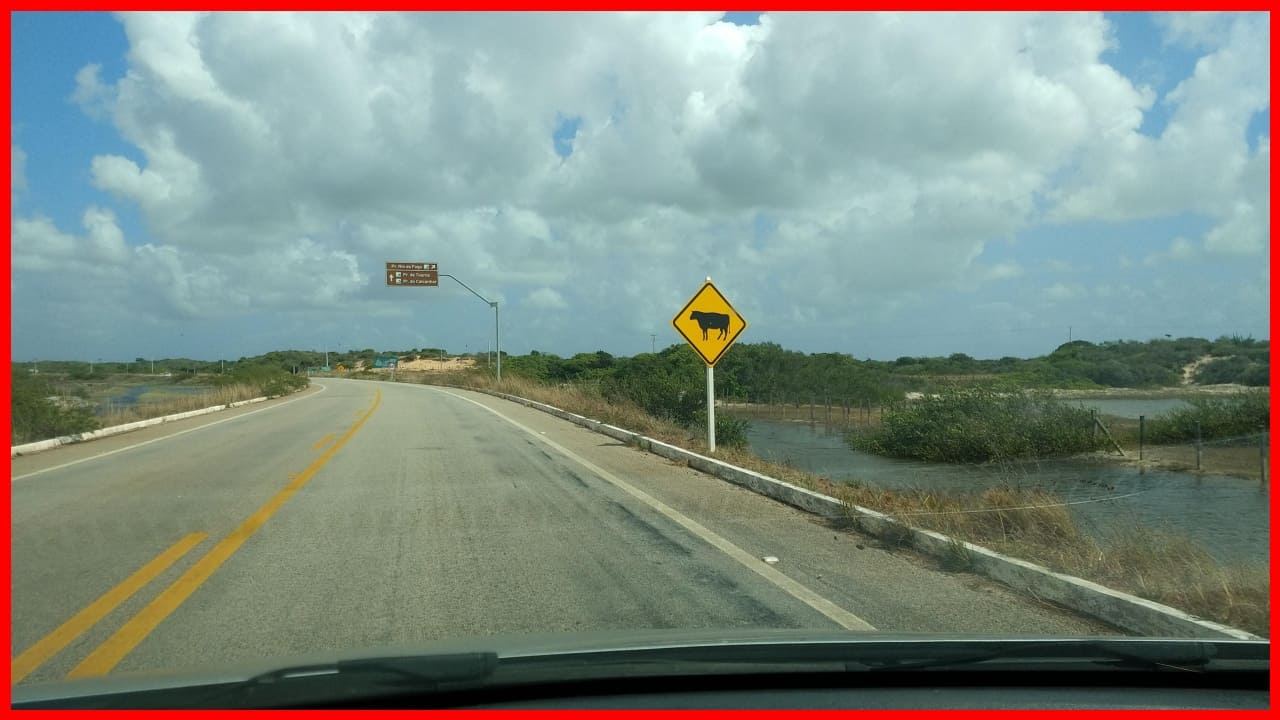}
    \end{subfigure}
    \begin{subfigure}[b]{\subfigwidth}
        \includegraphics[width=\textwidth]{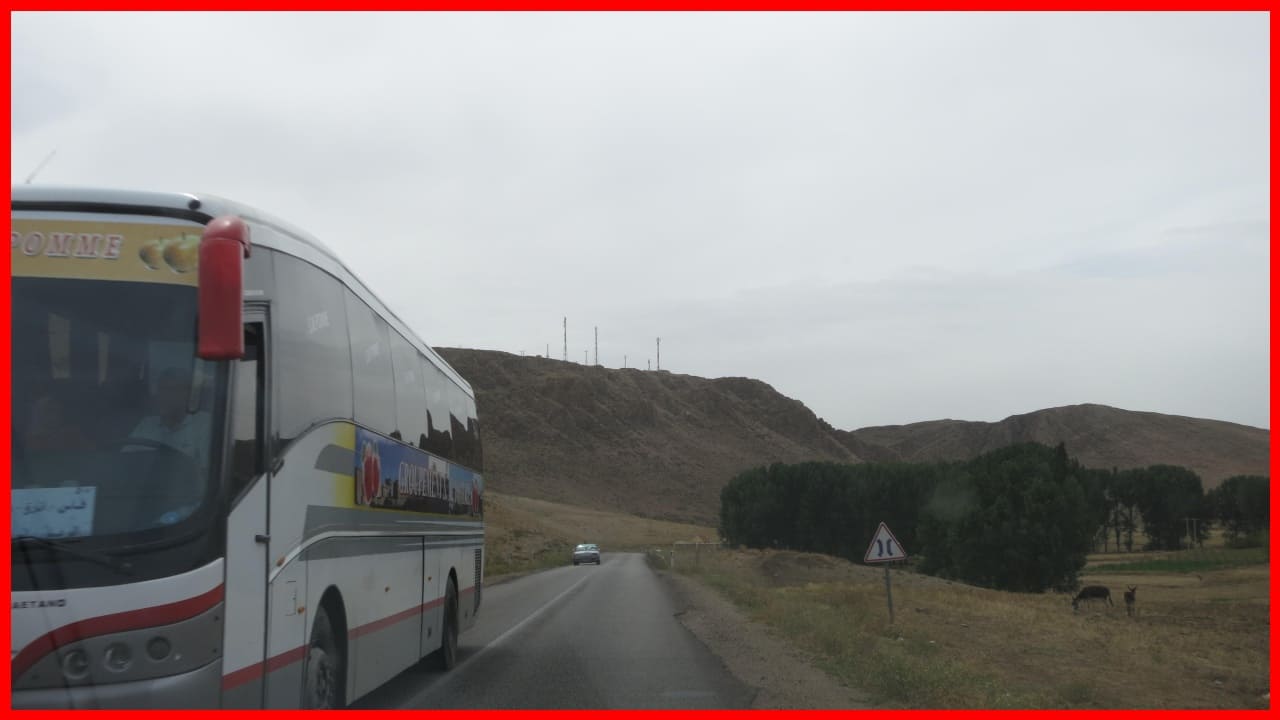}
    \end{subfigure}
    \begin{subfigure}[b]{\subfigwidth}
        \includegraphics[width=\textwidth]{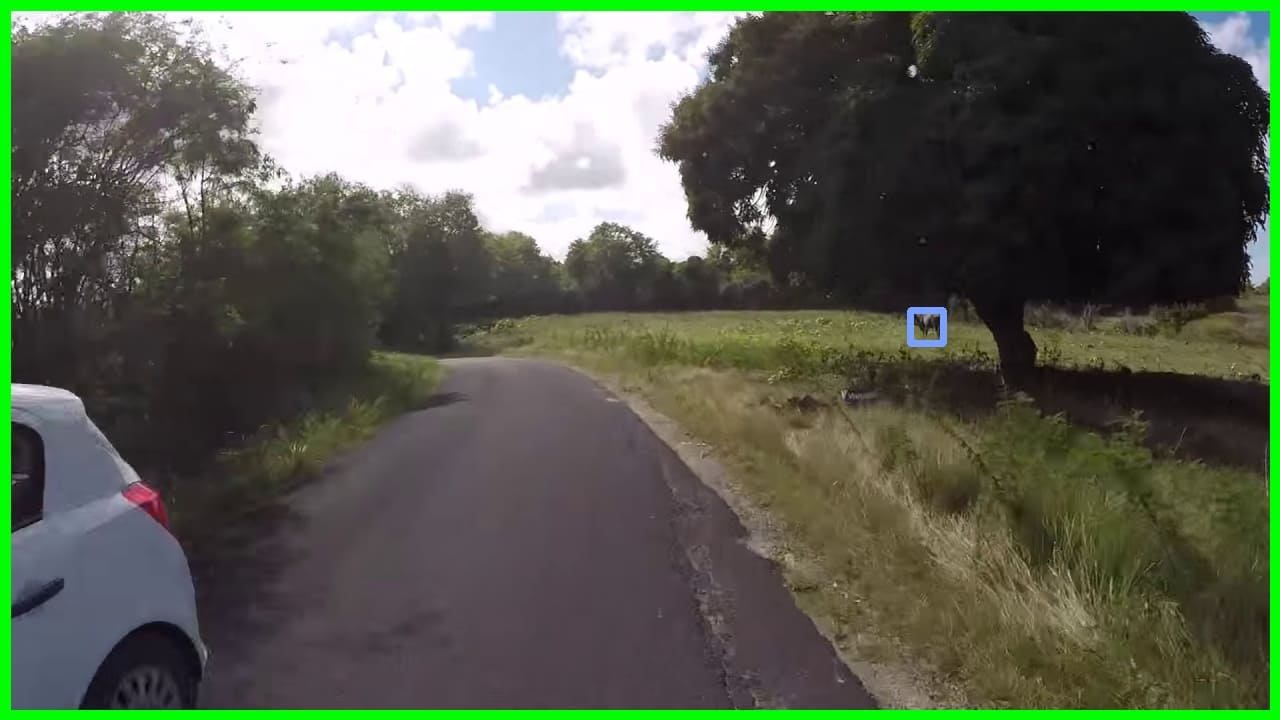}
    \end{subfigure}
    \begin{subfigure}[b]{\subfigwidth}
        \includegraphics[width=\textwidth]{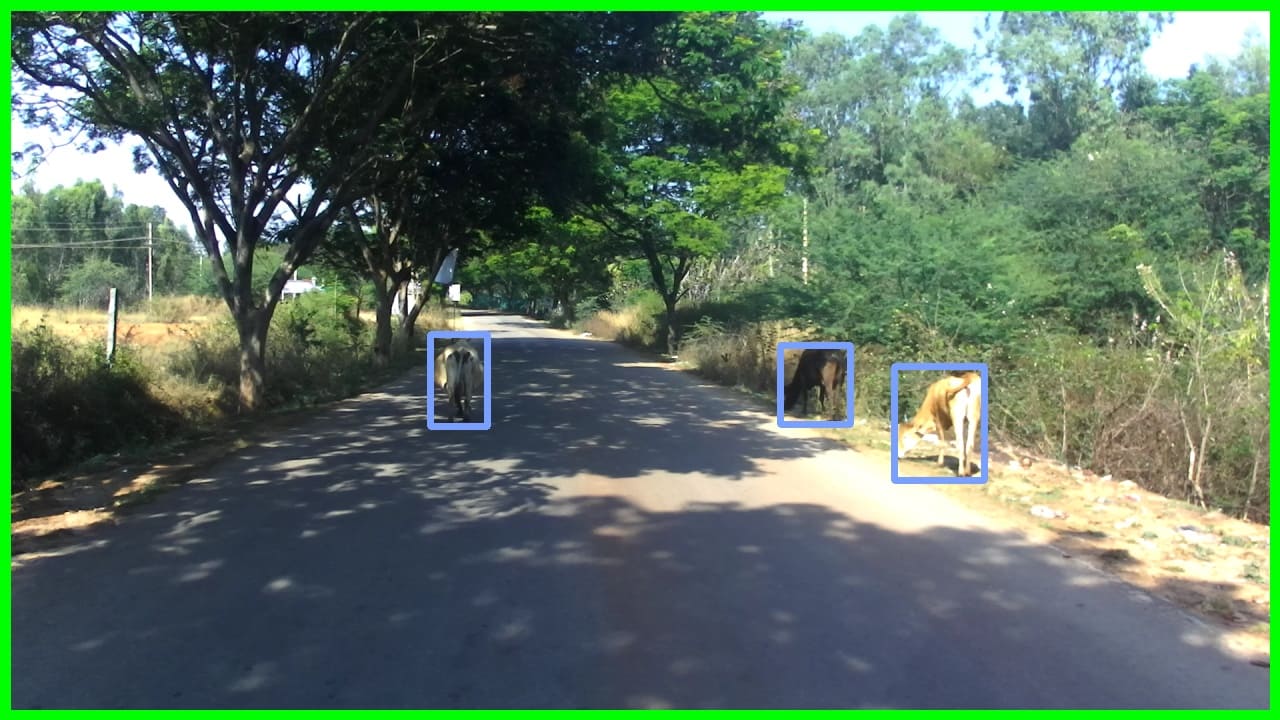}
    \end{subfigure}
    \begin{subfigure}[b]{\subfigwidth}
        \includegraphics[width=\textwidth]{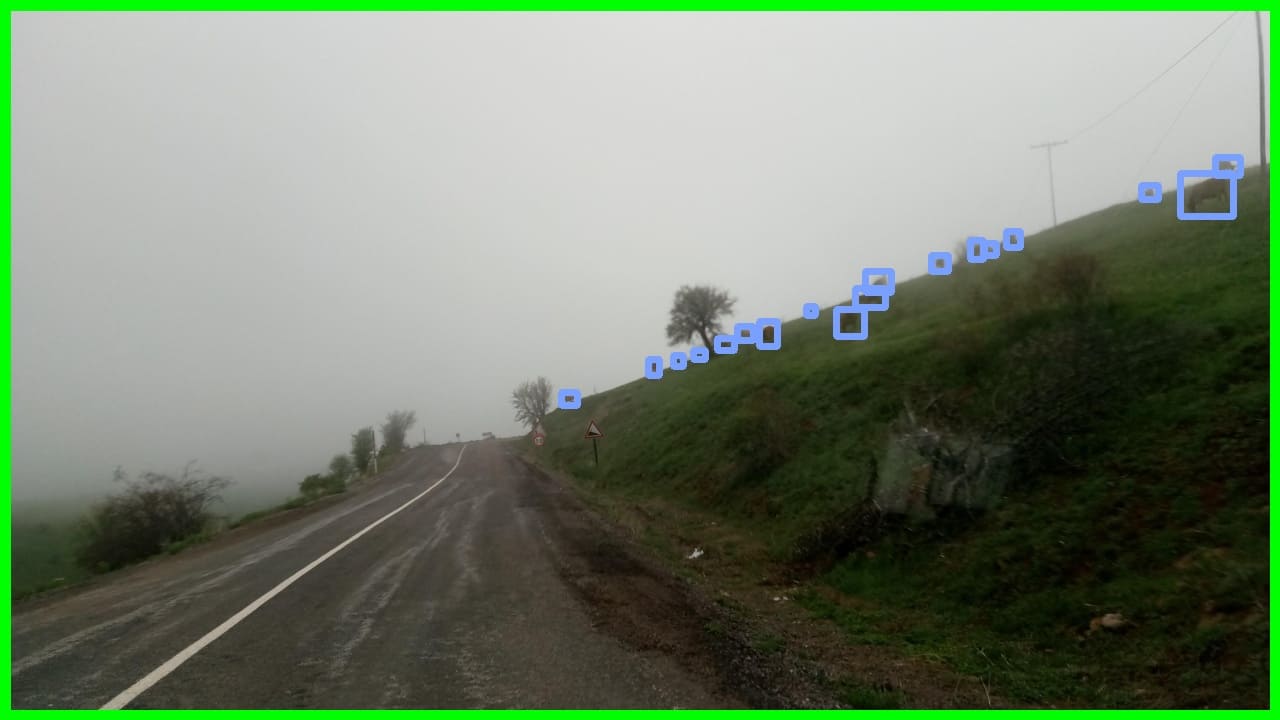}
    \end{subfigure} \\
    \caption{Text-based retrieval results on the validation split for the \emph{Animal-Real-Cow} class, showing top 5 ranked images for each model. Model references can be found in \Cref{tab:map_class_wise_text_to_image_retrieval}.}
    \label{fig:animal_cow_results}
    \endgroup
\end{figure*}
}

{
\def\rankOne{\bfseries Ranked 1st}
\def\rankTwo{\bfseries Ranked 2nd}
\def\rankThree{\bfseries Ranked 3rd}
\def\rankFour{\bfseries Ranked 4th}
\def\rankFive{\bfseries Ranked 5th}

\captionsetup[subfigure]{labelformat=empty}

\setlength{\subfigwidth}{0.18\textwidth} 

\begin{figure*}[ht]
    \begingroup
    \centering 
    \small
    \rotatebox[origin=left]{90}{\hspace{0.02cm} \textbf{GDINO}} 
    \begin{subfigure}[b]{\subfigwidth}
        \caption{\rankOne}
        \includegraphics[width=\textwidth]{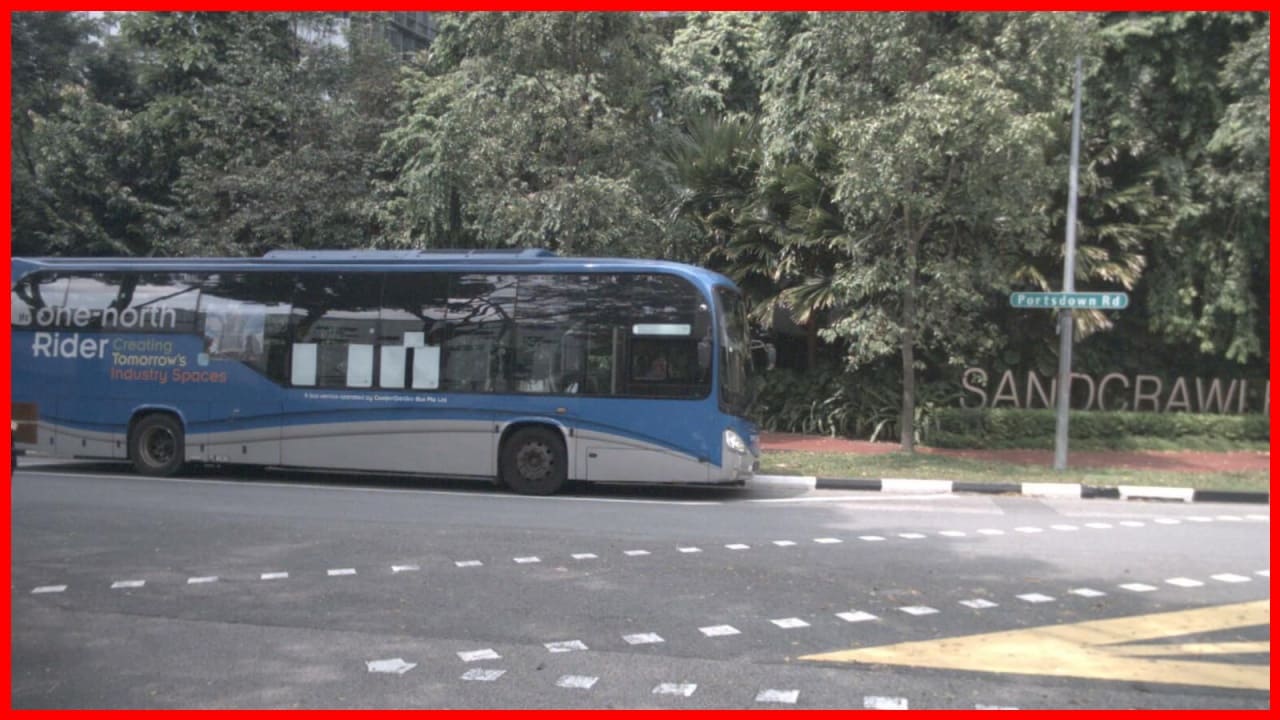}
    \end{subfigure}
    \begin{subfigure}[b]{\subfigwidth}
        \caption{\rankTwo}
        \includegraphics[width=\textwidth]{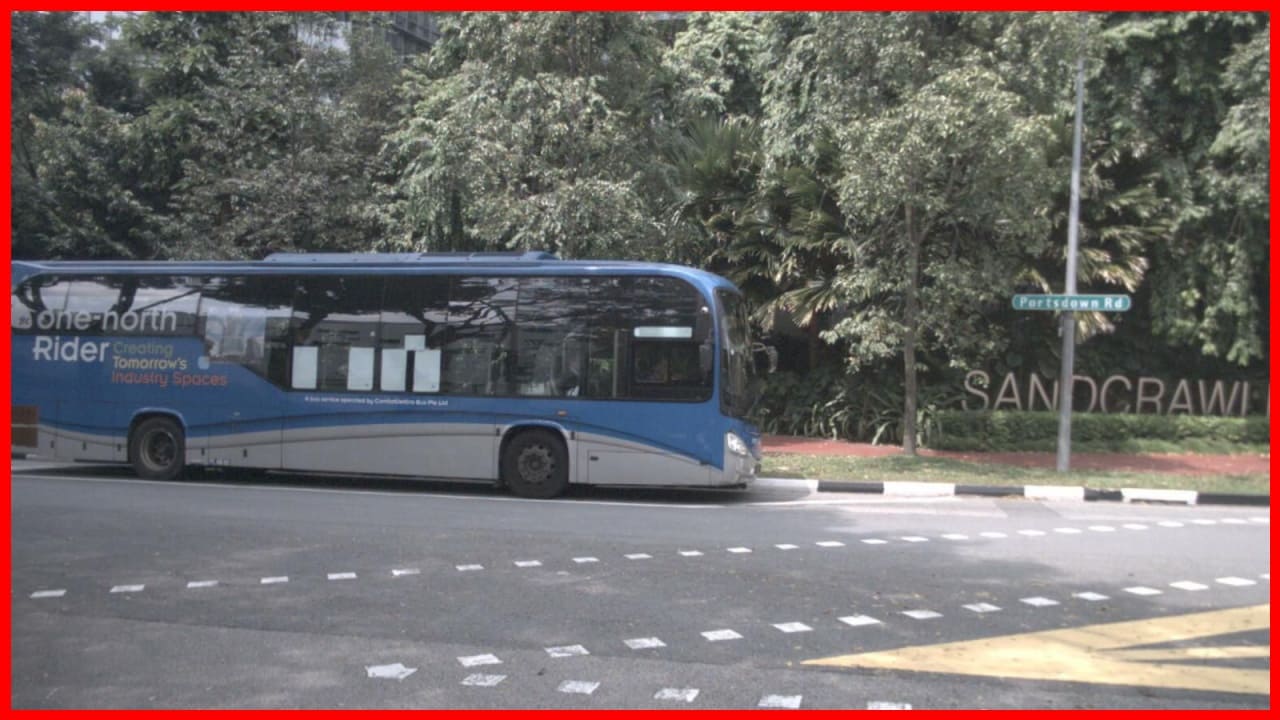}
    \end{subfigure}
    \begin{subfigure}[b]{\subfigwidth}
        \caption{\rankThree}
        \includegraphics[width=\textwidth]{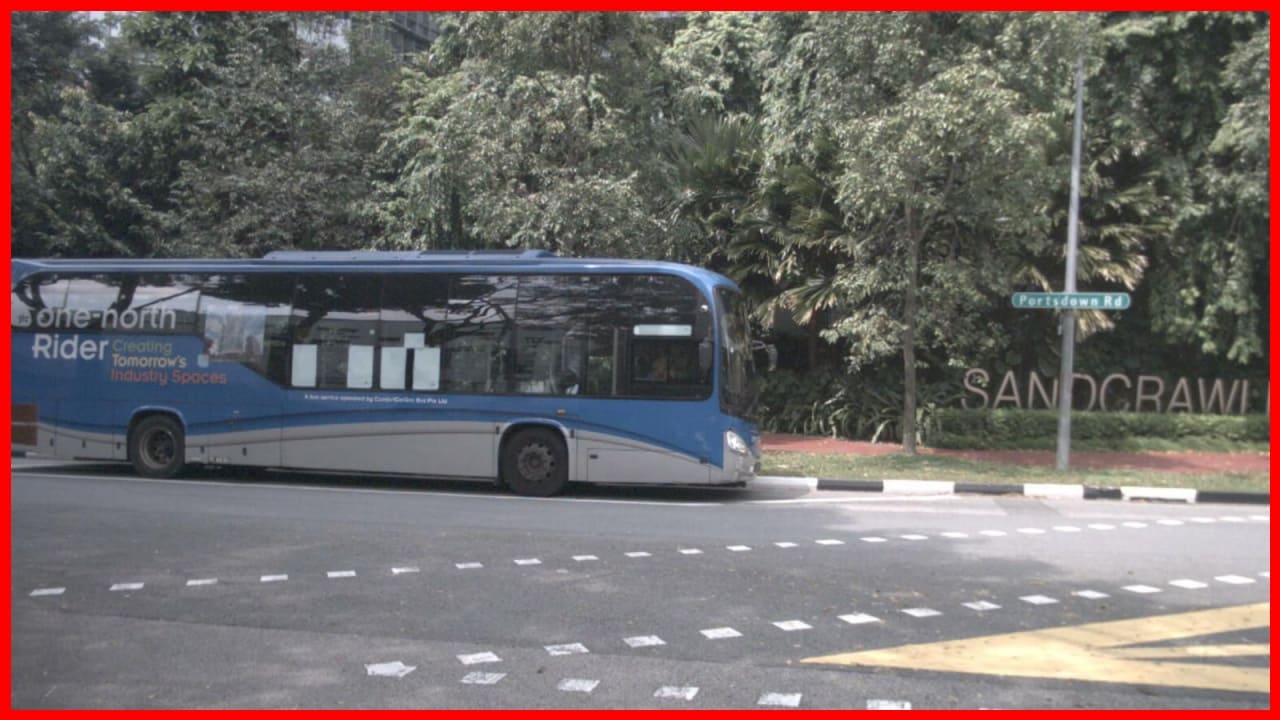}
    \end{subfigure}
    \begin{subfigure}[b]{\subfigwidth}
        \caption{\rankFour}
        \includegraphics[width=\textwidth]{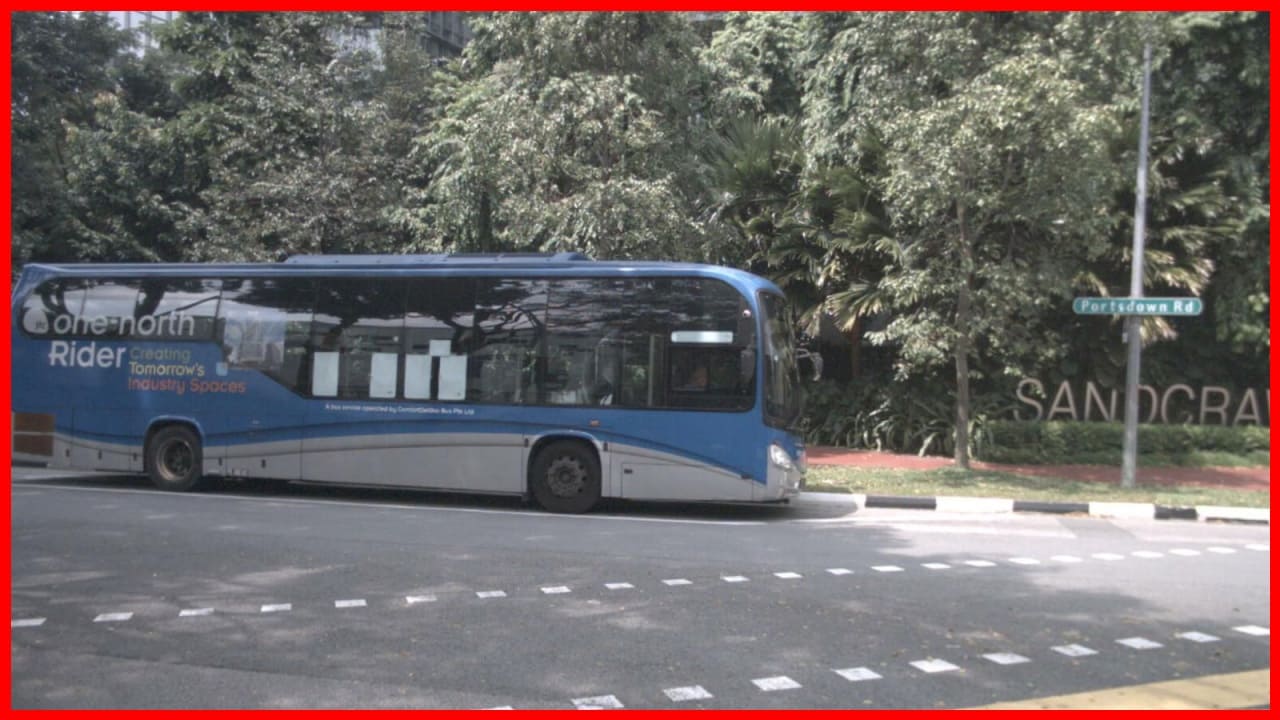}
    \end{subfigure}
    \begin{subfigure}[b]{\subfigwidth}
        \caption{\rankFive}
        \includegraphics[width=\textwidth]{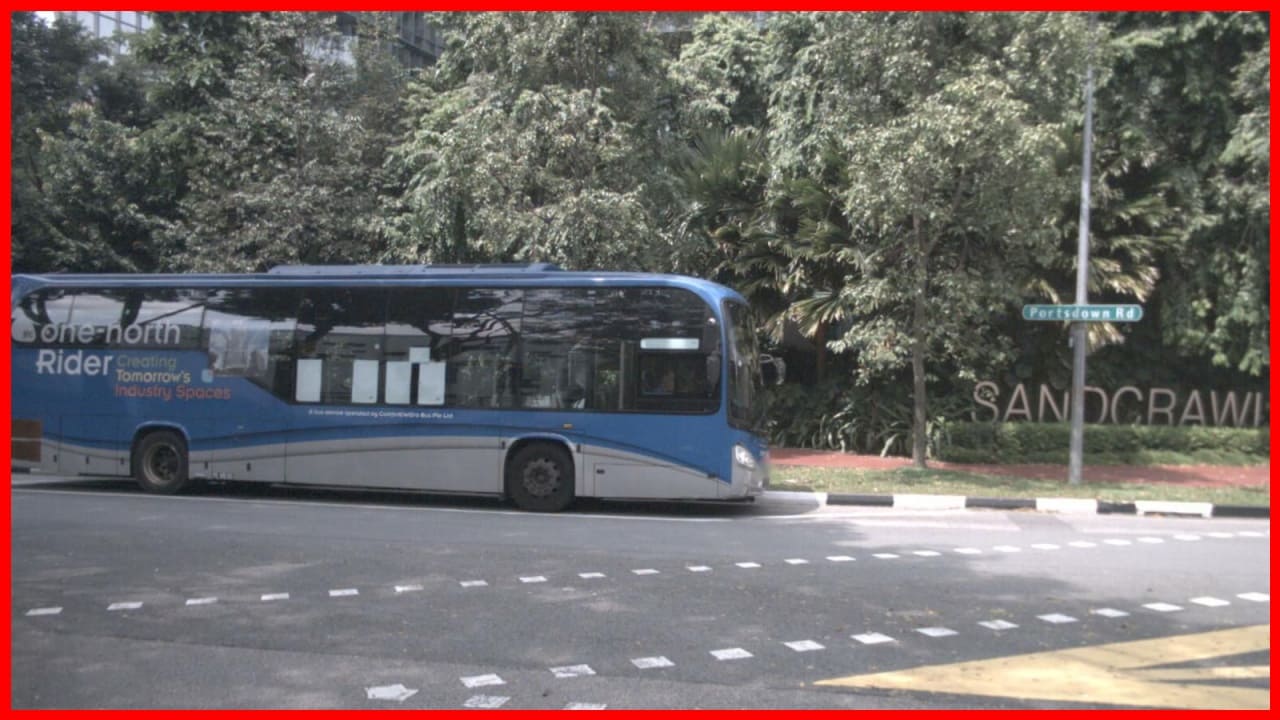}
    \end{subfigure} \\
    
    \rotatebox[origin=left]{90}{\hspace{0.17cm} \textbf{CLIP}} 
    \begin{subfigure}[b]{\subfigwidth}
        \includegraphics[width=\textwidth]{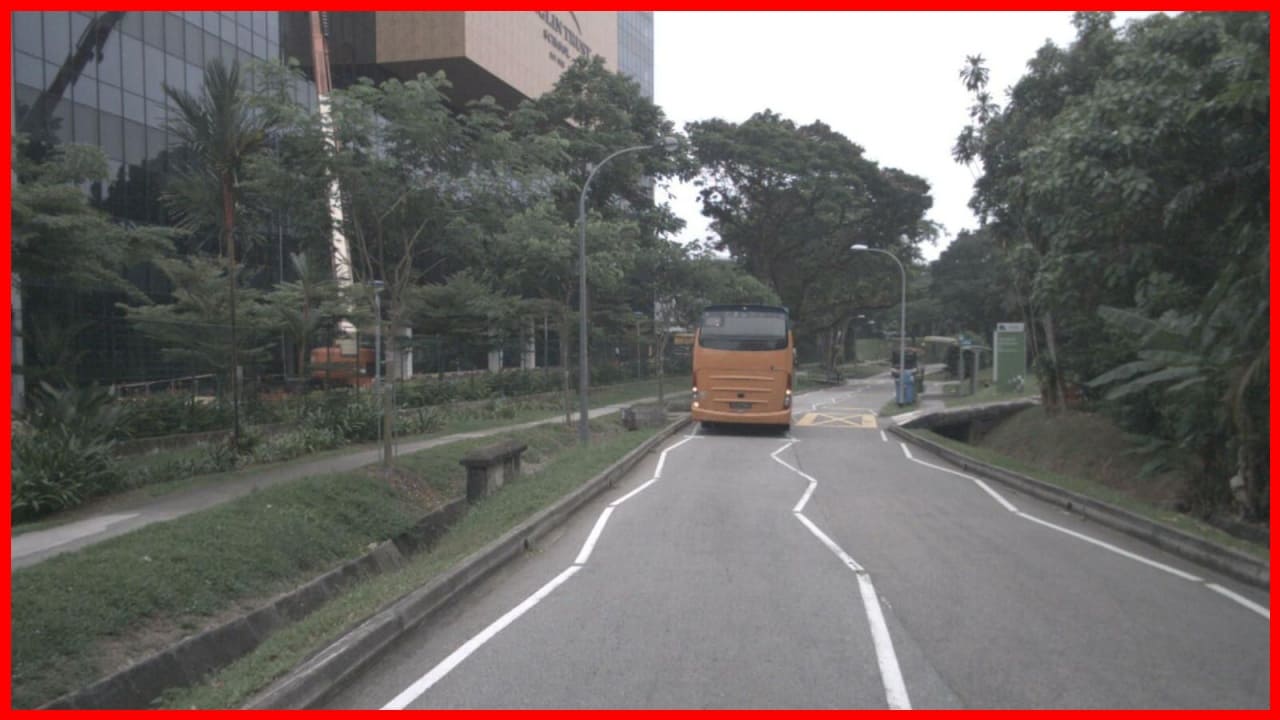}
    \end{subfigure}
    \begin{subfigure}[b]{\subfigwidth}
        \includegraphics[width=\textwidth]{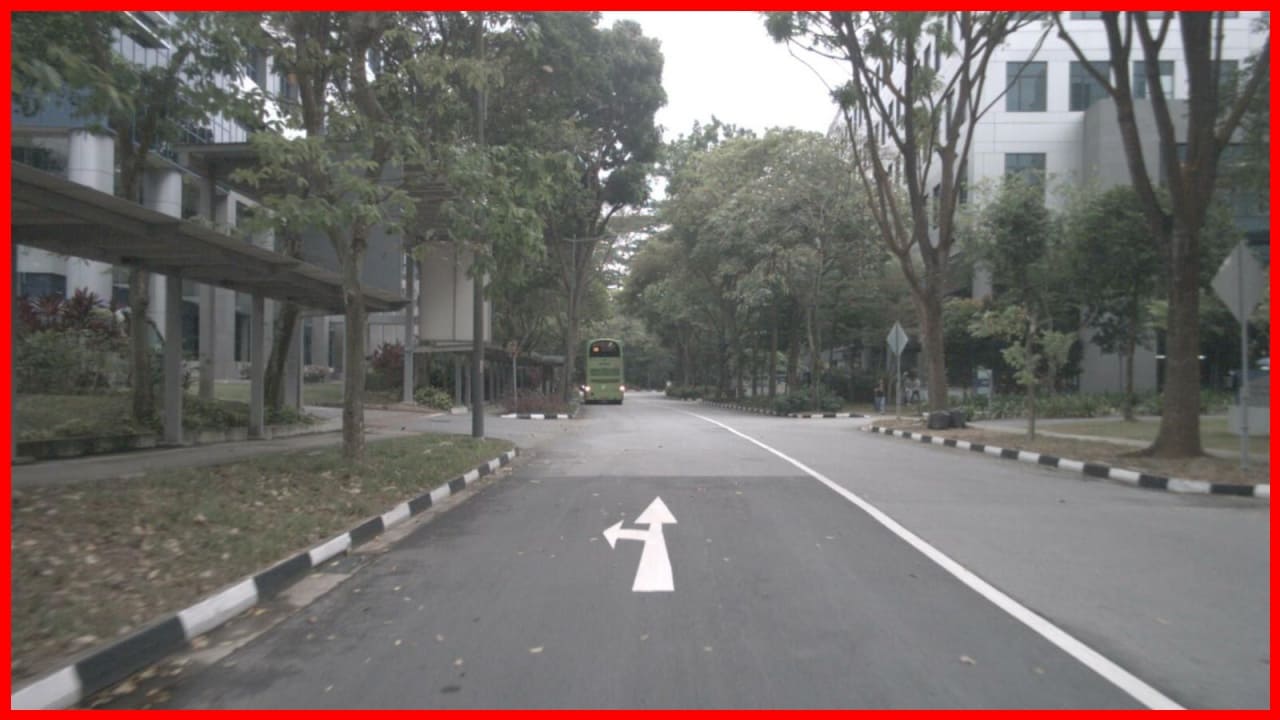}
    \end{subfigure}
    \begin{subfigure}[b]{\subfigwidth}
        \includegraphics[width=\textwidth]{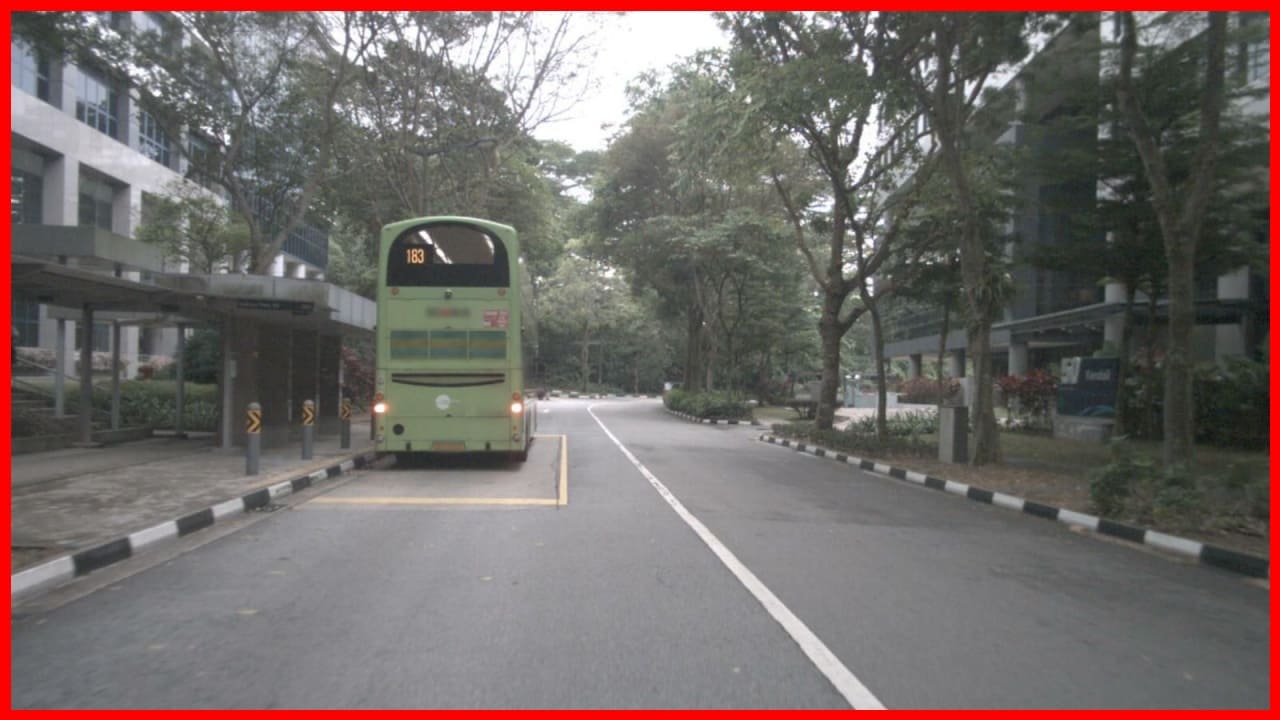}
    \end{subfigure}
    \begin{subfigure}[b]{\subfigwidth}
        \includegraphics[width=\textwidth]{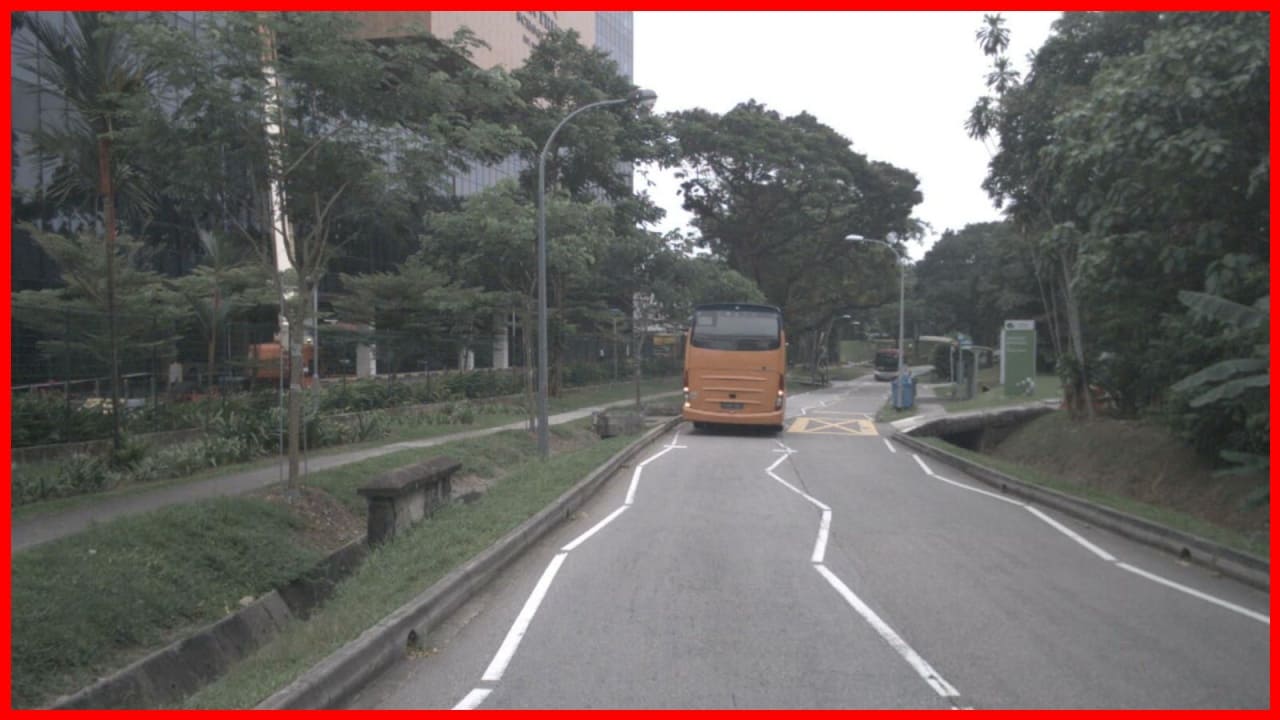}
    \end{subfigure}
    \begin{subfigure}[b]{\subfigwidth}
        \includegraphics[width=\textwidth]{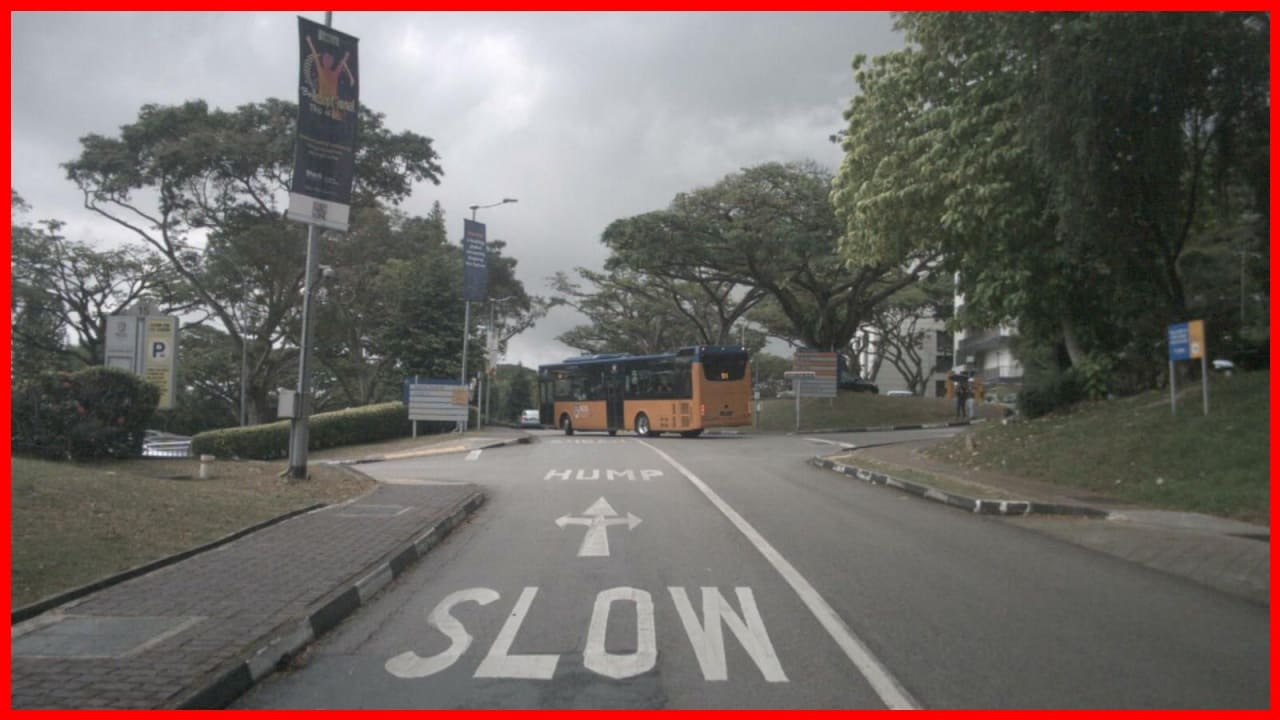}
    \end{subfigure} \\

    \rotatebox[origin=left]{90}{\hspace{0.07cm} \textbf{RADIO}} 
    \begin{subfigure}[b]{\subfigwidth}
        \includegraphics[width=\textwidth]{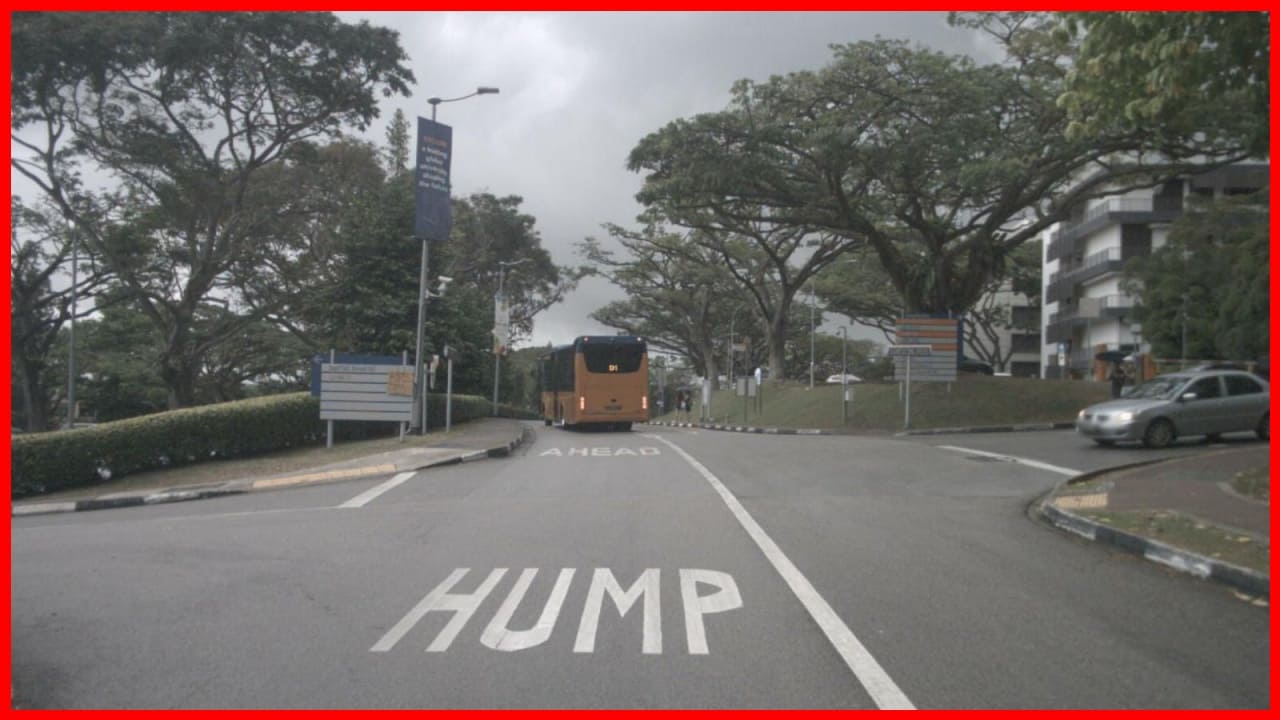}
    \end{subfigure}
    \begin{subfigure}[b]{\subfigwidth}
        \includegraphics[width=\textwidth]{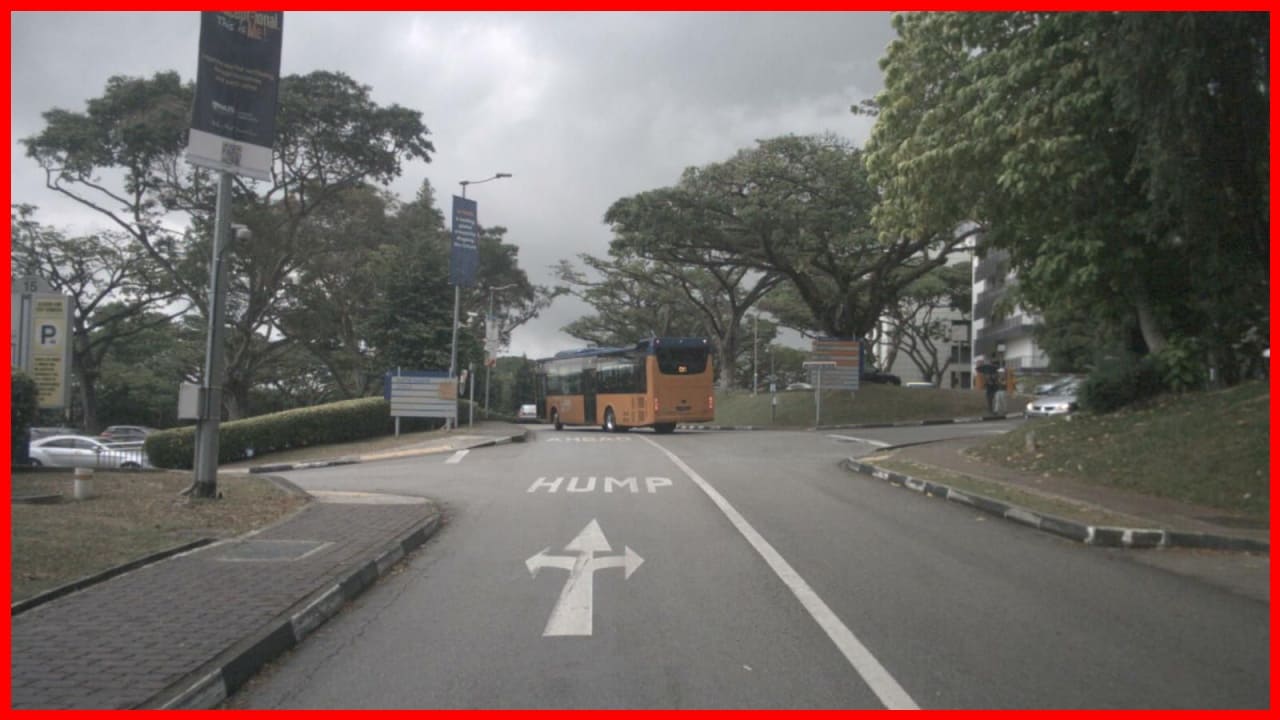}
    \end{subfigure}
    \begin{subfigure}[b]{\subfigwidth}
        \includegraphics[width=\textwidth]{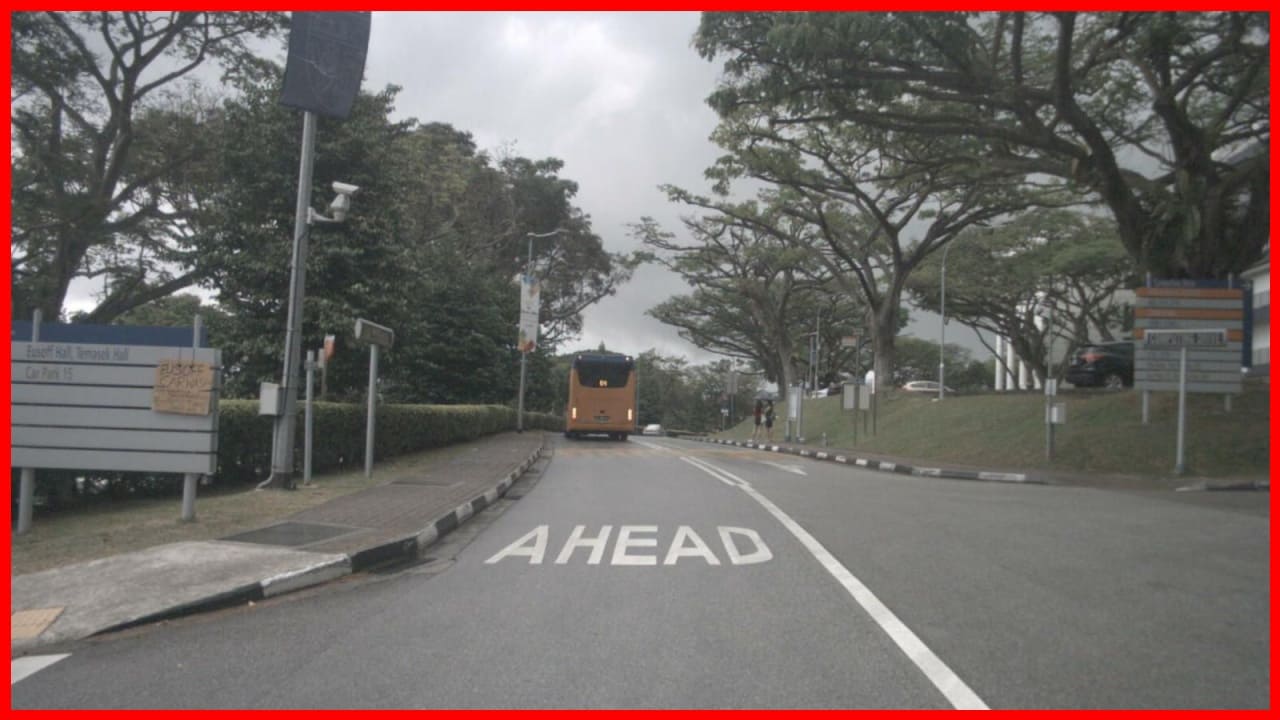}
    \end{subfigure}
    \begin{subfigure}[b]{\subfigwidth}
        \includegraphics[width=\textwidth]{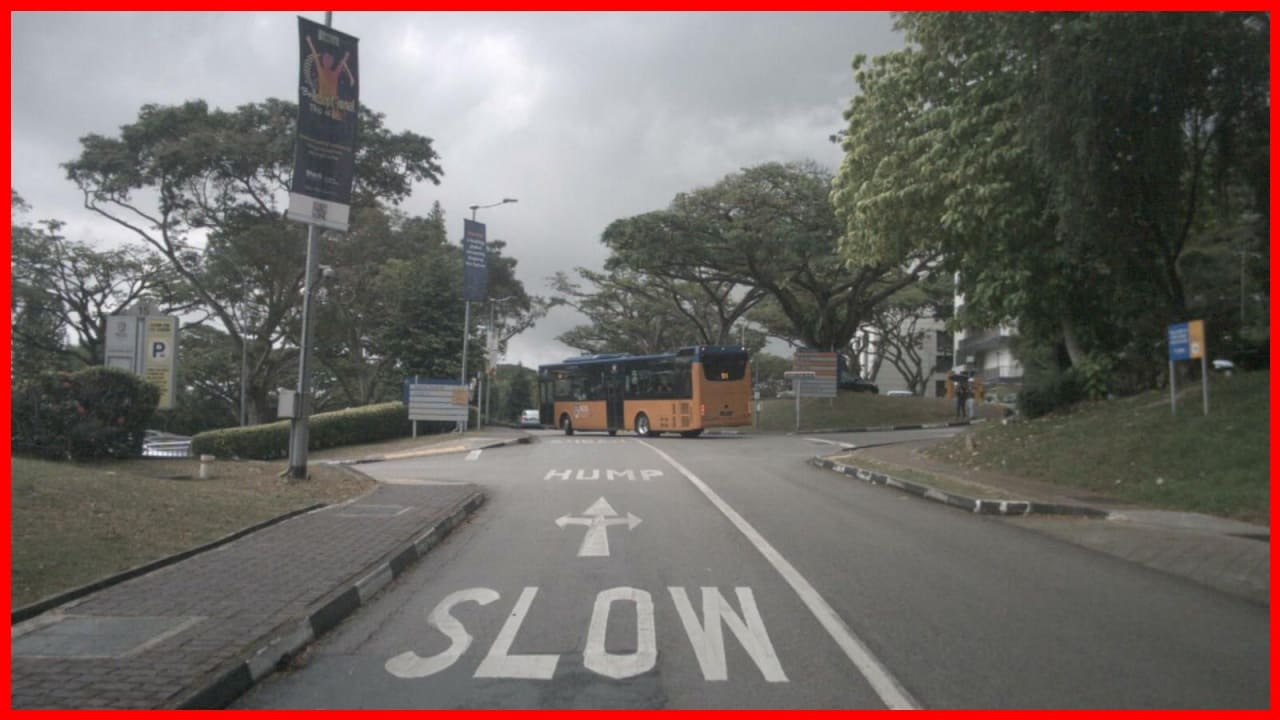}
    \end{subfigure}
    \begin{subfigure}[b]{\subfigwidth}
        \includegraphics[width=\textwidth]{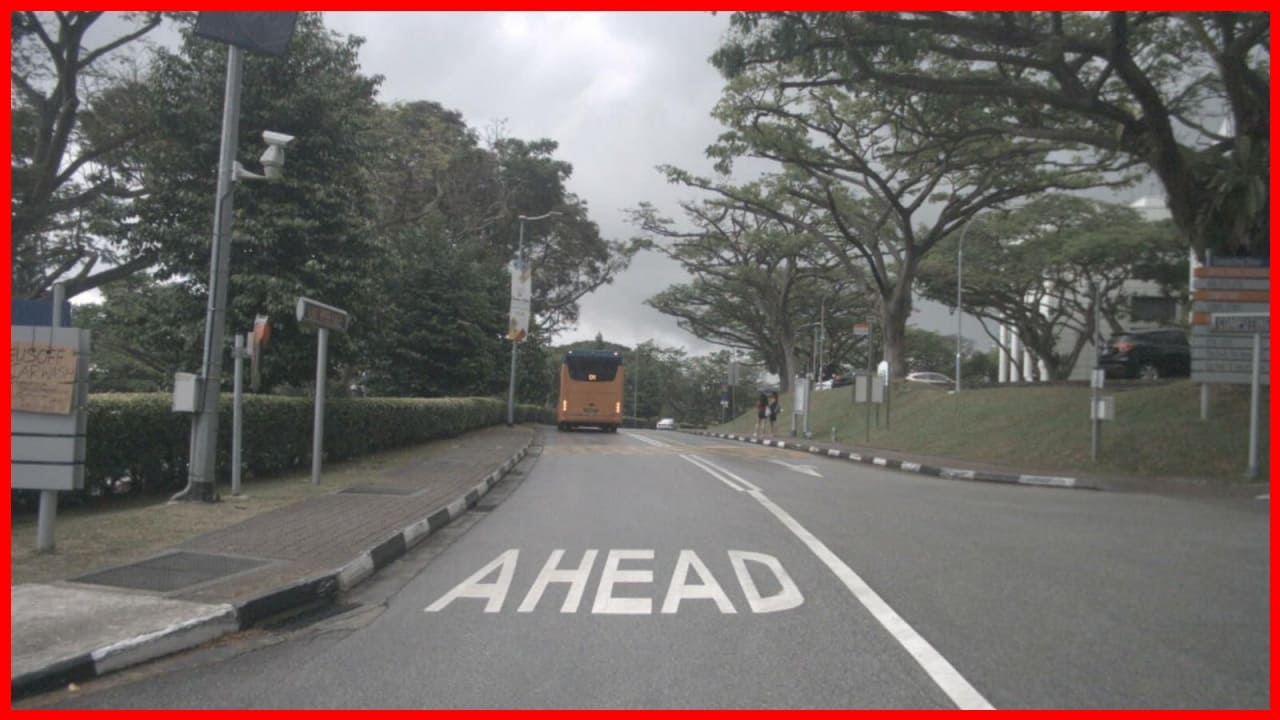}
    \end{subfigure} \\

    \rotatebox[origin=left]{90}{\hspace{0.15cm} \textbf{BLIP2}} 
    \begin{subfigure}[b]{\subfigwidth}
        \includegraphics[width=\textwidth]{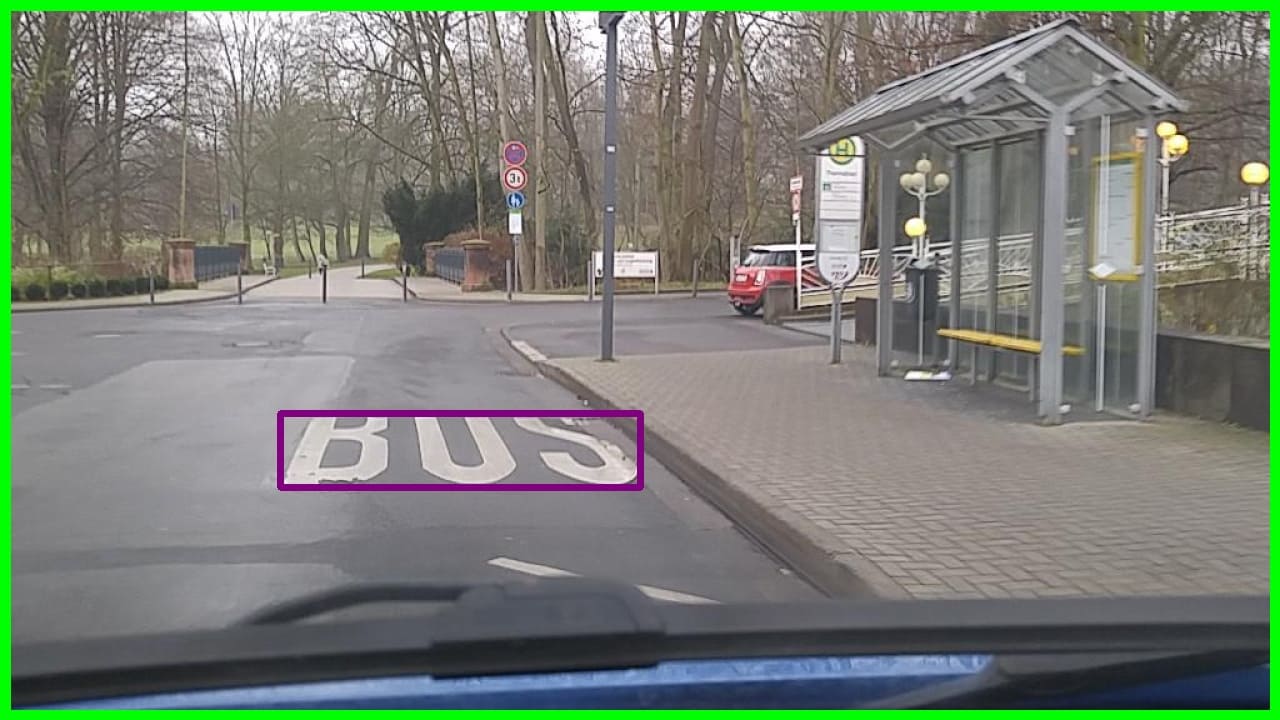}
    \end{subfigure}
    \begin{subfigure}[b]{\subfigwidth}
        \includegraphics[width=\textwidth]{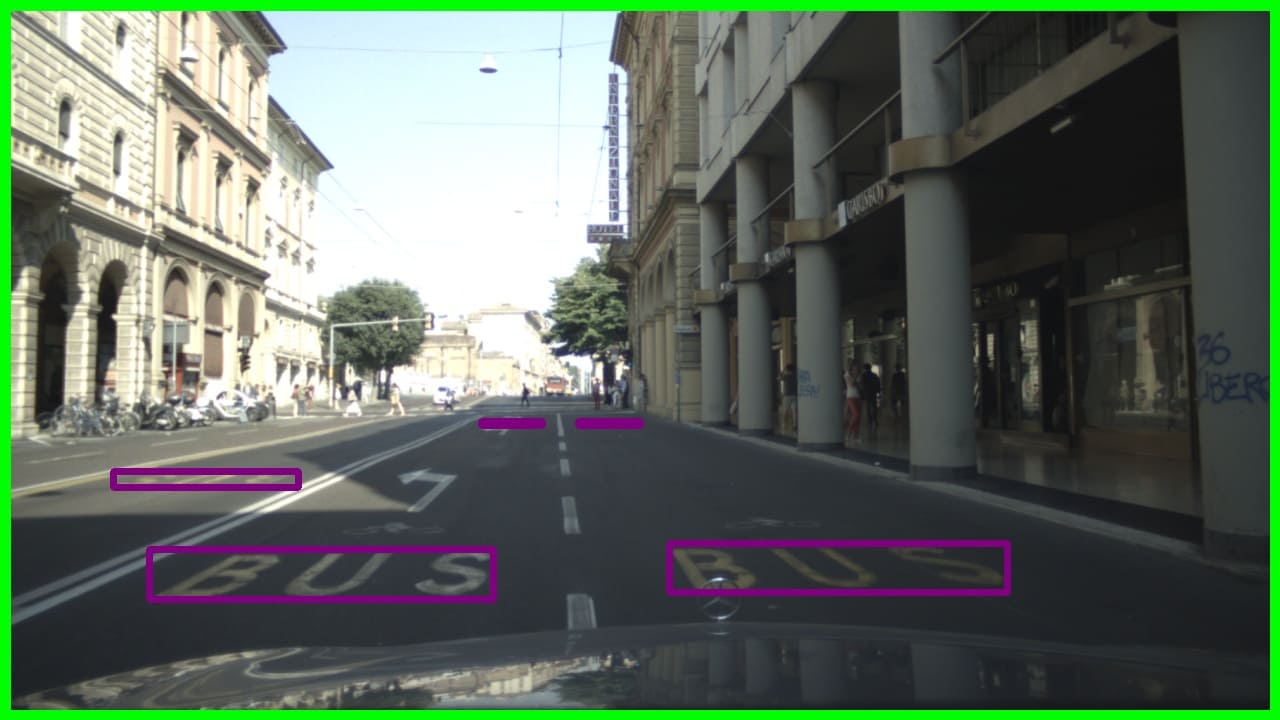}
    \end{subfigure}
    \begin{subfigure}[b]{\subfigwidth}
        \includegraphics[width=\textwidth]{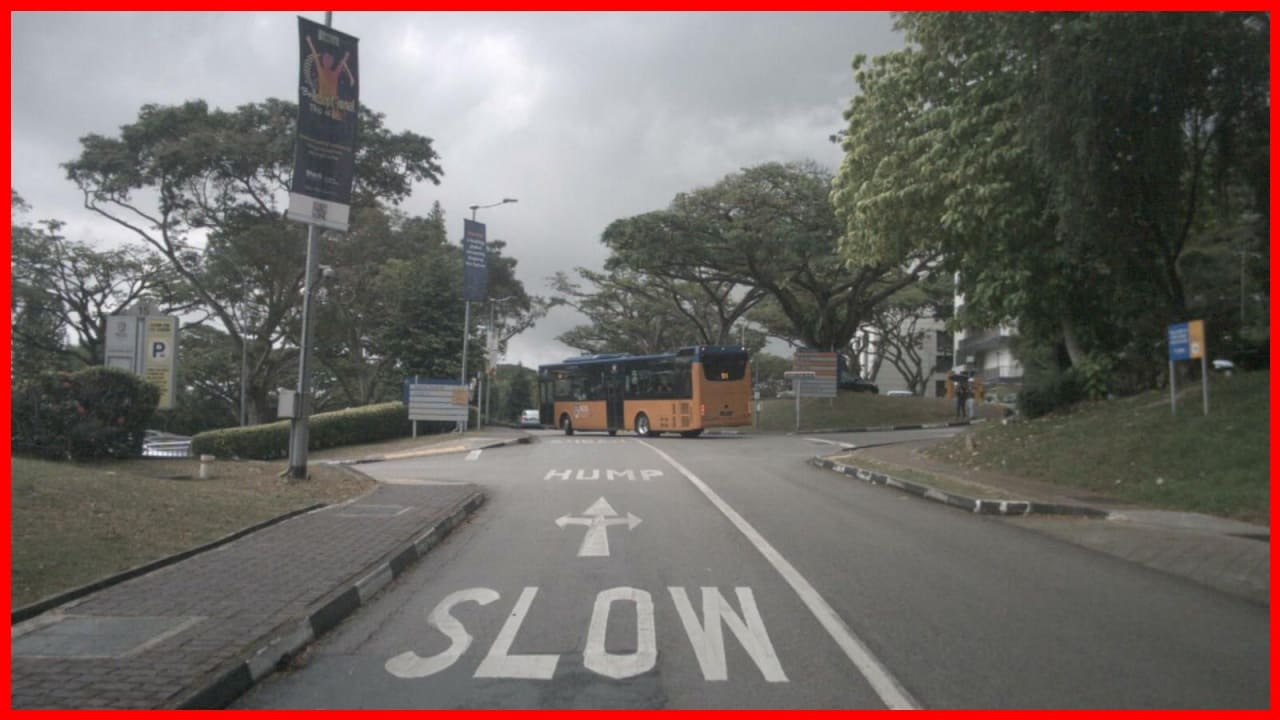}
    \end{subfigure}
    \begin{subfigure}[b]{\subfigwidth}
        \includegraphics[width=\textwidth]{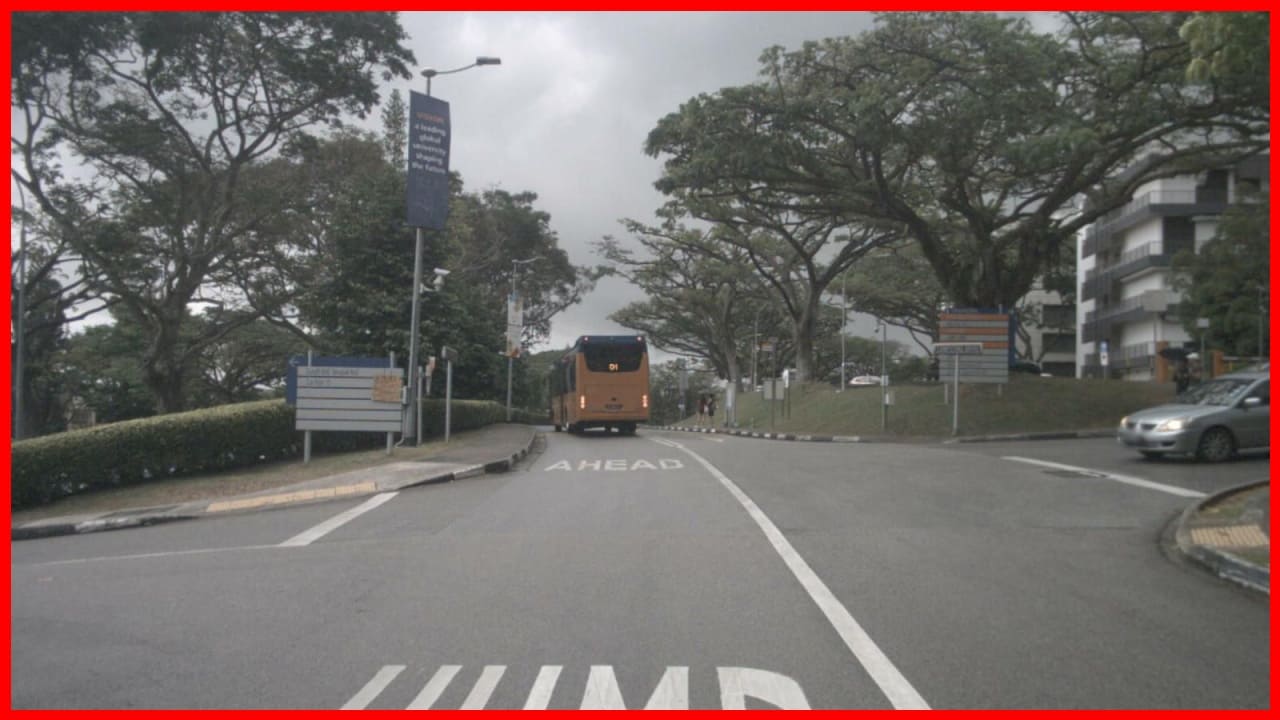}
    \end{subfigure}
    \begin{subfigure}[b]{\subfigwidth}
        \includegraphics[width=\textwidth]{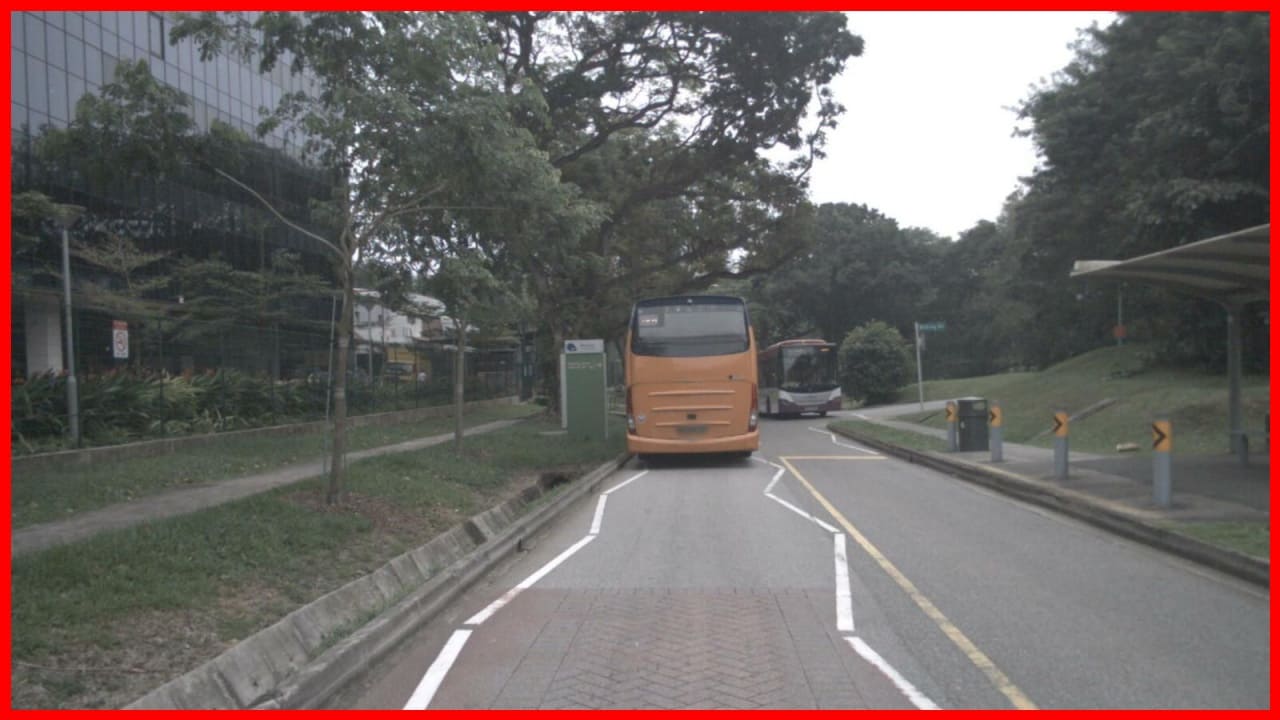}
    \end{subfigure} \\

    \rotatebox[origin=left]{90}{\hspace{-0.29cm} \textbf{METACLIP2}} 
    \begin{subfigure}[b]{\subfigwidth}
        \includegraphics[width=\textwidth]{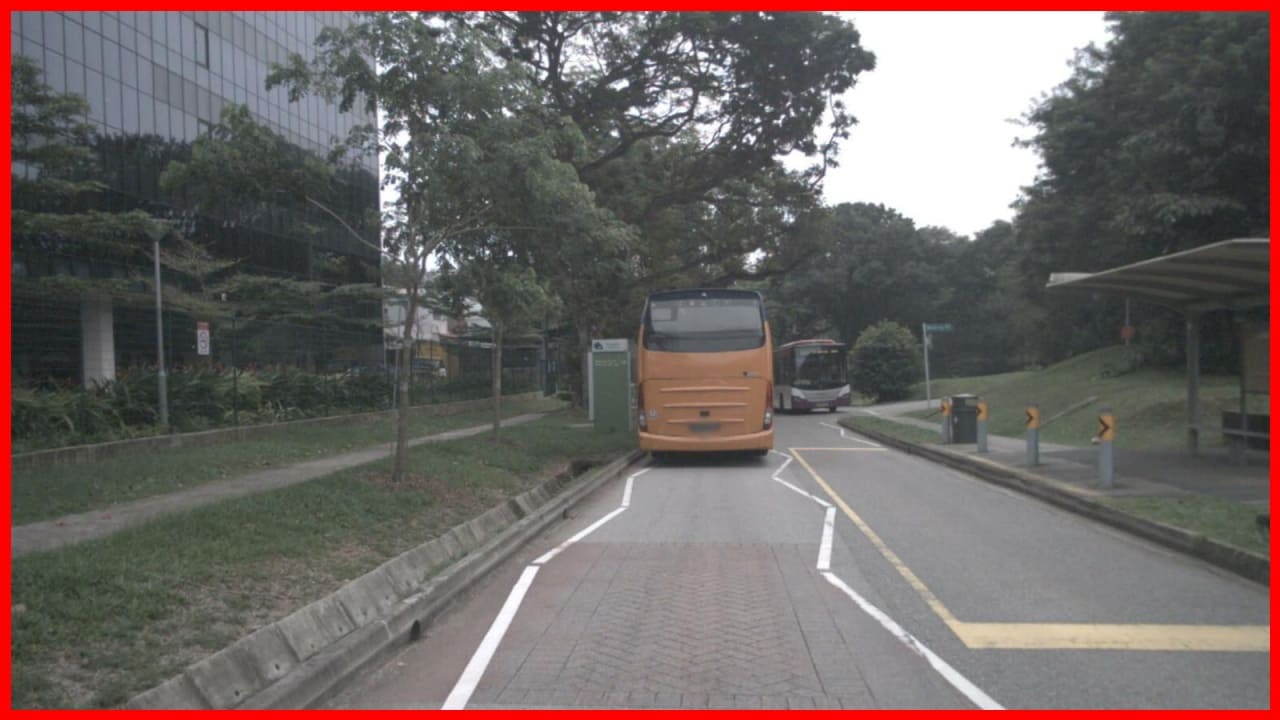}
    \end{subfigure}
    \begin{subfigure}[b]{\subfigwidth}
        \includegraphics[width=\textwidth]{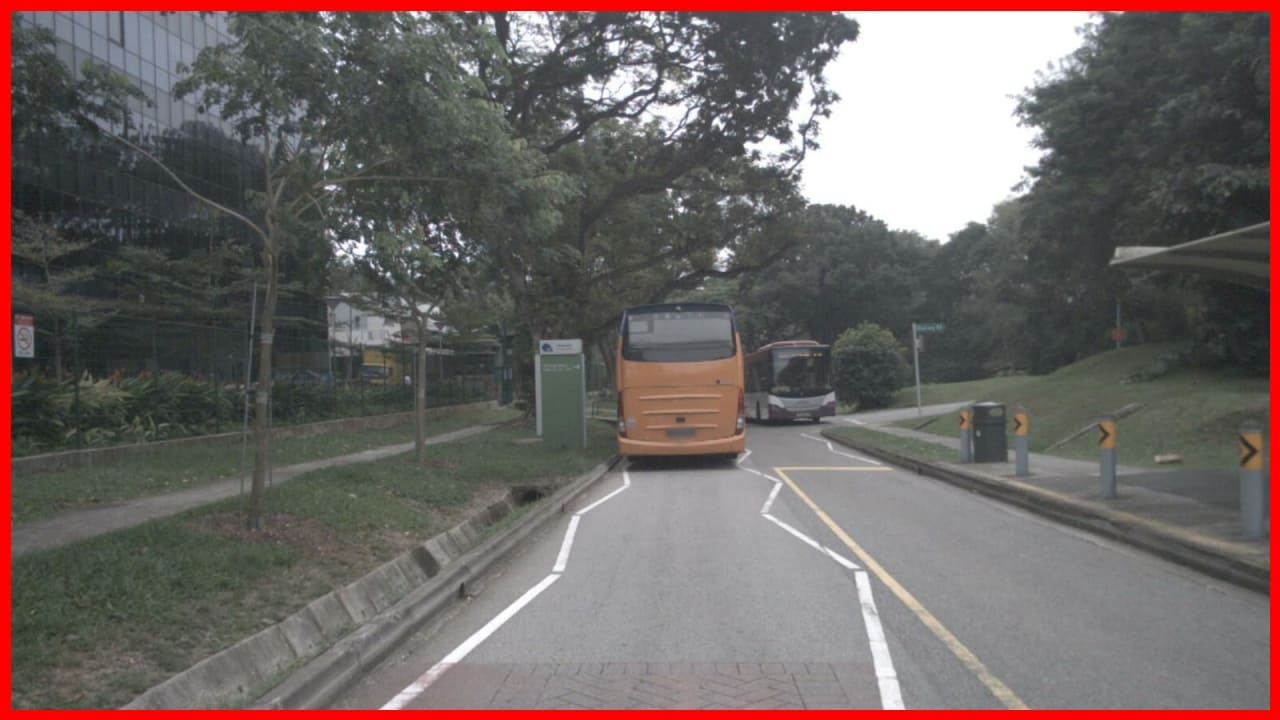}
    \end{subfigure}
    \begin{subfigure}[b]{\subfigwidth}
        \includegraphics[width=\textwidth]{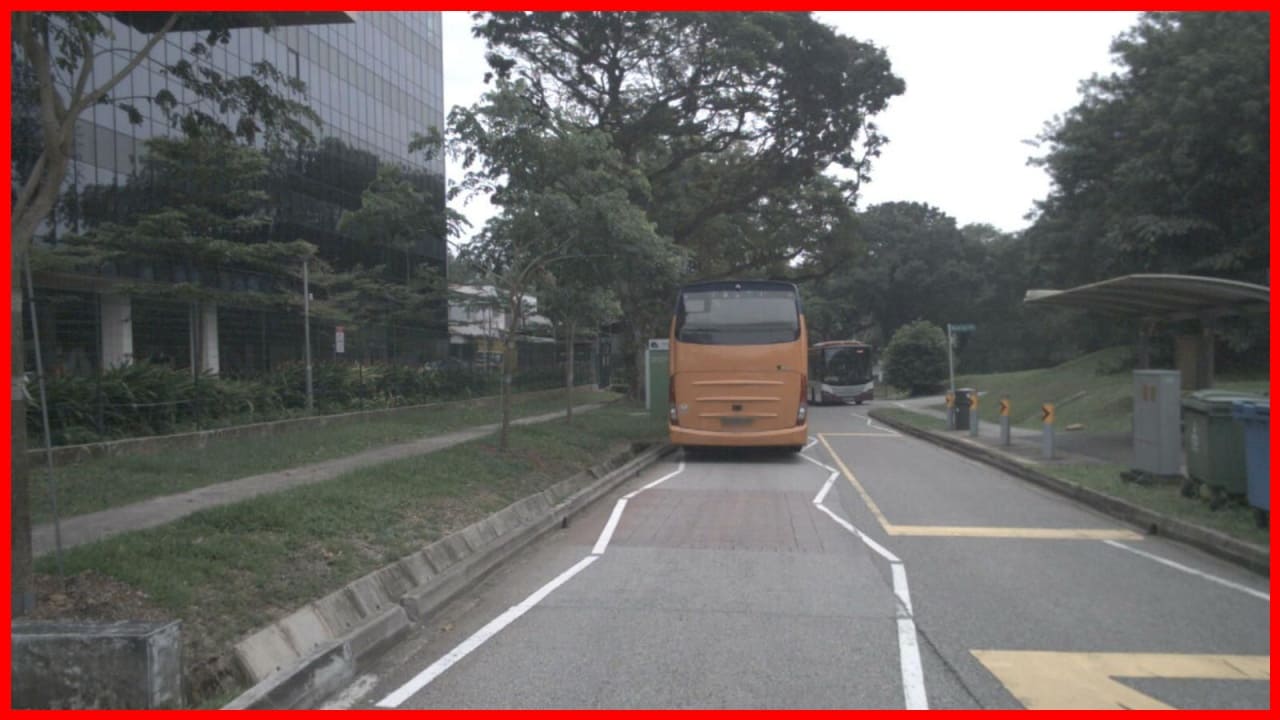}
    \end{subfigure}
    \begin{subfigure}[b]{\subfigwidth}
        \includegraphics[width=\textwidth]{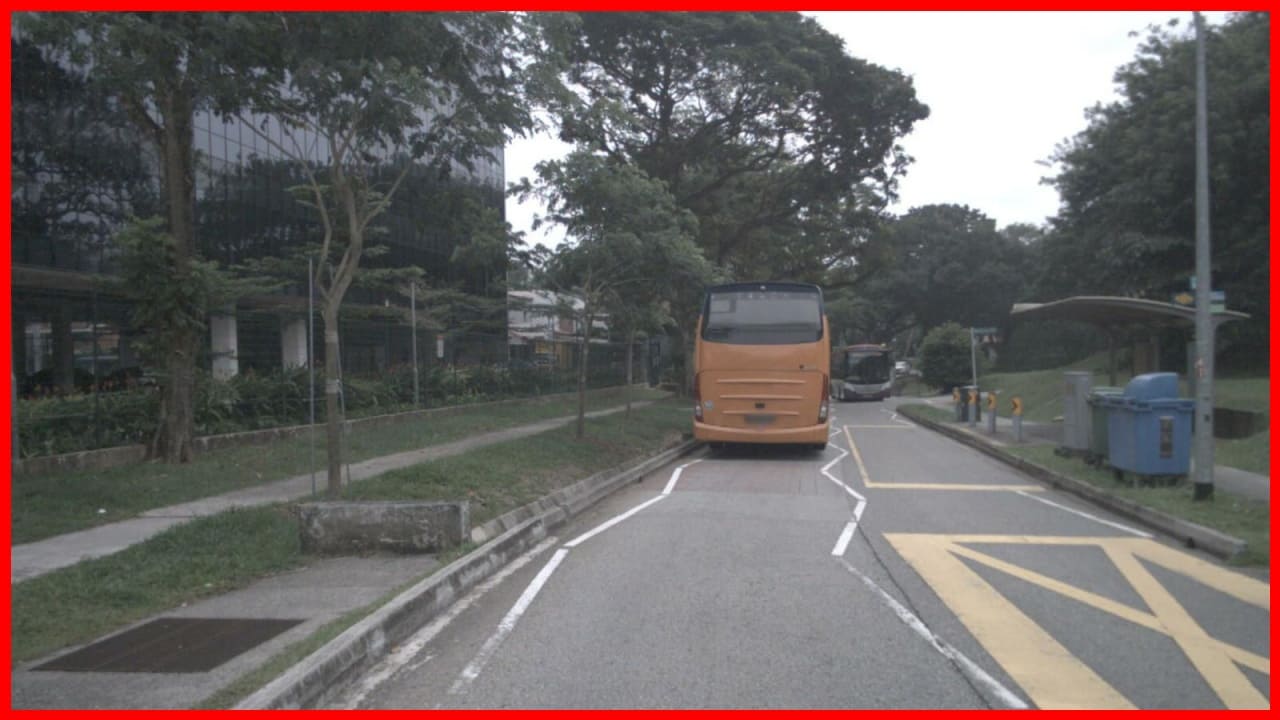}
    \end{subfigure}
    \begin{subfigure}[b]{\subfigwidth}
        \includegraphics[width=\textwidth]{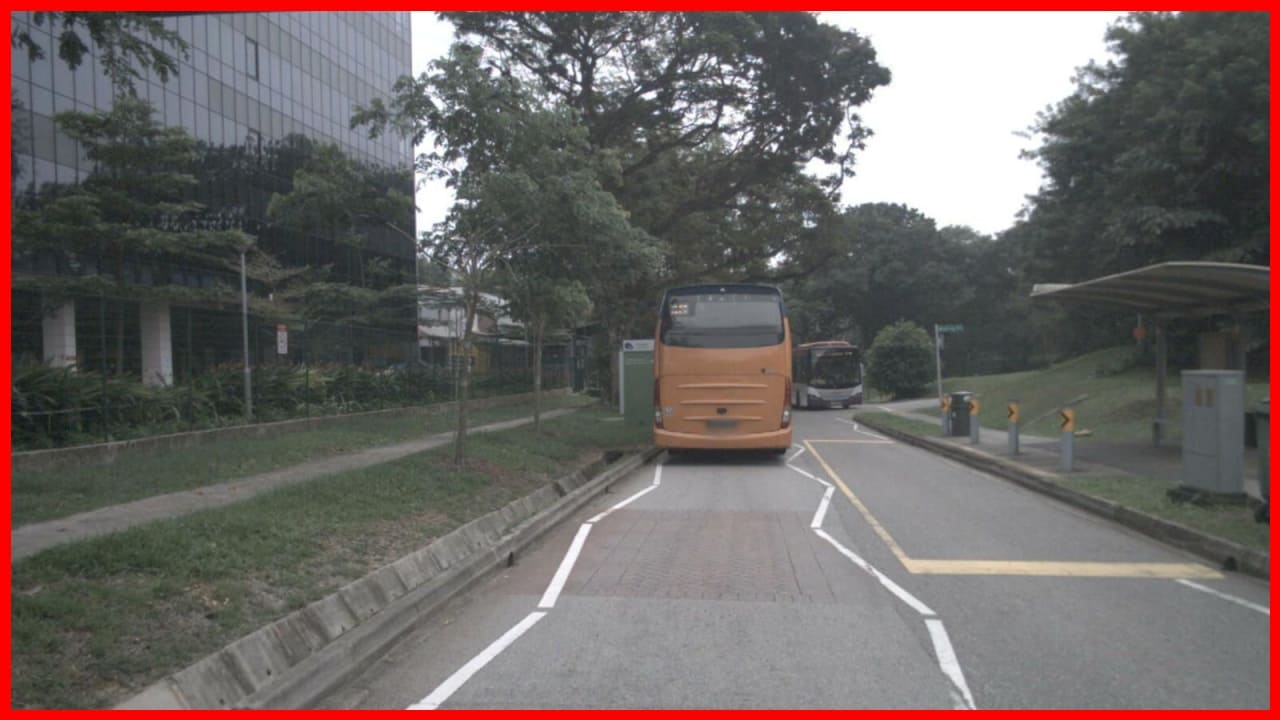}
    \end{subfigure} \\

    \rotatebox[origin=left]{90}{\hspace{0.02cm} \textbf{SIGLIP2}} 
    \begin{subfigure}[b]{\subfigwidth}
        \includegraphics[width=\textwidth]{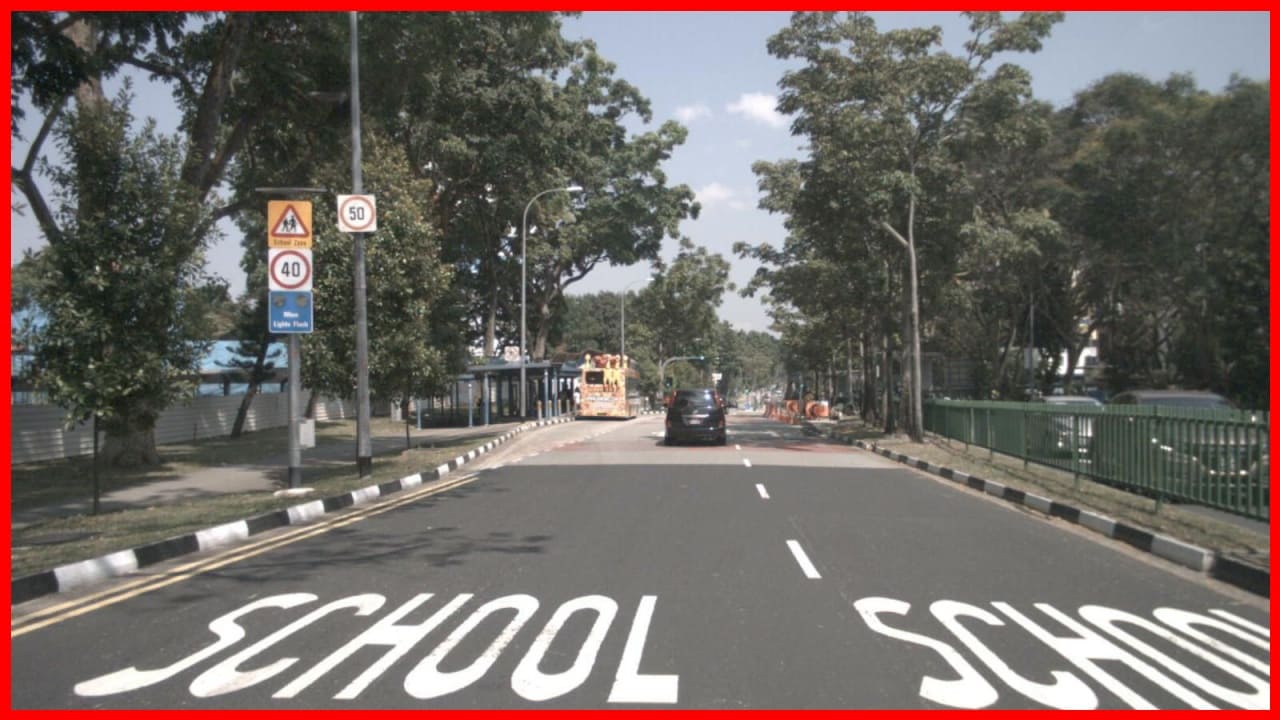}
    \end{subfigure}
    \begin{subfigure}[b]{\subfigwidth}
        \includegraphics[width=\textwidth]{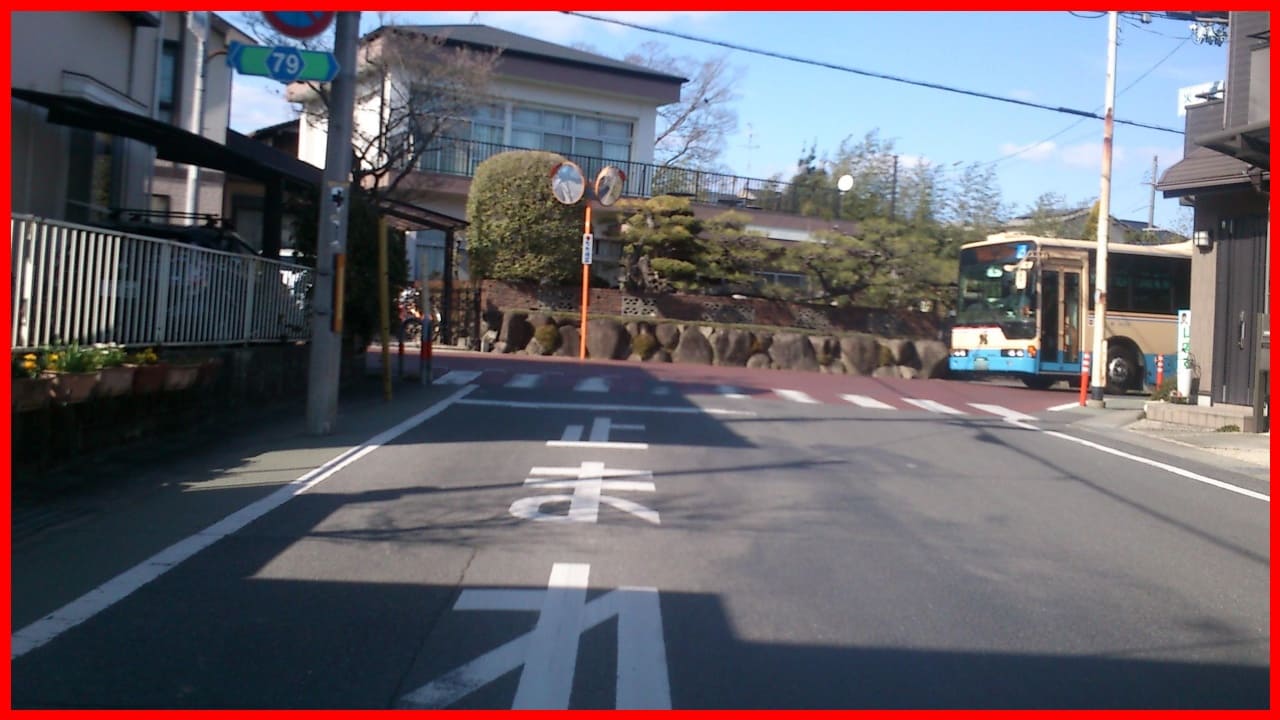}
    \end{subfigure}
    \begin{subfigure}[b]{\subfigwidth}
        \includegraphics[width=\textwidth]{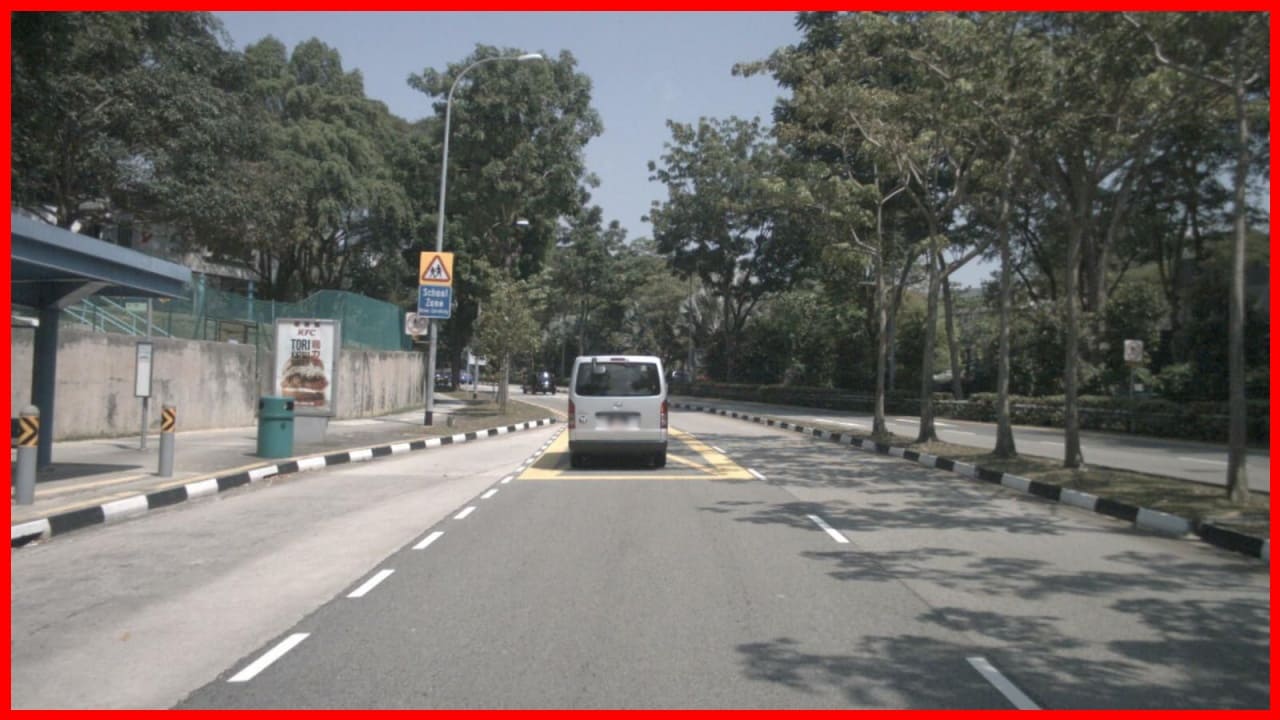}
    \end{subfigure}
    \begin{subfigure}[b]{\subfigwidth}
        \includegraphics[width=\textwidth]{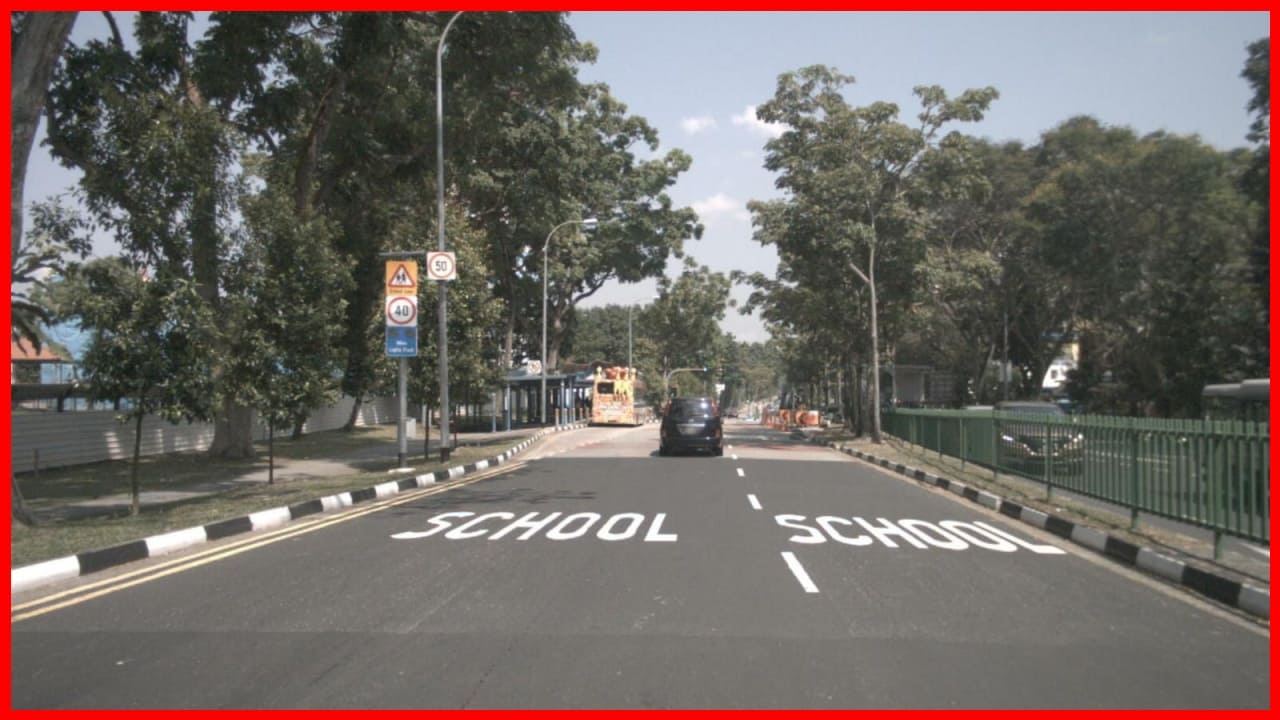}
    \end{subfigure}
    \begin{subfigure}[b]{\subfigwidth}
        \includegraphics[width=\textwidth]{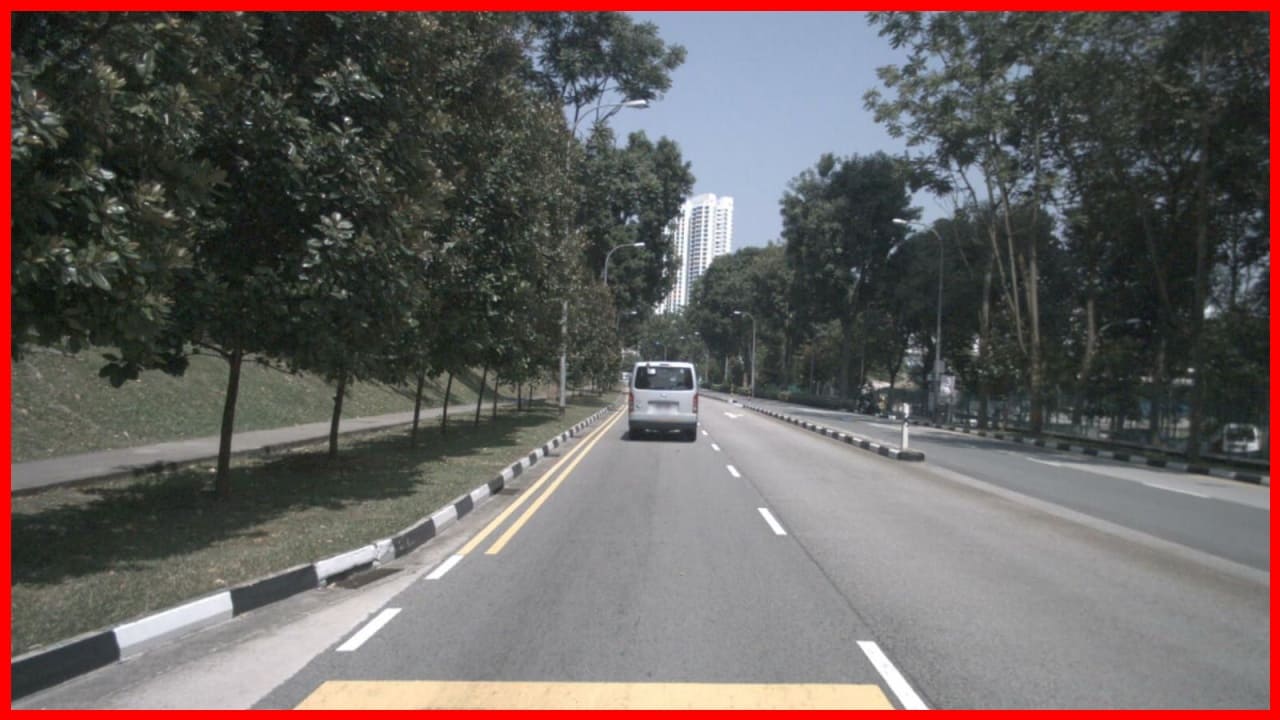}
    \end{subfigure} \\

    \rotatebox[origin=left]{90}{\hspace{0.04cm} \textbf{NACLIP}} 
    \begin{subfigure}[b]{\subfigwidth}
        \includegraphics[width=\textwidth]{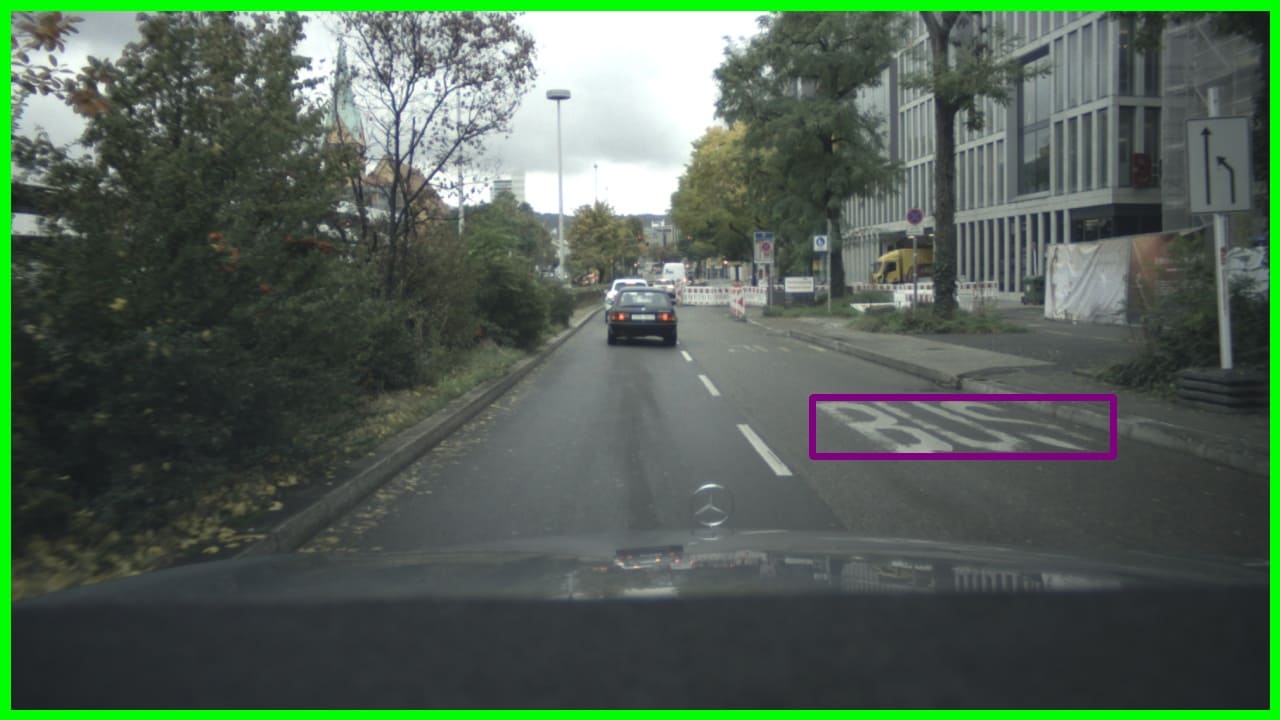}
    \end{subfigure}
    \begin{subfigure}[b]{\subfigwidth}
        \includegraphics[width=\textwidth]{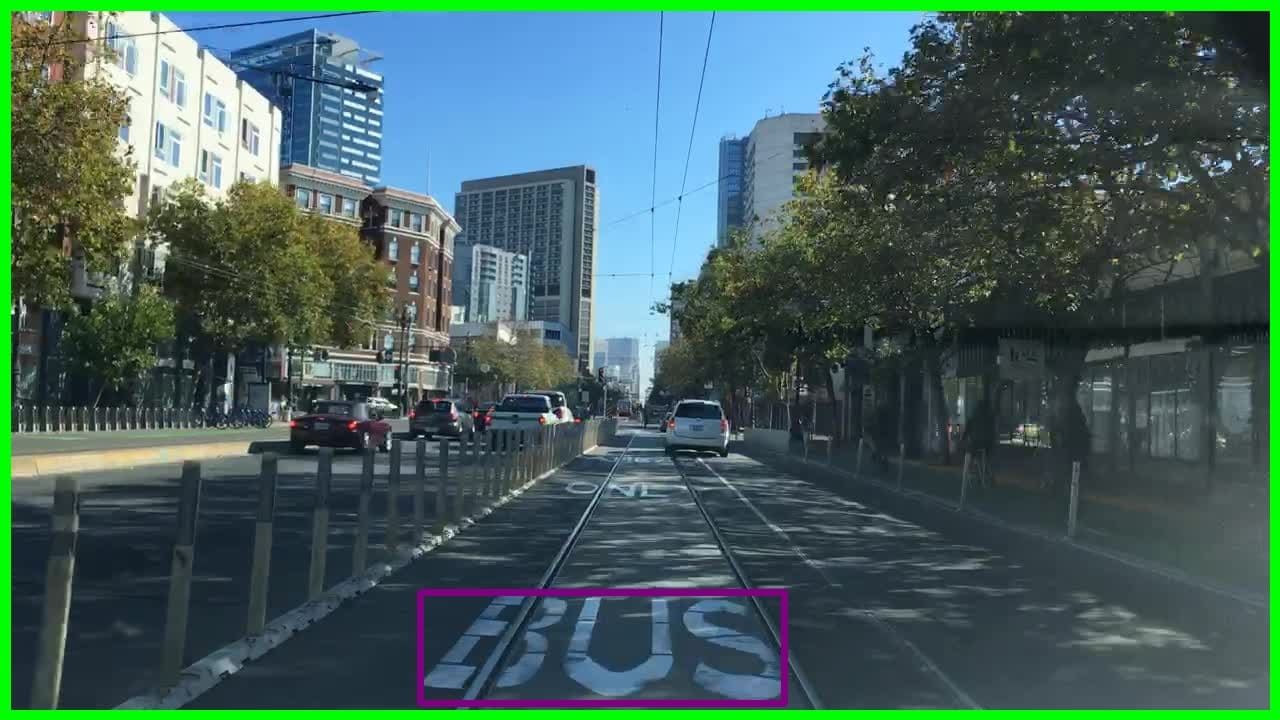}
    \end{subfigure}
    \begin{subfigure}[b]{\subfigwidth}
        \includegraphics[width=\textwidth]{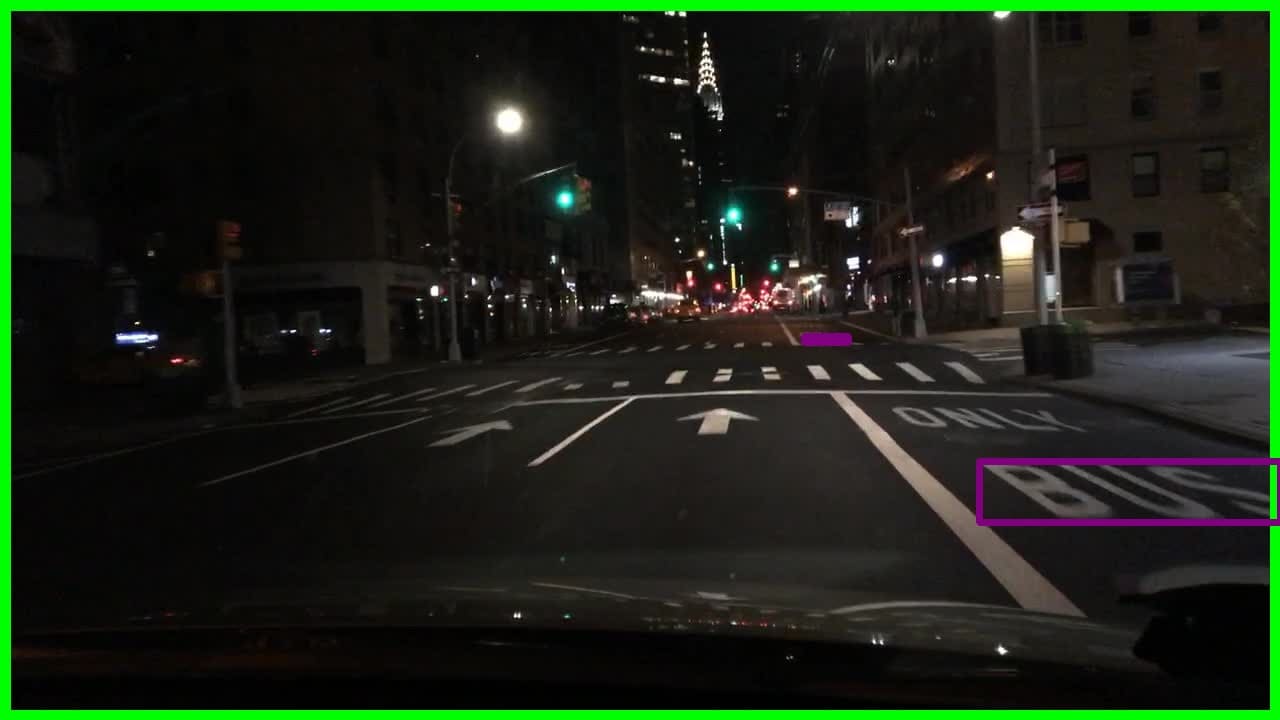}
    \end{subfigure}
    \begin{subfigure}[b]{\subfigwidth}
        \includegraphics[width=\textwidth]{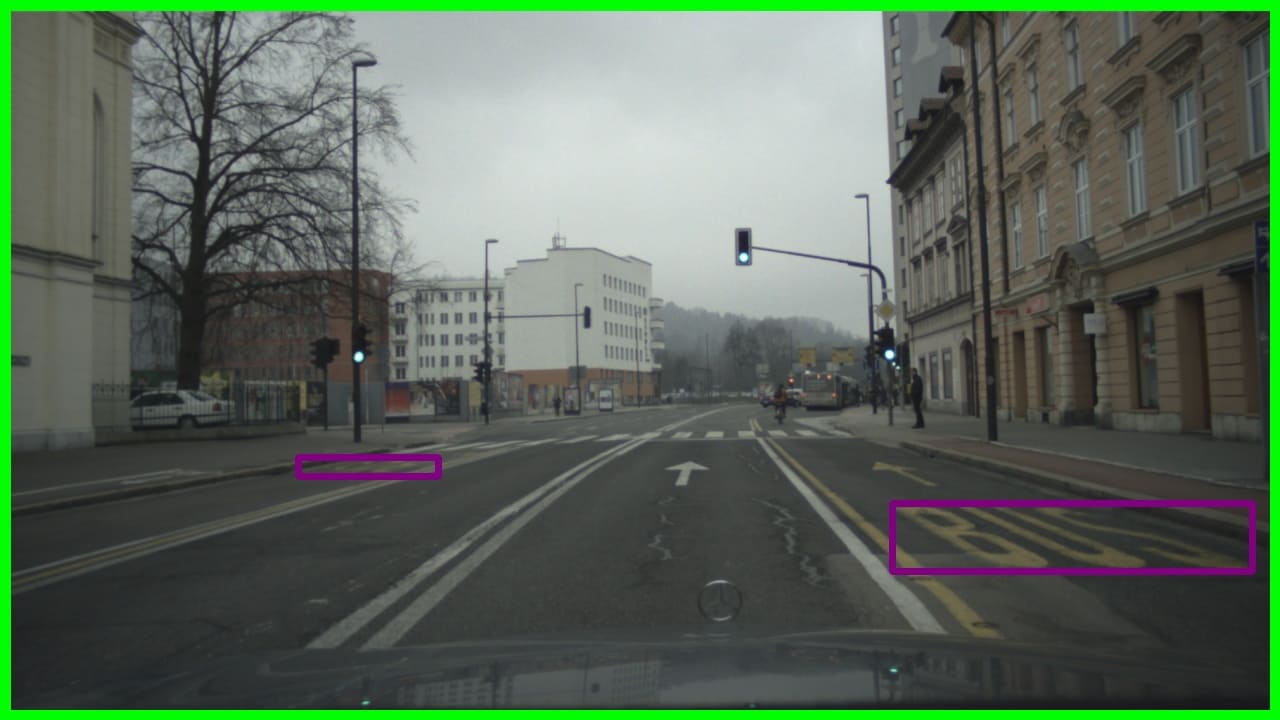}
    \end{subfigure}
    \begin{subfigure}[b]{\subfigwidth}
        \includegraphics[width=\textwidth]{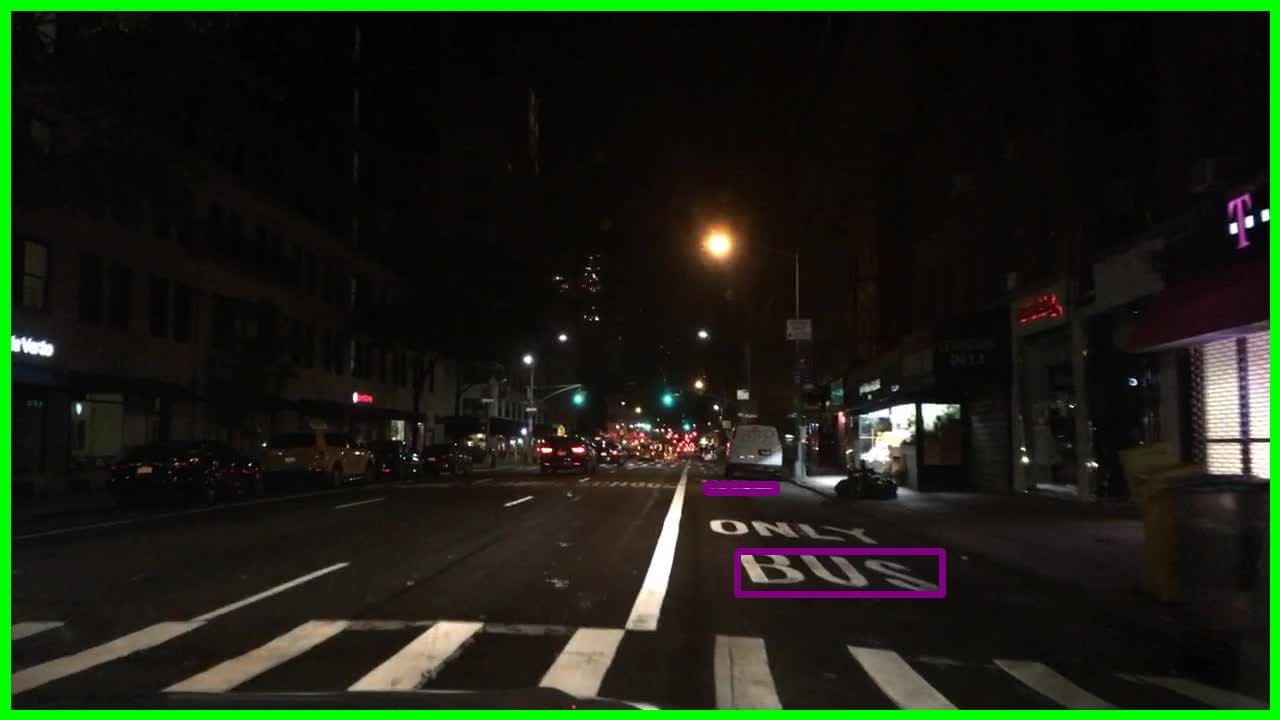}
    \end{subfigure} \\

    \rotatebox[origin=left]{90}{\hspace{-0.15cm} \textbf{NARADIO}} 
    \begin{subfigure}[b]{\subfigwidth}
        \includegraphics[width=\textwidth]{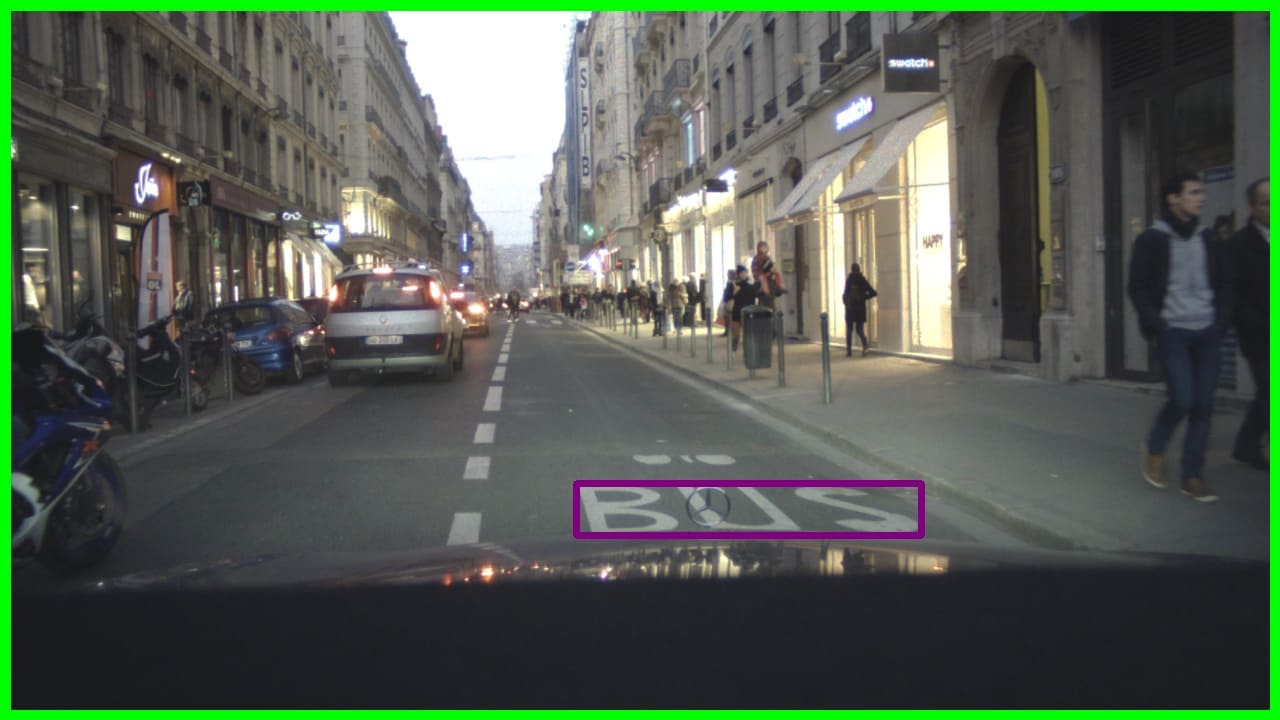}
    \end{subfigure}
    \begin{subfigure}[b]{\subfigwidth}
        \includegraphics[width=\textwidth]{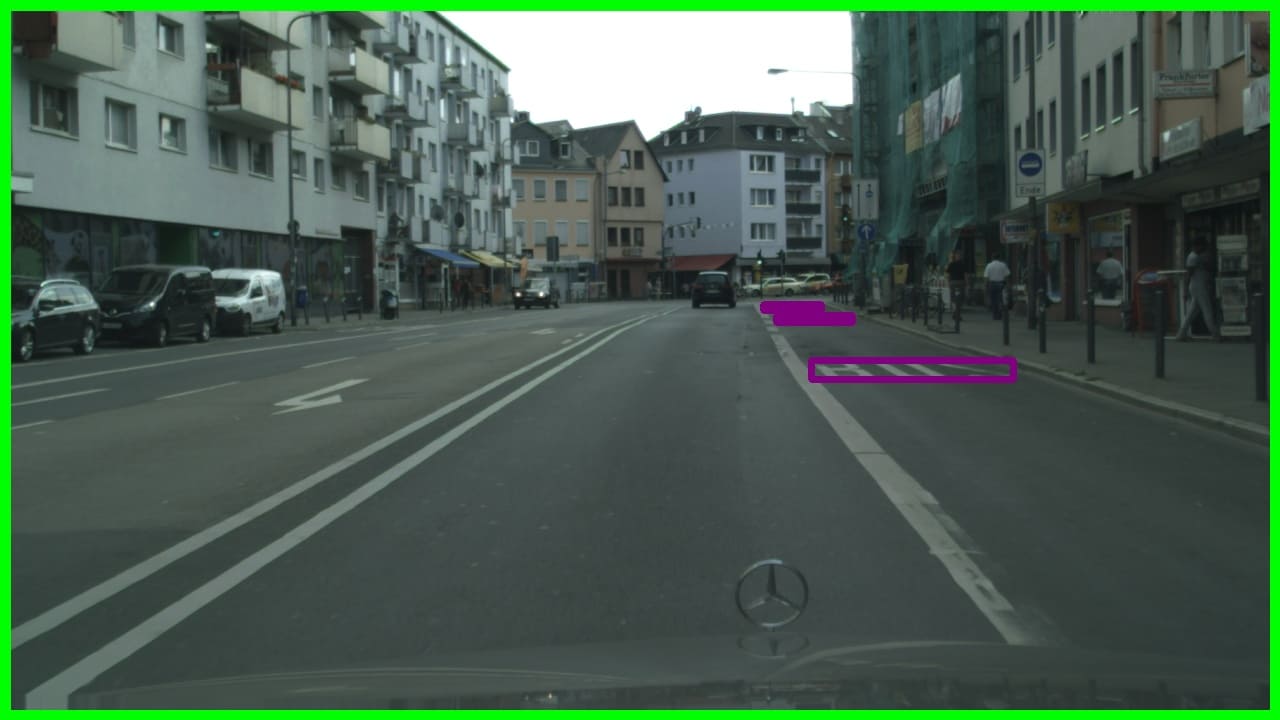}
    \end{subfigure}
    \begin{subfigure}[b]{\subfigwidth}
        \includegraphics[width=\textwidth]{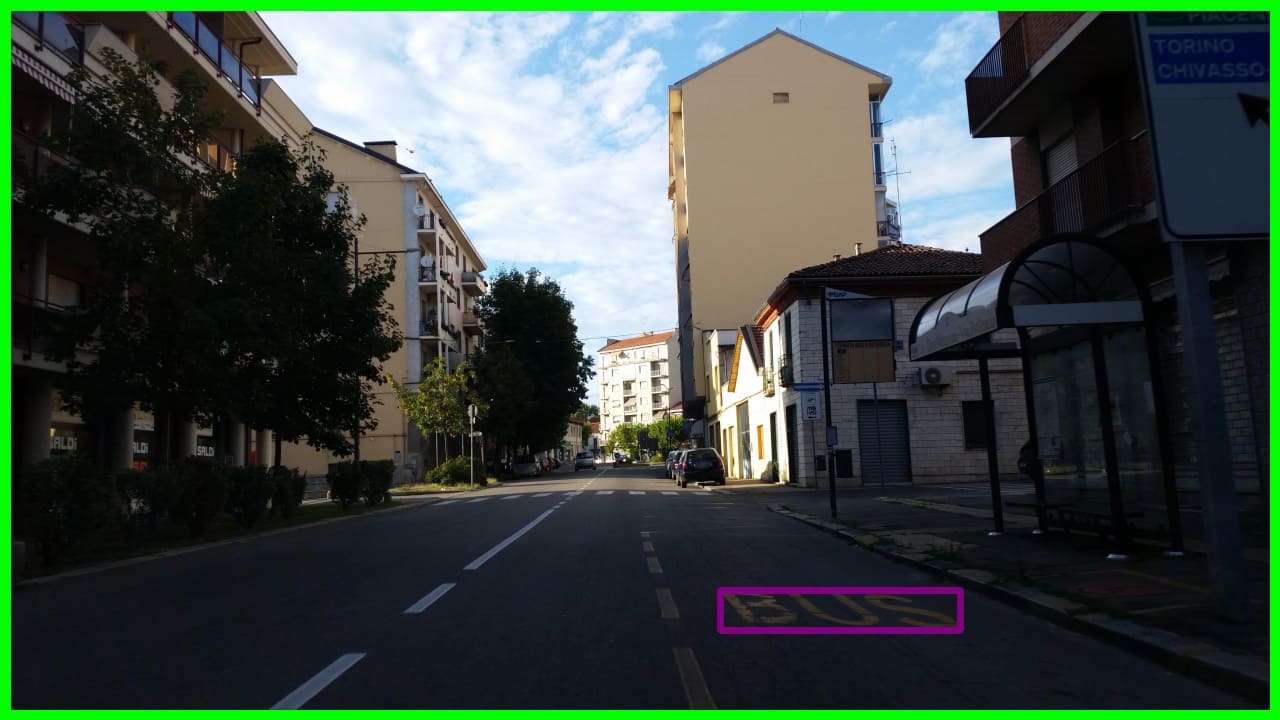}
    \end{subfigure}
    \begin{subfigure}[b]{\subfigwidth}
        \includegraphics[width=\textwidth]{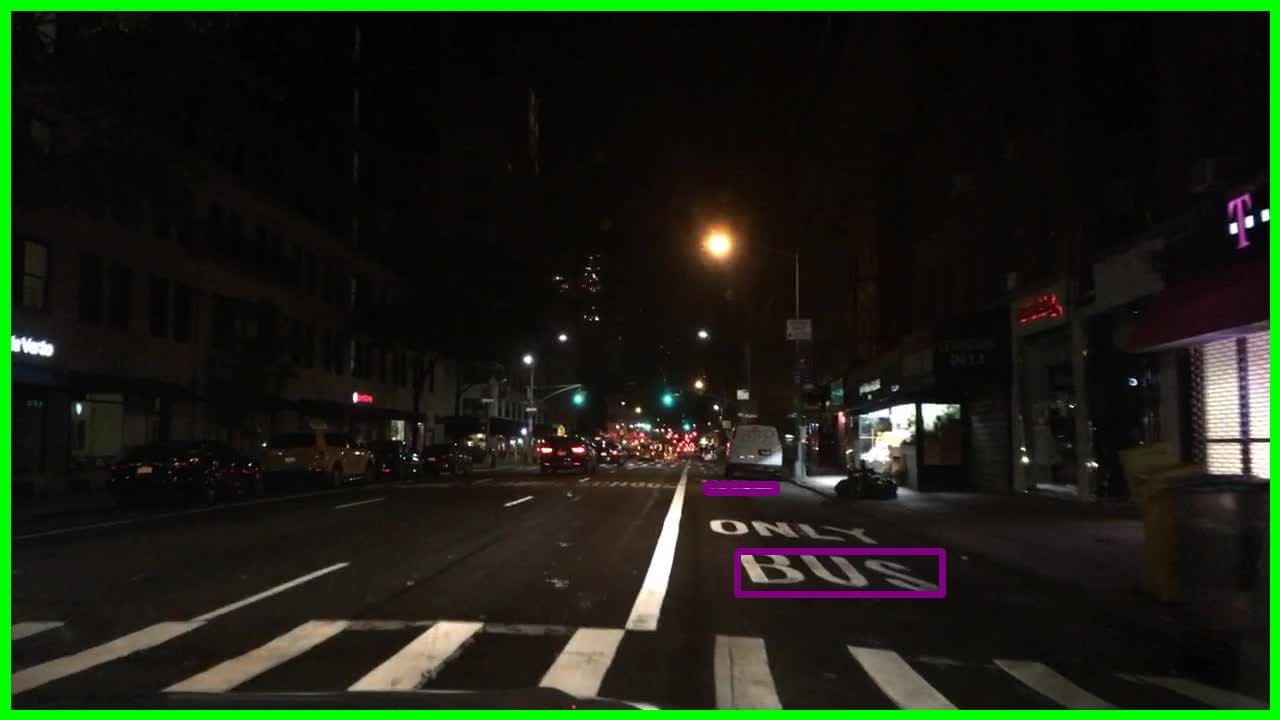}
    \end{subfigure}
    \begin{subfigure}[b]{\subfigwidth}
        \includegraphics[width=\textwidth]{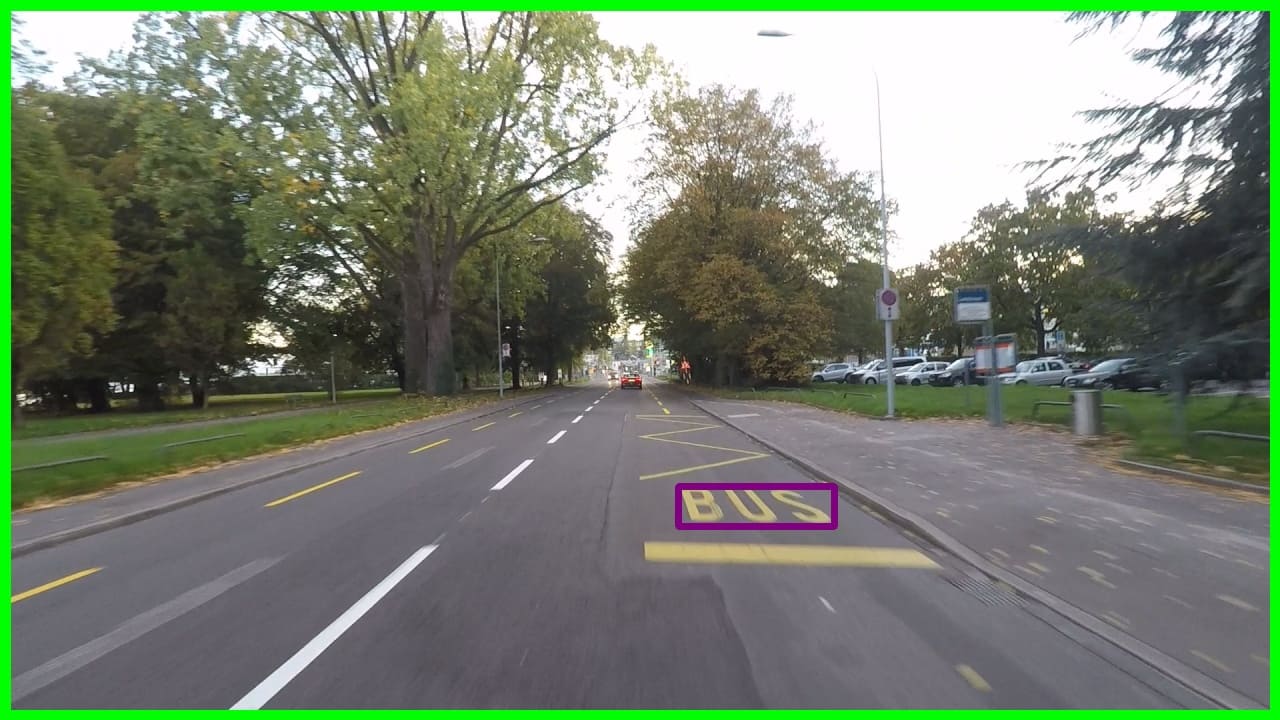}
    \end{subfigure} \\
    \caption{Text-based retrieval results on the validation split for the \emph{Marking-Bus-Text} class, showing top 5 ranked images for each model. Model references can be found in \Cref{tab:map_class_wise_text_to_image_retrieval}.}
    \label{fig:marking_bus_text_results}
    \endgroup
\end{figure*}
}

{
\def\rankOne{\bfseries Ranked 1st}
\def\rankTwo{\bfseries Ranked 2nd}
\def\rankThree{\bfseries Ranked 3rd}
\def\rankFour{\bfseries Ranked 4th}
\def\rankFive{\bfseries Ranked 5th}

\captionsetup[subfigure]{labelformat=empty}

\setlength{\subfigwidth}{0.18\textwidth} 

\begin{figure*}[ht]
    \begingroup
    \centering 
    \small
    \rotatebox[origin=left]{90}{\hspace{0.02cm} \textbf{GDINO}} 
    \begin{subfigure}[b]{\subfigwidth}
        \caption{\rankOne}
        \includegraphics[width=\textwidth]{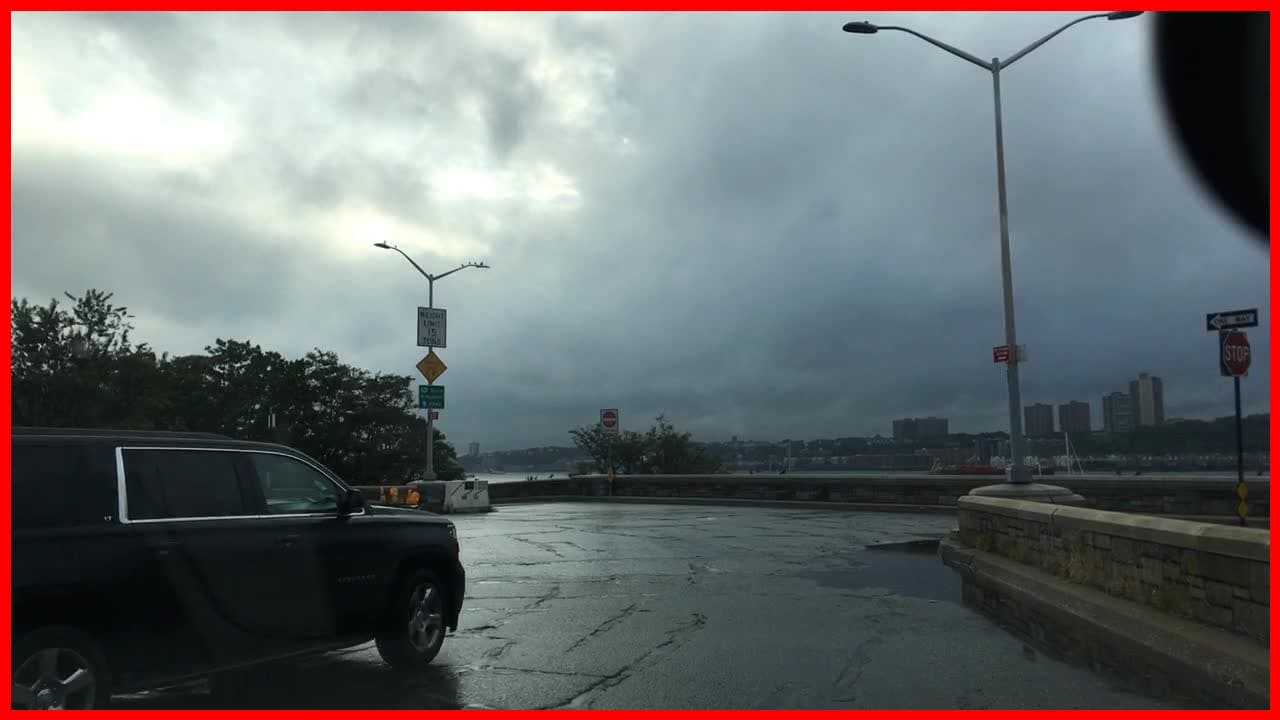}
    \end{subfigure}
    \begin{subfigure}[b]{\subfigwidth}
        \caption{\rankTwo}
        \includegraphics[width=\textwidth]{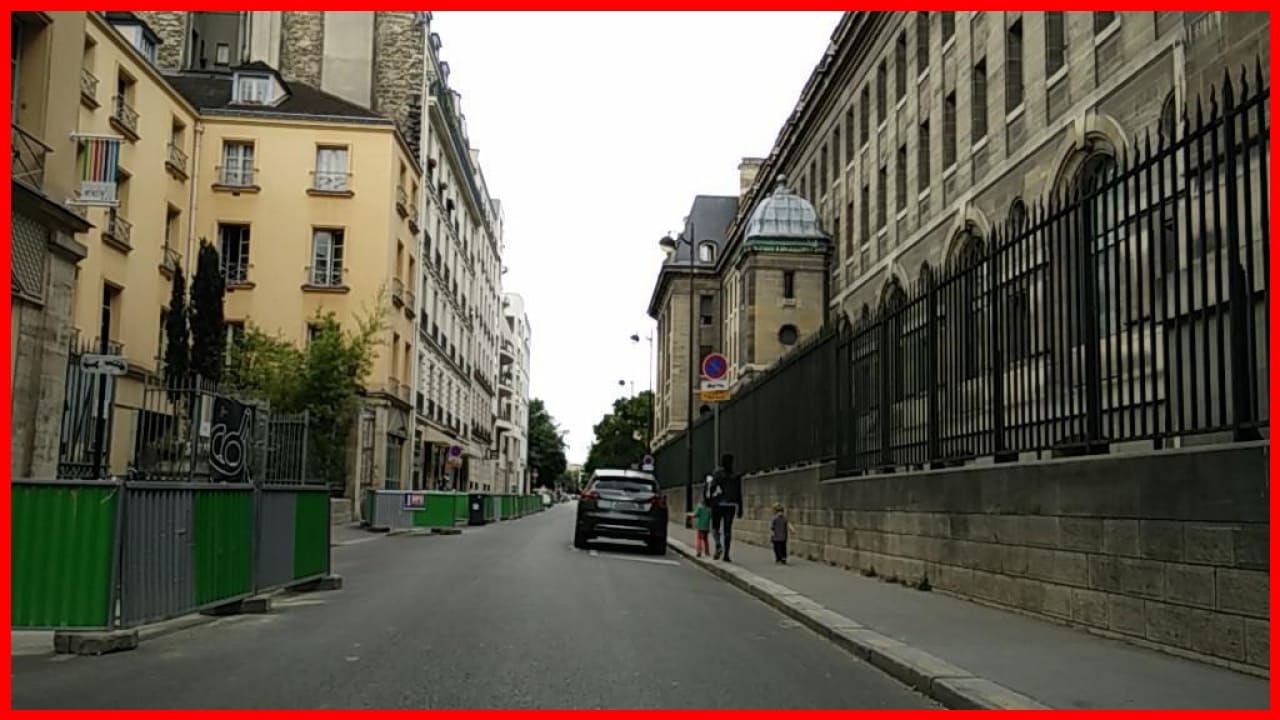}
    \end{subfigure}
    \begin{subfigure}[b]{\subfigwidth}
        \caption{\rankThree}
        \includegraphics[width=\textwidth]{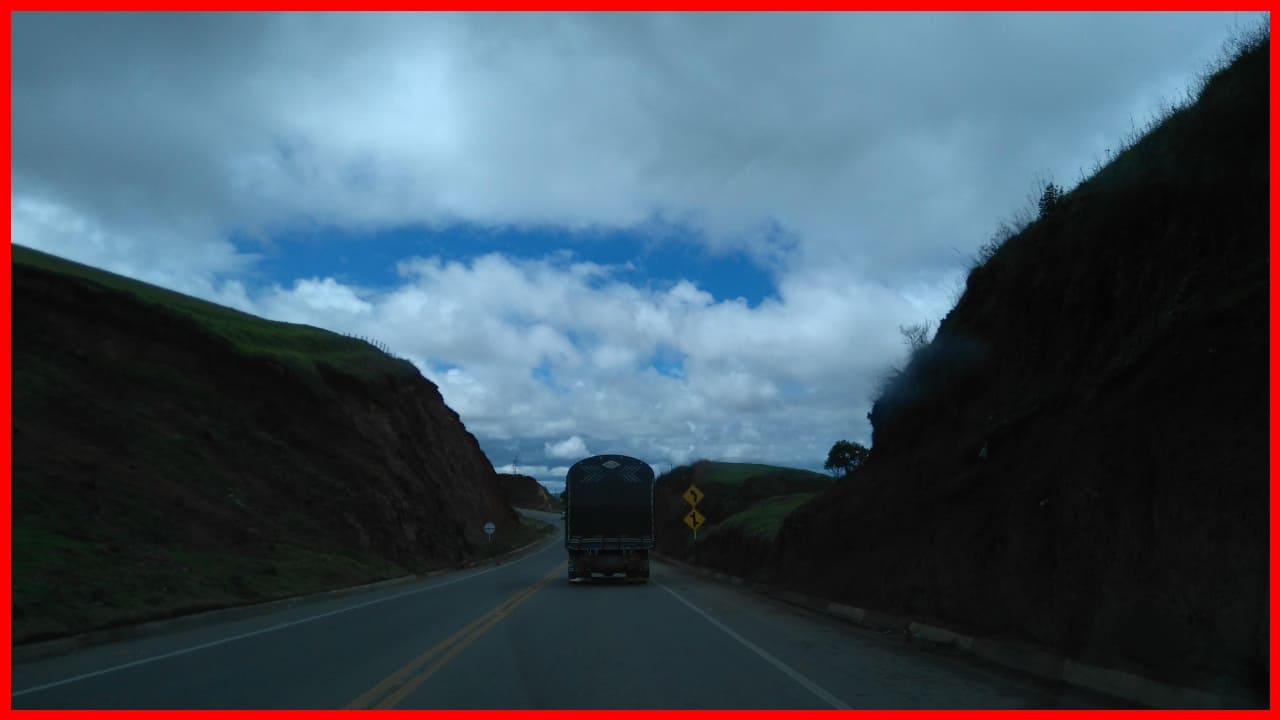}
    \end{subfigure}
    \begin{subfigure}[b]{\subfigwidth}
        \caption{\rankFour}
        \includegraphics[width=\textwidth]{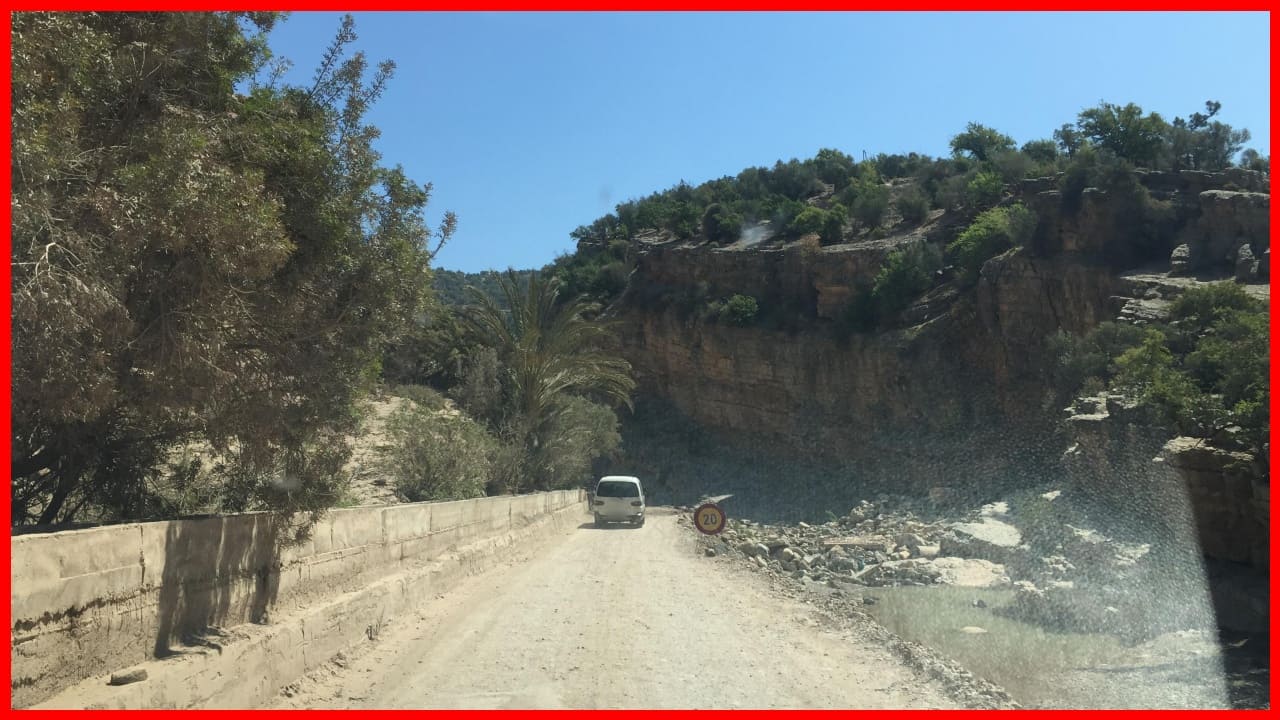}
    \end{subfigure}
    \begin{subfigure}[b]{\subfigwidth}
        \caption{\rankFive}
        \includegraphics[width=\textwidth]{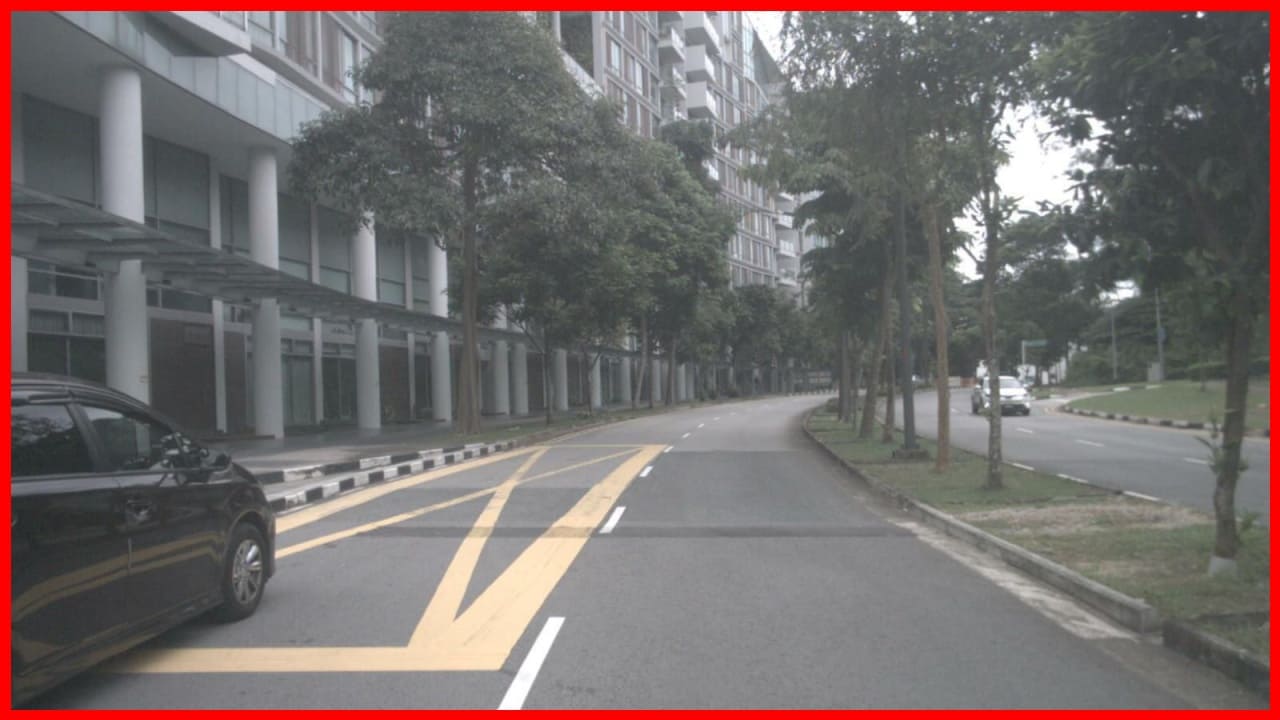}
    \end{subfigure} \\
    
    \rotatebox[origin=left]{90}{\hspace{0.17cm} \textbf{CLIP}} 
    \begin{subfigure}[b]{\subfigwidth}
        \includegraphics[width=\textwidth]{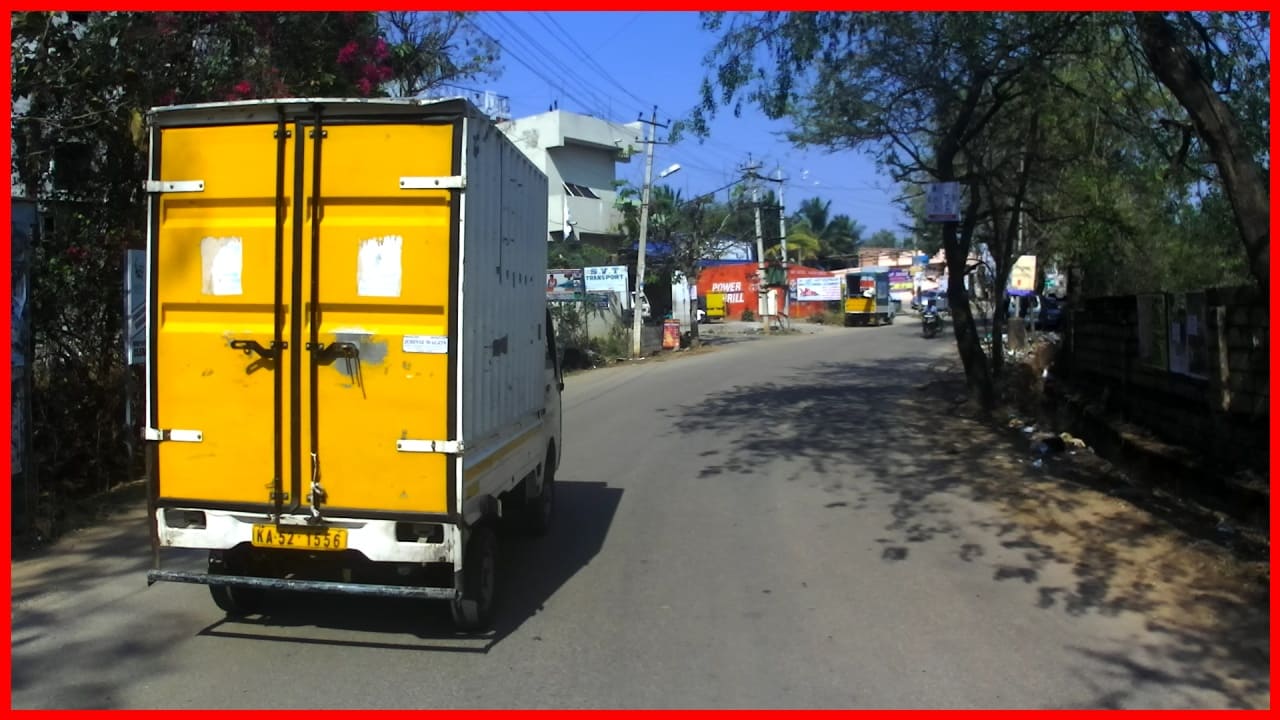}
    \end{subfigure}
    \begin{subfigure}[b]{\subfigwidth}
        \includegraphics[width=\textwidth]{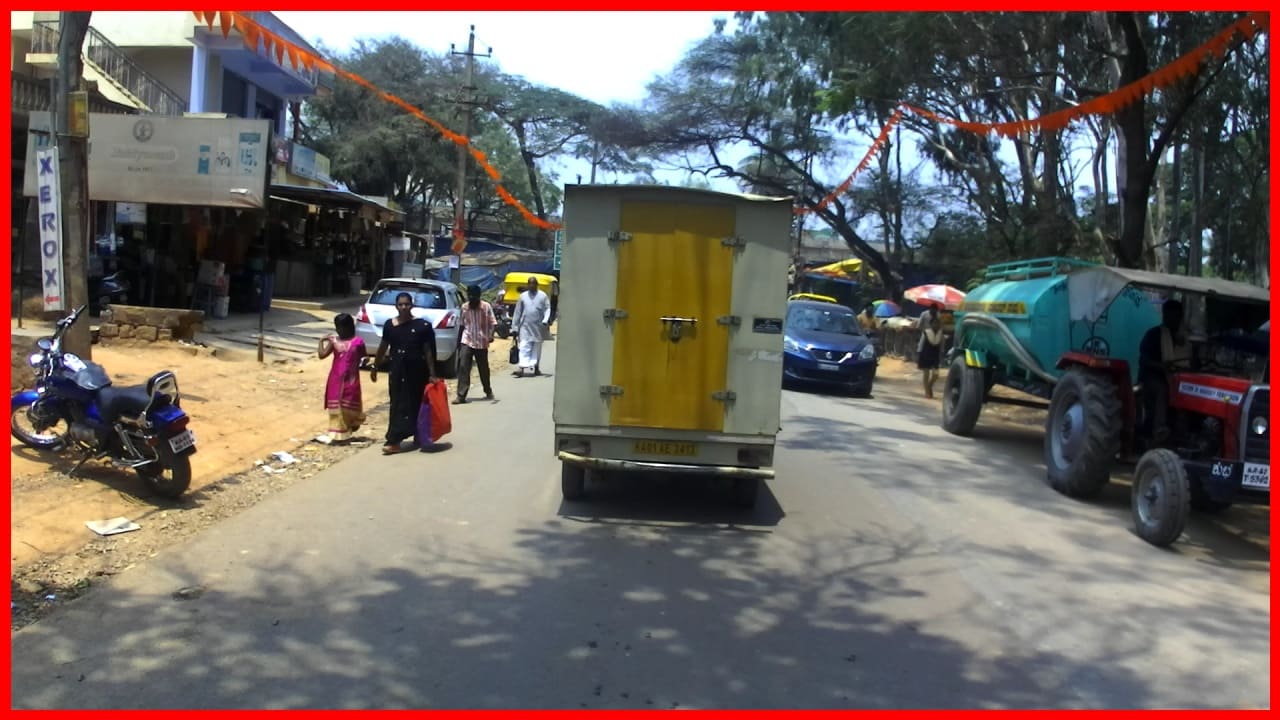}
    \end{subfigure}
    \begin{subfigure}[b]{\subfigwidth}
        \includegraphics[width=\textwidth]{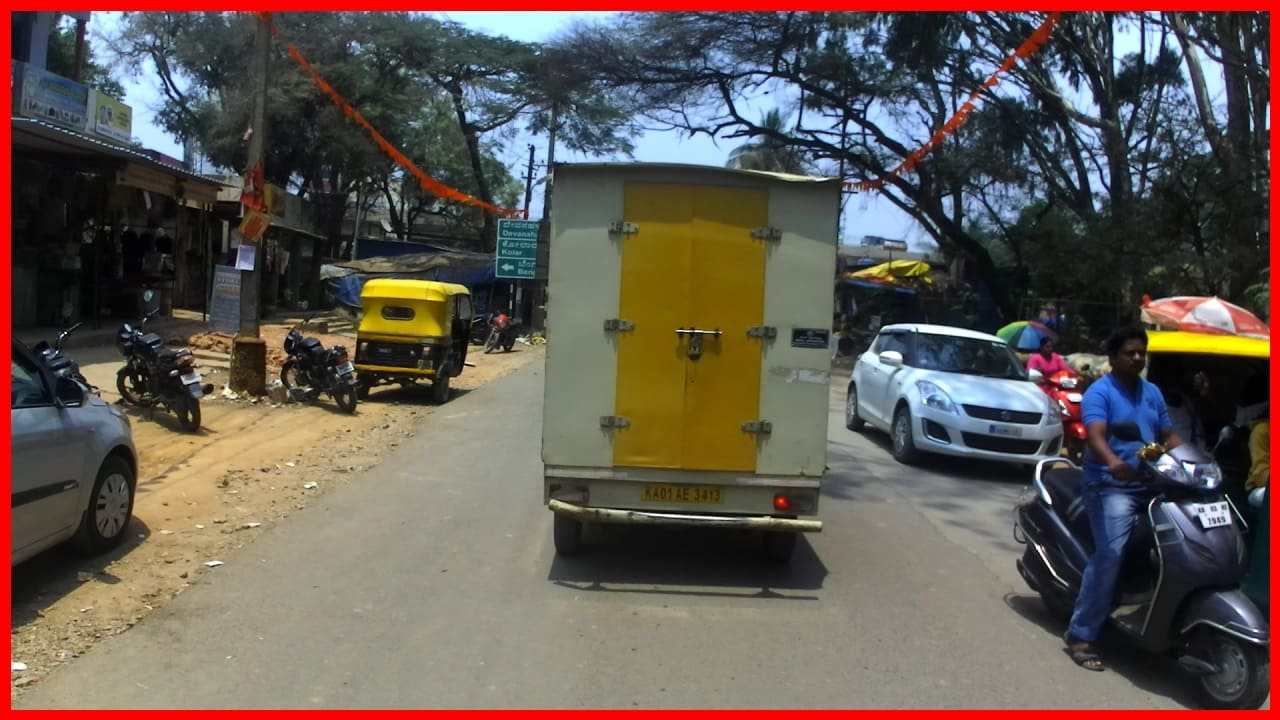}
    \end{subfigure}
    \begin{subfigure}[b]{\subfigwidth}
        \includegraphics[width=\textwidth]{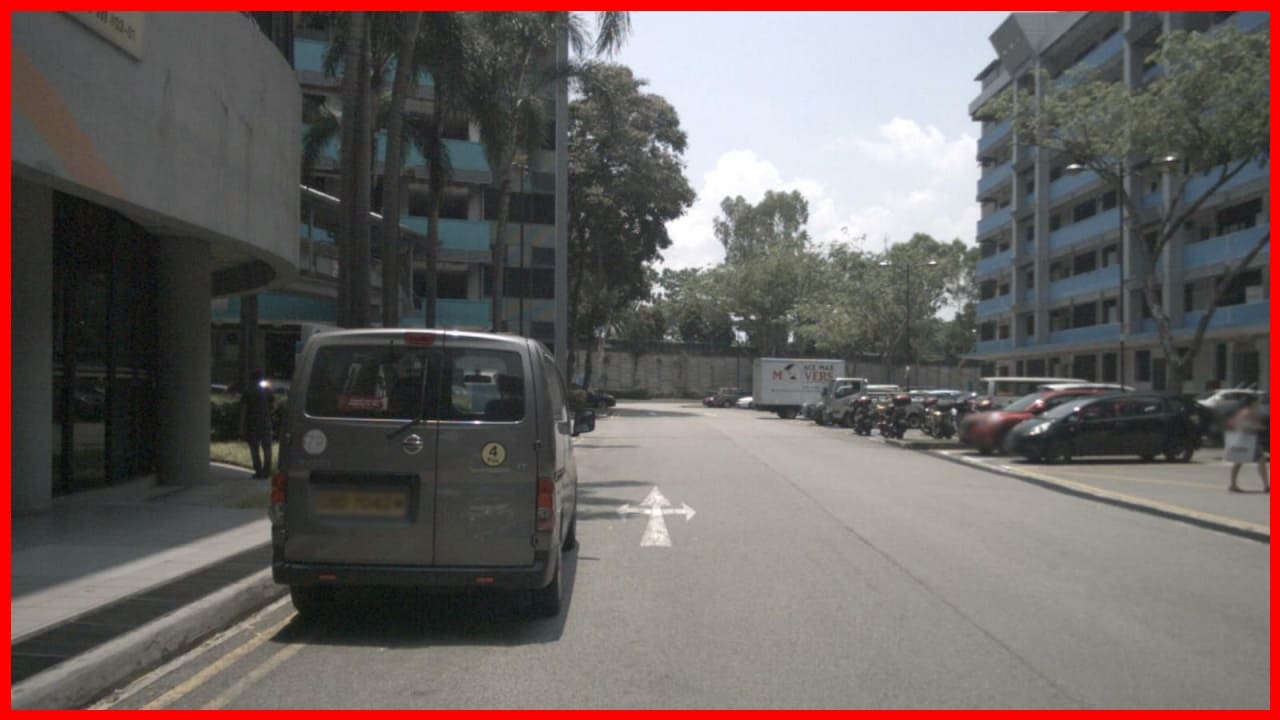}
    \end{subfigure}
    \begin{subfigure}[b]{\subfigwidth}
        \includegraphics[width=\textwidth]{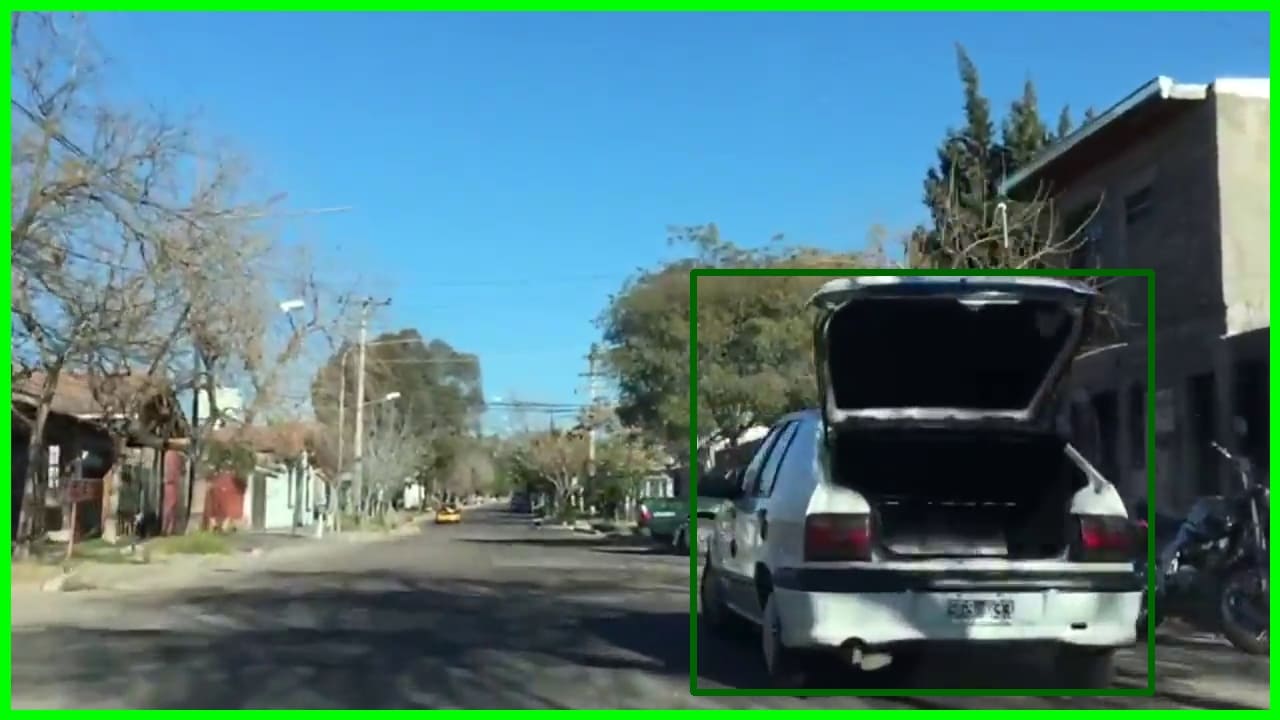}
    \end{subfigure} \\

    \rotatebox[origin=left]{90}{\hspace{0.07cm} \textbf{RADIO}} 
    \begin{subfigure}[b]{\subfigwidth}
        \includegraphics[width=\textwidth]{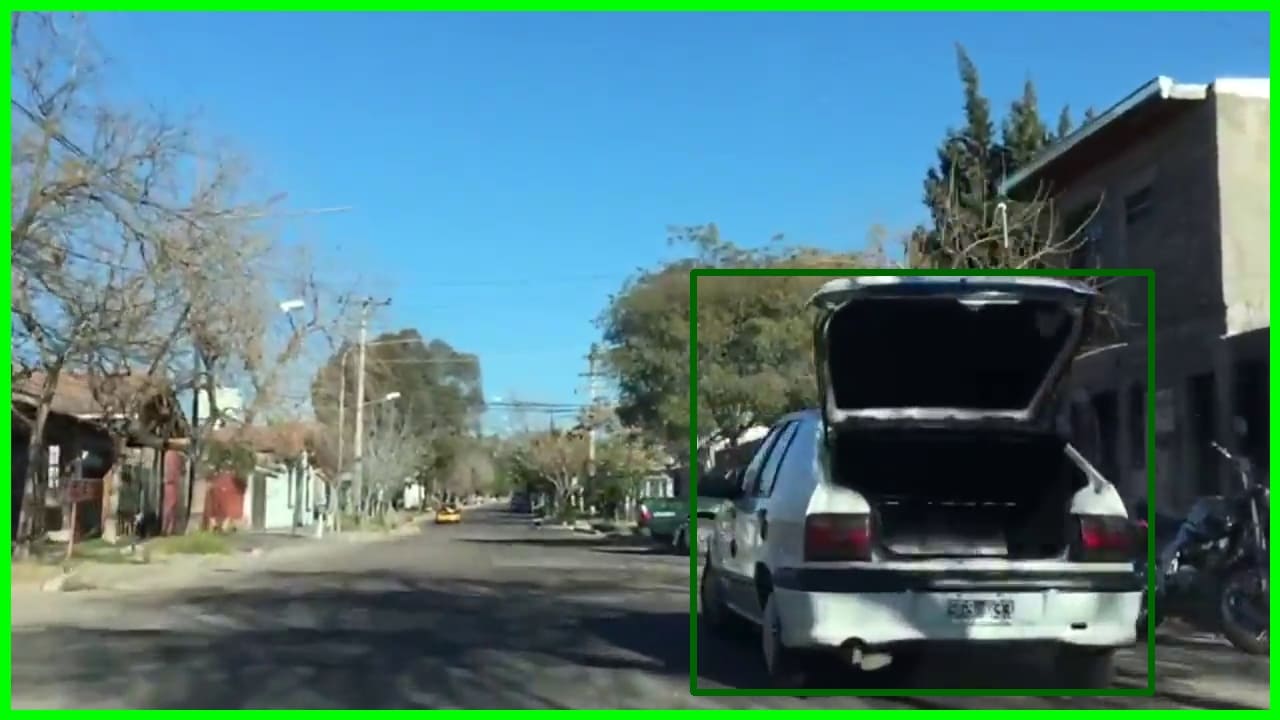}
    \end{subfigure}
    \begin{subfigure}[b]{\subfigwidth}
        \includegraphics[width=\textwidth]{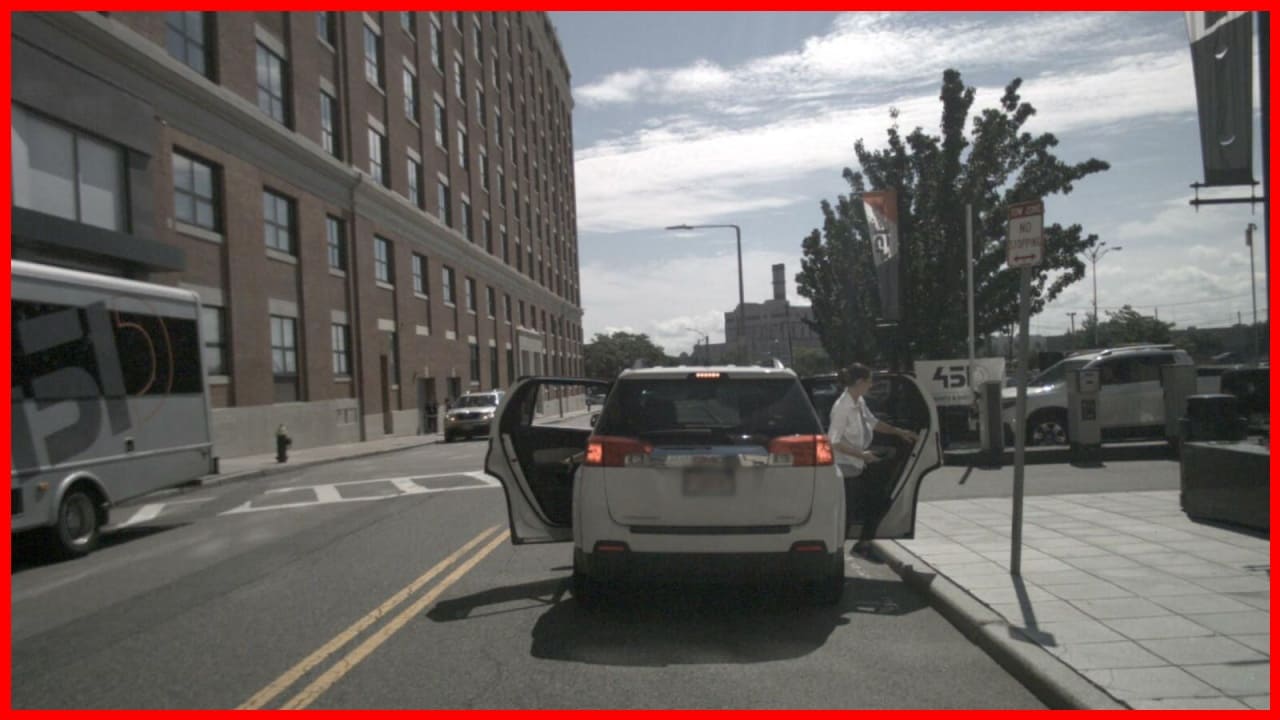}
    \end{subfigure}
    \begin{subfigure}[b]{\subfigwidth}
        \includegraphics[width=\textwidth]{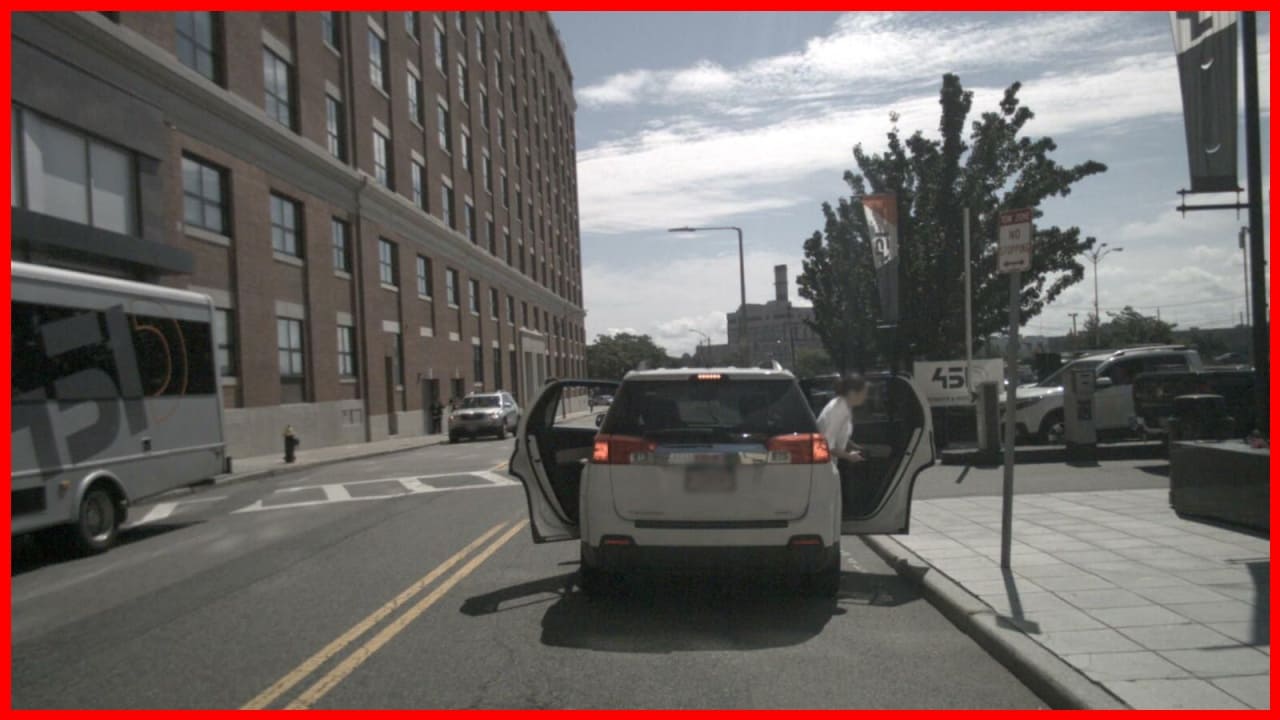}
    \end{subfigure}
    \begin{subfigure}[b]{\subfigwidth}
        \includegraphics[width=\textwidth]{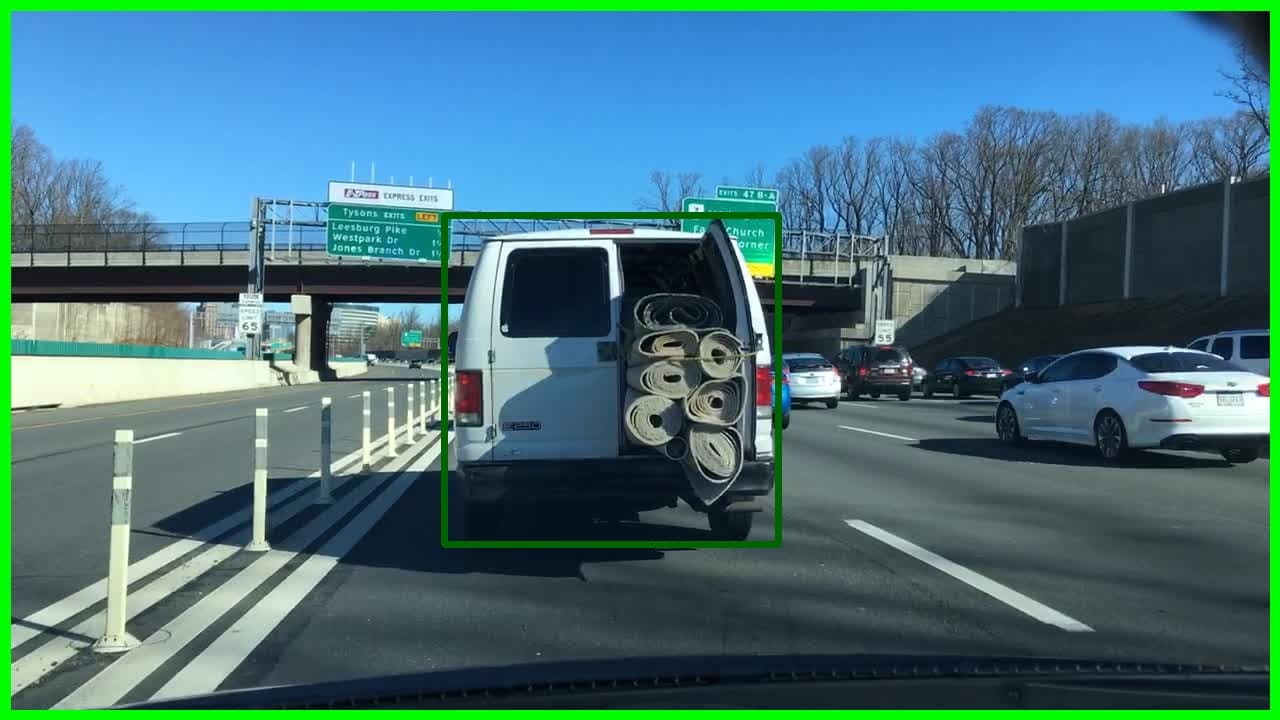}
    \end{subfigure}
    \begin{subfigure}[b]{\subfigwidth}
        \includegraphics[width=\textwidth]{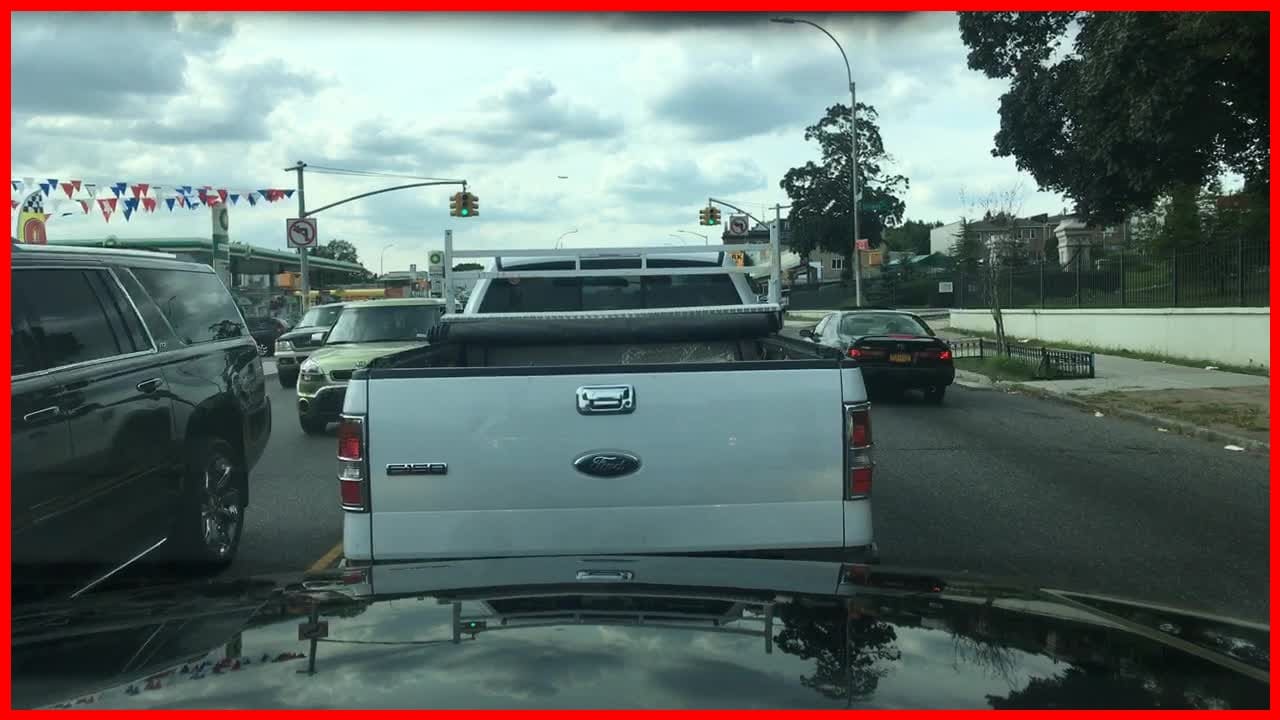}
    \end{subfigure} \\

    \rotatebox[origin=left]{90}{\hspace{0.15cm} \textbf{BLIP2}} 
    \begin{subfigure}[b]{\subfigwidth}
        \includegraphics[width=\textwidth]{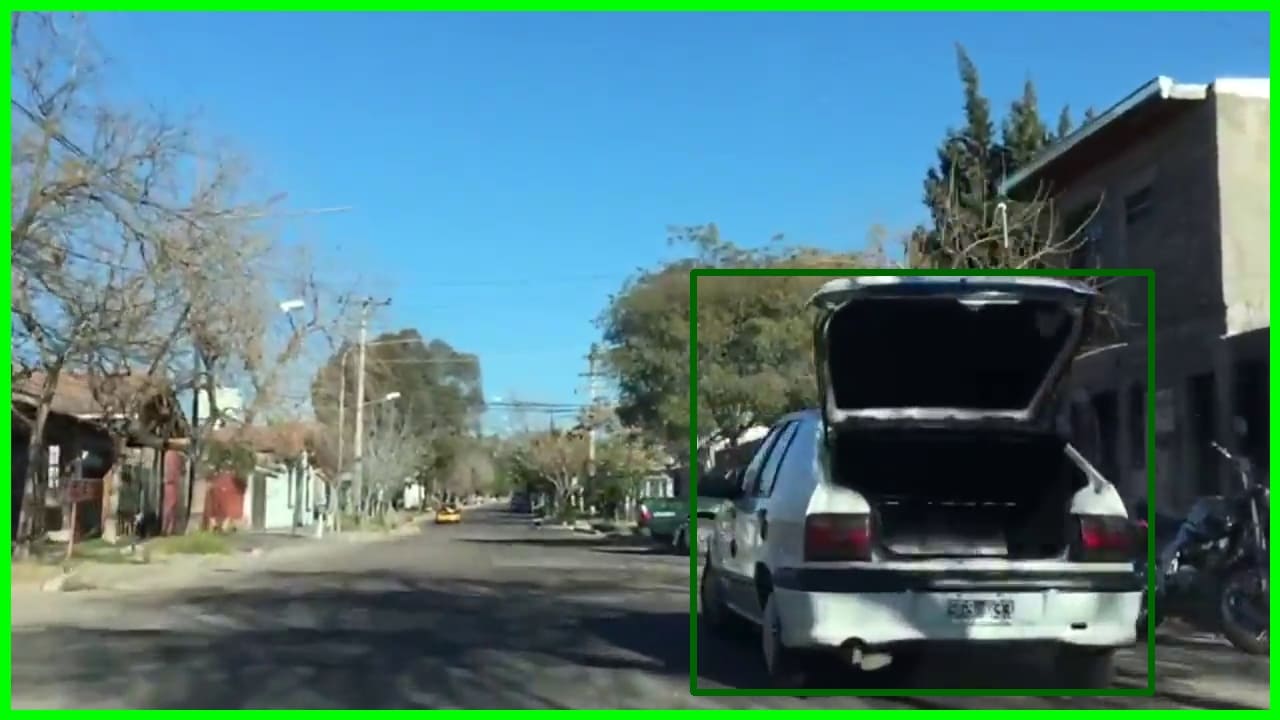}
    \end{subfigure}
    \begin{subfigure}[b]{\subfigwidth}
        \includegraphics[width=\textwidth]{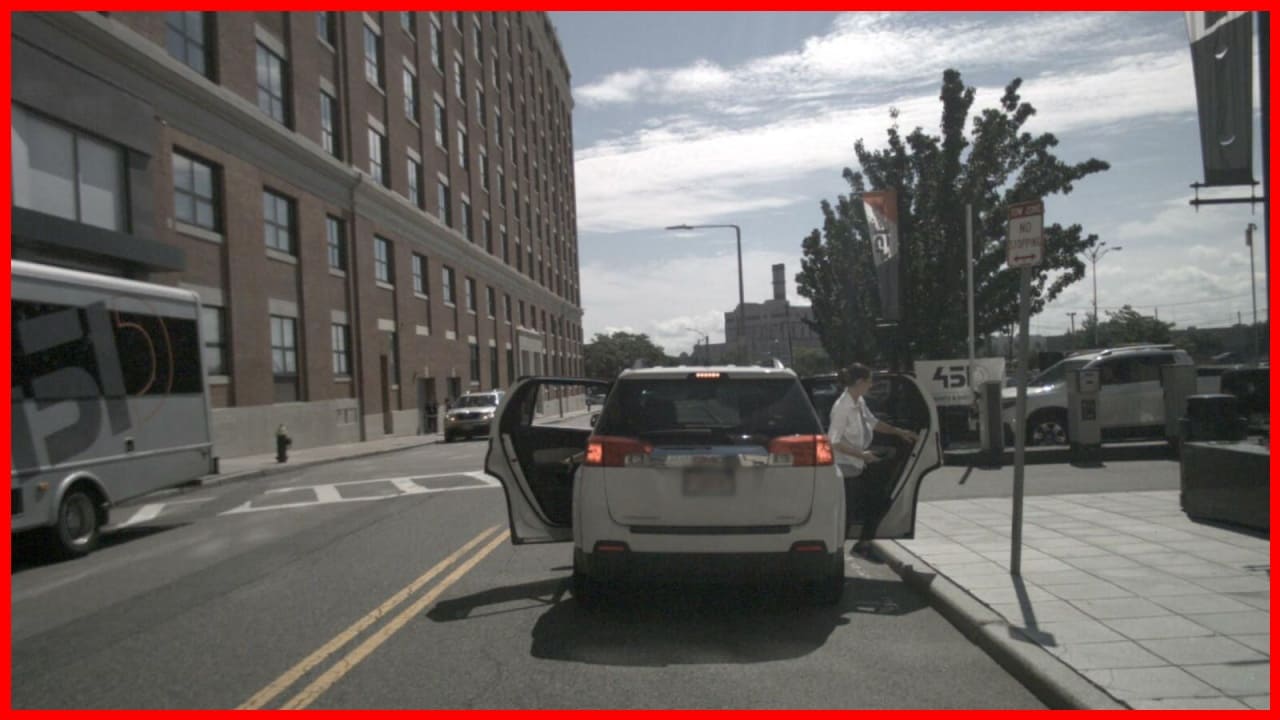}
    \end{subfigure}
    \begin{subfigure}[b]{\subfigwidth}
        \includegraphics[width=\textwidth]{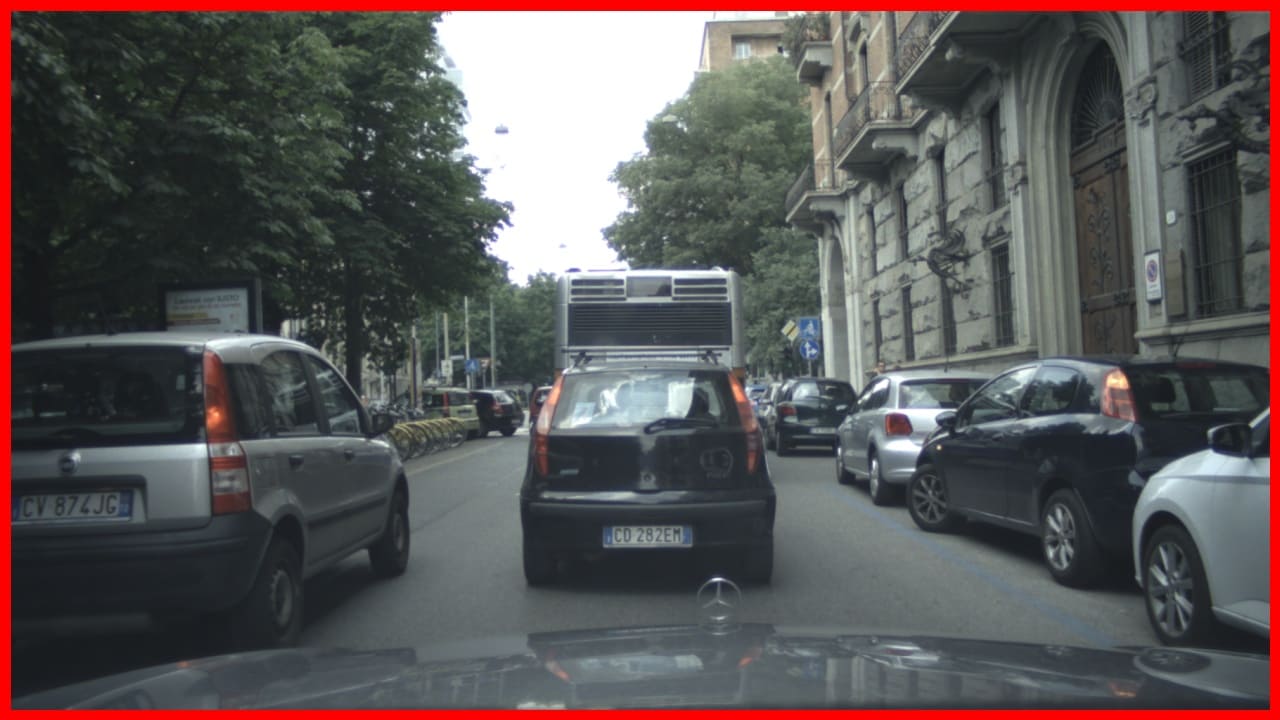}
    \end{subfigure}
    \begin{subfigure}[b]{\subfigwidth}
        \includegraphics[width=\textwidth]{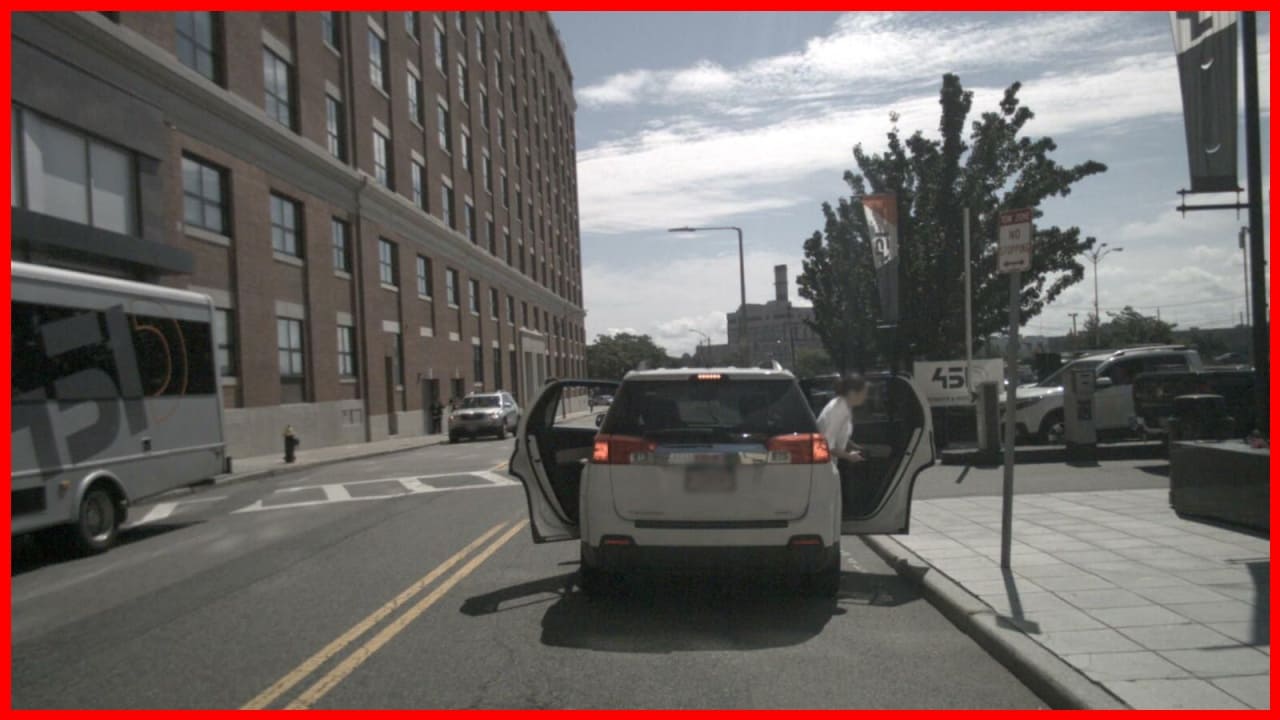}
    \end{subfigure}
    \begin{subfigure}[b]{\subfigwidth}
        \includegraphics[width=\textwidth]{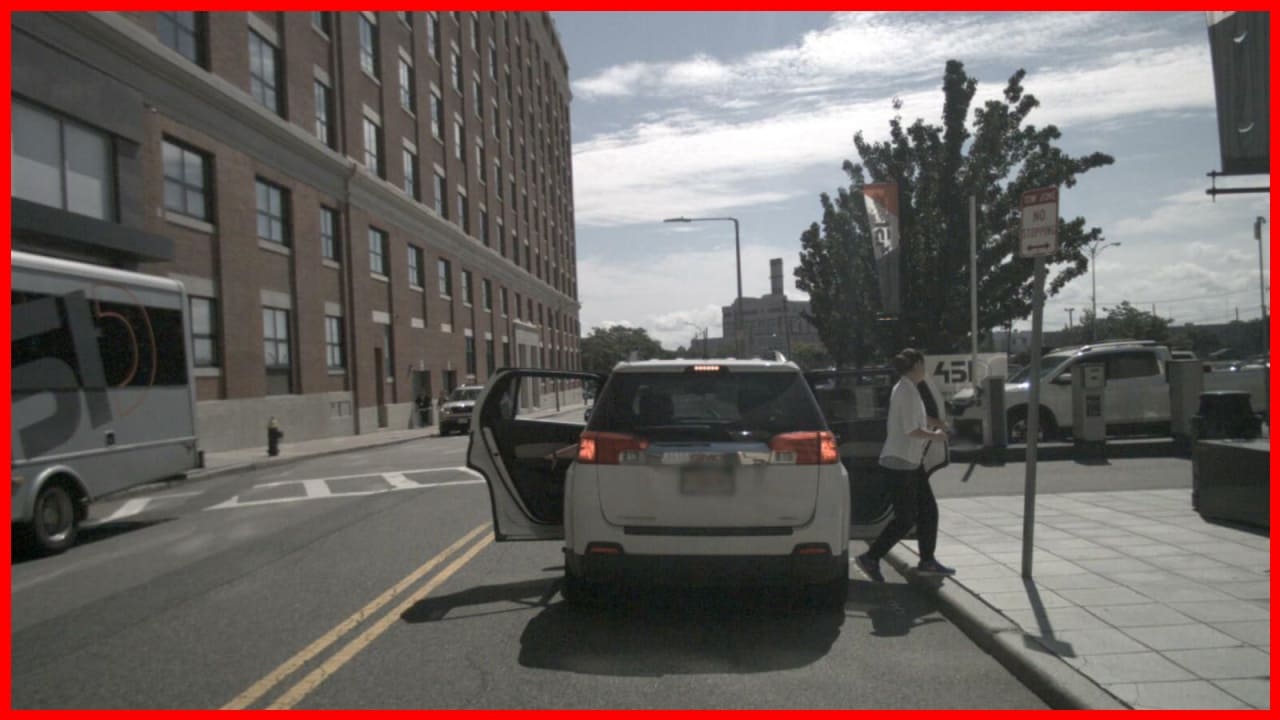}
    \end{subfigure} \\

    \rotatebox[origin=left]{90}{\hspace{-0.29cm} \textbf{METACLIP2}} 
    \begin{subfigure}[b]{\subfigwidth}
        \includegraphics[width=\textwidth]{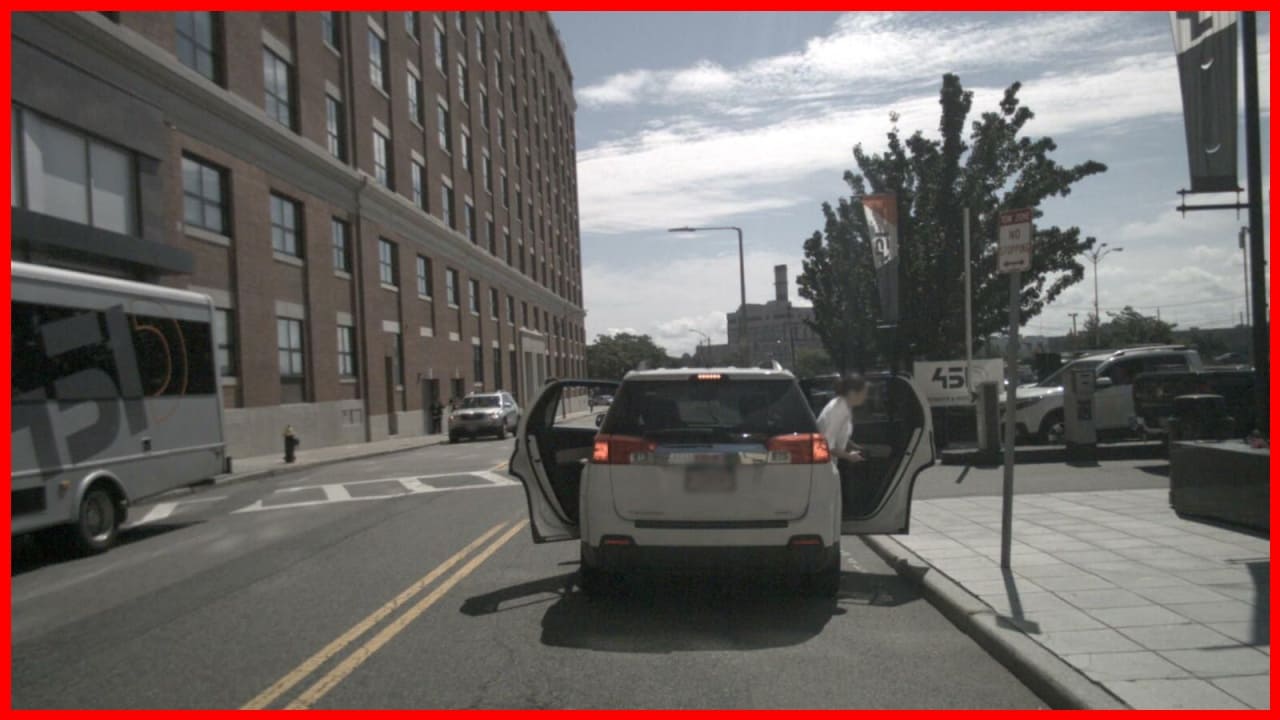}
    \end{subfigure}
    \begin{subfigure}[b]{\subfigwidth}
        \includegraphics[width=\textwidth]{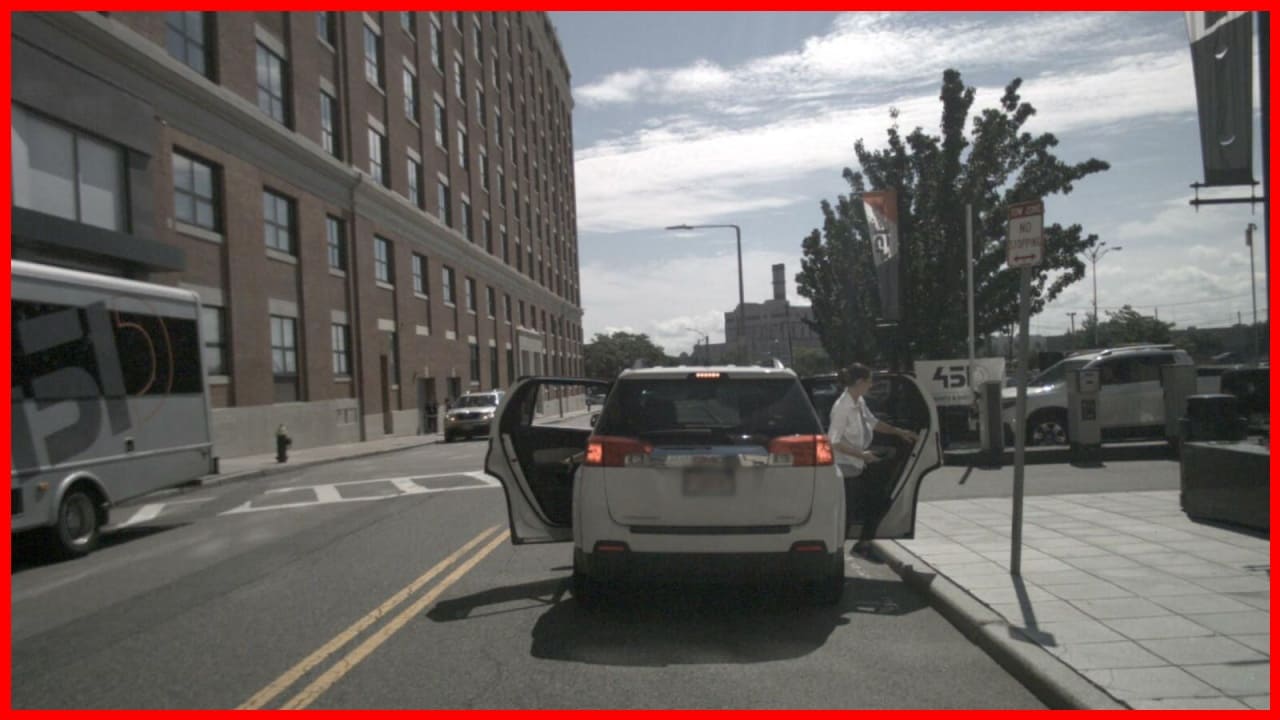}
    \end{subfigure}
    \begin{subfigure}[b]{\subfigwidth}
        \includegraphics[width=\textwidth]{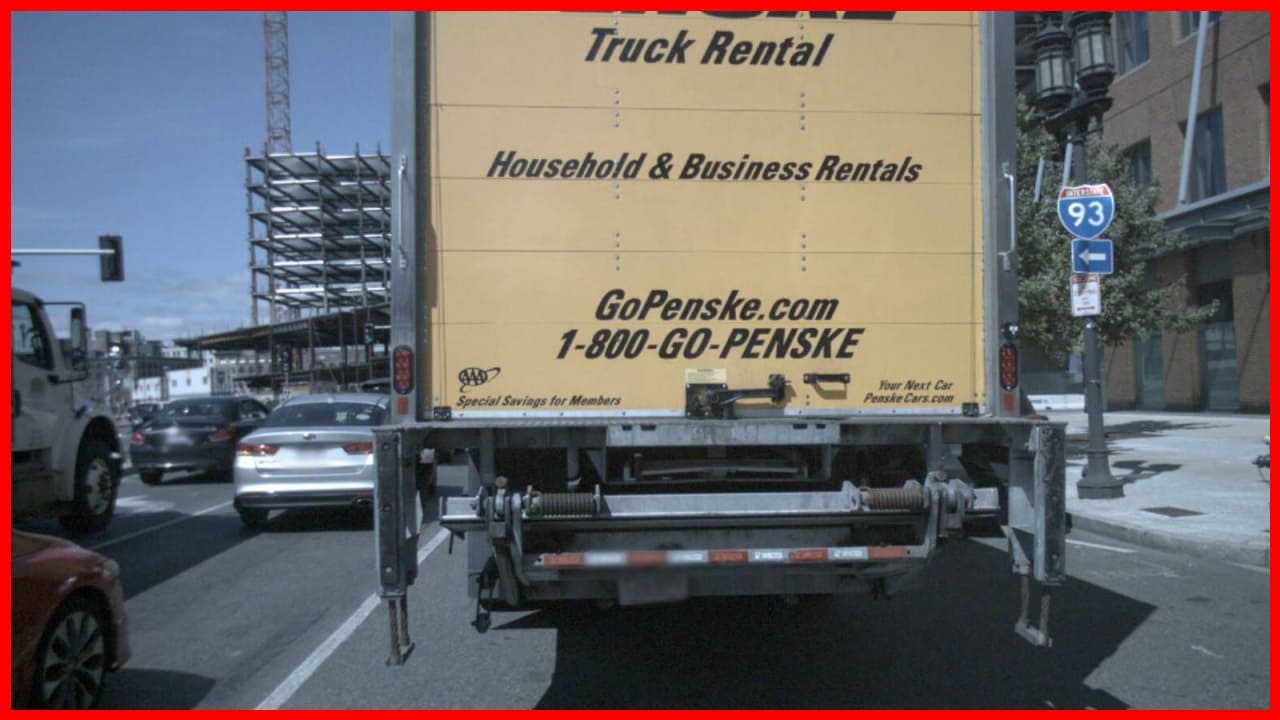}
    \end{subfigure}
    \begin{subfigure}[b]{\subfigwidth}
        \includegraphics[width=\textwidth]{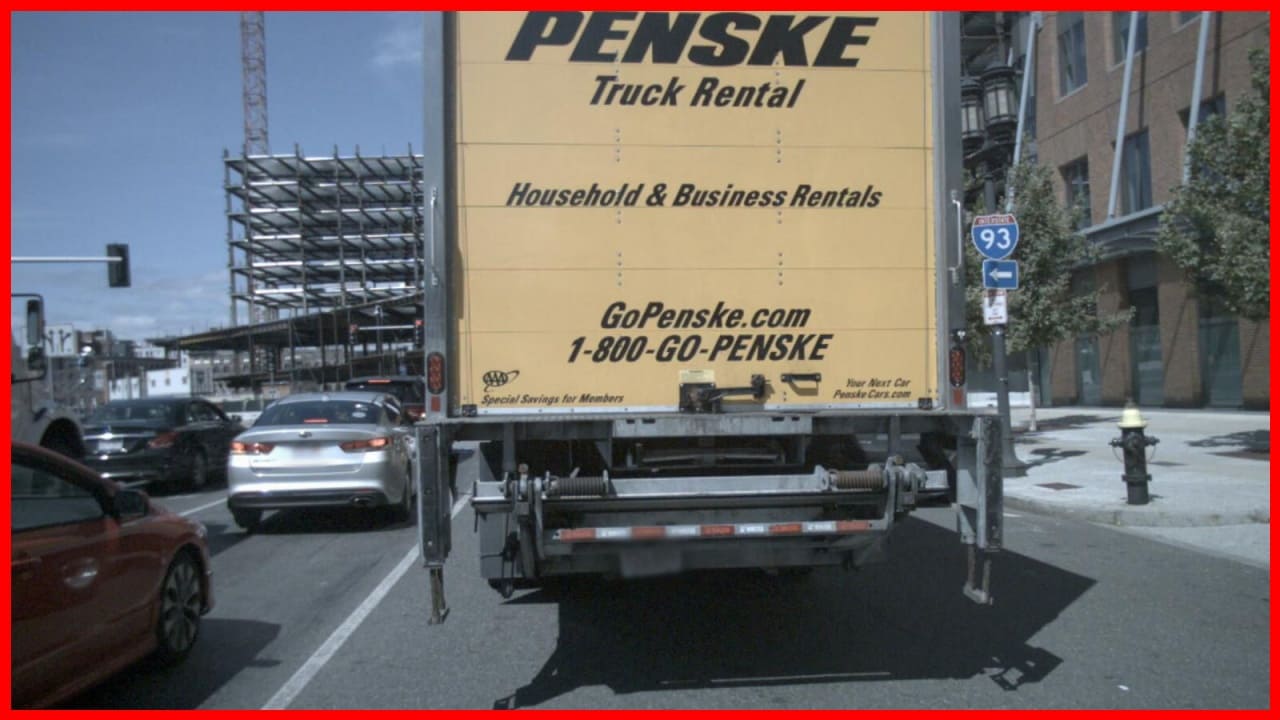}
    \end{subfigure}
    \begin{subfigure}[b]{\subfigwidth}
        \includegraphics[width=\textwidth]{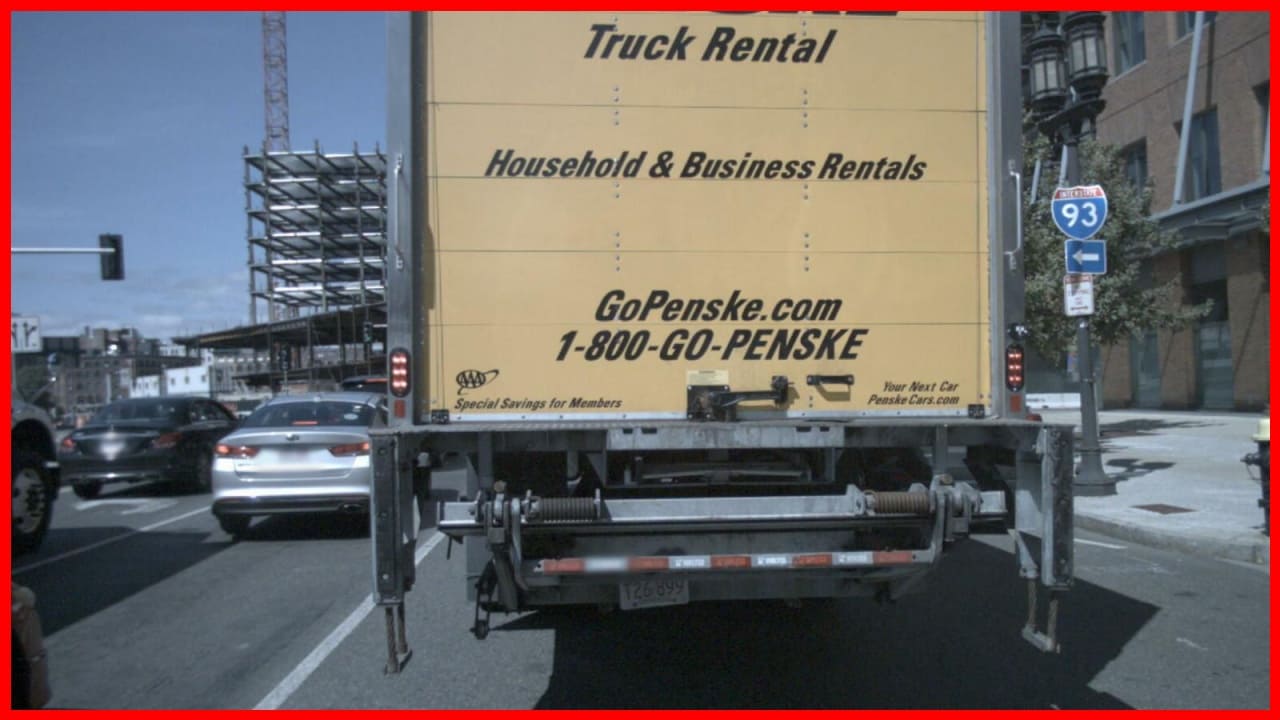}
    \end{subfigure} \\

    \rotatebox[origin=left]{90}{\hspace{0.02cm} \textbf{SIGLIP2}} 
    \begin{subfigure}[b]{\subfigwidth}
        \includegraphics[width=\textwidth]{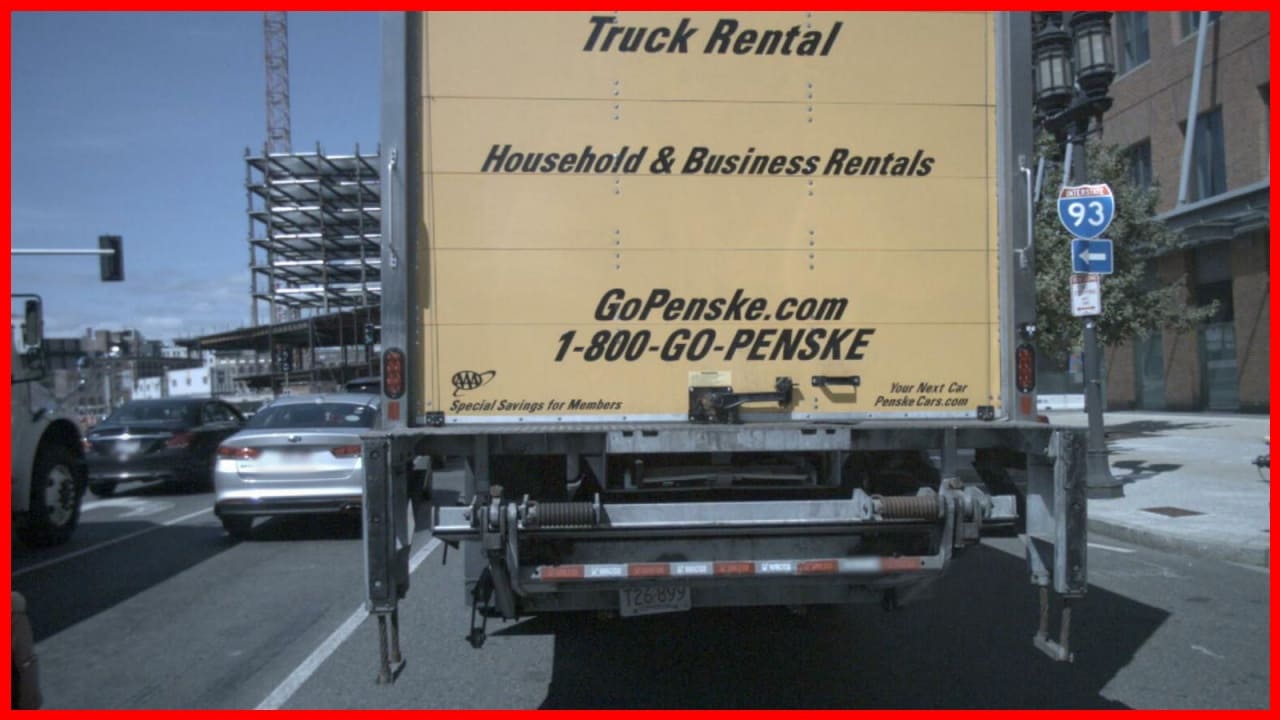}
    \end{subfigure}
    \begin{subfigure}[b]{\subfigwidth}
        \includegraphics[width=\textwidth]{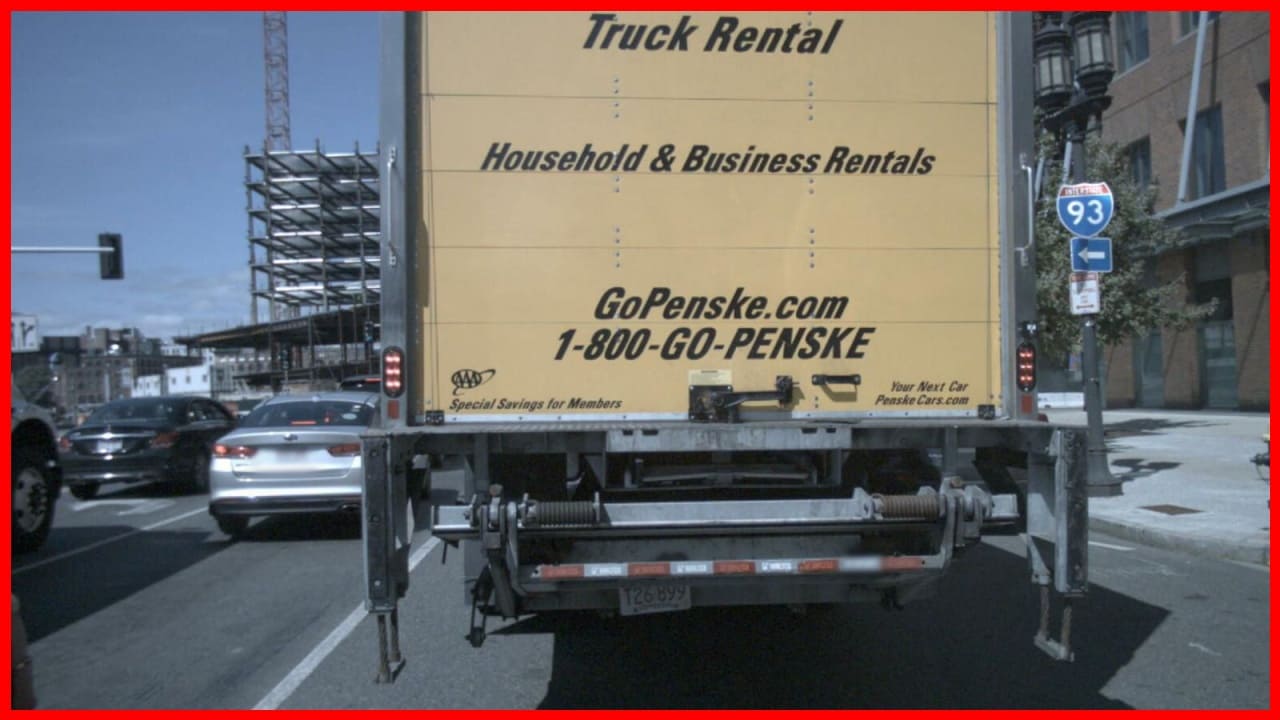}
    \end{subfigure}
    \begin{subfigure}[b]{\subfigwidth}
        \includegraphics[width=\textwidth]{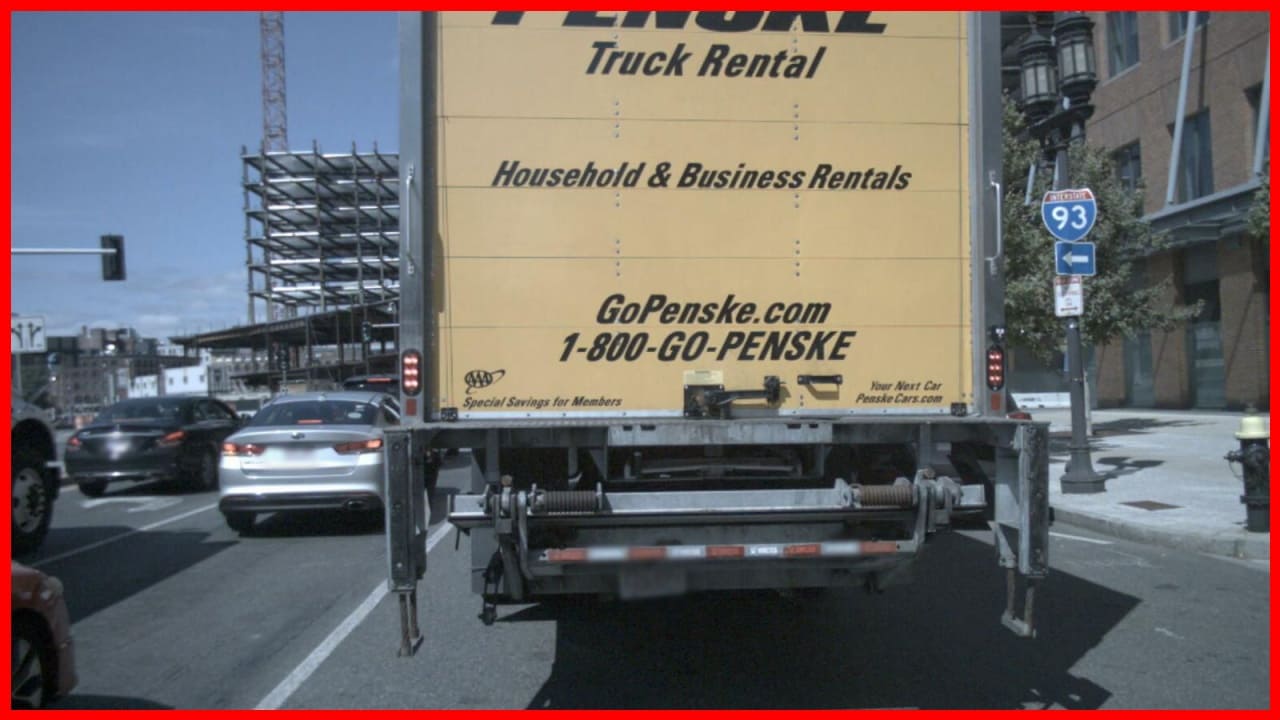}
    \end{subfigure}
    \begin{subfigure}[b]{\subfigwidth}
        \includegraphics[width=\textwidth]{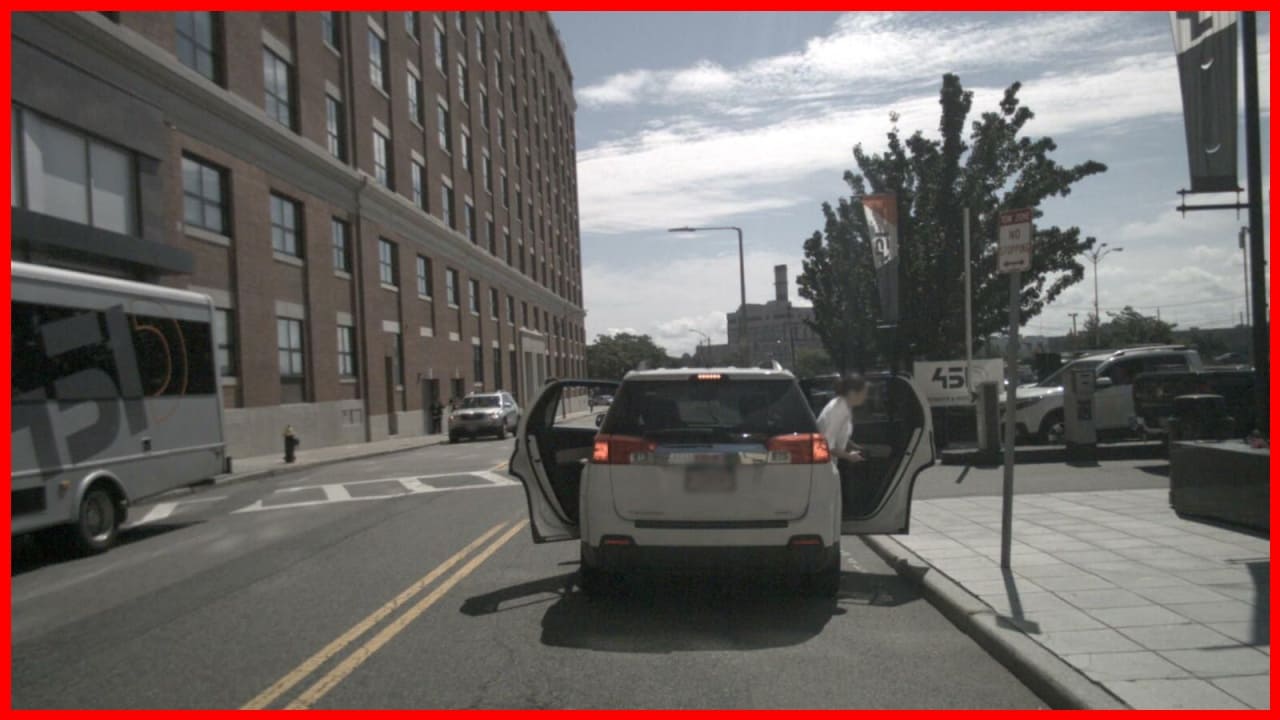}
    \end{subfigure}
    \begin{subfigure}[b]{\subfigwidth}
        \includegraphics[width=\textwidth]{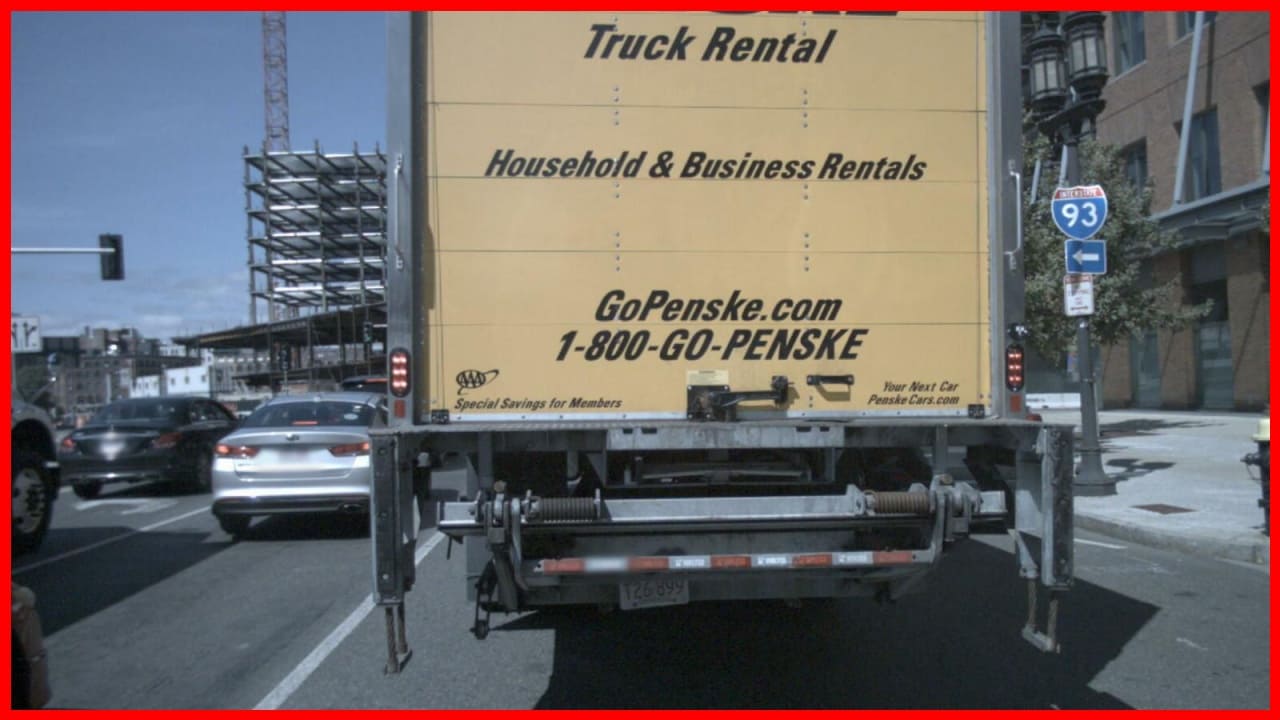}
    \end{subfigure} \\

    \rotatebox[origin=left]{90}{\hspace{0.04cm} \textbf{NACLIP}} 
    \begin{subfigure}[b]{\subfigwidth}
        \includegraphics[width=\textwidth]{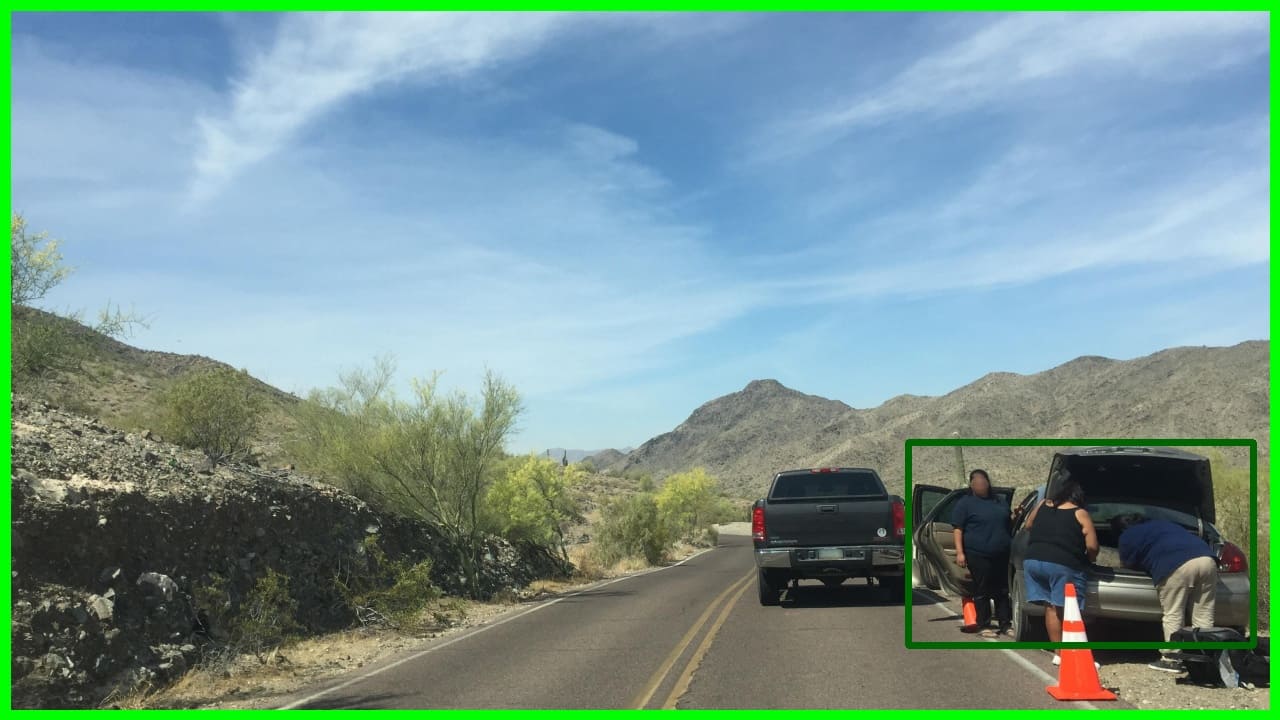}
    \end{subfigure}
    \begin{subfigure}[b]{\subfigwidth}
        \includegraphics[width=\textwidth]{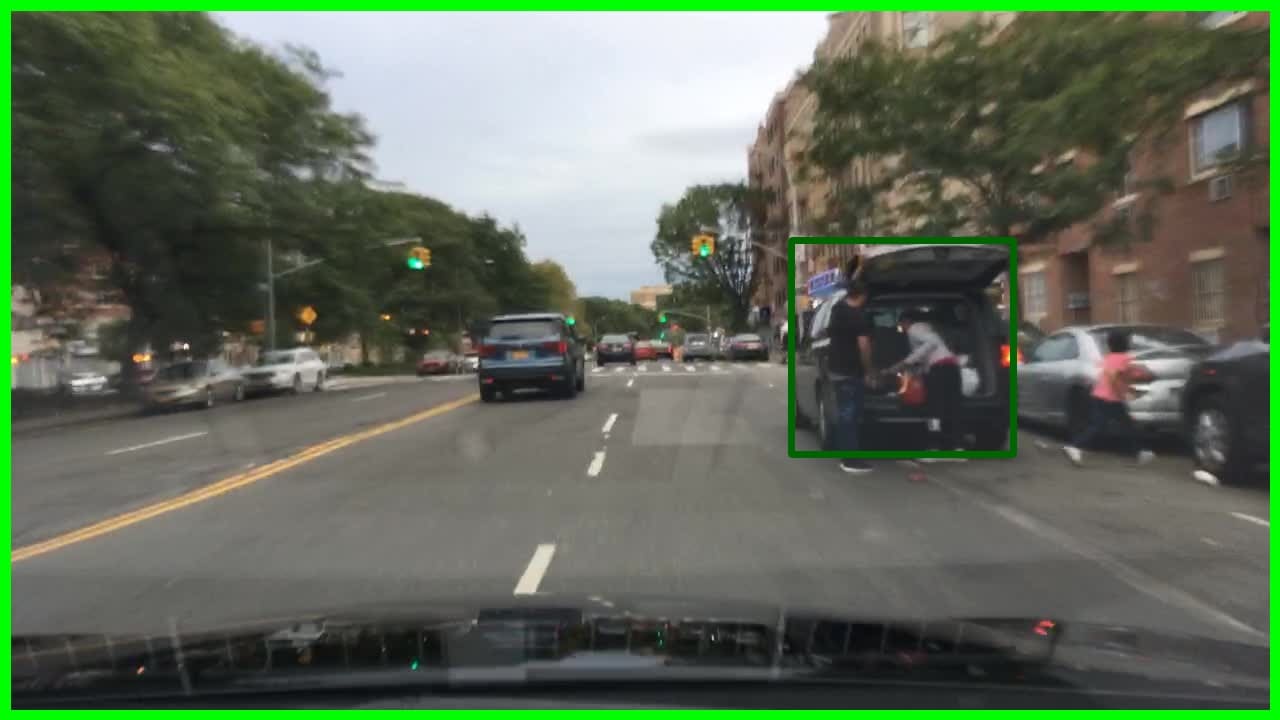}
    \end{subfigure}
    \begin{subfigure}[b]{\subfigwidth}
        \includegraphics[width=\textwidth]{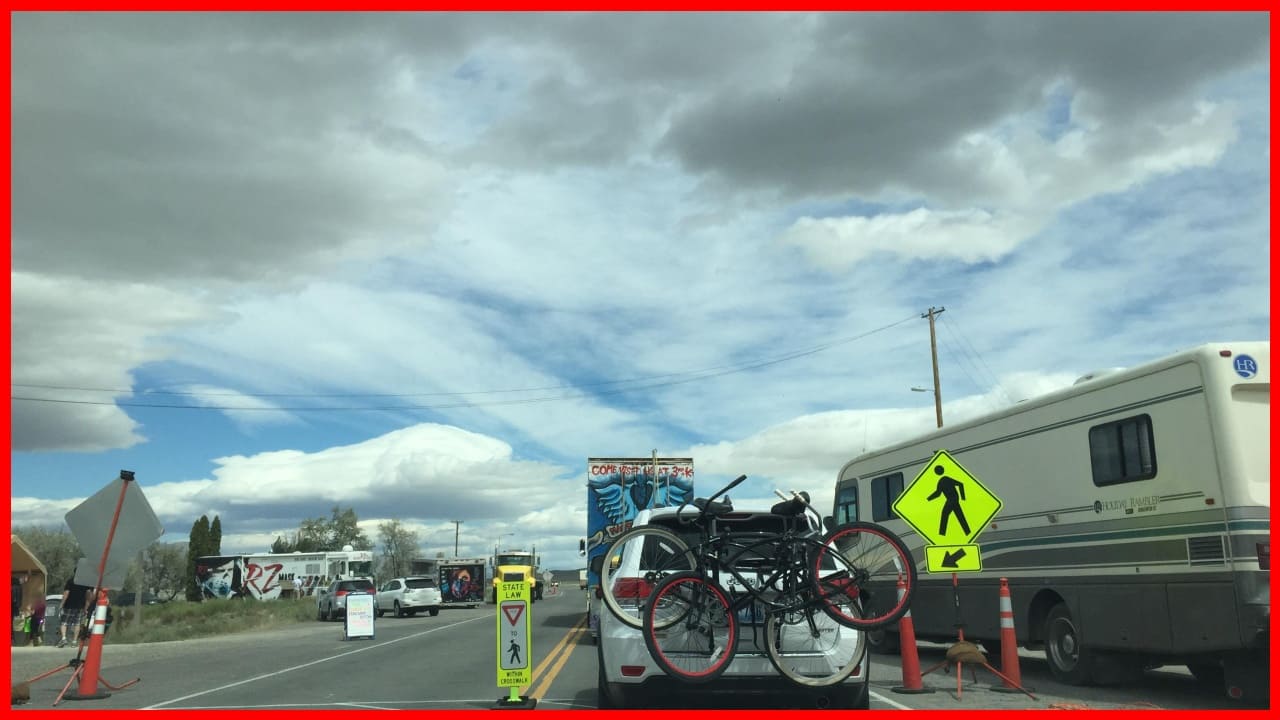}
    \end{subfigure}
    \begin{subfigure}[b]{\subfigwidth}
        \includegraphics[width=\textwidth]{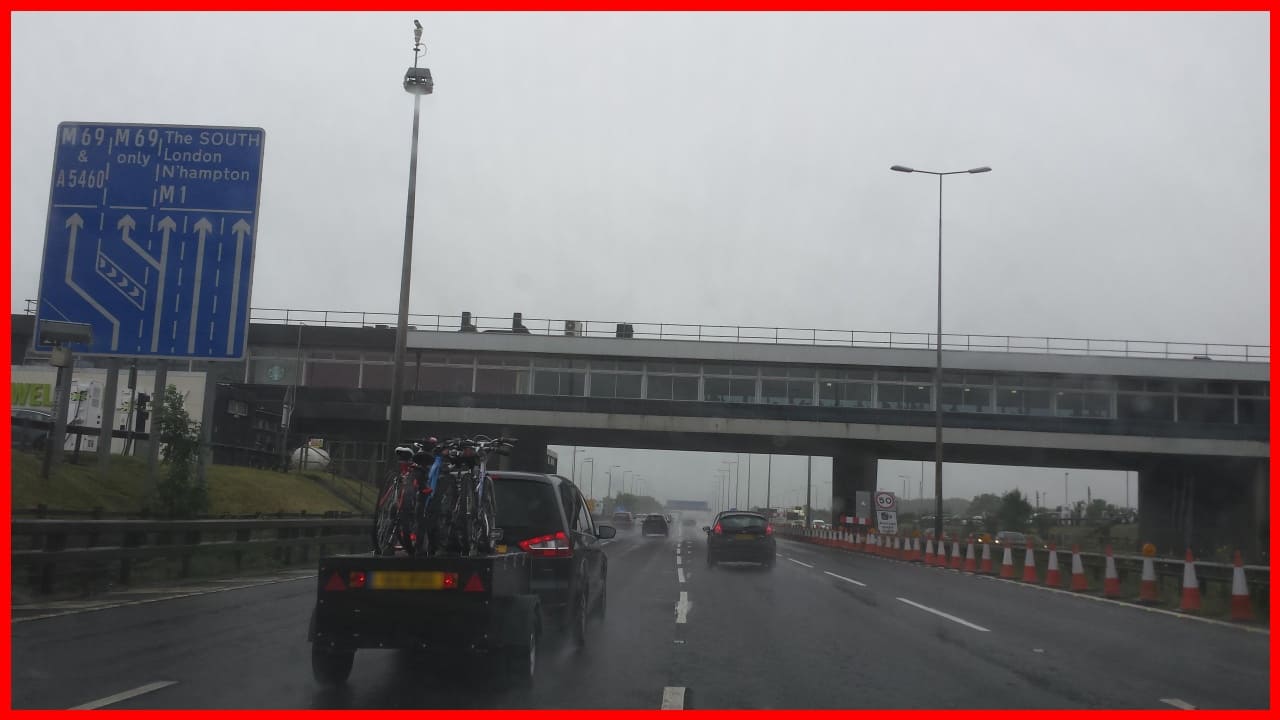}
    \end{subfigure}
    \begin{subfigure}[b]{\subfigwidth}
        \includegraphics[width=\textwidth]{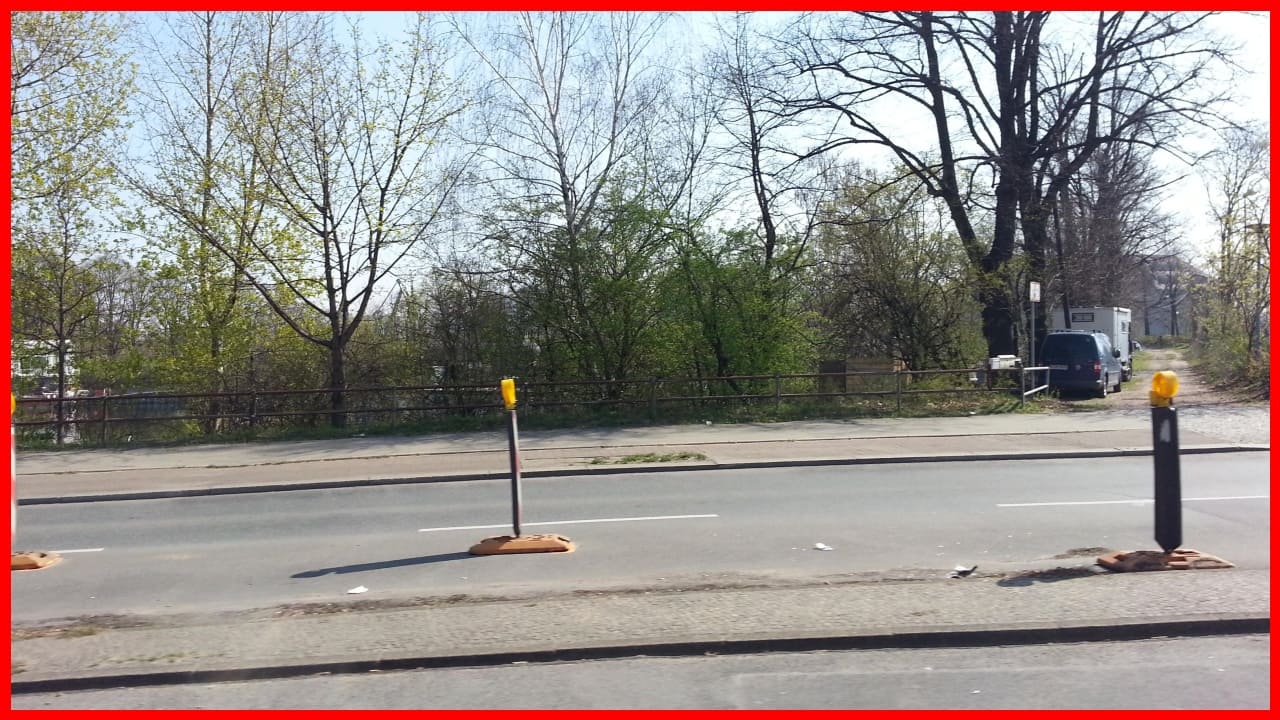}
    \end{subfigure} \\

    \rotatebox[origin=left]{90}{\hspace{-0.15cm} \textbf{NARADIO}} 
    \begin{subfigure}[b]{\subfigwidth}
        \includegraphics[width=\textwidth]{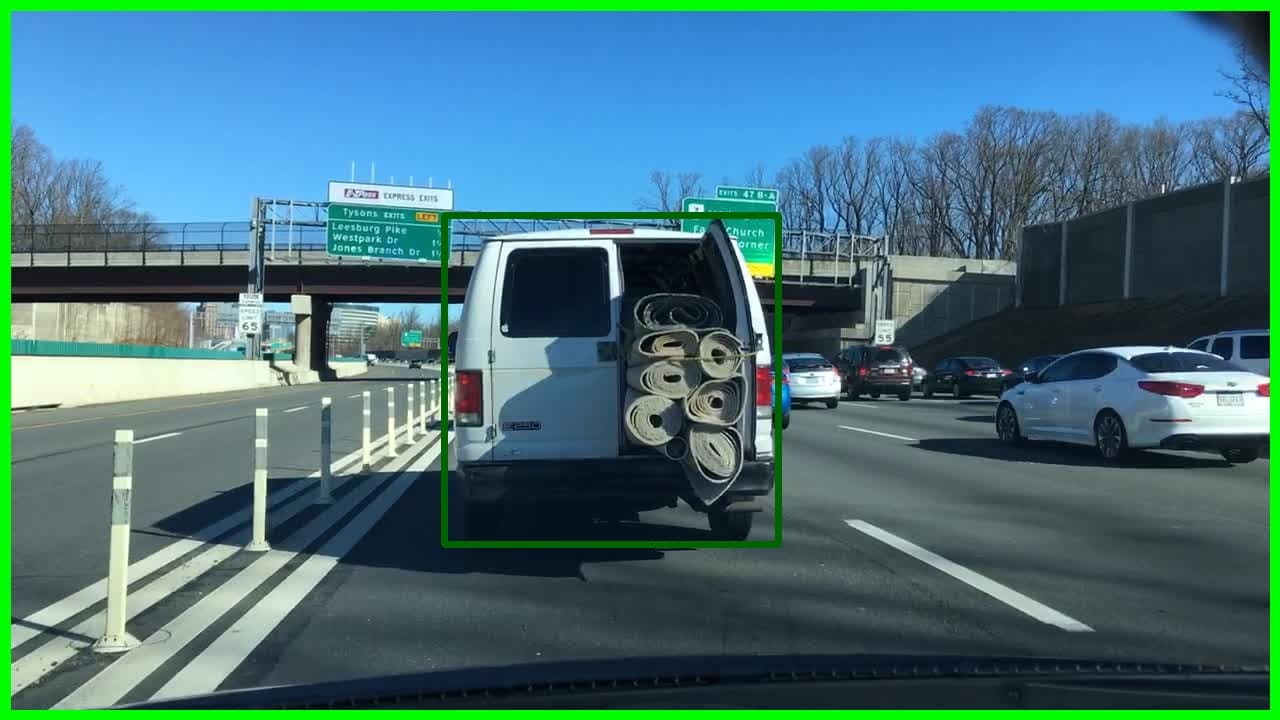}
    \end{subfigure}
    \begin{subfigure}[b]{\subfigwidth}
        \includegraphics[width=\textwidth]{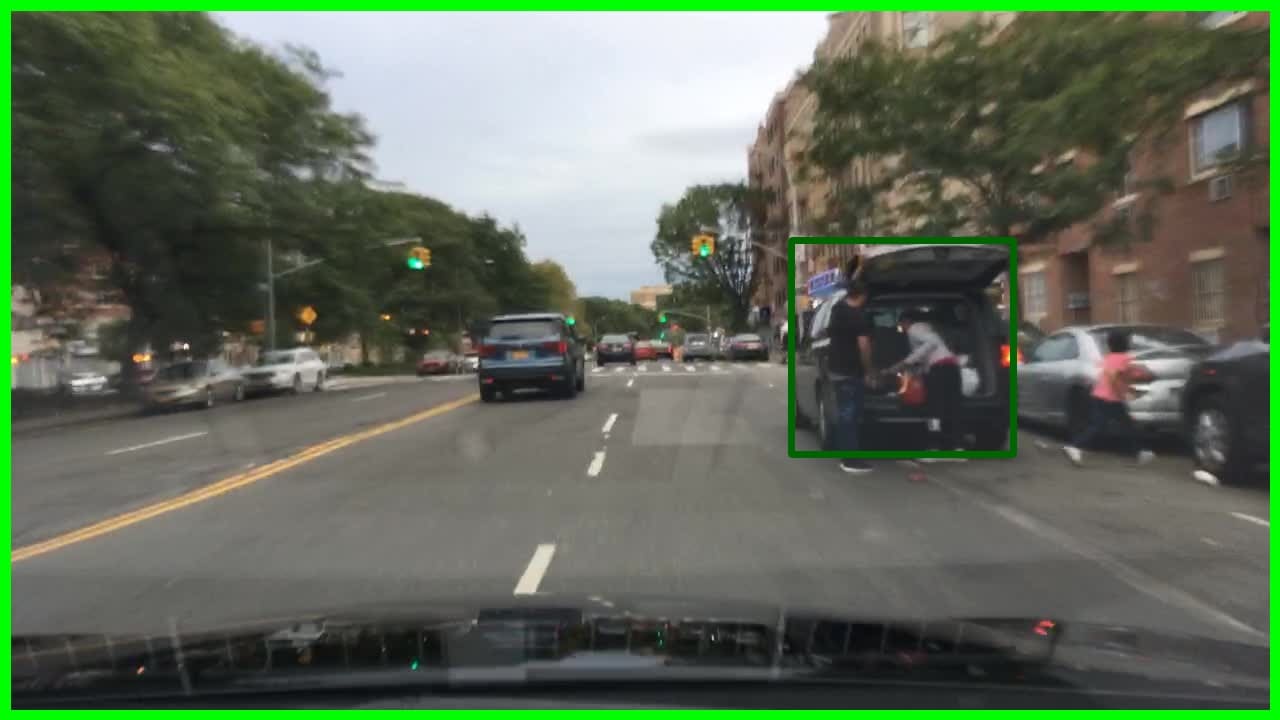}
    \end{subfigure}
    \begin{subfigure}[b]{\subfigwidth}
        \includegraphics[width=\textwidth]{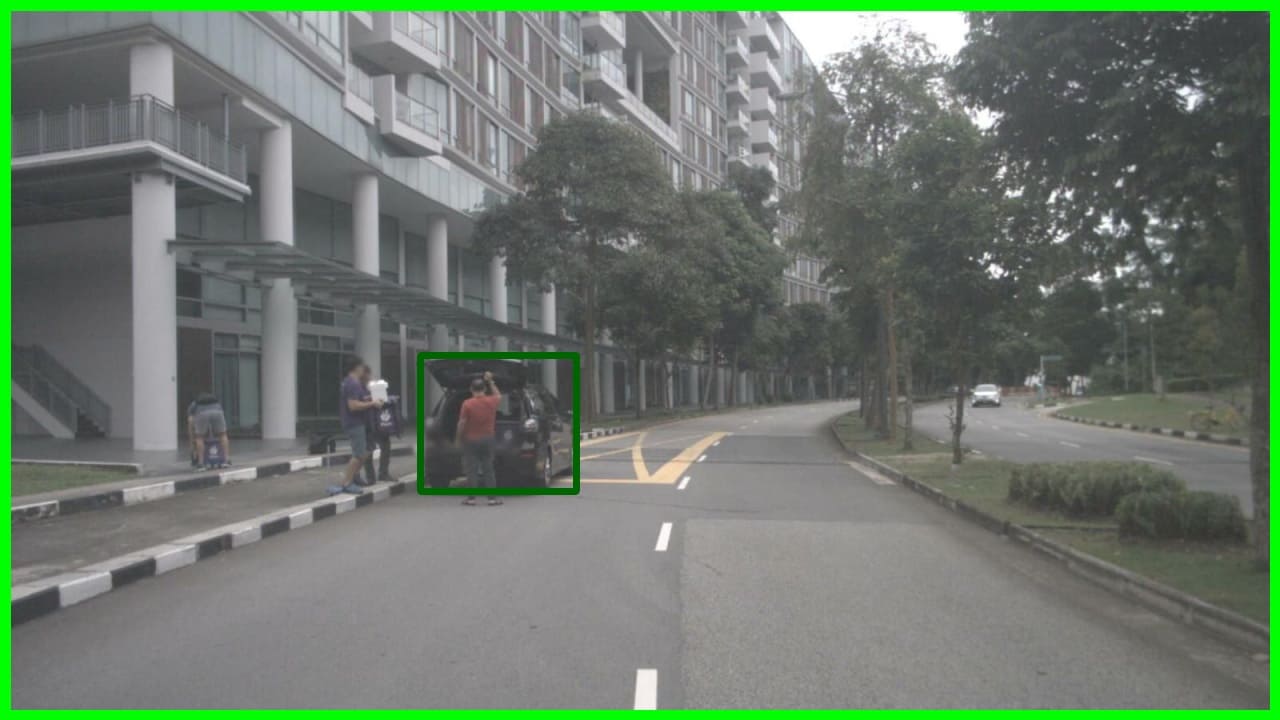}
    \end{subfigure}
    \begin{subfigure}[b]{\subfigwidth}
        \includegraphics[width=\textwidth]{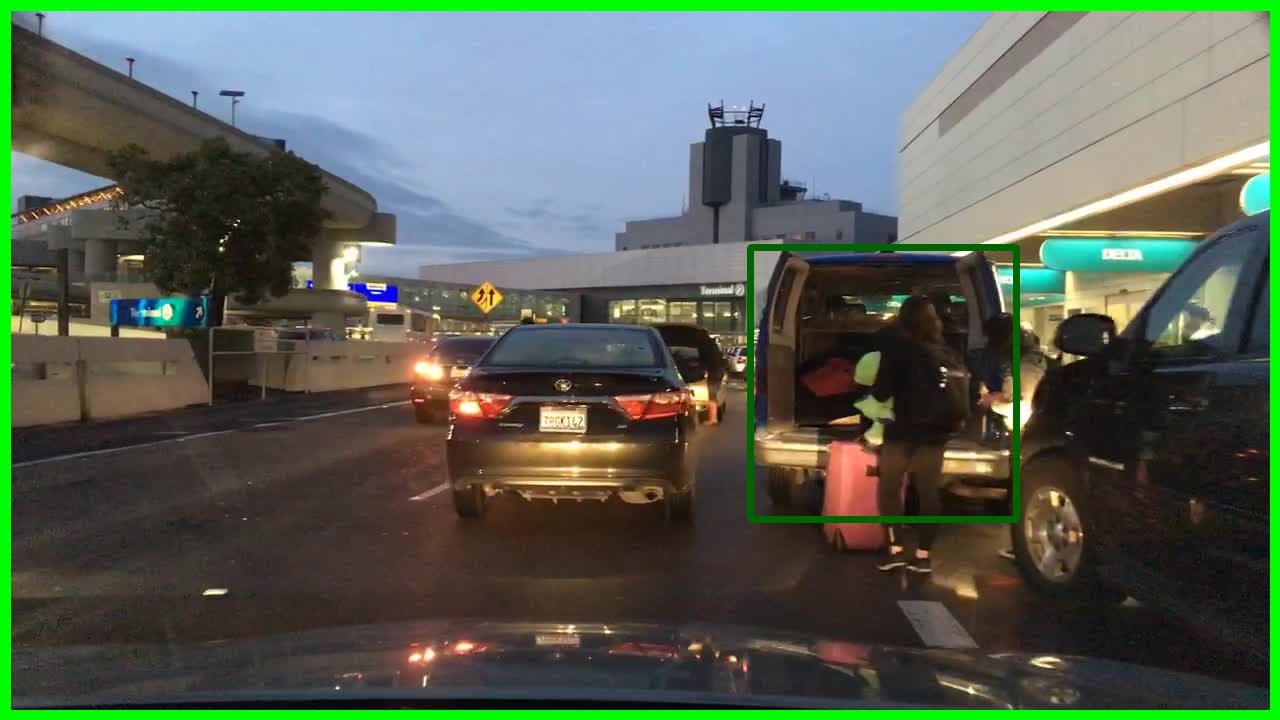}
    \end{subfigure}
    \begin{subfigure}[b]{\subfigwidth}
        \includegraphics[width=\textwidth]{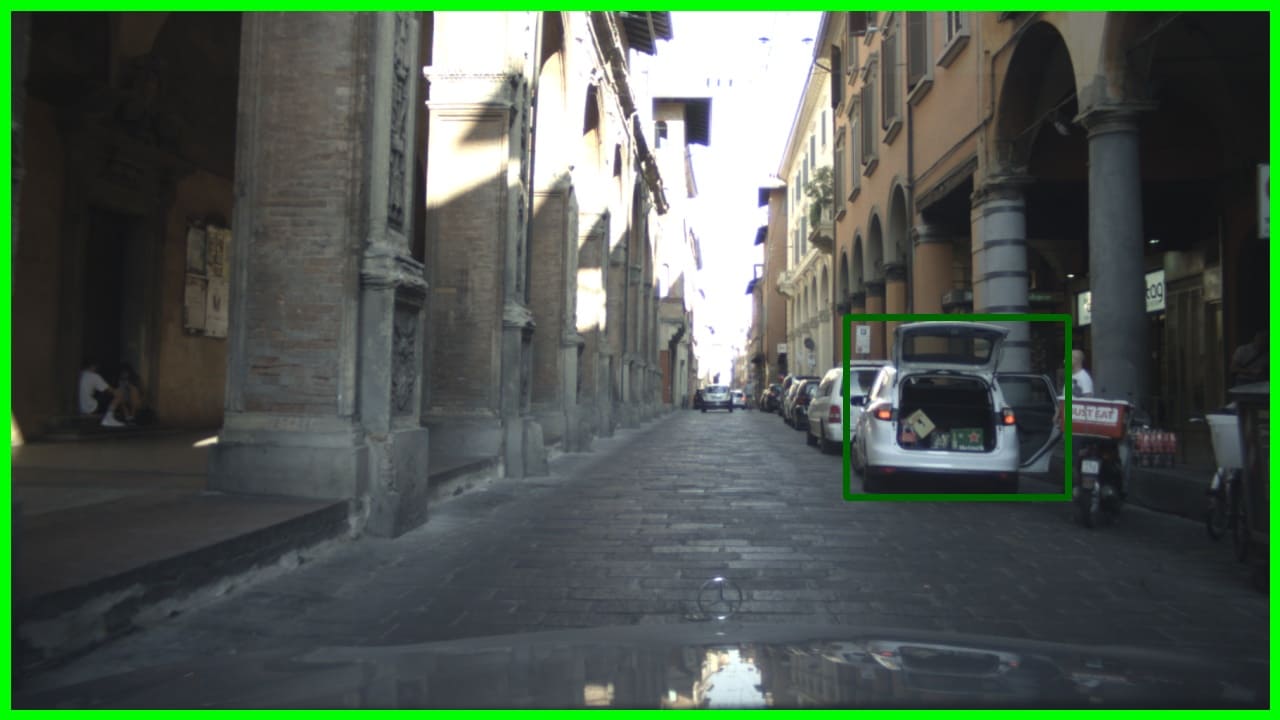}
    \end{subfigure} \\
    \caption{Text-based retrieval results on the validation split for the \emph{Scene-Open-Trunk} class, showing top 5 ranked images for each model. Model references can be found in \Cref{tab:map_class_wise_text_to_image_retrieval}.}
    \label{fig:scene_open_trunk_results}
    \endgroup
\end{figure*}
}

{
\onecolumn
\small
\centering

\begin{longtable}{>{\centering}p{0.6cm} p{5cm} p{4.0cm} p{4.0cm}}
    \toprule
    \textbf{Class} & \textbf{Label Description} & \textbf{Positive Examples} & \textbf{Negative Examples} \\
    \midrule
    \endfirsthead

    \multicolumn{4}{c}{\tablename\ \thetable\ -- Continued from previous page} \\
    \toprule
    \textbf{Class} & \textbf{Label Description} & \textbf{Positive Examples} & \textbf{Negative Examples} \\
    \midrule
    \endhead

    \midrule
    \multicolumn{4}{r}{\tablename\ \thetable\ -- \textit{Continued on next page}} \\
    \endfoot

    \endlastfoot

    \vspace{0cm}
    \rotatebox{90}{\textbf{Human-Construction-Worker}} &
    \vspace{0cm}
    \begin{itemize}
        \item Ideal case: Person with yellow vest and helmet, standing on a construction site
        \item Person with helmet
        \item Person with warning / high-visibility vest (vest does not have to be yellow or orange)
        \item To not label the person inside a (construction) vehicle. Only when it is a rideable vehicle or when the door is open and he/she is getting in or out
        \item People standing within the construction site or moving materials should also be labeled as Pedestrian Duty Construction.
        \item Hard negative example: Police officer or fire fighter with warning vest, other duty worker (\eg, garbage collector)
    \end{itemize}
    &
    {
    \vspace{0cm}
    \includegraphics[width=\linewidth]{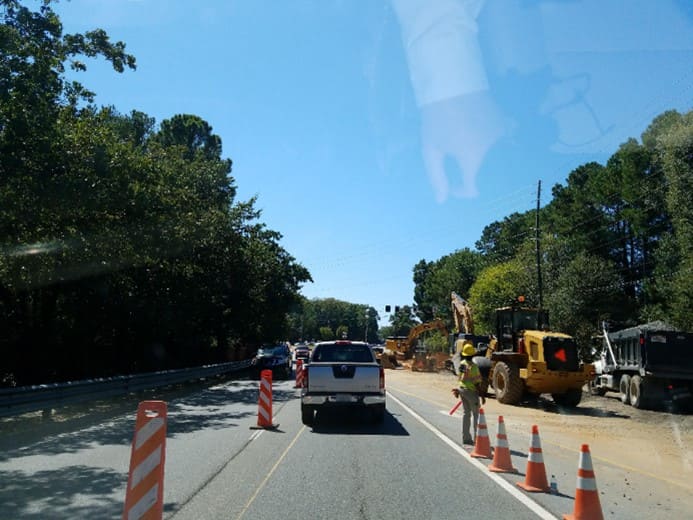} %
    \hfill 
    \includegraphics[width=\linewidth]{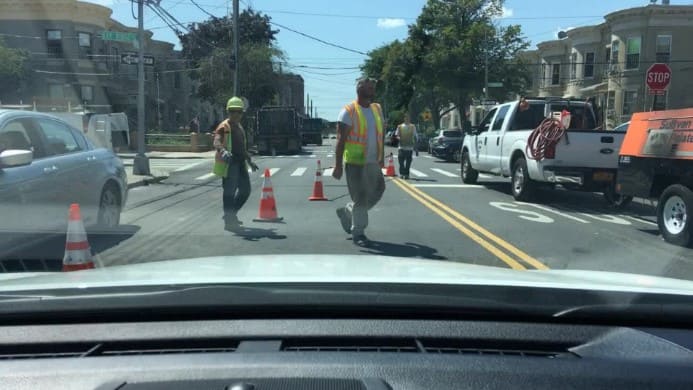}
    \hfill
    \includegraphics[width=0.5\linewidth]{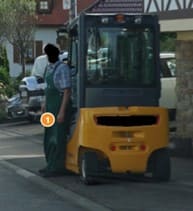}
    \hfill
    } &
    { 
    \vspace{0cm}
    \includegraphics[width=\linewidth]{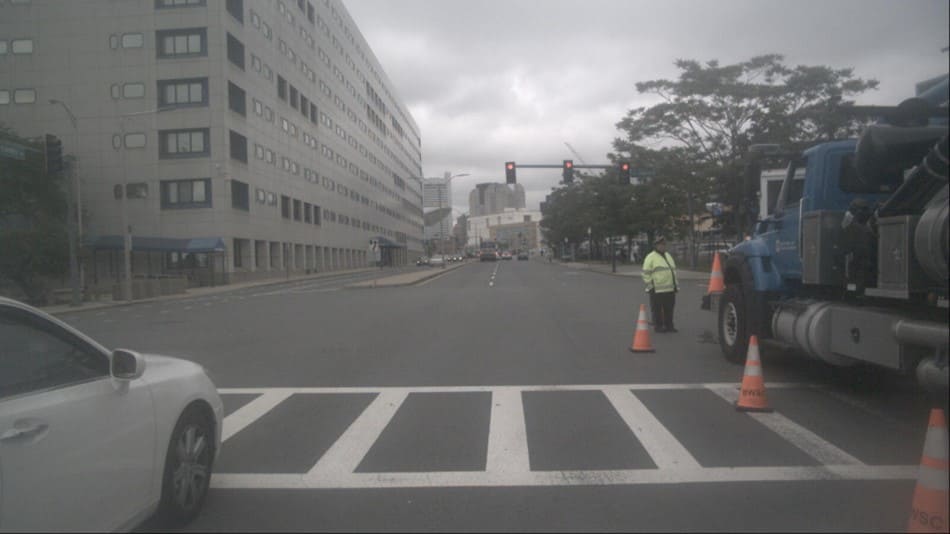} \hfill
    } \\
    \hline

    \vspace{0cm}
    \rotatebox{90}{\textbf{Animal-Real-Dog}} &
    \vspace{0cm}
    \begin{itemize}
        \item All kinds of dogs
        \item Artificial dog \eg, dog-dummy / puppet (\eg, in Lost and Found  \cite{LostFounddetectingPinggera2016} dataset) that is indistinguishable from a real dog should also be counted as a dog
        \item Any 3D dog dummies of dogs should be also counted as dog dataset)
        \item Hard negative example: a dog printed on a poster (2D dog)
    \end{itemize}
    &
    {
    \vspace{0cm}
    \includegraphics[width=\linewidth]{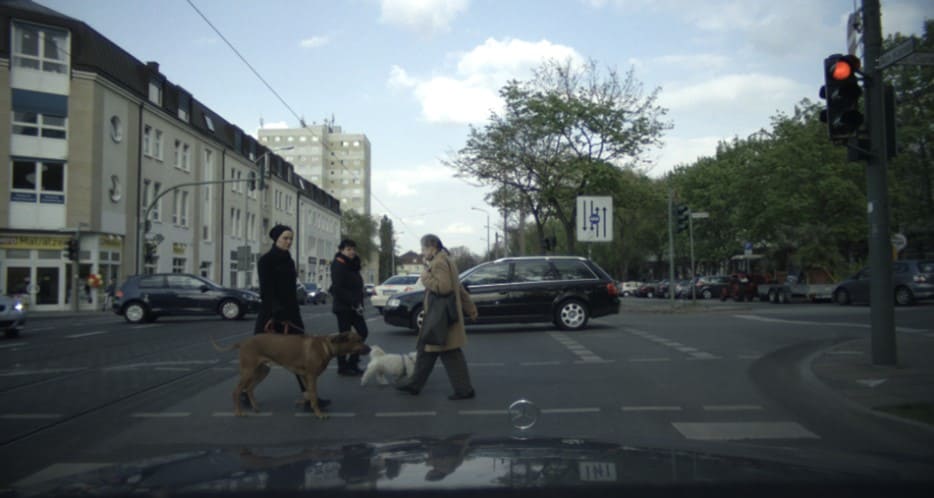} %
    \hfill 
    \includegraphics[width=\linewidth]{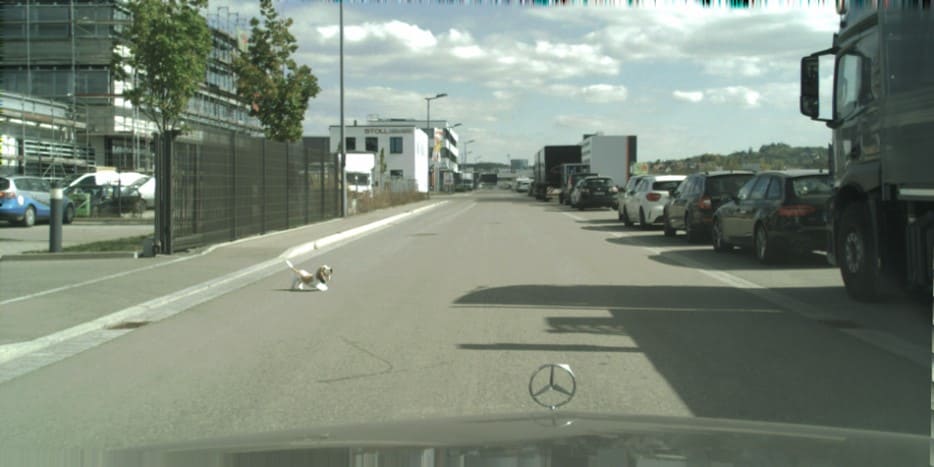}
    \hfill
    } &
    { 
    \vspace{0cm}
    \includegraphics[width=\linewidth]{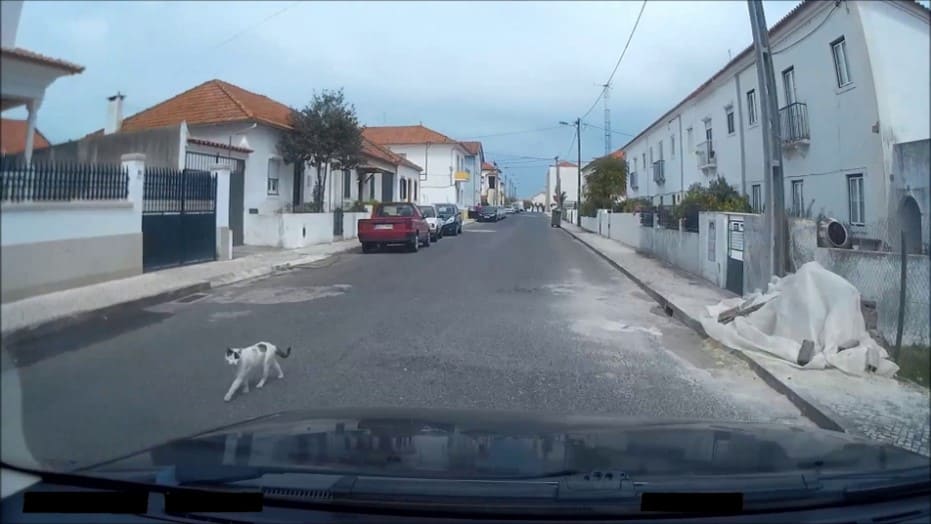} \hfill
    } \\
    \hline

    \vspace{0cm}
    \rotatebox{90}{\textbf{Marking-Bus-Text}} &
    \vspace{0cm}
    \begin{itemize}
        \item Find the text pattern “BUS” on the road
        \item The text pattern must be on the road, not on a traffic sign or somewhere else
        \item The text pattern “BUS” can be in any color
        \item It can be occluded, but it has to be clear that it is the “BUS” text pattern, \eg, if it is identifiable by the context
        \item Label based on context — for example, if there’s a nearby bus station or a bus stop sign, it should be annotated.
        \item Hard negative example: Any other text pattern, which looks similar to the “BUS” text pattern
    \end{itemize}
    &
    {
    \vspace{0cm}
    \includegraphics[width=\linewidth]{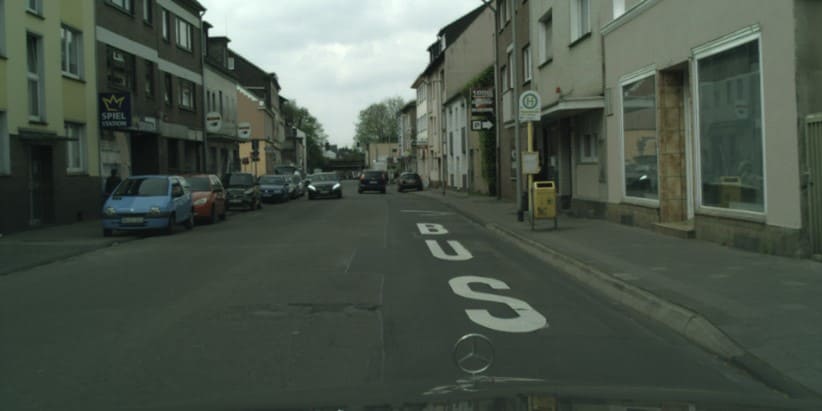} %
    \hfill 
    \includegraphics[width=\linewidth]{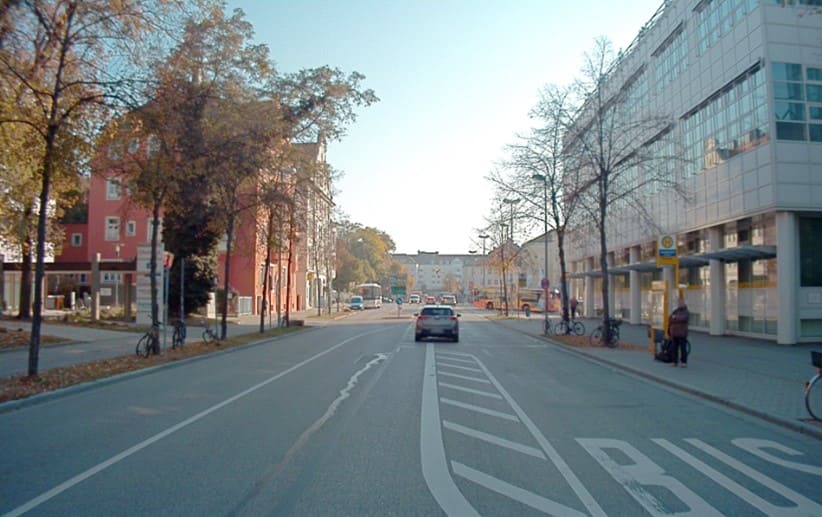}
    \hfill
    } &
    { 
    \vspace{0cm}
    \includegraphics[width=\linewidth]{images_supp/labeling_guidelines/marking_bus_text/marking_bus_text_neg1.jpg}
    \hfill
    \includegraphics[width=\linewidth]{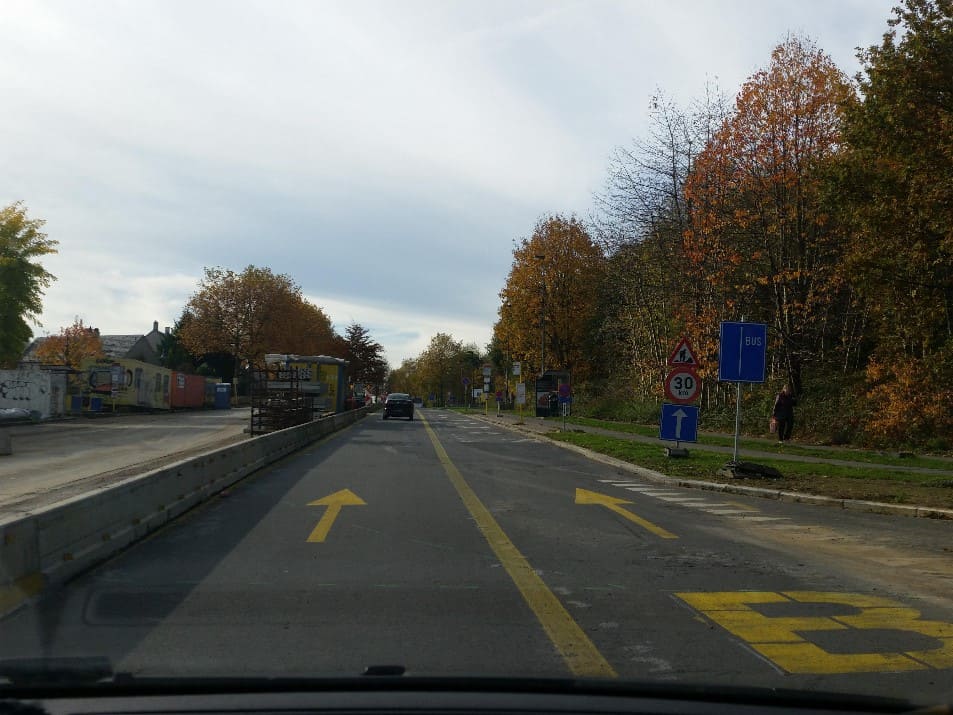}
    \hfill
    } \\

    \vspace{0cm}
    \rotatebox{90}{\textbf{Object-Ball}} &
    \vspace{0cm}
    \begin{itemize}
        \item All kinds of balls (football ball, tennis ball, basketball ball, …)
        \item The ball must be real (in 3D) and not shown on a poster
        \item Bounding box should include the person carrying or playing the ball if there is one. Otherwise label only the ball itself
        \item Hard negative examples: Ball printed on a poster, football field, where you cannot see a football, helmets, any other spherical objects, \eg, round street lamps
    \end{itemize}
    &
    {
    \vspace{0cm}
    \includegraphics[width=\linewidth]{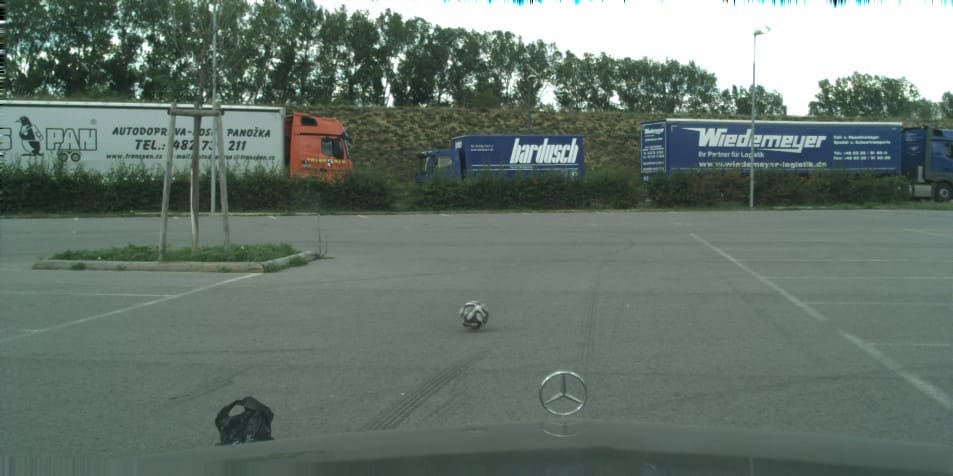} %
    \hfill 
    \includegraphics[width=\linewidth]{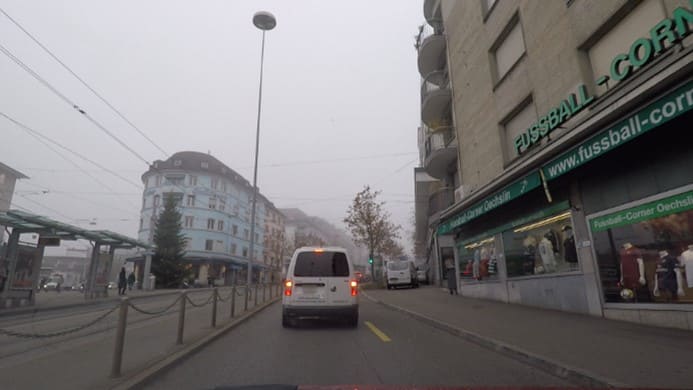}
    \hfill
    \includegraphics[width=\linewidth]{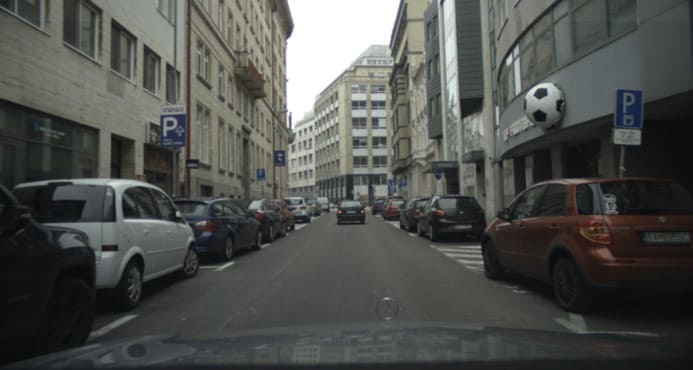}
    \hfill
    } &
    { 
    \vspace{0cm}
    \includegraphics[width=\linewidth]{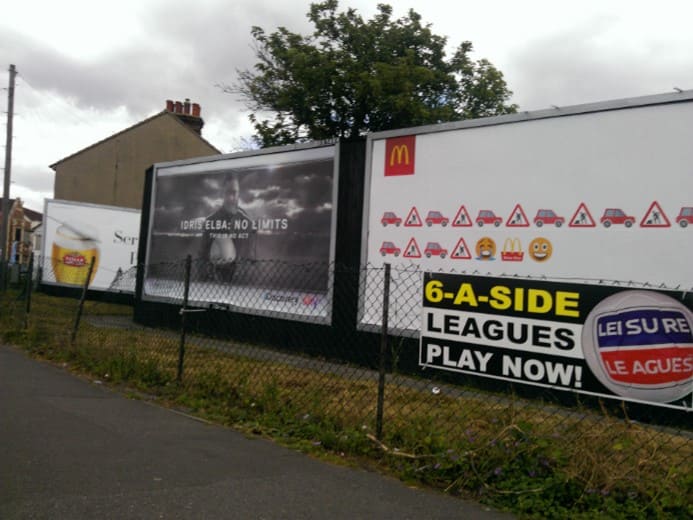}
    \hfill
    } \\
    \hline

    \vspace{0cm}
    \rotatebox{90}{\textbf{Rideable-Wheelchair}} &
    \vspace{0cm}
    \begin{itemize}
        \item Wheelchair usually has four wheels and is designed for people who cannot walk
        \item Also if there is no person on the wheelchair
        \item Usually the wheelchair is pushed by someone but some also have a motor
        \item Bounding box should include the person in the wheelchair if there is one. Otherwise label only the wheelchair itself
        \item Hard positive example: Wheelchair without any person on the sidewalk
        \item Hard negative example: Rollator, Shopping trolley, Stroller, Shopping cart
    \end{itemize}
    &
    {
    \vspace{0cm}
    \includegraphics[width=\linewidth]{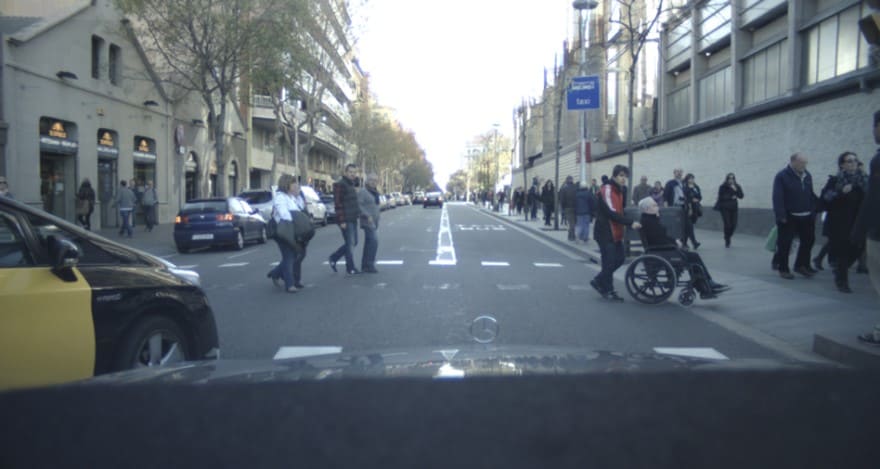} %
    \hfill 
    \includegraphics[width=\linewidth]{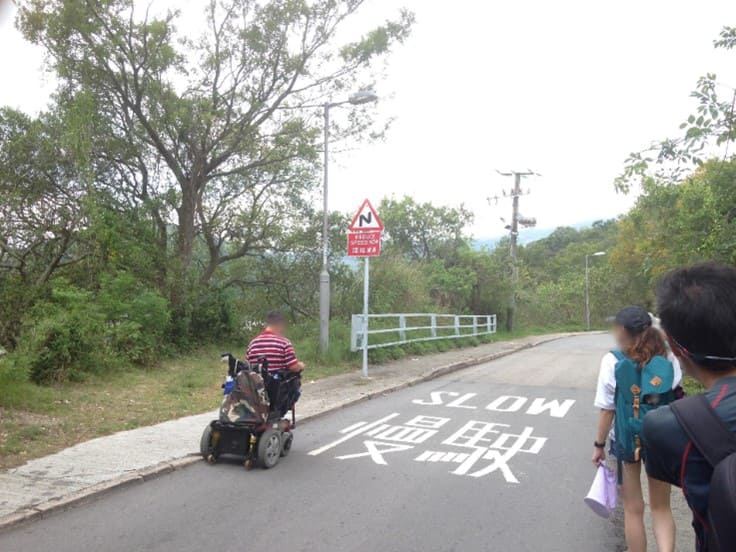}
    \hfill
    \includegraphics[width=\linewidth]{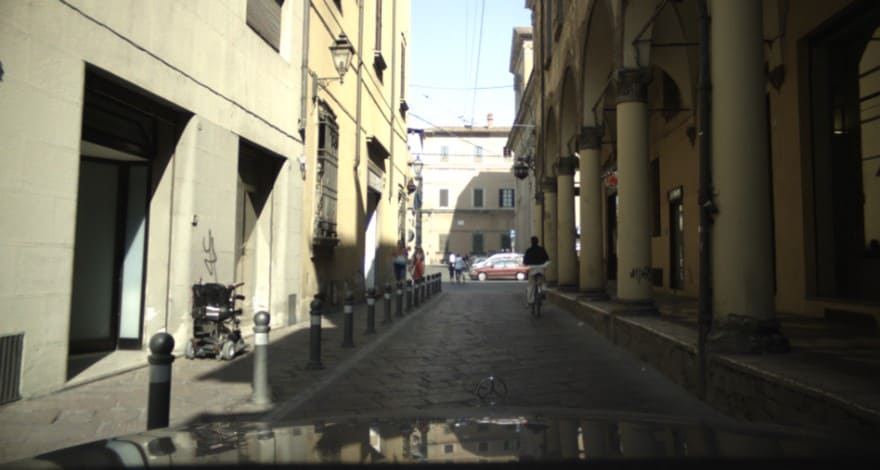}
    \hfill
    } &
    { 
    \vspace{0cm}
    \includegraphics[width=\linewidth]{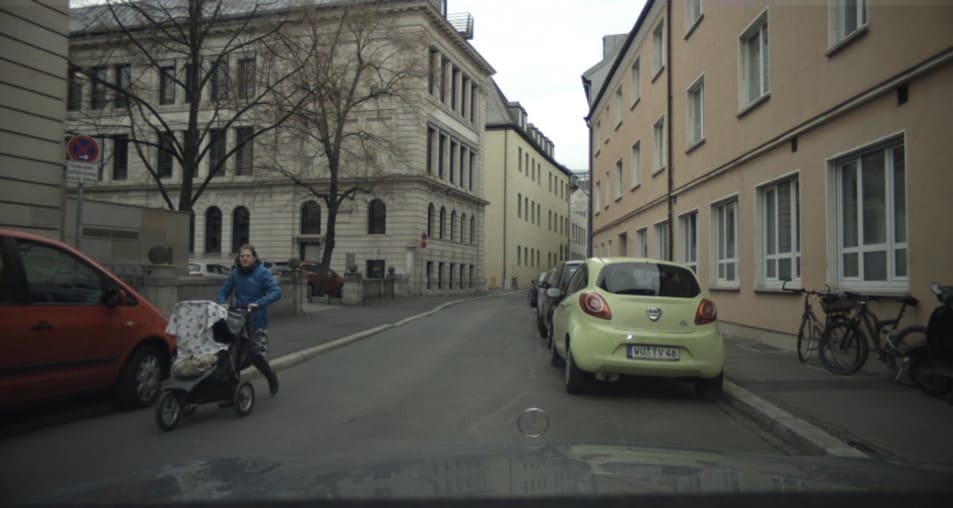} \hfill
    \includegraphics[width=\linewidth]{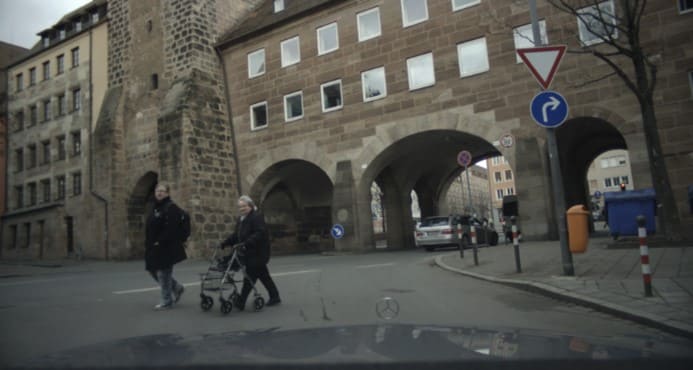} \hfill
    } \\
    \hline

    \vspace{0cm}
    \rotatebox{90}{\textbf{Scene-Fog}} &
    \vspace{0cm}
    \begin{itemize}
        \item Definition of fog: When visibility is less than one kilometer
        \item Otherwise it is only haze
        \item Hard negative example: Poor visibility due to heavy rain
        \item Choose a large part of the image where you can also see the effect of the fog and use this as bounding box (see example)

    \end{itemize}
    &
    {
    \vspace{0cm}
    \includegraphics[width=\linewidth]{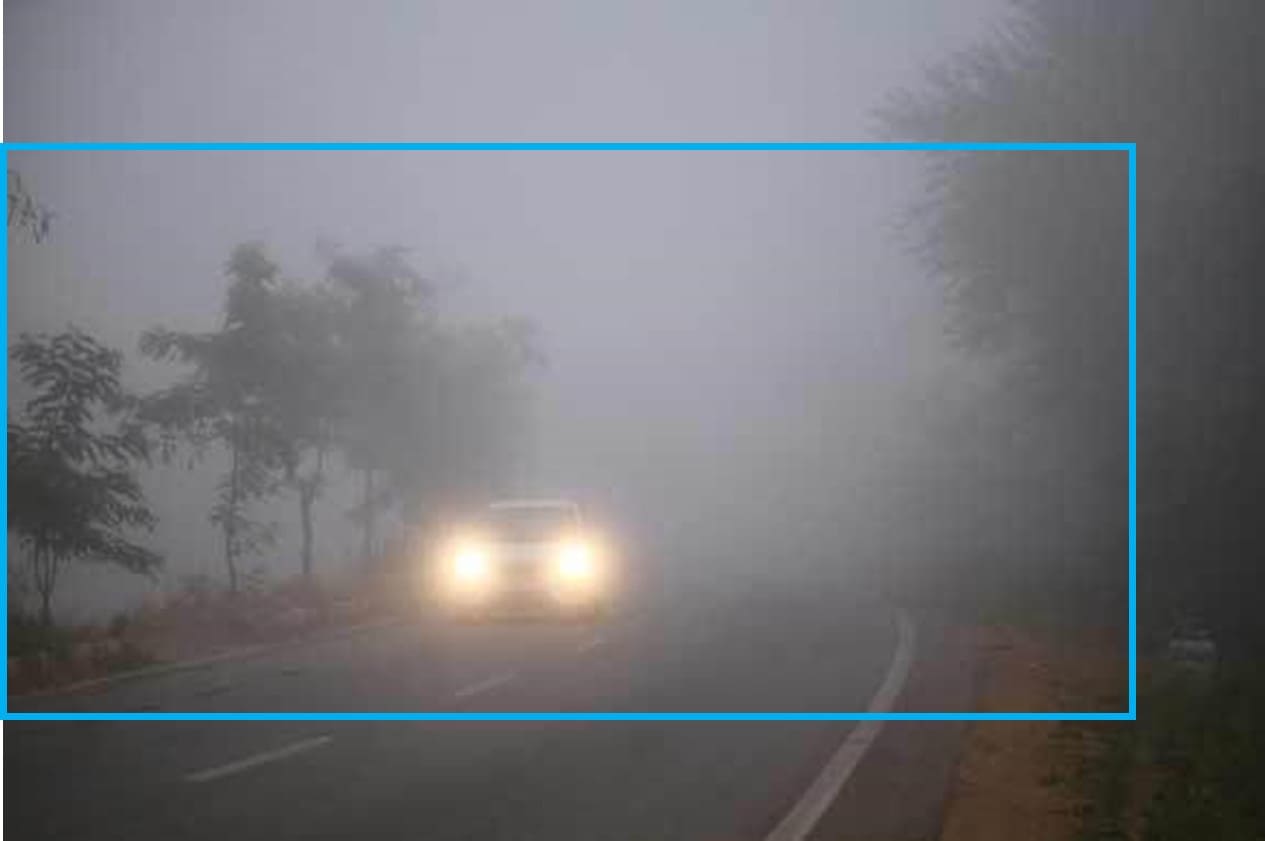} %
    \hfill 
    \includegraphics[width=\linewidth]{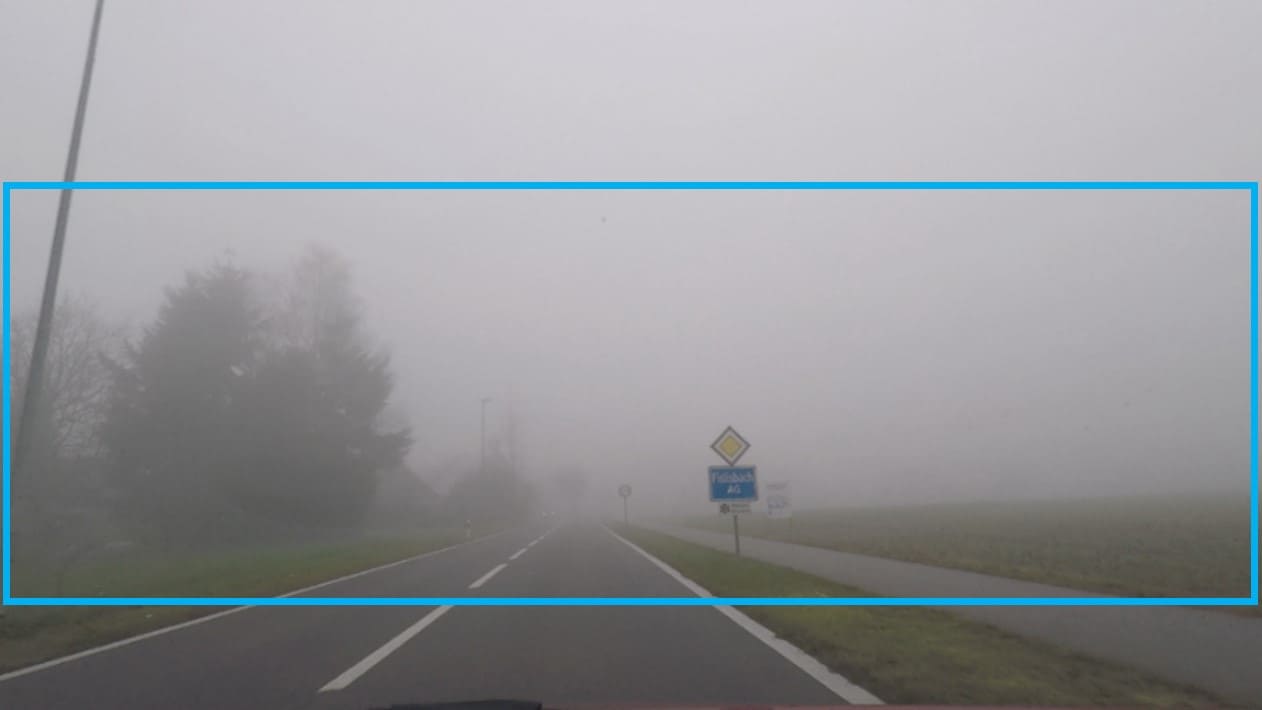}
    \hfill
    } &
    { 
    \vspace{0cm}
    \includegraphics[width=\linewidth]{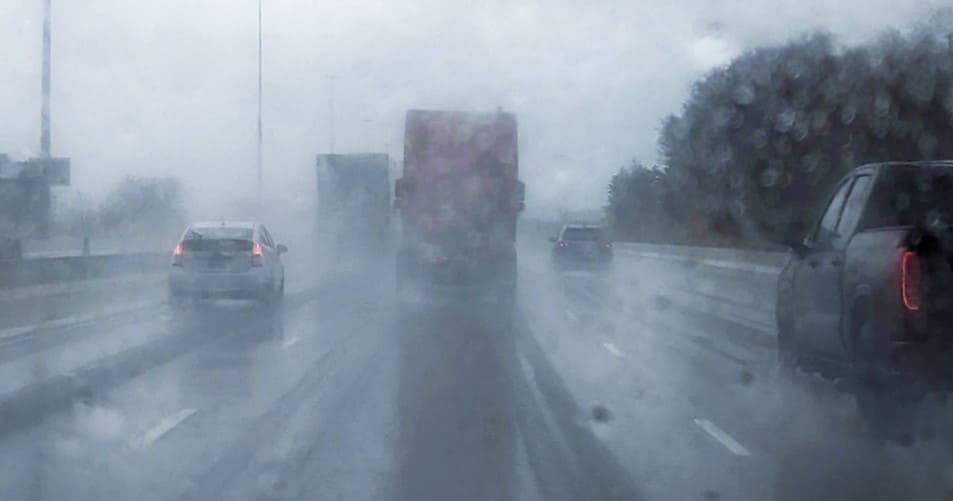} \hfill
    \includegraphics[width=\linewidth]{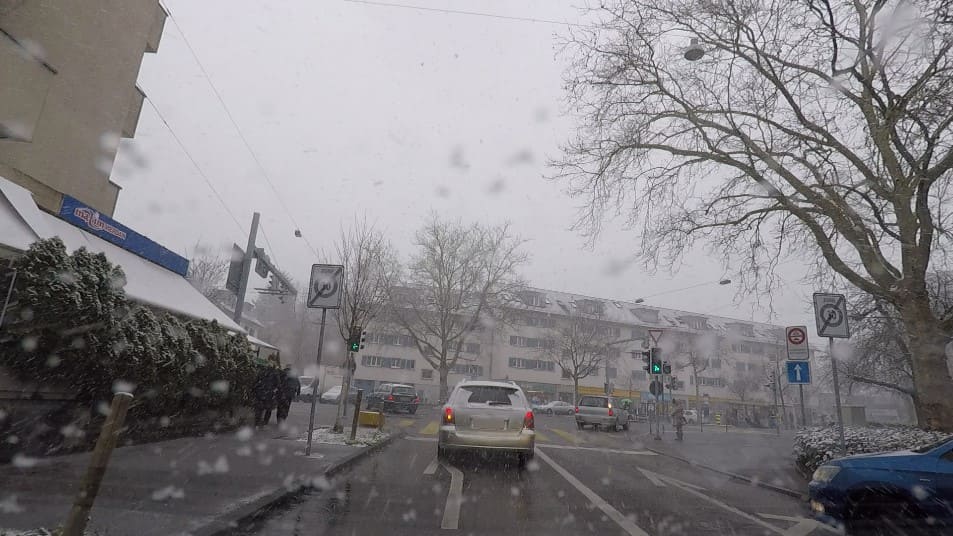} \hfill
    } \\

    \vspace{0cm}
    \rotatebox{90}{\textbf{Trailer-Bicycle-Trailer}} &
    \vspace{0cm}
    \begin{itemize}
        \item Bicycle trailer for children
        \item Bicycle trailer to transport things
        \item Also if there is no bicycle in the front (\eg, if the bicycle trailer stands on the sidewalk)
        \item Trailer must be intended for bicycles/E-bikes and not for vehicles and motorcycles
        \item Bicycle trailer has to be a trailer and no cargo bike (see negative example)
    \end{itemize}
    &
    {
    \vspace{0cm}
    \includegraphics[width=\linewidth]{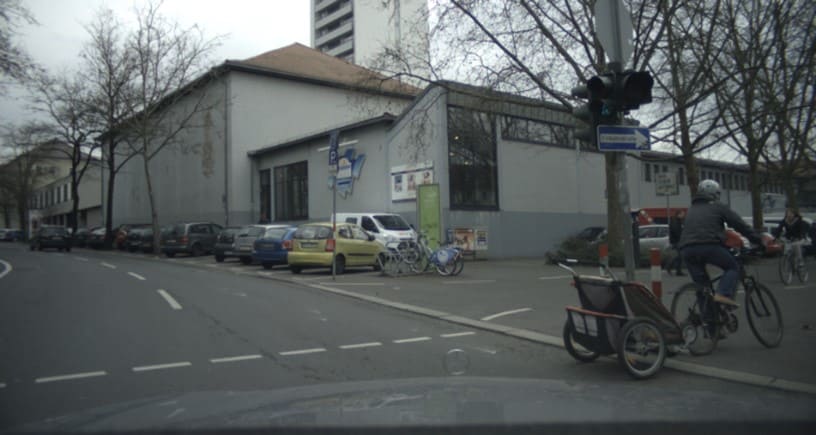} %
    \hfill 
    \includegraphics[width=\linewidth]{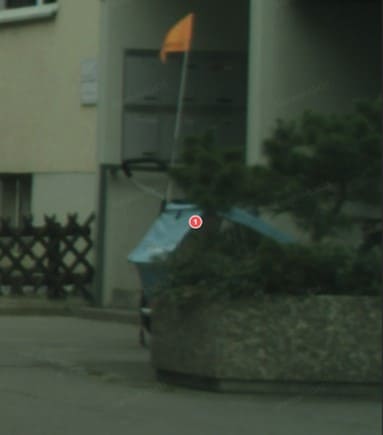}
    \hfill
    } &
    { 
    \vspace{0cm}
    \includegraphics[width=\linewidth]{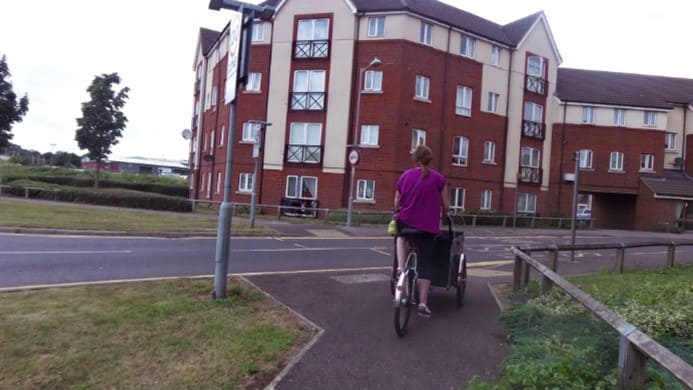} \hfill
    \includegraphics[width=\linewidth]{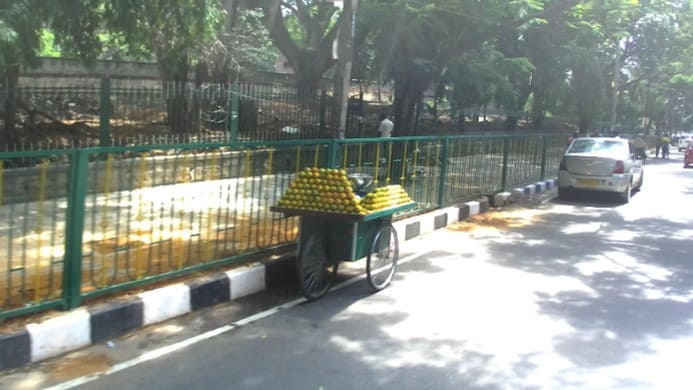} \hfill
    \includegraphics[width=\linewidth]{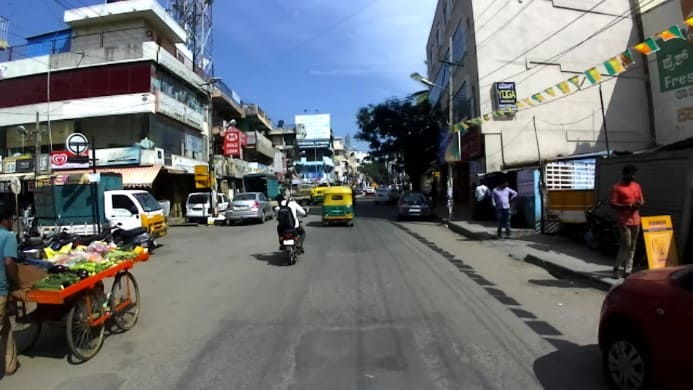} \hfill
    } \\
    \hline

    \vspace{0cm}
    \rotatebox{90}{\textbf{Sign-Train-Sign}} &
    \vspace{0cm}
    \begin{itemize}
        \item There has to be a train symbol on the traffic sign
        \item Symbols could be a modern train, steam train or a tram
        \item Only traffic signs and no real trains
        \item Hard negative examples: Any other traffic signs that warn of a railway crossing, \eg, Andrew cross
    \end{itemize}
    &
    {
    \vspace{0cm}
    \includegraphics[width=\linewidth]{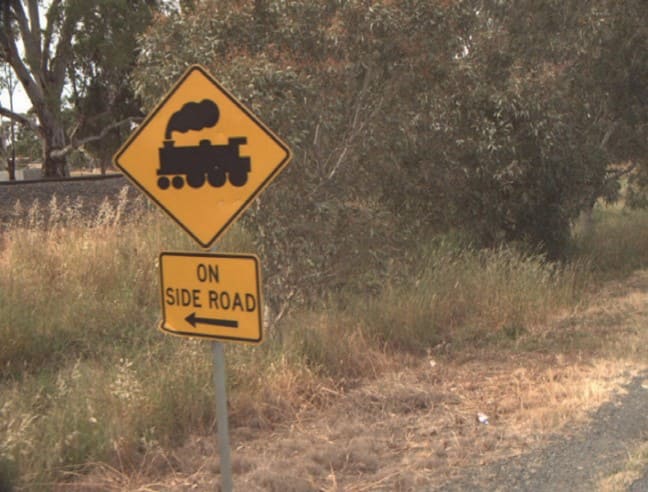} %
    \hfill 
    \includegraphics[width=\linewidth]{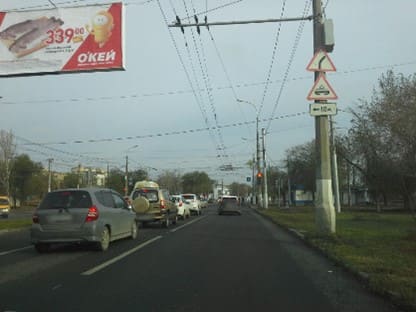}
    \hfill
    \includegraphics[width=\linewidth]{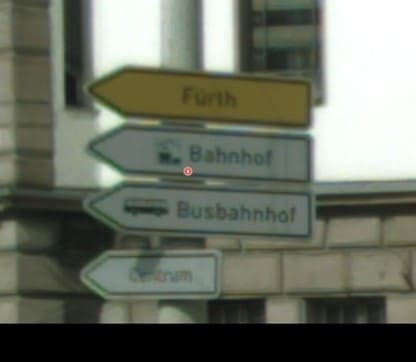}
    \hfill
    } &
    { 
    \vspace{0cm}
    \includegraphics[width=\linewidth]{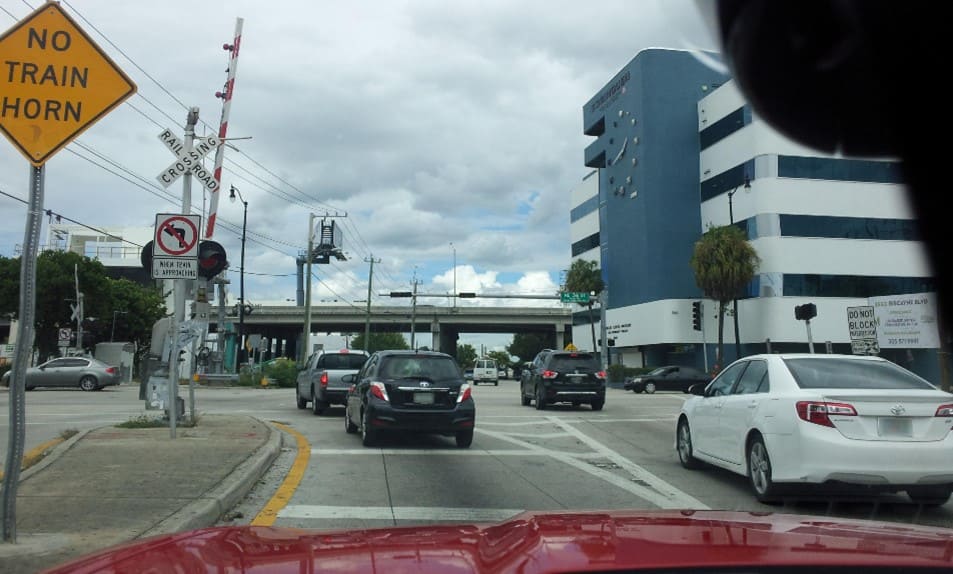} \hfill
    \includegraphics[width=\linewidth]{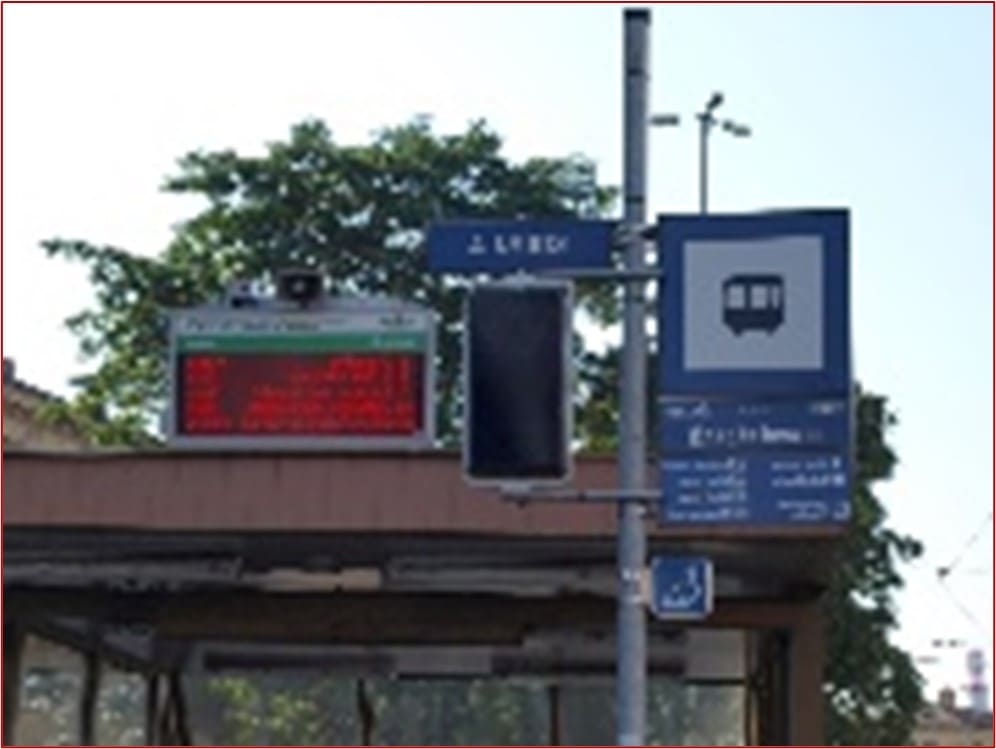} \hfill
    } \\
    
    \vspace{0cm}
    \rotatebox{90}{\textbf{Vehicle-Medical}} &
    \vspace{0cm}
    \begin{itemize}
        \item All kinds of ambulance vehicles, usually red/orange and white colored
        \item Ambulances can usually be identified by their text lettering (\eg, Ambulance, Ambulancia, Krankenwagen, Emergency Medical Service, American Red Cross, Rettungsdienst, Rotes Kreuz)
        \item Some vehicles are both police and ambulance vehicles
        \item Some vehicles come from both the fire department and the ambulance
        \item Hard negative example: Small fire truck
    \end{itemize}
    &
    {
    \vspace{0cm}
    \includegraphics[width=\linewidth]{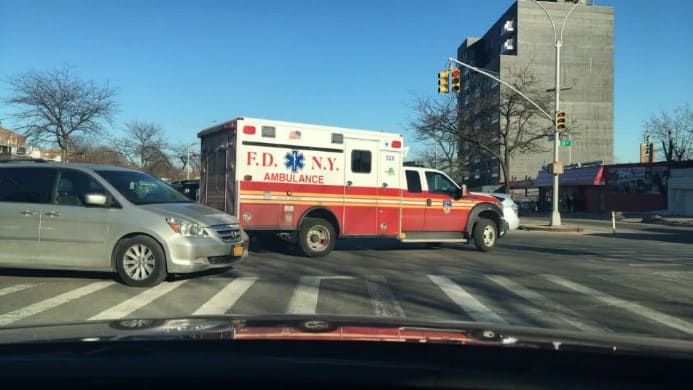} %
    \hfill 
    \includegraphics[width=\linewidth]{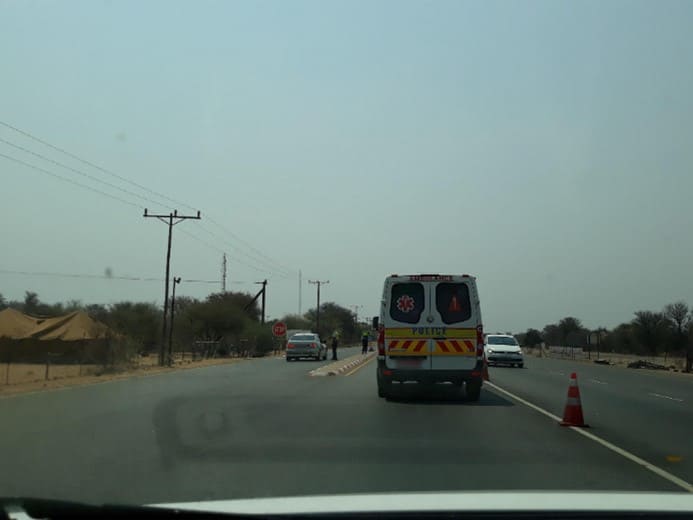}
    \hfill
    \includegraphics[width=\linewidth]{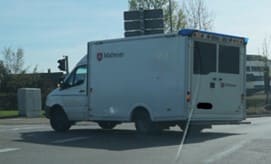}
    \hfill
    } &
    { 
    \vspace{0cm}
    \includegraphics[width=\linewidth]{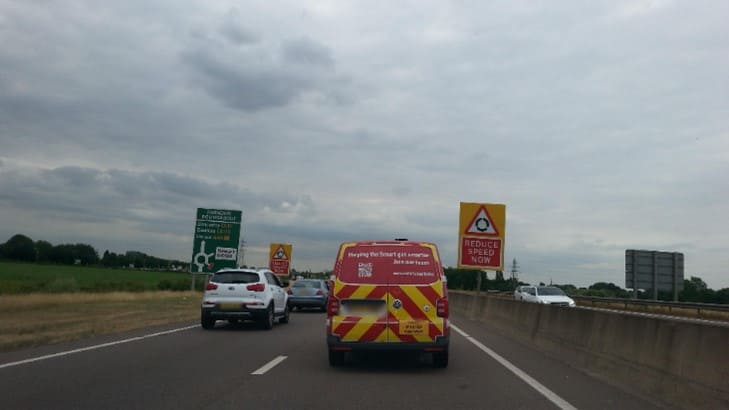} \hfill
    } \\
    
    \bottomrule
    \caption{SearchAD labeling guidelines: For each category, we provide one example class to showcase the labeling guidelines. All positive and negative images are extracted from the eleven SearchAD datasets \cite{LostFounddetectingPinggera2016, UnifyingPanopticSegmentationZendel2022, ACDCAdverseConditionsSakaridis2021, IDDDatasetExploringVarma2019, VisionmeetsroboticsGeiger2013, CityscapesDatasetSemanticCordts2016, MapillaryVistasDatasetNeuhold2017, EuroCityPersonsNovelBraun2019, nuScenesMultimodalDatasetCaesar2020, BDD100KDiverseDrivingYu2020, MapillaryTrafficSignErtler2020}.}
    \label{tab:labeling_guidelines}
\end{longtable}
}

%% file: tikz/confusion_matrix.tikz
\begin{tikzpicture}
    \begin{axis}[
            scale=0.275, 
            colormap={bluewhite}{color=(white) rgb255=(90,96,191)},
            xlabel=Searched, 
            xlabel style={at={(axis description cs:0.5,1)}, anchor=south, yshift=0pt}, 
            ylabel=Found, 
            xticklabels={Cow, Horse, Sign},
            xtick={0,...,3},
            yticklabels={Cow, Horse, Sign},
            ytick={0,...,3},
            ytick style={draw=none},
            enlargelimits=false,
            xticklabel style={rotate=45.0,anchor=east},
            xtick style={draw=none},
            nodes near coords={\pgfmathprintnumber\pgfplotspointmeta},
            nodes near coords style={
                yshift=-7pt
            },
        ]
        \addplot[
            matrix plot,
            mesh/cols=3,
            point meta=explicit,draw=gray
        ] table [meta=C] {
            x y C
            0 0 23
            1 0 10
            2 0 0

            0 1 2
            1 1 6
            2 1 0

            0 2 5
            1 2 7
            2 2 3

        };
    \end{axis}
\end{tikzpicture}

%% file: tikz/confusion_matrix2.tikz
\begin{tikzpicture}
    \begin{axis}[
            scale=0.275, 
            colormap={bluewhite}{color=(white) rgb255=(90,96,191)},
            xlabel=Searched, 
            xlabel style={at={(axis description cs:0.5,1)}, anchor=south, yshift=0pt}, 
            xticklabels={Door, Hood, Trunk},
            xtick={0,...,3},
            xtick style={draw=none},
            yticklabels={Door, Hood, Trunk},
            ytick={0,...,3},
            ytick style={draw=none},
            enlargelimits=false,
            colorbar,
            xticklabel style={rotate=45.0,anchor=east},
            nodes near coords={\pgfmathprintnumber\pgfplotspointmeta},
            nodes near coords style={
                yshift=-7pt
            },
        ]
        \addplot[
            matrix plot,
            mesh/cols=3,
            point meta=explicit,draw=gray
        ] table [meta=C] {
            x y C
            0 0 42
            1 0 8
            2 0 9

            0 1 0
            1 1 0
            2 1 5

            0 2 11
            1 2 12
            2 2 44

        };
    \end{axis}
\end{tikzpicture}